\documentclass{article}

\usepackage[preprint]{neurips_2026}

\usepackage[utf8]{inputenc} % allow utf-8 input
\usepackage[T1]{fontenc}    % use 8-bit T1 fonts
\usepackage{hyperref}       % hyperlinks
\usepackage{url}            % simple URL typesetting
\usepackage{booktabs}       % professional-quality tables
\usepackage{amsfonts}       % blackboard math symbols
\usepackage{nicefrac}       % compact symbols for 1/2, etc.
\usepackage{microtype}      % microtypography
\usepackage[table]{xcolor}        % colors

\usepackage{microtype}
\usepackage{graphicx}
\usepackage{amsmath}
\usepackage{caption}
\usepackage{subcaption}
\usepackage{microtype}
\usepackage{longtable}
\usepackage{setspace}
\usepackage{makecell}
\usepackage{booktabs} % for professional tables
\usepackage{multirow}
\usepackage{siunitx} 
\usepackage{algorithm}
\usepackage{algorithmic}

\usepackage{amsmath}
\usepackage{amssymb}
\usepackage{mathtools}
\usepackage{amsthm}
\usepackage{multirow}
\title{Geometry-Aware Deep Congruence Networks for Manifold Learning in Cross-Subject Motor Imagery}

\author{
Sanjeev Manivannan \\
Department of Biotechnology\\
Indian Institute of Technology Madras\\
\texttt{sanjeev.manivannan@gmail.com}
\And
Chandra Shekar Lakshminarayan \\
Department of Data Science and Artificial Intelligence\\
Indian Institute of Technology Madras\\
\texttt{chandrashekar@cse.iitm.ac.in}
}

\begin{document}

\maketitle

\begin{abstract}
Cross-subject motor-imagery decoding remains a major challenge in EEG-based brain--computer interfaces. To mitigate inter-subject variability, recent work has emphasized manifold-based approaches operating on covariance representations. Yet dispersion scaling and orientation alignment remain largely unaddressed in existing methods. In this paper, we address both issues through congruence transforms and introduce three complementary geometry-aware models: (i) Discriminative Congruence Transform (DCT), (ii) Deep Linear DCT (DLDCT), and (iii) Deep DCT--UNet (DDCT--UNet). These models are evaluated both as pre-alignment modules for downstream classifiers and as end-to-end discriminative systems trained via cross-entropy backpropagation with a custom logistic-regression head. Across challenging benchmarks, our framework improves transductive cross-subject accuracy by $2$--$3\%$, demonstrating the value of geometry-aware congruence learning.
\end{abstract}

\section{Introduction}
Motor-imagery (MI) brain–computer interfaces (BCIs) translate neural activity into control signals ~\cite{wolpaw2002bci}, enabling interaction without overt muscular movement. Electroencephalography (EEG) is the most widely used non-invasive recording modality due to its high temporal resolution ~\cite{blankertz2006bbc}. In MI-BCIs, subjects execute or imagine executing specific motor actions (e.g., left/right hand or foot movements) which are decoded from EEG recordings~\cite{wang2012motor}. From a clinical perspective, MI-BCIs have been explored for post-stroke rehabilitation, with studies reporting motor improvements using BCI–driven robotic therapy ~\cite{ang2015rehab}.

Decoding brain signals requires machine learning to uncover high-level representations from noisy EEG data (see Appendix, Fig.~\ref{fig:1020})~\cite{saeidi2021neural}. In typical MI-BCI datasets, subjects follow a structured experimental protocol in which they either perform or imagine motor tasks. For instance, \textit{BCI Competition IV-2a (BCI-IV-2a)} \cite{blankertz2007bci} contain only real movements  whereas \textit{PhysioNet} \cite{goldberger2000physionet} includes both real and imagined movements. EEG signals are recorded using $d$ electrodes over $T$ time samples per trial, yielding a matrix $X \in \mathbb{R}^{d \times T}$.

MI-BCI methods are evaluated under increasingly challenging regimes, ranging from \emph{within-subject, within-session} evaluation, to \emph{within-subject, cross-session} evaluation that accounts for session variability, and finally to \emph{cross-subject evaluation}, in which the models trained on source subjects are tested on unseen target subjects, reflecting real-world scenarios~\cite{lotte2018review}.In cross-subject setting, a commonly used \emph{relaxed} protocol is \emph{transfer learning}, where a few \emph{labeled} target-subject trials are used for adaptation~\cite{jayaram2016transfer}, which in real world scenarios increases calibration burden. In contrast, our method adopts a \emph{transductive cross-subject} protocol, using unlabeled target statistics \textit{\textbf{only for normalization}} without any test-time adaptation or training~\cite{duan2020zero}.

The \emph{most important challenge} is that EEG signals are non-robust due to
inherent physiological variability~\cite{lotte2018review}. Recently, covariance
representations have emerged as a robust alternative~\cite{koles1990csp}. Given
an EEG trial $X \in \mathbb{R}^{d \times T}$, its covariance
$C = XX^\top \in \mathbb{R}^{d \times d}$ is symmetric positive definite (SPD)
and therefore lies on a Riemannian manifold rather than in Euclidean
space~\cite{barachant2012riemannian}. Geometrically, SPD covariance represents an ellipsoid, where eigenvalues encode dispersion and eigenvectors define orientation. So, they rely on affine-invariant metrics (e.g., geodesic distances) rather than Euclidean norms~\cite{barachant2013riemannian}, necessitating Riemannian geometry-aware learning. Classical Riemannian methods such as Minimum-Distance-to-Mean (MDM), Tangent Space Logistic Regression (TSLR), and Tangent Space Analysis with LDA (TSA-LDA)~\cite{barachant2012mdm} uses affine-invariant metrics and project covariances from the SPD manifold to local tangent spaces via logarithmic maps, with predictions transferred back through exponential maps~\cite{barachant2012mdm}. In contrast, most existing \emph{deep MI decoding models} operate in Euclidean space and do not preserve SPD geometry ~\cite{bronstein2017geometric}.

\textbf{Our Contributions:} We summarize the key algorithmic and experimental contributions as follows:

$\bullet$ \textbf{Algorithmic:} We propose three geometry-aware congruence transformations. The first, \textbf{Discriminative Congruence Transformation (DCT)}, (i) learns class-specific orientation and dispersion scaling via a single congruent transformation, and (ii) incorporates Fisher discriminant penalties to separate between- and within-class variability. The deep variants, \textbf{Deep Linear DCT (DLDCT)} and \textbf{Deep DCT--UNet (DDCT--UNet)}, extend DCT beyond tangent-space representations via deep congruence-based networks, with DDCT--UNet adopting a U-Net encoder--decoder with multi-scale fusion to introduce nonlinearity. We further introduce End-to-End (E2E) variants, where cross-entropy gradients from logistic regression are backpropagated to pre-aligner (DCT, Deep DCTs) parameters.

%\textbf{Our work:}
%We propose geometry-aware congruence transformations on the SPD manifold~\cite{pennec2006riemannian} to overcome limitations of Riemannian Alignment (RA)~\cite{barachant2012riemannian}, which performs mean (whitening) alignment but ignores residual \emph{orientation} and \emph{dispersion} distortions, limiting cross-subject discriminability. We extend this to discriminative, multi-scale manifold transformations, with novelty in the geometry-aware application of congruence transforms and Fisher objectives (discussed below) to EEG covariances for cross-subject alignment (Table~\ref{tab:novelty_congruence}).

%$\bullet$ \textbf{Discriminative Congruence Transformation (DCT):} We fill the gap in RA using two steps: (i) learning class-specific orientation and dispersion scaling via a single congruent transformation, and (ii) incorporating Fisher discriminant penalties to explicitly separate between- and within-class variability. Thus, DCT improves action discriminability over RA by suppressing subject-dependent~distortions.

%$\bullet$ \textbf{Deep Discriminative Congruence Models:} We generalize DCT beyond tangent-space representations via deep stacks of \emph{intermediate}-dimension linear congruence transforms on the SPD manifold, termed \textbf{Deep Linear DCT (DLDCT)}. We further propose \textbf{Deep DCT--UNet (DDCT--UNet)}, an encoder--decoder architecture with multi-scale feature fusion to introduce nonlinearity and incorporates subject-wise variance regularization to mitigate identity drift in deep congruence stacks.
\vspace{-2mm}
\begin{table}[h]
\centering
\footnotesize
\caption{Novelty progression of proposed models.}
\setlength{\tabcolsep}{3pt}
\begin{tabular}{lccccccc}
\toprule
\textbf{Method} & \textbf{SPD} & \textbf{Mean Align} & \textbf{Orientation} & \textbf{Dispersion} & \textbf{Tangent-free} & \textbf{Intermediate dimensions} & \textbf{Nonlinear} \\
\midrule
Vanilla Congruence & \checkmark & $\times$ & $\times$ & $\times$ & $\times$ & $\times$ & $\times$ \\
Vanilla RA & \checkmark & \checkmark & $\times$ & $\times$ & $\times$ & $\times$ & $\times$ \\
DCT & \checkmark & \checkmark & \checkmark & \checkmark & $\times$ & $\times$ & $\times$ \\
Deep DCTs & \checkmark & \checkmark & \checkmark & \checkmark &\checkmark & \checkmark & \checkmark \\
\bottomrule
\end{tabular}
\label{tab:novelty_congruence}
\end{table}
$\bullet$ \textbf{Novelty and geometric validation:} While Fisher objectives and dispersion scaling individually are well established, our novelty lies in their unified, geometry-aware integration for cross-subject EEG decoding. A comparison with prior geometry-aware methods, namely Riemannian Alignment~\cite{barachant2012riemannian} and congruence transforms~\cite{pennec2006riemannian}, is given in Table~\ref{tab:novelty_congruence}. We also provide synthetic validation that support the proposed transformations (Table~\ref{tab:synthetic-main}, Appendix~\ref{app:synthetic}).

$\bullet$ \textbf{Experiments:} We consider the following downstream classifiers—MDM, TSLR, and TSA-LDA.
We evaluate our proposed models—DCT, DLDCT, and DDCT--UNet—both as pre-aligners feeding these classifiers and as end-to-end discriminative models under a leave-one-subject-out (LOSO) protocol in the transductive setting~\cite{yger2017riemannian}.
Across multiple MI-EEG benchmarks, our pre-aligners and their E2E-TSLR variants yield $2$--$3\%$ accuracy improvements and outperform both classical methods (MDM, TSLR, TSA-LDA), and also deep baselines (SPD-CNN \cite{gao2022cnn_riemannian, xiong2023riemannian_tangent_cfc}). Detailed comparisons with other deep Euclidean baselines in Appendix~\ref{app:deep_baselines},~\ref{app:novelty_congruence}.

\section{Notation}
EEG trials are represented as $X_{i,e}\in\mathbb{R}^{d\times T}$ for subject $i$
and trial $e$, with $d$ channels and $T$ temporal samples, and are associated
with class labels $y_{i,e}\in\{1,\dots,K\}$ where $K$ denotes the number of classes. From each trial we compute the covariance representation $C_{i,e} \in \mathbb{S}_{++}^d$ as
\[
C_{i,e}
=
\frac{1}{T-1} X_{i,e}X_{i,e}^\top
\;\in\;
\mathbb{S}_{++}^d .
\]
\textbf{Tangent-Space Mapping:} (see
Appendix, Fig.~\ref{fig:riemannian_manifold_ts}) Let $C_t\in\mathbb{S}_{++}^d$ denote a reference matrix with tangent space
$T_{C_t}\mathbb{S}_{++}^d=\{Z\in\mathbb{R}^{d\times d}\mid Z=Z^\top\}$. The log- and exp- maps at $C_t$ are defined as:
\begin{subequations}
\begin{align}
\label{eq:log-map}
Z=\log_{C_t}(C)
&=
C_t^{1/2}
\log\!\big(C_t^{-1/2} C C_t^{-1/2}\big)
C_t^{1/2},
\\
\label{eq:exp-map}
\exp_{C_t}(Z)
&=
C_t^{1/2}
\exp\!\big(C_t^{-1/2} Z C_t^{-1/2}\big)
C_t^{1/2}.
\end{align}
\end{subequations}
Throughout this paper, $X_{i,e}$ denotes trial-level EEG signals,
$C_{i,e}\in\mathbb{S}_{++}^d$ their covariance, and
$Z_{i,e}:=\log_{C_t}(C_{i,e})\in T_{C_t}\mathbb{S}_{++}^d$ denotes their tangent (Euclidean) space
representation.

\section{Generic Fisher Discriminants On Tangent-Space Matrices}
In this section, we define the Fisher discriminant loss to separate within- and
between-class variability, where the class $h$ corresponds to either subject
$(S\equiv i)$ or action $(A\equiv y_{i,e})$.
Let $\{\mathcal{G}_m\}_{m=1}^M$ be a partition of the index set $\{(i,e)\}$ into $M$
groups. The number of trials in group $m$ is $n_m = |\mathcal{G}_m|$, with the total number of trials $N = \sum_{m=1}^M n_m$. Define the group-wise means and global mean as $\bar{Z}_m
=
\frac{1}{n_m}
\sum_{(i,e)\in\mathcal{G}_m}
Z_{i,e},
\,
\bar{Z}
=
\frac{1}{N}
\sum_{m=1}^M
n_m\,\bar{Z}_m .
$ The within- and between-group scatters are
\begin{subequations}
\label{eq:generic_fisher_scatters}
\begin{align}
W_h(\{Z_{i,e}\},\{\mathcal{G}_m\})
=
\frac{1}{N}
\sum_{m=1}^M
\sum_{\mathcal{G}_m}
\|Z_{i,e}-\bar{Z}_m\|_F^2 ,
\quad
B_h(\{Z_{i,e}\},\{\mathcal{G}_m\})
=
\frac{1}{N}
\sum_{m=1}^M
n_m\,\|\bar{Z}_m-\bar{Z}\|_F^2 .
\label{eq:generic_fisher}
\end{align}
\end{subequations}
where $Z_{i,e}\in T_{C_t}\mathbb{S}_{++}^d$ in the 
tangent-space which allow efficient Frobenius-norm discriminants.

\section{Discriminative Congruence Transform (DCT)}
\label{sec:dct}
In this section, we first discuss Riemannian Alignment (RA) and its geometric
limitations. We then introduce DCT, which augments RA with two learnable
tangent-space operations: a global dispersion scaling that regulates eigenvalue
structure, and an orientation-preserving rotation that realigns
action-discriminative eigendirections (see Table~\ref{tab:synthetic-main} for synthetic validation). We subsequently formulate a Fisher-style
objective to learn these parameters directly from data, together with auxiliary
regularizers that maintain geometric consistency. The complete procedure is
summarized in Algorithm~\ref{alg:dct}.

\subsection{Riemannian Alignment (RA)}
A principal obstacle to transductive cross-subject learning is that each subject exhibits distinct covariance distribution. RA reduces this inter-subject shift
by subject-wise whitening with respect to a subject-specific reference
covariance $C^{\mathrm{ref}}_i\in\mathbb{S}_{++}^d$. This reference is defined as the affine-invariant Riemannian mean~\cite{barachant2012riemannian}, computed by
\[
C^{\mathrm{ref}}_i
=
\arg\min_{C\in\mathbb{S}_{++}^d}
\sum_{e}
\big\|
\log\!\big(
C^{-1/2}\,C_{i,e}\,C^{-1/2}
\big)
\big\|_F^2 ,
\]
where $\|\cdot\|_F$ denotes the Frobenius norm and $\log(\cdot)$ is the matrix
logarithm. This objective selects $C\in\mathbb{S}_{++}^d$ which minimizes the sum of
squared geodesic distances to subject’s trials. Once Riemannian mean is computed, RA applies congruence (whitening) transform~\cite{pennec2006riemannian}.
\[
C'_{i,e}
=
\big(C^{\mathrm{ref}}_i\big)^{-1/2}\,
C_{i,e}\,
\big(C^{\mathrm{ref}}_i\big)^{-1/2},
\qquad
C'_{i,e}\in\mathbb{S}_{++}^d.
\]
This mapping centers all subjects at a common reference point:
the Riemannian mean of $\{C'_{i,e}\}_e$ equals the identity matrix $I$.
This is why RA removes subject-wise mean offsets, a major source of
inter-subject variability \cite{barachant2012riemannian} but it does not ensure alignment of
class-conditional covariance \emph{shapes} across subjects which is the key problem tackled in our work.

\paragraph{Covariance ellipsoids:}
Each aligned covariance $C'_{i,e}\in\mathbb{S}_{++}^d$ defines a covariance ellipsoid whose
\emph{principal axes} and \emph{axis lengths} are governed by its
eigendecomposition $C'_{i,e}
=
U_{i,e}\,\Lambda_{i,e}\,U_{i,e}^{\top}$,
where $
\Lambda_{i,e}=\mathrm{diag}\big(\lambda^{(1)}_{i,e},\lambda^{(2)}_{i,e},\ldots,\lambda^{(d)}_{i,e}\big)$ and $U_{i,e}=\big[u^{(1)}_{i,e},u^{(2)}_{i,e},\ldots,u^{(d)}_{i,e}\big]$. Here, $\{\lambda^{(j)}_{i,e}\}_{j=1}^d$ are the eigenvalues capturing \emph{\textbf{dispersion}}, and $\{u^{(j)}_{i,e}\}_{j=1}^d$ are the eigenvectors capturing \emph{\textbf{orientation}}. Although RA aligns the mean geometry (mean-to-identity), action-specific
ellipsoids can remain anisotropically scaled and rotated across subjects,
inflating within-class scatter and degrading separability.

\begin{algorithm}[tb]
\caption{Discriminative Congruence Transform (DCT)}
\label{alg:dct}
\begin{algorithmic}
\setlength{\baselineskip}{13pt}

\STATE \textbf{Input:} Covariance matrices $C_{i,e}$; labels $y_{i,e}$

\STATE \textbf{Riemannian Alignment:}
compute $C^{\mathrm{ref}}_i$ and whiten $C'_{i,e}$

\STATE Compute tangent representations $L'_{i,e}=\log(C'_{i,e})$

\STATE Initialize generator $A=0$, rotation $R=\exp(A-A^\top)$, and dispersion scale $\gamma$

\FOR{each training iteration}
\STATE \textit{Forward pass:} update rotation $R=\exp(A-A^\top)$
\STATE Dispersion scaling (Eq.~\eqref{eq:dispersion_gamma}): $L^{\mathrm{DS}}_{i,e}=\gamma L'_{i,e}$ and rotation (Eq.~\eqref{eq:orientation_r}): $L^{O}_{i,e}=R^\top L^{\mathrm{DS}}_{i,e} R$
\STATE Compute $\mathcal{L}_{\mathrm{Fisher}}(R,\gamma)$ (Eq.~\eqref{eq:fisher_loss_dct}), $\mathcal{R}(R,\gamma)$ (Eq.~\eqref{eq:reg_loss_dct}), and total loss $\mathcal{L}_{\mathrm{DCT}}$ (Eq.~\eqref{eq:dct_loss})
\STATE \textit{Backpropagate:} update $A$ and $\gamma$

\ENDFOR

\STATE Finalize rotation $R^\star=\exp(A-A^\top)$ and Apply final transform:
$
{C}^{out}_{i,e}
=
\exp\big(R^{\star\top}\,
\gamma^{\star} L'_{i,e}\,
R^\star\big)$

\STATE \textbf{Output:} trained $R^\star$, $\gamma^{\star}$ and ${C}^{out}_{i,e}$

\end{algorithmic}
\end{algorithm}

\subsection{Proposed Method (DCT)}
To address this challenge of dispersion scaling and orientation alignment, we introduce the \emph{Discriminative Congruence Transform (DCT)} to enhance RA
by explicitly modeling dispersion and orientation in a shared space as shown in Figure ~\ref{fig:prealigners} in the Appendix. In RA \cite{barachant2012riemannian}, we obtain tangent-space representations at the identity by mapping aligned covariances via the Riemannian logarithm map defined in~\eqref{eq:log-map}, yielding the simplified projection
\begin{equation}
L'_{i,e}
=
\log_I(C'_{i,e})
=
\log(C'_{i,e})
\label{eq:log-map-identity}
\end{equation}
\paragraph{Dispersion Scaling:}
RA can suppress action-dependent variability ~\cite{zanini2018transfer} in the tangent space. DCT reintroduces a learnable global dispersion
via
\begin{equation}
L^{\mathrm{DS}}_{i,e} (\gamma)
=
\gamma\,L'_{i,e},
\qquad \gamma>0,
\label{eq:dispersion_gamma}
\end{equation}
which expands/contract class spread while preserving the shared
mean geometry induced by RA.

\paragraph{Orientation Alignment:}
Even after dispersion, action information may remain embedded in
inconsistent principal directions across subjects. In DCT we learn a
global orthogonal rotation $R\in SO(d)$ which is a tangent-space congruence transform,
\begin{equation}
L^{\mathrm{O}}_{i,e}(\gamma)
=
R^{\top}\,L^{\mathrm{DS}}_{i,e}(\gamma)\,R.
\label{eq:orientation_r}
\end{equation}
This aligns action-discriminative axes across subjects while decoupling class
geometry from residual subject-specific rotations. The rotation is constrained, introducing no scaling beyond dispersion scale~$\gamma>0$. We enforce this constraint by parameterizing
$R=\exp(A-A^\top),$
where $A\in\mathbb{R}^{d\times d}$ is unconstrained so that $A-A^\top$ is
skew-symmetric and $\exp(A-A^\top)$ is orthogonal, yielding $R^\top R=I$
throughout training; formal proof provided in
Appendix~\ref{lem:skewexp_in_SO}.

\paragraph{Fisher Objective:}
From the action labels $y_{i,e}$ we define the action index sets
$\{\mathcal{I}_k\}_{k=1}^K$ with $\mathcal{I}_k=\{(i,e):y_{i,e}=k\}$.
The generic within- and between-group scatter operators in
Eq~\eqref{eq:generic_fisher} are
instantiated using the \emph{final} oriented outputs
$L^{\mathrm{O}}_{i,e}(\gamma)$ together with the action index set
$\{\mathcal{I}_k\}$, i.e.,
$W_{A}\coloneqq W_A(\{L^{\mathrm{O}}_{i,e}(\gamma)\},\{\mathcal{I}_k\})$ and
$B_{A}\coloneqq B_A (\{L^{\mathrm{O}}_{i,e}(\gamma)\},\{\mathcal{I}_k\})$.
The resulting Fisher objective is
\begin{equation}
\mathcal{L}_{\mathrm{Fisher}}(R,\gamma)
=
\delta\,
\frac{W_A}{B_A+\varepsilon},
\label{eq:fisher_loss_dct}
\end{equation}
where $\delta>0$ weights the term and $\varepsilon>0$ ensures stability.

\paragraph{Regularization:}
Since RA already collapses subject-wise mean shifts, we regularize the
dispersion parameter to remain close to unity and constrain the learned
rotation to stay near the identity to prevent initial deviations affecting performance:
\begin{equation}
\mathcal{R}(R,\gamma)
=
\alpha(\gamma-1)^2
+
\beta_t\,\|R-I\|_F^2,
\label{eq:reg_loss_dct}
\end{equation}
where $\alpha>0$ controls dispersion and $\beta_t>0$ may be optionally scheduled during training.

\paragraph{Total DCT loss:}
The overall objective includes Fisher criterion and regularizers:
\begin{equation}
\begin{aligned}
\mathcal{L}_{\mathrm{DCT}}(R,\gamma)
&=
\mathcal{L}_{\mathrm{Fisher}}(R,\gamma)
+
\mathcal{R}(R,\gamma).
\end{aligned}
\label{eq:dct_loss}
\end{equation}
The complete DCT optimization procedure is summarized in
Algorithm~\ref{alg:dct}. In the next section, we move beyond rotational
alignment and introduce a deeper congruence-based architecture to handle more complex geometric distortions.

\section{Deep Linear Discriminative Congruence Transform (DLDCT)}
\label{sec:dldct}

While DCT projects covariances onto the tangent space at the identity using log-maps and applies dispersion scaling and rotational alignment in that domain, this single tangent-plane correction may discard information present on the original SPD manifold. To address this limitation, in \textit{Deep Linear DCT }(DLDCT) we avoid relying on a single tangent-space transformation and instead operate directly on SPD matrices through a stack of congruent transforms that preserve SPD geometry. DLDCT is based on the assumption that, while a single congruence transform in the tangent space may suffice for DCT, multiple congruence transforms are required while operating on the SPD manifold to gradually move covariances toward a more discriminative configuration (Refer to Synthetic geometric validation of this assumption in Table~\ref{tab:synthetic-main} (Appendix~\ref{app:synthetic}). To enable repeated manifold-valued corrections, DLDCT permits intermediate dimensionality changes across layers, mapping covariances between SPD manifolds $\mathbb{S}_{++}^{d_\ell}$ of different sizes. Each DLDCT layer is explicitly linear; depth arises from composing linear congruence transforms across dimensions. We next describe the SPD congruence layers, geometry-aware weight initialization, and Fisher objectives defining DLDCT.

\paragraph{SPD Congruence Layers:}
Each DLDCT layer operates using congruence transform~\cite{bhatia2009positive}.
\[
\Phi_\ell(C)
=
W_\ell^\top C W_\ell,
\qquad
W_\ell\in\mathbb{R}^{d_{\ell}\times d_{\ell+1}},
\]
where $W_\ell$ is constrained to be full column rank and $\ell$ refers to the SPD linear layer. For
$C\in\mathbb{S}_{++}^{d_{\ell}}$, the congruence map yields a positive
\emph{semi}-definite matrix in $\mathbb{R}^{d_{\ell+1}\times d_{\ell+1}}$; to
guarantee strict positive definiteness for subsequent Riemannian
operations, we add $\varepsilon I$ ($\varepsilon>0$) to ensure numerical stability.
\[
\Phi_\ell^{\varepsilon}(C)
=
W_\ell^\top C W_\ell + \varepsilon I,
\qquad 
\Phi^{(\ell)}(C)
=
\Phi_\ell^{\varepsilon}
\circ
\Phi_{\ell-1}^{\varepsilon}
\circ \cdots \circ
\Phi_1^{\varepsilon}(C),
\]
Stacking $\ell$ such layers as shown above defines the full DLDCT mapping with intermediate dimensions
$(d_0=d),d_1,\dots,d_L$.
With this, DLDCT reshapes covariance structure, adjusts
dimensionality, and learns hierarchical geometry-aware representations on the
SPD manifold.

\begin{algorithm}[tb]
\caption{Deep Linear Discriminative Congruence Transform (DLDCT)}
\label{alg:dldct}
\begin{algorithmic}
\setlength{\baselineskip}{13pt}

\STATE \textbf{Input:} Covariances $C_{i,e}$; labels $y_{i,e}$

\STATE Set hyperparameters 
$\lambda_{act}, \lambda_{sub},
\lambda_w, \lambda_b, \lambda_{rec}$

\STATE \textbf{Riemannian Alignment:}
compute $C^{\mathrm{ref}}_i$ and whiten $C'_{i,e}$

\STATE Initialize $\{W_1,\dots,W_\ell\}$ stacks using Eq.~\eqref{eq:weight_init}

\FOR{each training iteration}

    % ------------------------------------------------------------
    \STATE \textit{Forward pass:} $C^{\mathrm{out}}_{i,e} = \Phi^{(\ell)}(C'_{i,e})$

    % ------------------------------------------------------------
    \STATE Compute
    $L^{out}_{i,e} = \log C^{\mathrm{out}}_{i,e}$ and 
    $L'_{i,e} = \log C'_{i,e}$ 

    % ------------------------------------------------------------
    \STATE Compute Fisher loss $\mathcal{L}_{\mathrm{Fisher}}$ using Eq.~\eqref{eq:fisher_dldct_loss}, and reconstruction loss $\mathcal{L}_{\mathrm{Rec}}$ using Eq.~\eqref{eq:dldct_rec_loss}.
    
    % ------------------------------------------------------------
    \STATE Compute total loss $\mathcal{L}_{\mathrm{DLDCT}}$ using Eq.~\eqref{eq:dldct_loss}

    % ------------------------------------------------------------
    \STATE \textit{Backpropagate:} update the stacks $\{W_\ell\}$ in $\Phi^{(\ell)}$

\ENDFOR

\STATE Save $\{W_\ell^\star\}$ and perform \textit{Forward pass for inference:}
$C^{\mathrm{out}}_{i,e} = \Phi^{(\ell)}(C'_{i,e})$ 

\STATE \textbf{Output:} trained parameters $\{W_\ell^\star\}$ in $\Phi^{(\ell)\star}$ and $C^{\mathrm{out}}_{i,e}$

\end{algorithmic}
\end{algorithm}

\paragraph{Geometry-Aware Weight Initialization.}
To stabilize deep congruence learning, we estimate a global covariance structure as the Euclidean mean of the RA-aligned covariances,
\(
M = \frac{1}{N}\sum_{i,e} C'_{i,e},
\)
and compute its eigendecomposition \(M = U \Lambda U^\top\), where
\(U\in\mathbb{R}^{d\times d}\) is orthonormal. For a layer
\(d_\ell \!\to\! d_{\ell+1}\), we form a lifted basis \(U'\) by column-wise
concatenation of \(U\) repeated \(k\) times defined as:
\begin{equation}
\label{eq:lifted_U_main}
U'
=
\frac{1}{\sqrt{k}}
\big[\, \underbrace{U \ \cdots \ U}_{k\ \text{times}} \,\big],
\qquad
k=\left\lceil\frac{d_{\ell+1}}{d}\right\rceil .
\end{equation}
We then initialize the SPD linear transformation as
\begin{equation}
W_\ell = U'[1{:}d_\ell,\,1{:}d_{\ell+1}].
\label{eq:weight_init}
\end{equation}
This empirically yields well-conditioned, near full-rank operators at
initialization and reduces reliance on \(\varepsilon I\) regularization; full
details are provided in Appendix~\ref{app:init}.

\paragraph{Fisher Objective:}
\label{sec:dldct_fisher}
While DLDCT applies congruence transforms directly on SPD matrices, optimizing Fisher-style objectives on $\mathbb{S}^{d}_{++}$ is costly and would require recomputing Riemannian means at every layer, and adding $\varepsilon I$ can induce subject-dependent drifts from the identity-centered geometry. We therefore evaluate all discriminative objectives in the tangent space for stable training. When we compute the tangent-space representations of the input and output
covariances at the identity using Eq.~\eqref{eq:log-map-identity} with reference
$C_t=I$, we obtain
$
L^{\mathrm{out}}_{i,e}=\log(C^{\mathrm{out}}_{i,e})
$
and
$
L'_{i,e}=\log(C'_{i,e}).
$
Action labels $y_{i,e}$ define index sets
$
\mathcal{I}_k=\{(i,e):y_{i,e}=k\},
$
and subject identities define trial sets
$
\mathcal{E}_s=\{(i,e):i=s\}.
$
Action-wise scatters $W_A,B_A$ are obtained by instantiating
\eqref{eq:generic_fisher} with
$\{L^{\mathrm{out}}_{i,e}\}$ and $\{\mathcal{I}_k\}$, while subject-wise
scatters $W_S,B_S$ are computed analogously using $\{\mathcal{E}_s\}$.
Using these quantities, we form a Fisher-style objective that promotes action
separability while suppressing subject-specific variability:
\begin{equation}
\begin{aligned}
\mathcal{L}_{\mathrm{Fisher}}^{\mathrm{DLDCT}}
&=
\lambda_{act}\big(\lambda_{w}W_A-\lambda_{b}B_A\big)
+\lambda_{sub}\big(\lambda_{b}B_S-\lambda_{w}W_S\big),
\end{aligned}
\label{eq:fisher_dldct_loss}
\end{equation}
where $\lambda_{act}, \lambda_{sub} > 0$ weight action and subject terms, and $\lambda_w,\lambda_b > 0$ balance within- and between-class scatter terms.

\paragraph{Reconstruction Loss:}
To preserve geometric fidelity across DLDCT layers, we use a reconstruction term that penalizes deviations from the layer input. When dimensionality is preserved, we define
\begin{equation}
\mathcal{L}_{\mathrm{Rec}}
=
\frac{1}{N}
\sum_{i,e}
\big\|
L^{\mathrm{out}}_{i,e}-L'_{i,e}
\big\|_F^2 .
\label{eq:dldct_rec_loss}
\end{equation}
When dimensionality changes, reconstruction is evaluated on the shared
subspace by comparing principal blocks, ensuring aligned variances and
consistent covariance centers.

\paragraph{Total DLDCT Loss:}
The final objective includes Fisher criterion with reconstruction
regularizer:
\begin{equation}
\mathcal{L}_{\mathrm{DLDCT}}
=
\mathcal{L}_{\mathrm{Fisher}}^{\mathrm{DLDCT}}
+
\lambda_{rec}\mathcal{L}_{\mathrm{Rec}} .
\label{eq:dldct_loss}
\end{equation}
where $\lambda_{rec}>0$
control the relative importance of each term, and $\lambda_w,\lambda_b>0$
balance within- and between-scatter contributions.

The complete optimization procedure is summarized in
Algorithm~\ref{alg:dldct}. We next introduce a deep congruence-based model that incorporates nonlinear manifold operations.

\section{Deep Discriminative Congruence Transform U-Net (DDCT--UNet)}
While DLDCT employs deep stacks of geometry-preserving linear congruence transforms, we replace it with a U-Net--style encoder--decoder architecture in DDCT--UNet (as shown in Figure~\ref{fig:classifiers} in the Appendix), using the same SPD layers but introducing nonlinearity through multi-scale skip--merge fusion \cite{ronneberger2015unet} (see Table~\ref{app:synthetic}, Appendix~\ref{app:synthetic} for synthetic validation). The encoder and decoder are compositions of linear congruence maps with residual-style skip paths where nonlinearity arises through skip-merge from log--exp fusion acting as a manifold-valued activation.
\paragraph{Skip--Merge:}
ResNet-style skip connections \cite{he2016resnet} pass intermediate SPD features linearly to the decoder, while fusion is performed using the log--Euclidean merge
\[
\mathrm{merge}(C_1,C_2)
=
\exp\!\Big(\tfrac{1}{2}\big(\log C_1+\log C_2\big)\Big),
\]
where log- and exp-maps at identity ($C_t=I$) are computed via
Eqs.~\eqref{eq:log-map}--\eqref{eq:exp-map}, introducing a nonlinear
manifold-valued transformation that causes skip fusion to
act as a geometry-aware activation.
\paragraph{Encoder--Decoder Map:}
The encoder is formulated as
\[
B = E(C)
=
\Phi_{enc}^{(\ell)} \circ \cdots \circ \Phi_{enc}^{(1)}(C),
\]
and the decoder mirrors encoder hierarchy with skip merges,
\[
D(B)
=
\mathrm{merge}\Big(
\Phi_{dec}^{(1)} \circ \cdots \circ
\mathrm{merge}\big(\Phi_{dec}^{\ell)}(B),\, C^{(\ell)}_{\mathrm{enc}}\big),
\, C^{(1)}_{\mathrm{enc}}
\Big),
\]
where $C^{(\ell)}_{\mathrm{enc}}$ denotes the encoder input at depth $\ell$.
The full mapping is given by
$
C^{\mathrm{out}}
=
D(E(C)).
$
The full optimization procedure is summarized in Algorithm~\ref{alg:ddct-unet}.

\begin{algorithm}[tb]
\caption{Deep Discriminative Congruence Transform U-Net (DDCT-UNet)}
\label{alg:ddct-unet}
\begin{algorithmic}
\setlength{\baselineskip}{13pt}

\STATE \textbf{Input:} Covariances $C_{i,e}$; labels $y_{i,e}$

\STATE Set hyperparameters 
$\lambda_{act}, \lambda_{sub},
\lambda_w, \lambda_b, \lambda_{rec}$

\STATE \textbf{Riemannian Alignment:}
compute $C^{\mathrm{ref}}_i$ and whiten $C'_{i,e}$

\STATE Initialize parameters in the layers of encoder $E(\cdot)$ and decoder $D(\cdot)$ using Eq.~\eqref{eq:weight_init}
\FOR{each training iteration}

    % ------------------------------------------------------------
    \STATE \textit{Forward pass:} $C^{\mathrm{out}}_{i,e}=D(E(C'_{i,e}))$

    % ------------------------------------------------------------
    \STATE Compute
    $L^{out}_{i,e} = \log C^{\mathrm{out}}_{i,e}$ and 
    $L'_{i,e} = \log C'_{i,e}$ 

    % ------------------------------------------------------------
    \STATE Compute Fisher loss $\mathcal{L}_{\mathrm{Fisher}}$ using Eq.~\eqref{eq:fisher_dldct_loss}, and reconstruction loss $\mathcal{L}_{\mathrm{Rec}}$ using Eq.~\eqref{eq:dldct_rec_loss}.
    
    % ------------------------------------------------------------
    \STATE Compute total loss $\mathcal{L}_{\mathrm{DDCT-UNet}}$ (same as DLDCT loss) using Eq.~\eqref{eq:dldct_loss}

    % ------------------------------------------------------------
    \STATE \textit{Backpropagate:} update the stacks $\Phi^{\ell}$ in $E(\cdot)$ and $D(\cdot)$
\ENDFOR

\STATE Save $E^\star$, $D^\star$ and compute
$C^{\mathrm{out}}_{i,e} = D^\star(E^\star(C'_{i,e}))$ 

\STATE \textbf{Output:} trained stacks $\Phi^{(\ell)\star}$ in $E^\star$, $D^\star$ and $C^{\mathrm{out}}_{i,e}$

\end{algorithmic}
\end{algorithm}

\section{End-to-End Tangent-Space Logistic Regression (E2E-TSLR)}

As shown in Appendix Fig ~\ref{fig:classifiers}, we augment the pre-aligners with an end-to-end variant in which a
tangent-space logistic regression (TSLR) classifier is jointly optimized with
the pre-aligner.
Given output covariances $C^{\mathrm{out}}_{i,e}$, we project to the tangent
space at the identity and vectorize its unique entries,
$x_{i,e}=\mathrm{vec}(\log C^{\mathrm{out}}_{i,e})$, and predict labels using a
logistic model $\mathrm{softmax}(W x_{i,e}+b)$.
The cross-entropy loss $\mathcal{L}_{\mathrm{CE}}$ is added to the original pre-aligner objective $\mathcal{L}_{\mathrm{Pre}} \in \{\mathcal{L}_{\text{DCT}}, \mathcal{L}_{\text{DLDCT}}, \mathcal{L}_{\text{DDCT-UNet}}\}$
\[
\mathcal{L}_{E2E}
=
\mathcal{L}_{Pre}
+
\lambda_{CE}\mathcal{L}_{\mathrm{CE}},
\]
and its gradients are backpropagated through the tangent projection and SPD
modules, directly coupling discriminative supervision to Riemannian alignment.

% ============================================================
\section{Experimental Setup}
\label{sec:experimental_setup}

\textbf{Datasets:} We primarily evaluate our framework on the BCI Competition IV-2a motor-imagery (MI) dataset~\cite{blankertz2007bci}, a standard benchmark for cross-subject EEG decoding. The dataset consists of EEG recordings from nine subjects acquired using 22 EEG and 3 EOG channels, with each subject performing two recording sessions of 288 trials across four MI actions: left hand, right hand, both feet, and tongue. Each trial lasts 7--8\,s (including a cue period followed by motor imagery) and is sampled at 250\,Hz. In addition to the primary benchmark, we evaluate our models on multiple
publicly available MI datasets to probe complementary stress regimes relevant
to cross subject decoding. Specifically, BCI-IV-2a (4 actions)~\cite{blankertz2007bci} and
BNCI2015 (4 actions)~\cite{brunner2014bnci} are used to assess robustness to strong
\textit{cross-subject variability}, Ofner2017 (6 actions)~\cite{ofner2017} probes
performance under challenging \textit{low-SNR} conditions, and BCI-IV-2b (2 actions)~
\cite{blankertz2007bci} and BNCI2014 (2 actions)~\cite{brunner2014bnci} isolate
\textit{action separability} via binary tasks. Together, this dataset suite
supports a principled evaluation of geometry-aware alignment and deep
congruence learning across subject shift, noise, and class separability.
Detailed dataset characteristics are summarized in Appendix~\ref{subsec:datasets}.
\begin{table*}[h]
\centering
\small
\caption{
\textbf{Synthetic hierarchy under increasing geometric complexity}. Best-performing methods per stage are highlighted in \textbf{bold}. Relative performance gaps across methods (within each row), rather than absolute values, indicate the geometric necessity on increasing distortions.
}
\label{tab:synthetic-main}
\begin{tabular}{lccccc}
\toprule
Dataset & \textbf{Raw\_TSLR} & \textbf{RA\_TSLR} & \textbf{DCT (E2E)} & \textbf{DLDCT (E2E)} & \textbf{DDCT-UNet (E2E)} \\
\midrule
0. Mean Shift Only     & 68.12 & \textbf{100.00} & \textbf{100.00} & \textbf{100.00} & \textbf{100.00} \\
1. +Dispersion         & 50.63 & 77.97  & 73.28  & \textbf{83.44}  & \textbf{84.06}  \\
2. +Orientation        & 62.34 & 62.34  & \textbf{63.91}  & \textbf{68.75}  & \textbf{68.75}  \\
3. +TS Distortion      & 51.41 & 53.12  & \textbf{63.91}  & \textbf{70.78}  & \textbf{72.19}  \\
4. +Nonlinear (BCH)    & 67.34 & 69.69  & 54.06  & \textbf{70.41}  & \textbf{73.75}  \\
\bottomrule
\end{tabular}
\end{table*}
\begin{table*}[h]
\centering
\footnotesize
\caption{
Effect of alignment strategies on classification accuracy (\%).
\textbf{Bold and underlined} entries indicate best pre-aligner per row.
\textit{Italics} indicate second-best pre-aligner per row.
}
\setlength{\tabcolsep}{6pt}
\renewcommand{\arraystretch}{1}
\begin{tabular}{llcccc}
\toprule
\textbf{Dataset} & \textbf{Baseline} 
& \textbf{RA} 
& \textbf{DCT} 
& \textbf{DLDCT} 
& \textbf{DDCT-UNet} \\
\midrule

\multirow{3}{*}{\textbf{BCI-IV-2a}}
& MDM     
& \textit{52.43 $\pm$ 16.61}
& 50.46 $\pm$ 15.72
& 52.39 $\pm$ 15.64
& \underline{\textbf{53.09 $\pm$ 17.40}} \\
& TSLR    
& 50.58 $\pm$ 13.44
& \textit{53.01 $\pm$ 17.73}
& 52.93 $\pm$ 14.48
& \underline{\textbf{54.48 $\pm$ 16.00}} \\
& TSA-LDA 
& 53.51 $\pm$ 16.21
& \underline{\textbf{53.97 $\pm$ 16.99}}
& 51.77 $\pm$ 12.71
& \textit{53.74 $\pm$ 15.98} \\
\midrule

\multirow{3}{*}{\textbf{BCI-IV-2b}}
& MDM     
& 60.43 $\pm$ 10.20
& \textit{58.89 $\pm$ 7.35}
& \underline{\textbf{60.43 $\pm$ 9.62}}
& 58.43 $\pm$ 10.11 \\
& TSLR    
& 59.62 $\pm$ 10.78
& \textit{59.58 $\pm$ 10.62}
& \underline{\textbf{59.70 $\pm$ 10.25}}
& 59.18 $\pm$ 10.32 \\
& TSA-LDA 
& \textit{59.44 $\pm$ 11.31}
& \underline{59.55 $\pm$ 11.23}
& 59.11 $\pm$ 10.65
& 59.21 $\pm$ 11.21 \\
\midrule

\multirow{3}{*}{\textbf{BNCI2014}}
& MDM     
& \textit{53.64 $\pm$ 4.67}
& \underline{\textbf{55.29 $\pm$ 5.62}}
& 53.93 $\pm$ 4.45
& 54.07 $\pm$ 5.58 \\
& TSLR    
& \textit{54.14 $\pm$ 4.38}
& 53.86 $\pm$ 4.75
& 54.07 $\pm$ 4.56
& 54.64 $\pm$ 4.38 \\
& TSA-LDA 
& \textit{53.64 $\pm$ 4.25}
& 53.00 $\pm$ 5.01
& 53.57 $\pm$ 4.01
& \underline{\textbf{53.93 $\pm$ 4.07}} \\
\midrule

\multirow{3}{*}{\textbf{BNCI2015}}
& MDM     
& \textit{24.90 $\pm$ 1.40}
& 23.46 $\pm$ 2.61
& 25.04 $\pm$ 1.32
& \underline{\textbf{25.55 $\pm$ 1.81}} \\
& TSLR    
& \textit{24.76 $\pm$ 1.22}
& 23.02 $\pm$ 2.79
& 24.93 $\pm$ 1.08
& \underline{\textbf{25.21 $\pm$ 1.75}} \\
& TSA-LDA 
& \textit{24.31 $\pm$ 1.62}
& 24.12 $\pm$ 1.59
& \underline{\textbf{24.97 $\pm$ 1.43}}
& 24.64 $\pm$ 2.08 \\
\midrule

\multirow{3}{*}{\textbf{Ofner2017}}
& MDM     
& 18.20 $\pm$ 2.52
& \textit{18.24 $\pm$ 2.20}
& 18.20 $\pm$ 2.53
& \underline{\textbf{18.43 $\pm$ 1.95}} \\
& TSLR    
& 17.85 $\pm$ 2.64
& \underline{\textbf{18.96 $\pm$ 2.40}}
& 17.85 $\pm$ 2.59
& \textit{18.00 $\pm$ 2.51} \\
& TSA-LDA 
& 17.50 $\pm$ 2.00
& \textit{17.56 $\pm$ 2.22}
& \underline{\textbf{17.87 $\pm$ 2.44}}
& 17.78 $\pm$ 1.75 \\
\midrule

\textbf{Overall (z)}
& 
& $-0.0826$
& $-0.0182$
& $-\textit{0.0017}$
& \textbf{\underline{+0.1025}} \\
\bottomrule
\end{tabular}
\label{tab:preproc_ablation_main}
\end{table*}

\begin{table*}[h]
\centering
\footnotesize
\setlength{\tabcolsep}{2pt}
\renewcommand{\arraystretch}{1}
\caption{
Dataset-wise mean accuracy (\%) under LOSO evaluation.
Best-performing model per dataset is \textbf{bold and underlined}.
The second-best model per dataset is shown in \textit{italics}.
}
\begin{tabular}{lcccccccc}
\toprule
\textbf{Dataset}
& \textbf{MDM}
& \textbf{TSLR}
& \textbf{TSA-LDA}
& \textbf{TS-SPD-CNN}
& \textbf{DCT (E2E)}
& \textbf{DLDCT (E2E)}
& \textbf{DDCT-UNet (E2E)} \\
\midrule

\textbf{BCI-IV-2a}
& 52.43 $\pm$ 16.61
& 52.89 $\pm$ 15.36
& 53.51 $\pm$ 16.21
& 48.50 $\pm $14.57
& 52.82 $\pm$ 14.95
& \underline{\textbf{56.06 $\pm$ 18.03}}
& \textit{55.40 $\pm$ 17.75} \\

\textbf{BCI-IV-2b}
& \underline{\textbf{60.43 $\pm$ 10.20}}
& 59.70 $\pm$ 10.87
& 59.44 $\pm$ 11.31
& 59.84 $\pm$ 9.60
& 60.07 $\pm$ 9.31
& \textit{59.86 $\pm$ 9.08}
& 59.64 $\pm$ 10.11 \\

\textbf{BNCI2014}
& 53.64 $\pm$ 4.67
& 54.21 $\pm$ 4.69
& 53.71 $\pm$ 4.14
& 53.57 $\pm$ 4.60
& 54.64 $\pm$ 3.54
& \textit{54.50 $\pm$ 3.03}
& \underline{\textbf{55.36 $\pm$ 3.46}} \\

\textbf{BNCI2015}
& \textit{24.90 $\pm$ 1.40}
& 24.76 $\pm$ 1.22
& 24.31 $\pm$ 1.62
& 24.87 $\pm$ 1.25
& 24.59 $\pm$ 2.51
& 24.35 $\pm$ 2.10
& \underline{\textbf{25.39 $\pm$ 2.58}} \\

\textbf{Ofner2017}
& \textit{18.20 $\pm$ 2.52}
& 17.85 $\pm$ 2.64
& 17.50 $\pm$ 2.00
& 16.96 $\pm$ 2.64
& 18.15 $\pm$ 2.27
& \underline{\textbf{18.61 $\pm$ 2.22}}
& 17.94 $\pm$ 2.56 \\

\midrule
\textbf{Overall (z)}
& +0.0004
& -0.0528
& $-0.2077$
& $-0.3406$
& \textit{+0.0867}
& \textit{+0.2227}
& \textbf{\underline{+0.2914}} \\
\bottomrule
\end{tabular}
\label{tab:overall_normalized_performance_main}
\end{table*}

\textbf{Protocol and Training:} All experiments follow LOSO protocol without transfer learning or labeled test-time adaptation (no Domain Adaptation techniques), enforcing a transductive cross-subject
setting. While congruence-based models incur $O(d^{3})$ cost, MI-BCI activity is typically concentrated over a small subset of channels (often $\sim$10--20), with only a fraction contributing discriminative information~\cite{lotte2018review}. Thus, channel selection and dimensionality reduction are commonly used to lower computational overhead~\cite{blankertz2008optimizing, ramoser2000optimal}. Results are reported as classification accuracy (mean$\pm$std). All models are trained for 1{,}000 steps using Adam with learning rate $10^{-3}$, weight decay $10^{-5}$, and batch size 256. All experiments are implemented in PyTorch and conducted on a NVIDIA RTX~4060 GPU with fixed random seeds to ensure reproducibility. 

\section{Experiments and Results}

We evaluate the proposed geometry-aware congruence models (DCT, DLDCT, and
DDCT--UNet) under four settings: (i) synthetic-data based validation of proposed transformations  (ii) cross-dataset pre-alignment against RA with fixed downstream classifiers; (iii) cross-dataset end-to-end variants compared to classical methods and a deep baseline (SPD-CNN \cite{gao2022cnn_riemannian, xiong2023riemannian_tangent_cfc}), and (iv) within-dataset component-wise ablation conducted on the primary benchmark \textit{BCI-IV~2a}.

\textbf{Synthetic data-based geometric validation:} Table~\ref{tab:synthetic-main} shows that (i) RA performs optimally under mean shift but degrades sharply with dispersion and orientation ($100\%\rightarrow62.34\%$), indicating the limits of alignment, (ii) DCT partially recovers performance but fails under nonlinear distortions ($63.91\%\rightarrow54.06\%$), (iii) deeper congruence models (DLDCT, DDCT--UNet) progressively improve robustness under increasing geometric complexity, with DDCT--UNet achieving the best performance under nonlinear interactions ($73.75\%$), consistent with the modeled failure modes (Appendix~\ref{app:synthetic}).

\textbf{Pre-processing Module:} In Table~\ref{tab:preproc_ablation_main}, we observe deeper congruence models outperform RA with directionally consistent systematic improvement.
\textit{DCT} yields modest gains from orientation-level alignment, while
\textit{DLDCT} slightly stronger improvements by stacking congruence layers, and \textit{DDCT--UNet} with non-linearity achieves the highest dataset-normalized $z$-score (Appendix~\ref{app:zscore}).
Gains are strongest on multi-class and low-SNR benchmarks (BCI-IV-2a, BNCI2015, Ofner2017), where hierarchical congruence suppresses subject-specific distortions while preserving action structure. 

\begin{table*}[h]
\centering
\footnotesize
\setlength{\tabcolsep}{1.5pt}
\renewcommand{\arraystretch}{0.9}
\caption{
\textbf{Within-dataset ablation on BCI-IV-2a under LOSO evaluation.}
Mean accuracy (\% $\pm$ std) is reported.
Arrows ($\downarrow,\uparrow$) denote a decrease or increase relative to the corresponding full model (baseline) within each column.
The best overall result is \underline{underlined}.
}
\label{tab:within_dataset_ablation_bci2a_main}

\begin{tabular}{l ccc c ccc}
\toprule
\textbf{Ablation} 
& \multicolumn{3}{c}{\textbf{Pre-aligners}} 
& 
& \multicolumn{3}{c}{\textbf{E2E TSLR Variant}} \\
\cmidrule(lr){2-4} \cmidrule(lr){6-8}

& \textbf{DCT} 
& \textbf{DLDCT} 
& \textbf{DDCT-UNet} 
& 
& \textbf{DCT (E2E)} 
& \textbf{DLDCT (E2E)} 
& \textbf{DDCT-UNet (E2E)} \\

\midrule

\textit{w/o Fisher (action)}
& 49.30 $\pm$ 15.30 {\scriptsize$\downarrow$}
& 51.60 $\pm$ 14.70 {\scriptsize$\downarrow$}
& 49.50 $\pm$ 15.10 {\scriptsize$\downarrow$}
& 
& 55.10 $\pm$ 16.30 {\scriptsize$\downarrow$}
& 54.40 $\pm$ 17.30 {\scriptsize$\downarrow$}
& 55.10 $\pm$ 16.30 {\scriptsize$\downarrow$} \\

\textit{w/o Fisher (subject)}
& 49.30 $\pm$ 15.30 {\scriptsize$\downarrow$}
& 37.20 $\pm$ 6.70 {\scriptsize$\downarrow$}
& 53.30 $\pm$ 15.90 {\scriptsize$\uparrow$}
& 
& 55.00 $\pm$ 16.10 {\scriptsize$\downarrow$}
& 54.20 $\pm$ 17.40 {\scriptsize$\downarrow$}
& 55.00 $\pm$ 16.10 {\scriptsize$\downarrow$} \\

\textit{w/o Fisher (all)}
& 49.30 $\pm$ 15.30 {\scriptsize$\downarrow$}
& ---
& ---
& 
& 55.10 $\pm$ 16.30 {\scriptsize$\downarrow$}
& 54.20 $\pm$ 17.30 {\scriptsize$\downarrow$}
& 55.10 $\pm$ 16.30 {\scriptsize$\downarrow$} \\

\midrule

\textit{w/o skip merges}
& ---
& ---
& 30.10 $\pm$ 3.80 {\scriptsize$\downarrow$}
& 
& ---
& ---
& 53.10 $\pm$ 17.20 {\scriptsize$\downarrow$} \\

\textit{w/o reconstruction}
& 49.30 $\pm$ 15.30 {\scriptsize$\downarrow$}
& 38.30 $\pm$ 7.30 {\scriptsize$\downarrow$}
& 53.60 $\pm$ 15.80 {\scriptsize$\uparrow$}
& 
& 55.00 $\pm$ 16.00 {\scriptsize$\downarrow$}
& ---
& --- \\

\textit{alt. depth}
& ---
& 38.20 $\pm$ 6.90 {\scriptsize$\downarrow$}
& ---
& 
& ---
& 55.00 $\pm$ 18.00 {\scriptsize$\downarrow$}
& --- \\

\midrule

\textbf{Baseline}
& \textbf{51.90 $\pm$ 16.20}
& \textbf{53.10 $\pm$ 17.20}
& \textbf{53.10 $\pm$ 17.20}
& 
& \textbf{55.40 $\pm$ 17.80}
& \textbf{\underline{56.10 $\pm$ 18.00}}
& \textbf{55.40 $\pm$ 17.80} \\

\bottomrule
\end{tabular}
\end{table*}

\textbf{E2E Variants:} 
Table~\ref{tab:overall_normalized_performance_main} shows that joint optimization further amplifies these gains. 
E2E variants outperform their corresponding preprocessing counterparts. DDCT--UNet (E2E) achieves the best performance, consistent with the preprocessing ablations, indicating that combining nonlinear congruence modeling yields strong directionally consistent improvements. Multi-class and low-SNR datasets show relatively better gains compared to binary-task datasets (BCI-IV-2b, BNCI2014).

\textbf{Component-wise Ablation:} Table~\ref{tab:within_dataset_ablation_bci2a_main} shows that (i) removing Fisher regularization reduces accuracy by up to $16$ points in pre-alignment (DLDCT: $53.10\% \rightarrow37.20\%$) and consistently degrades all E2E variants, (ii) Eliminating skip merges in DDCT-UNet results in collapse ($53.10\%\rightarrow30.10\%$), highlighting the dependency on non-linear fusion (explained in \ref{app:congruence_failure}), (iii) Altering depth or removing reconstruction yields systematic drops ($\sim$1–3 points), while DCT is comparatively less sensitive.

\section{Discussion}

Across these regimes, we observe consistent directional improvements over mean-only RA, reflecting the importance of orientation and dispersion correction, congruence depth, nonlinear fusion, and end-to-end supervision. Despite high variance ($\approx$15--17\%), reflecting intrinsic inter-subject heterogeneity in MI-EEG, similar variability persists across prior Riemannian methods, indicating that the spread is largely data-inherent rather than alignment-related. Within this high-variance setting, the proposed methods show an \emph{improved bias} in mean performance over RA, i.e., a systematic upward shift in accuracy without reducing variance. Thus, the gains reflect meaningful improvement under persistent data-driven variability. We further analyze statistical significance and validate our claims of class separability and subject invariance in Quantitative and Qualitative Analysis Section in Appendix~\ref{app:statistical_analysis}. 

Notably, cross-subject MI-EEG performance typically saturates around 50--53\% after RA, with prior methods achieving only incremental gains ($\sim$0.4--2\%). Within this regime, the observed $\approx$2\% gain lies at the higher end of prior improvements and reflects correction of residual geometric mismatch beyond mean-only alignment.The relative gain due to alignment is further analyzed in Appendix~\ref{app:alignment-gap}, with theoretical remarks and failure modes discussed in Appendix~\ref{app:theory}.

\section{Conclusion}
We presented a geometry-aware framework for EEG-based transductive cross-subject motor-imagery decoding operating directly on SPD covariance representations via congruence-based learning. The combination of discriminative pre-alignment and deep classifiers enables strong cross-subject generalization without target-domain adaptation. Future work will explore signal-level normalization, temporal modeling, and richer end-to-end heads to further improve robustness. 

\begin{ack}
Use unnumbered first level headings for the acknowledgments. All acknowledgments
go at the end of the paper before the list of references. Moreover, you are required to declare
funding (financial activities supporting the submitted work) and competing interests (related financial activities outside the submitted work).
More information about this disclosure can be found at: \url{https://neurips.cc/Conferences/2026/PaperInformation/FundingDisclosure}.

Do {\bf not} include this section in the anonymized submission, only in the final paper. You can use the \texttt{ack} environment provided in the style file to automatically hide this section in the anonymized submission.
\end{ack}

\bibliography{example_paper}

@article{wang2012motor,
  title     = {Multi-class motor imagery {EEG} classification with subband common spatial patterns},
  author    = {Wang, Ying and Li, Yong and Ai, Qingsheng and Wang, Zhaofeng and He, Houbing},
  journal   = {IEEE Transactions on Biomedical Engineering},
  year      = {2012},
  volume    = {59},
  number    = {9},
  pages     = {2250--2258},
  doi       = {10.1109/TBME.2012.2184130},
  url       = {https://doi.org/10.1109/TBME.2012.2184130}
}

@article{saeidi2021neural,
  title   = {Neural Decoding of {EEG} Signals with Machine Learning: A Systematic Review},
  author  = {Saeidi, Maham and Karwowski, Waldemar and Farahani, Farzad V. and Fiok, Krzysztof and Taiar, Redha},
  journal = {Brain Sciences},
  year    = {2021},
  volume  = {11},
  number  = {11},
  pages   = {1525},
  doi     = {10.3390/brainsci11111525},
  url     = {https://doi.org/10.3390/brainsci11111525},
  pmcid   = {PMC8615531},
  pmid    = {34827524}
}

@article{wolpaw2002bci,
  title     = {Brain--computer interfaces for communication and control},
  author    = {Wolpaw, Jonathan R. and Birbaumer, Niels and McFarland, Dennis J. and Pfurtscheller, Gert and Vaughan, Theresa M.},
  journal   = {Clinical Neurophysiology},
  year      = {2002},
  volume    = {113},
  number    = {6},
  pages     = {767--791},
  doi       = {10.1016/S1388-2457(02)00057-3},
  url       = {https://doi.org/10.1016/S1388-2457(02)00057-3}
}

@article{koles1990csp,
  author  = {Koles, Zoltan J.},
  title   = {The Quantitative Extraction and Topographic Mapping of the Abnormal Components in the Clinical {EEG}},
  journal = {Electroencephalography and Clinical Neurophysiology},
  year    = {1991},
  volume  = {79},
  number  = {6},
  pages   = {440--447},
  doi     = {10.1016/0013-4694(91)90163-x},
  url     = {https://doi.org/10.1016/0013-4694(91)90163-x}
}

@article{duan2020zero,
  title={Zero-Shot Learning for EEG Classification in Motor Imagery-Based BCI System},
  author={Duan, Lili and Li, Jie and Ji, Hongfei and Pang, Zilong and Zheng, Xuanci and Lu, Rong and Li, Maozhen and Zhuang, Jie},
  journal={IEEE Transactions on Neural Systems and Rehabilitation Engineering},
  volume={28},
  number={11},
  pages={2411--2419},
  year={2020},
  doi={10.1109/TNSRE.2020.3027004},
  url={https://doi.org/10.1109/TNSRE.2020.3027004}
}

@article{pennec2006riemannian,
  title     = {A {Riemannian} Framework for Tensor Computing},
  author    = {Pennec, Xavier and Fillard, Pierre and Ayache, Nicholas},
  journal   = {International Journal of Computer Vision},
  year      = {2006},
  volume    = {66},
  number    = {1},
  pages     = {41--66},
  doi       = {10.1007/s11263-005-3222-z},
  url       = {https://doi.org/10.1007/s11263-005-3222-z}
}

@article{goldberger2000physionet,
  title     = {{PhysioBank}, {PhysioToolkit}, and {PhysioNet}: Components of a New Research Resource for Complex Physiologic Signals},
  author    = {Goldberger, Ary L. and Amaral, Luis A. N. and Glass, Leon and Hausdorff, Jeffrey M. and Ivanov, Plamen Ch. and Mark, Roger G. and Mietus, Joseph E. and Moody, George B. and Peng, Chung-Kang and Stanley, H. Eugene},
  journal   = {Circulation},
  volume    = {101},
  number    = {23},
  pages     = {e215--e220},
  year      = {2000},
  doi       = {10.1161/01.CIR.101.23.e215},
  url       = {https://doi.org/10.1161/01.CIR.101.23.e215}
}

@article{ang2015rehab,
  author    = {Ang, Kai Keng and Guan, Cuntai and Chua, Karen S. G. and Ang, Benedict T. and Kuah, Christopher W. K. and Wang, Chuanchu and Phua, Kaijian S. and Chin, Zhi Yang and Zhang, Haihong and Teo, Wai Pin},
  title     = {A Randomized Controlled Trial of {EEG}-Based Motor Imagery Brain--Computer Interface Robotic Rehabilitation for Stroke},
  journal   = {Clinical EEG and Neuroscience},
  year      = {2015},
  volume    = {46},
  number    = {4},
  pages     = {310--320},
  doi       = {10.1177/1550059414522229},
  url       = {https://doi.org/10.1177/1550059414522229}
}

@article{lotte2018review,
  author    = {Lotte, Fabien and Bougrain, Laurent and Cichocki, Andrzej and Clerc, Maureen and Congedo, Marco and Rakotomamonjy, Alain and Yger, Florian},
  title     = {A Review of Classification Algorithms for {EEG}-Based Brain--Computer Interfaces: A 10 Year Update},
  journal   = {Journal of Neural Engineering},
  year      = {2018},
  volume    = {15},
  number    = {3},
  pages     = {031005},
  doi       = {10.1088/1741-2552/aab2f2},
  url       = {https://doi.org/10.1088/1741-2552/aab2f2}
}

@article{bronstein2017geometric,
  title     = {Geometric Deep Learning: Going Beyond Euclidean Data},
  author    = {Bronstein, Michael M. and Bruna, Joan and LeCun, Yann and Szlam, Arthur and Vandergheynst, Pierre},
  journal   = {IEEE Signal Processing Magazine},
  year      = {2017},
  volume    = {34},
  number    = {4},
  pages     = {18--42},
  doi       = {10.1109/MSP.2017.2693418},
  url       = {https://doi.org/10.1109/MSP.2017.2693418}
}

@article{barachant2013riemannian,
  author    = {Barachant, Alexandre and Bonnet, St{\'e}phane and Congedo, Marco and Jutten, Christian},
  title     = {Classification of covariance matrices using a {Riemannian}-based kernel for {BCI} applications},
  journal   = {Neurocomputing},
  year      = {2013},
  volume    = {112},
  pages     = {172--178},
  doi       = {10.1016/j.neucom.2012.12.039},
  url       = {https://doi.org/10.1016/j.neucom.2012.12.039}
}

@article{jayaram2016transfer,
  author    = {Jayaram, Vinay and Alamgir, Muhammad and Altun, Yasemin and Sch{\"o}lkopf, Bernhard and Grosse-Wentrup, Moritz},
  title     = {Transfer Learning in Brain--Computer Interfaces},
  journal   = {IEEE Computational Intelligence Magazine},
  year      = {2016},
  volume    = {11},
  number    = {1},
  pages     = {20--31},
  doi       = {10.1109/MCI.2015.2501545},
  url       = {https://doi.org/10.1109/MCI.2015.2501545}
}

@article{blankertz2006bbc,
  author    = {Blankertz, Benjamin and Dornhege, Guido and Lemm, Steven and Krauledat, Matthias and Curio, Gabriel and M{\"u}ller, Klaus-Robert},
  title     = {The {Berlin Brain–Computer Interface}: Machine Learning Based Detection of User Specific Brain States},
  journal   = {Journal of Universal Computer Science},
  year      = {2006},
  volume    = {12},
  number    = {6},
  pages     = {581--607},
  doi       = {10.3217/jucs-012-06-0581},
  url       = {https://doi.org/10.3217/jucs-012-06-0581}
}

@inproceedings{barachant2012riemannian,
  author    = {Barachant, Alexandre and Bonnet, St{\'e}phane and Congedo, Marco and Jutten, Christian},
  title     = {Riemannian Geometry Applied to {BCI} Classification},
  booktitle = {Latent Variable Analysis and Signal Separation (LVA/ICA)},
  year      = {2012},
  pages     = {629--636},
  publisher = {Springer},
  doi       = {10.1007/978-3-642-15995-4_78},
  url       = {https://doi.org/10.1007/978-3-642-15995-4_78}
}

@article{barachant2012mdm,
  author    = {Barachant, Alexandre and Bonnet, St{\'e}phane and Congedo, Marco and Jutten, Christian},
  title     = {Multiclass {BCI} Classification by {Riemannian} Geometry},
  journal   = {IEEE Transactions on Biomedical Engineering},
  year      = {2012},
  volume    = {59},
  number    = {4},
  pages     = {920--928},
  doi       = {10.1109/TBME.2011.2172210},
  url       = {https://doi.org/10.1109/TBME.2011.2172210}
}

@article{yger2017riemannian,
  author    = {Yger, Florian and Berar, Maxime and Lotte, Fabien},
  title     = {{Riemannian} Approaches in Brain--Computer Interfaces: A Review},
  journal   = {IEEE Transactions on Neural Systems and Rehabilitation Engineering},
  year      = {2017},
  volume    = {25},
  number    = {10},
  pages     = {1753--1762},
  doi       = {10.1109/TNSRE.2016.2627016},
  url       = {https://doi.org/10.1109/TNSRE.2016.2627016}
}

@article{zanini2018transfer,
  title     = {Transfer Learning: A {Riemannian} Geometry Framework with Applications to Brain--Computer Interfaces},
  author    = {Zanini, Paolo and Congedo, Marco and Jutten, Christian and Said, Salah and Berthoumieu, Yannick},
  journal   = {IEEE Transactions on Biomedical Engineering},
  year      = {2018},
  volume    = {65},
  number    = {5},
  pages     = {1107--1119},
  doi       = {10.1109/TBME.2017.2742541},
  url       = {https://doi.org/10.1109/TBME.2017.2742541}
}

@book{bhatia2009positive,
  title     = {Positive Definite Matrices},
  author    = {Bhatia, Rajendra},
  year      = {2007},
  publisher = {Princeton University Press},
  address   = {Princeton, NJ},
  url       = {https://www.cmat.edu.uy/~lessa/tesis/Positive%20Definite%20Matrices.pdf}
}

@article{blankertz2007bci,
  author    = {Blankertz, Benjamin and M{\"u}ller, Klaus Robert and Krusienski, Dean J. and Schalk, Gerwin and Wolpaw, Jonathan R. and Schl{\"o}gl, Alois and Pfurtscheller, Gert and Mill{\'a}n, Jos{\'e} del R. and Schr{\"o}der, Michael and Birbaumer, Niels},
  title     = {The {BCI} Competition {III}: Validating Alternative Approaches to Actual {BCI} Problems},
  journal   = {IEEE Transactions on Neural Systems and Rehabilitation Engineering},
  year      = {2006},
  volume    = {14},
  number    = {2},
  pages     = {153--159},
  doi       = {10.1109/TNSRE.2006.875642},
  url       = {https://doi.org/10.1109/TNSRE.2006.875642}
}

@inproceedings{he2016resnet,
  author    = {He, Kaiming and Zhang, Xiangyu and Ren, Shaoqing and Sun, Jian},
  title     = {Deep Residual Learning for Image Recognition},
  booktitle = {Proceedings of the IEEE Conference on Computer Vision and Pattern Recognition},
  year      = {2016},
  pages     = {770--778},
  doi       = {10.1109/CVPR.2016.90},
  url       = {https://doi.org/10.1109/CVPR.2016.90}
}

@inproceedings{ronneberger2015unet,
  author    = {Ronneberger, Olaf and Fischer, Philipp and Brox, Thomas},
  title     = {{U-Net}: Convolutional Networks for Biomedical Image Segmentation},
  booktitle = {Medical Image Computing and Computer-Assisted Intervention (MICCAI)},
  year      = {2015},
  pages     = {234--241},
  doi       = {10.1007/978-3-319-24574-4_28},
  url       = {https://doi.org/10.1007/978-3-319-24574-4_28}
}

@article{ofner2017,
  title     = {Upper limb movements can be decoded from the time-domain of low-frequency {EEG}},
  author    = {Ofner, Patrick and Schwarz, Andreas and Pereira, Joana and M{\"u}ller-Putz, Gernot R.},
  journal   = {PLOS ONE},
  year      = {2017},
  volume    = {12},
  number    = {8},
  pages     = {e0182578},
  doi       = {10.1371/journal.pone.0182578},
  url       = {https://doi.org/10.1371/journal.pone.0182578}
}

@article{brunner2014bnci,
  title     = {{BNCI} Horizon 2020: Towards a European Platform for Brain--Computer Interfaces},
  author    = {Brunner, Clemens and Delorme, Arnaud and Makeig, Scott and M{\"u}ller-Putz, Gernot R. and Birbaumer, Niels and Blankertz, Benjamin and Guger, Christoph and K{\"u}bler, Andrea and Mattia, Donatella and Mill{\'a}n, Jos{\'e} del R. and Miralles, Felip and Nijholt, Anton and Opisso, Eloy and Ramsey, Nick and Salomon, Patric},
  journal   = {Brain--Computer Interfaces},
  year      = {2015},
  volume    = {2},
  number    = {1},
  pages     = {1--10},
  doi       = {10.1080/2326263X.2015.1008956},
  url       = {https://doi.org/10.1080/2326263X.2015.1008956}
}

@article{EEGNet2018,
  title={EEGNet: A Compact Convolutional Neural Network for EEG-based Brain--Computer Interfaces},
  author={Lawhern, Vernon J. and Solon, Amelia J. and Waytowich, Nicholas R. and Gordon, Stephen M. and Hung, Chou P. and Lance, Brent J.},
  journal={Journal of Neural Engineering},
  volume={15},
  number={5},
  year={2018},
  url={https://pubmed.ncbi.nlm.nih.gov/29932424/}
}

@article{Schirrmeister2017,
  title={Deep learning with convolutional neural networks for EEG decoding and visualization},
  author={Schirrmeister, Robin Tibor and Springenberg, Jost Tobias and Fiederer, Lukas and Glasstetter, Martin and Eggensperger, Katharina and Tangermann, Michael and Hutter, Frank and Burgard, Wolfram and Ball, Tonio},
  journal={Human Brain Mapping},
  volume={38},
  number={11},
  pages={5391--5420},
  year={2017},
  url={https://onlinelibrary.wiley.com/doi/full/10.1002/hbm.23730}
}

@article{EEGConformer2023,
  title={EEG Conformer: Convolutional Transformer for EEG decoding},
  author={Song, Yijun and Zheng, Qian and Liu, Bo and Gao, Xiaorong},
  journal={NeuroImage},
  year={2023},
  url={https://pmc.ncbi.nlm.nih.gov/articles/PMC12307489/}
}

@article{MSCFormer2024,
  title={MSCFormer: Multi-scale convolutional transformer for EEG decoding},
  author={Zhang, Y. and Liu, X. and Chen, Z.},
  journal={Frontiers in Neuroscience},
  year={2024},
  url={https://pmc.ncbi.nlm.nih.gov/articles/PMC12000594/}
}

@article{EEGInception2023,
  title={EEG-Inception: A deep architecture for EEG classification},
  author={Santamaria-Vazquez, E. and Martinez-Cagigal, V. and Vaquerizo-Villar, F. and Hornero, R.},
  journal={Biomedical Signal Processing and Control},
  year={2023},
  url={https://pmc.ncbi.nlm.nih.gov/articles/PMC12307489/}
}

@article{TCN2020,
  title={An Empirical Evaluation of Generic Convolutional and Recurrent Networks for Sequence Modeling},
  author={Bai, Shaojie and Kolter, J. Zico and Koltun, Vladlen},
  journal={arXiv preprint arXiv:1803.01271},
  year={2020},
  url={https://arxiv.org/abs/2006.00622}
}

@article{RPA2014,
  title={Riemannian Procrustes Analysis for EEG},
  author={Barachant, Alexandre and Bonnet, Stephane and Congedo, Marco and Jutten, Christian},
  journal={IEEE Transactions on Biomedical Engineering},
  volume={60},
  number={6},
  pages={1730--1738},
  year={2014},
  url={https://ieeexplore.ieee.org/document/6871319}
}

@article{DDAF2023,
  title={Domain adaptation framework for EEG classification},
  author={Zhang, X. and Li, Y. and Chen, M.},
  journal={Computers in Biology and Medicine},
  year={2023},
  url={https://www.sciencedirect.com/science/article/pii/S001048252300700X}
}

@article{SDDA2023,
  title={Semi-supervised domain adaptation for EEG},
  author={Liu, Y. and Wang, H. and Zhang, J.},
  journal={Neurocomputing},
  year={2023},
  url={https://www.sciencedirect.com/science/article/pii/S0925231223007828}
}

@article{DAMSDAF2021,
  title={Domain adaptation multi-source framework},
  author={Wang, Z. and Chen, X. and Liu, S.},
  journal={IEEE},
  year={2021},
  url={https://pmc.ncbi.nlm.nih.gov/articles/PMC9402410/}
}

@article{ADFR2023,
  title={Adversarial domain feature representation for EEG},
  author={Li, J. and Zhao, Q. and Huang, D.},
  journal={IEEE},
  year={2023},
  url={https://pmc.ncbi.nlm.nih.gov/articles/PMC11686875/}
}

@article{gao2022cnn_riemannian,
  title = {Convolutional neural network and Riemannian geometry hybrid approach for motor imagery classification},
  author = {Gao, Chang and Liu, Wenchao and Yang, Xian},
  journal = {Neurocomputing},
  volume = {507},
  pages = {180--190},
  year = {2022},
  publisher = {Elsevier},
  doi = {10.1016/j.neucom.2022.08.024}
}

@article{xiong2023riemannian_tangent_cfc,
  title = {Enhancing Motor Imagery Decoding in Brain--Computer Interfaces using Riemannian Tangent Space Mapping and Cross Frequency Coupling},
  author = {Xiong, Xiong and Su, Li and Guo, Jinjie and Song, Tianyuan and Wang, Ying and Huang, Jinguo and Kang, Guixia},
  journal = {Biomedical Signal Processing and Control},
  year = {2023},
  publisher = {Elsevier},
  url={https://arxiv.org/abs/2310.19198}
}

@article{blankertz2008optimizing,
  title={Optimizing spatial filters for robust EEG single-trial analysis},
  author={Blankertz, Benjamin and others},
  journal={IEEE Signal Processing Magazine},
  year={2008},
  url={https://doc.ml.tu-berlin.de/bbci/publications/BlaTomLemKawMue08.pdf}
}

@article{ramoser2000optimal,
  title={Optimal spatial filtering of single trial EEG during imagined hand movement},
  author={Ramoser, Herbert and Muller-Gerking, Johannes and Pfurtscheller, Gert},
  journal={IEEE Transactions on Rehabilitation Engineering},
  year={2000},
  url={https://pubmed.ncbi.nlm.nih.gov/11204034/}
}
\bibliographystyle{plainnat}   % if available

\appendix
\newpage
\section{EEG Data Representation and Covariance Construction}
\label{app:eeg_cov}

This section specifies the construction of trial-wise symmetric positive definite (SPD) covariance matrices from raw multichannel EEG recordings. Only factors that affect covariance dimensionality, numerical conditioning, or cross-subject comparability are described, as these directly determine the validity of subsequent Riemannian manifold operations.

\subsection{International 10--20 System and Channel Space}
\label{app:A1_channels}

\textbf{EEG recording and channels.}
EEG measures scalp-level voltage fluctuations generated by synchronized post-synaptic potentials in cortical pyramidal neuron populations. During acquisition, conductive electrodes placed on the scalp record voltage differences relative to a reference electrode, which are amplified and sampled at a fixed frequency to produce multichannel time series. Each electrode therefore yields one EEG channel, and the set of electrode locations determines the number of channels $d$, their ordering, and the ambient space $\mathbb{R}^{d\times d}$ in which trial-wise covariance matrices are defined.

\textbf{International 10--20 electrode system.}
Electrodes are positioned according to the international 10--20 system, which specifies locations using proportional distances (10\% or 20\%) between anatomical landmarks (nasion, inion, and preauricular points). This standardization ensures reproducible electrode placement and consistent channel indexing across subjects, enabling cross-subject comparison of covariance representations. The international 10--10 system is a higher-density extension of this scheme; in the present work, only the 10--20 configuration is used.

\begin{figure}[h]
    \centering
    \includegraphics[width=0.75\linewidth]{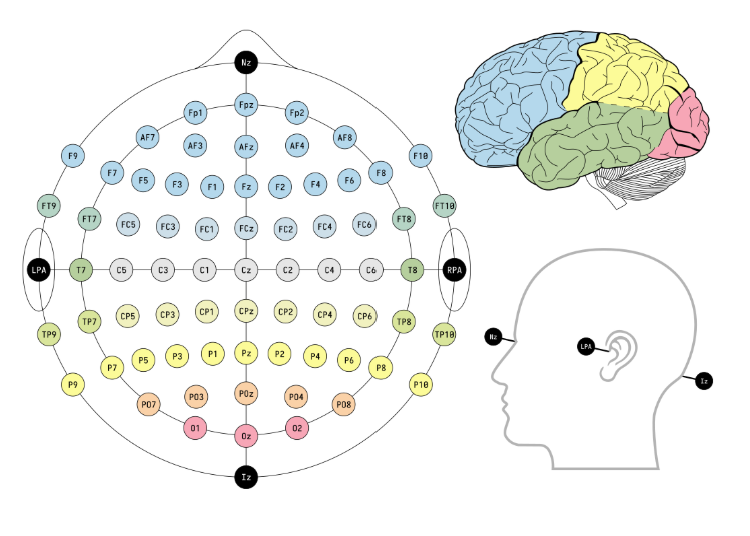}
    \caption{International 10--20 electrode placement illustrating the EEG channel locations used in BCI-IV 2a. Channels over frontal-central, central, and centro-parietal regions capture dominant motor imagery-related activity and define the spatial structure encoded by trial-wise covariance matrices.}
    \label{fig:1020}
\end{figure}

\textbf{BCI Competition IV-2a channel configuration.}

BCI Competition IV dataset~2a records EEG using the international 10--20 system with $d=22$ EEG channels sampled at $f_s = 250$~Hz; three additional EOG channels are provided for artifact monitoring and excluded from analysis.
The selected EEG channels densely cover frontal-central, central, and parietal regions associated with motor imagery, yielding a configuration well suited for covariance-based representations of MI-related spatial structure.
The ordered channel set
\[
\{\texttt{Fz}, \texttt{FC3}, \texttt{FC1}, \texttt{FCz}, \texttt{FC2}, \texttt{FC4},
\texttt{C5}, \texttt{C3}, \texttt{C1}, \texttt{Cz}, \texttt{C2}, \texttt{C4}, \texttt{C6},
\texttt{CP3}, \texttt{CP1}, \texttt{CPz}, \texttt{CP2}, \texttt{CP4},
\texttt{P1}, \texttt{Pz}, \texttt{P2}, \texttt{POz}\},
\]
which is treated as a fixed ordered basis across subjects for this dataset, so that each trial is represented by a covariance matrix in $\mathcal{S}_{++}^{22}$.

This channel configuration is \emph{not} imposed across datasets.
For each dataset, we instead use either the full available montage or a task-relevant subset, depending on recording protocol and channel availability.
For example, Ofner2017 provides 60 EEG channels; we select the subset corresponding to the BCI-IV-2a motor imagery channels where available.
One channel (\texttt{POz}) is absent in Ofner2017, resulting in a $d=21$ channel configuration and covariance matrices in $\mathcal{S}_{++}^{21}$.
No channel interpolation or padding is performed, and channel ordering is fixed per dataset to preserve anatomical consistency.

\subsection{Session Structure, Event Timing, and Trial Segmentation}
\label{app:A2_trials}

BCI recordings are provided as continuous multichannel time series with discrete event annotations marking the onset and identity of each motor imagery (MI) trial. This subsection specifies the temporal structure of trials and the construction of fixed-length epochs used for covariance-based representations, with BCI Competition IV dataset~2a serving as a concrete instantiation.

\textbf{Continuous recording and event annotations.}
For a given subject and session, the preprocessed EEG recording is represented as a discrete-time multivariate signal
\begin{equation}
x(t) \in \mathbb{R}^{d}, \qquad t = 1,\dots,T_{\mathrm{raw}},
\end{equation}
where $d$ denotes the number of EEG channels and $T_{\mathrm{raw}}$ the total number of recorded samples. Event annotations define a sequence of trial markers $\{(\tau_k, y_k)\}_{k=1}^{N}$, where $\tau_k$ is the sample index of the cue onset for the $k$-th trial and $y_k \in \{1,\dots,K\}$ is its class label. These cue onsets provide a common temporal reference for aligning trial-wise segments across subjects and sessions.

\textbf{Trial timing and cue-aligned indexing (BCI-IV 2a).}
In BCI-IV 2a, each trial follows a fixed temporal protocol consisting of a baseline period, a motor imagery interval, and a post-imagery inter-trial period. Specifically, a fixation/baseline phase spans $t \in [0,2]$~s, a visual cue is presented at $t=2$~s, and motor imagery is performed until $t=6$~s, yielding a nominal 4~s MI interval. Let $f_s$ denote the sampling frequency (BCI-IV 2a: $f_s=250$~Hz). A temporal offset of $\Delta t$ seconds relative to cue onset corresponds to $\lfloor \Delta t\, f_s \rfloor$ samples, so cue-aligned indices are expressed as $\tau_k + \lfloor \Delta t\, f_s \rfloor$.

\textbf{Active epoch selection.}
Although the MI interval spans 4~s following cue onset, covariance estimation benefits from restricting attention to a temporally stable sub-interval that excludes cue-locked transients and late trial-end effects. Accordingly, for each trial $k$, an active epoch is extracted strictly within the MI interval:
\begin{equation}
X_k^{\mathrm{act}}
=
\Big[
x\big(\tau_k + \lfloor t_{\min} f_s \rfloor\big),\,
\dots,\,
x\big(\tau_k + \lfloor t_{\max} f_s \rfloor - 1\big)
\Big]
\in \mathbb{R}^{d \times T},
\label{eq:active_epoch}
\end{equation}
with $(t_{\min}, t_{\max}) = (0.5, 3.5)$~s relative to cue onset. Under the BCI-IV 2a protocol, this corresponds to absolute trial time $t \in [2.5, 5.5]$~s, i.e., the central 3~s of the MI interval. At $f_s=250$~Hz, this yields
\begin{equation}
T = \lfloor (t_{\max} - t_{\min}) f_s \rfloor = 750
\end{equation}
samples per trial. This window captures sustained MI-related activity while providing sufficient sample support for stable covariance estimation.

\textbf{Rest epoch extraction (auxiliary construction).}
For auxiliary analyses such as rest-versus-action discrimination, a cue-preceding rest epoch of equal duration is extracted:
\begin{equation}
X_k^{\mathrm{rest}}
=
\Big[
x\big(\tau_k + \lfloor t_{\min}^{r} f_s \rfloor\big),\,
\dots,\,
x\big(\tau_k + \lfloor t_{\max}^{r} f_s \rfloor - 1\big)
\Big]
\in \mathbb{R}^{d \times T},
\label{eq:rest_epoch}
\end{equation}
with $(t_{\min}^{r}, t_{\max}^{r}) = (-3.5, -0.5)$~s relative to cue onset, also yielding $T=750$ samples. This choice produces rest and active epochs with matched temporal support while avoiding overlap with cue-locked activity. Rest epochs are used only for auxiliary constructions and do not alter the main multi-class MI evaluation protocol.

\textbf{Boundary constraints to validate epochs.}
Epochs are extracted only when their temporal extent lies fully within the recorded signal:
\begin{equation}
\tau_k + \lfloor t_{\min}^{r} f_s \rfloor \ge 1,
\qquad
\tau_k + \lfloor t_{\max} f_s \rfloor - 1 \le T_{\mathrm{raw}}.
\end{equation}
Trials violating these constraints are discarded to ensure a consistent $d \times T$ representation across all retained trials.

\textbf{Output of segmentation.}
After segmentation, each session yields a collection of labeled active epochs $\{(X_k^{\mathrm{act}}, y_k)\}_{k=1}^{N'}$, where $N' \le N$ accounts for boundary-based exclusions. These fixed-length epochs constitute the inputs to the covariance estimation procedure described in Section~\ref{app:A4_cov}.

\subsection{Preprocessing Prior to Covariance Estimation}
\label{app:A3_preproc}

This subsection specifies the preprocessing operators applied to EEG signals prior to covariance construction. Only operations that affect channel dimensionality, spectral content, or numerical conditioning of covariance matrices are described.

\textbf{Channel selection.}
Let $x(t) \in \mathbb{R}^{C_{\mathrm{raw}}}$ denote the recorded multichannel signal. We retain EEG channels only and discard EOG channels, yielding
\begin{equation}
x_{\mathrm{EEG}}(t) \in \mathbb{R}^{d},
\end{equation}
with fixed $d=22$ across subjects and datasets (Section~\ref{app:A1_channels}). This fixes the covariance dimensionality and excludes non-neural ocular activity that is not informative for motor imagery decoding.

\textbf{Frequency-band filtering.}
EEG signals are band-pass filtered to retain task-relevant frequency components and suppress slow drifts and high-frequency noise. Let $h_{\mathrm{bp}}(t)$ denote the impulse response of the band-pass filter. The filtered signal is
\begin{equation}
x_{\mathrm{bp}}(t) = (h_{\mathrm{bp}} * x_{\mathrm{EEG}})(t).
\end{equation}
To remove narrowband line interference, a notch filter with impulse response $h_{\mathrm{notch}}(t)$ is applied,
\begin{equation}
x_{\mathrm{filt}}(t) = (h_{\mathrm{notch}} * x_{\mathrm{bp}})(t).
\end{equation}
Both operations are linear and time-invariant, preserve channel dimensionality, and shape the eigen-spectrum and conditioning of the resulting covariance matrices without inducing cross-channel mixing.

\textbf{Artifact handling policy.}
No ICA-based artifact removal is applied. Data-dependent decompositions such as ICA can alter covariance rank and orientation in a subject-specific manner, complicating consistent geometry-aware learning across subjects and datasets.

\subsection{Covariance Estimation}
\label{app:A4_cov}

Motor--imagery EEG trials are denoted by
$X_{i,e}\in\mathbb{R}^{d\times T}$, where $i$ indexes the subject, $e$ the trial,
$d$ the number of EEG channels, and $T$ the number of temporal samples after
preprocessing (Appendix~\ref{app:A3_preproc}).
Each trial is associated with an action label
$y_{i,e}\in\{1,\dots,K\}$.

Following standard Riemannian BCI practice, temporal de--meaning is performed
during preprocessing, and the spatial covariance representation of each trial
is computed as
\begin{equation}
C_{i,e}
=
\frac{1}{T-1}\,X_{i,e}X_{i,e}^{\top}
\;\in\;
\mathbb{S}_{++}^{d}.
\label{eq:trial_cov}
\end{equation}

The resulting symmetric positive--definite (SPD) matrices encode the
second--order spatial statistics of EEG channels and constitute the basic
objects manipulated by all geometry--aware alignment and classification
modules throughout this work.
Unless otherwise stated, we use $X_{i,e}$ to denote trial--level EEG signals and
$C_{i,e}$ for their associated covariance representations.

By construction, $C_{i,e}$ is symmetric positive definite. In practice, finite-sample effects and inter-channel correlations may lead to rank-deficient covariance estimates, which is addressed in the subsequent mathematical formulation.

\subsection{Synthetic Evidence for Geometry-Aware Congruence Modeling}
\label{app:synthetic}

To explicitly characterize the limitations of Riemannian Alignment (RA) and tangent-space methods, we construct a controlled synthetic EEG covariance generator on the SPD manifold. The goal is not merely to report performance, but to provide mechanistic evidence for the proposed geometry-aware congruence transformations by isolating distinct sources of cross-subject variability.

\paragraph{Synthetic data construction.}
The default generator uses $9$ subjects, $4$ classes, $40$ samples per class per subject, and $20$ channels, yielding $360$ samples per subject. The four class prototypes are arranged in a tetrahedral configuration within a $3$-dimensional signal subspace, ensuring that classes are equidistant in their undistorted form. Each sample is generated by first constructing a clean class-centered signal in the low-dimensional subspace, mapping it to the SPD manifold using the matrix exponential, and then applying stage-specific subject distortions through rotations and congruence transforms of the form $W C W^\top$. Finally, each covariance matrix is symmetrized, stabilized with jitter $10^{-5}$, and trace-normalized, matching standard covariance preprocessing used in EEG pipelines.

For a trial EEG matrix $X_{i,e}\in\mathbb{R}^{d\times T}$ from subject $i$ and trial $e$, we represent the signal by its covariance
\[
C_{i,e}=\frac{1}{T-1}X_{i,e}X_{i,e}^\top \in\mathcal{S}_{++}^{d}.
\]
EEG class structure is naturally encoded in SPD geometry through: 
(i) eigenvalues, corresponding to power/dispersion, 
(ii) eigenvectors, corresponding to orientations of informative subspaces, and 
(iii) subject-specific congruence warps, corresponding to domain shift.

We model each covariance as
\begin{equation}
\begin{aligned}
C_{i,e}
=
G_i^{1/2}
\Big(
U_s \Lambda_{i,y_{i,e},e} U_i^\top
+
U_\perp \Lambda_{\mathrm{bg}} U_\perp^\top
\Big)
G_i^{1/2}, 
\quad
U_i = U R_i, \\
\Lambda_{i,e}=\mathrm{diag}\big(\exp(d_i\odot \mu_{y_{i,e}}+\varepsilon)\big),
\end{aligned}
\label{eq:synthetic_model}
\end{equation}
where $G_i$ is a subject-specific warp, $R_i$ is a subject-specific rotation of the signal subspace, and $d_i$ controls subject-dependent dispersion.

\paragraph{Why RA and tangent-space methods are insufficient.}
Riemannian Alignment removes the outer subject warp:
\[
\widetilde{C}_{i,y_{i,e},e}=\bar{C}_i^{-1/2}C_{i,y_{i,e},e}\bar{C}_i^{-1/2}.
\]
This is effective when domain shift is primarily congruence-based. However, RA does not remove subject dependence that persists \emph{within} the aligned geometry, namely:
\begin{itemize}
    \item dispersion scaling ($\Lambda_{i,y_{i,e},e}$),
    \item orientation mismatch ($R_i$).
\end{itemize}

Tangent-space methods further rely on first-order approximations:
\[
T(C)=\log\!\big(\bar{C}^{-1/2}C\bar{C}^{-1/2}\big),
\]
which do not preserve higher-order nonlinear interactions on the SPD manifold.

\paragraph{Synthetic hierarchy.}
We construct a progressive hierarchy of datasets by incrementally introducing sources of variability:

\begin{enumerate}
    \item \textbf{Mean Shift Only.}
    \[
    C_{i,y_{i,e},e}^{(0)} = G_i^\alpha \exp\big(U\,\mathrm{diag}(\mu_{y_{i,e}})\,U^\top + U_\perp \Lambda_{\mathrm{bg}}U_\perp^\top\big) G_i^\alpha.
    \]
    This isolates removable domain shift; RA is expected to perform optimally.

    \item \textbf{+ Dispersion.}
    \[
    C_{i,y_{i,e},e}^{(1)} = G_i^\alpha \exp\big(U\,\mathrm{diag}(d_i\odot\mu_{y_{i,e}}+\varepsilon)\,U^\top + U_\perp \Lambda_{\mathrm{bg}}U_\perp^\top\big) G_i^\alpha.
    \]
    Introduces subject-dependent anisotropic scaling.

    \item \textbf{+ Orientation.}
    \begin{equation}
    \begin{aligned}
    C_{i,y_{i,e},e}^{(2)} = G_i^\alpha \Big(U_i\,\mathrm{diag}(\exp(d_i\odot\mu_{y_{i,e}}+\varepsilon))\,U_i^\top \\
    + U_\perp \Lambda_{\mathrm{bg}}U_\perp^\top\Big) G_i^\alpha,
    \quad U_i=UR_i.
    \end{aligned}
    \end{equation}
    Adds subject-specific rotations of discriminative subspaces.

    \item \textbf{+ Tangent-Space Distortion.}
    Starting from log-SPD representations $L_{i,y_{i,e},e}$:
    \begin{equation}
    \begin{aligned}
    L'_{i,y_{i,e},e}=L_{i,y_{i,e},e}+\gamma_1\,\mathrm{sym}(S_iL_{i,y_{i,e},e}+L_{i,y_{i,e},e}S_i) \\
    +\gamma_2\,\mathrm{sym}(L_{i,y_{i,e},e}^2),
    \quad
    \Sigma_{i,y_{i,e},e}^{(3)}=\exp(L'_{i,y_{i,e},e}).
    \end{aligned}
    \end{equation}
    Introduces nonlinear interactions not preserved by a single tangent chart.

    \item \textbf{+ Nonlinear (BCH-style fusion).}
    Construct two SPD branches:
    \[
    C_1=\exp(L_1), \quad C_2=\exp(L_2),
    \]
    and fuse them via:
    \begin{equation}
    \begin{aligned}
    C_{i,y_{i,e},e}^{(4)} = G_i^\rho \exp\Big(\log C_1+\log C_2 \\
    +\delta\mathrm{sym}\big((\log C_1+\log C_2)^2\big)\Big) G_i^\rho.
    \end{aligned}
    \end{equation}
    Models higher-order nonlinear covariance interactions.
\end{enumerate}

\begin{table*}[t]
\centering
\small
\caption{Synthetic hierarchy demonstrating failure modes under increasing geometric complexity.}
\label{tab:synthetic}
\begin{tabular}{lccccc}
\toprule
Dataset & \textbf{Raw\_TSLR} & \textbf{RA\_TSLR} & \textbf{DCT (E2E)} & \textbf{DLDCT (E2E)} & \textbf{DDCT-UNet (E2E)} \\
\midrule
0. Mean Shift Only     & 68.12 & 100.00 & 100.00 & 100.00 & 100.00 \\
1. +Dispersion         & 50.63 & 77.97  & 73.28  & 83.44  & 84.06  \\
2. +Orientation        & 62.34 & 62.34  & 63.91  & 68.75  & 68.75  \\
3. +TS Distortion      & 51.41 & 53.12  & 63.91  & 70.78  & 72.19  \\
4. +Nonlinear (BCH)    & 67.34 & 69.69  & 54.06  & 70.41  & 73.75  \\
\bottomrule
\end{tabular}
\end{table*}

The hierarchy reflects progressively realistic covariance distortions:
\[
\text{mean shift} \rightarrow \text{dispersion} \rightarrow \text{orientation} \rightarrow \text{TS distortion} \rightarrow \text{nonlinear fusion}.
\]

RA performs optimally under pure mean shift but degrades systematically as dispersion, orientation, and nonlinear interactions are introduced. This demonstrates that cross-subject variability extends beyond mean alignment. These observations motivate our model progression:
\[
\text{DCT} \rightarrow \text{DLDCT} \rightarrow \text{DDCT--UNet},
\]
which progressively addresses dispersion/orientation, manifold-level transformations, and nonlinear SPD interactions. Each stage corresponds to a violation of assumptions implicitly made by RA and tangent-space methods, providing a mechanistic explanation for their empirical limitations.

\section{Mathematical Preliminaries}
\label{app:math_prelims}

\subsection{Notation and Basic Definitions}
\label{app:notation}

Let $\mathbb{S}^d := \{A\in\mathbb{R}^{d\times d}: A=A^\top\}$ denote the real
symmetric matrices, and
\begin{equation}
\mathcal{S}_{++}^d := \{C\in \mathbb{S}^d: x^\top C x>0,\ \forall x\neq 0\}
\end{equation}
the cone of symmetric positive definite (SPD) matrices.
We write $C\succ 0$ (resp.\ $C\succeq 0$) for SPD (resp.\ PSD).
The identity is $I$.
The Frobenius norm is
$\|A\|_F := \sqrt{\mathrm{tr}(A^\top A)}$.
For SPD $C$, the eigendecomposition
$C = U \Lambda U^\top$
has $U$ orthogonal and
$\Lambda=\mathrm{diag}(\lambda_1,\dots,\lambda_d)$ with $\lambda_i>0$.
Matrix functions are defined spectrally:
\begin{equation}
C^\alpha := U\Lambda^\alpha U^\top,\qquad
\log C := U(\log\Lambda)U^\top,\qquad
\exp C := U(\exp\Lambda)U^\top.
\end{equation}
The tangent space at any $C\in\mathcal{S}_{++}^d$ is
$T_C\mathcal{S}_{++}^d \cong \mathbb{S}^d$.

\subsection{SPD Matrix Properties}
\label{app:spd}

\textbf{Lemma 1 (SPD $\Leftrightarrow$ spectral positivity).}
$C\in\mathcal{S}_{++}^d$ iff $C=U\Lambda U^\top$ with $\lambda_i>0$ for all $i$.

\textit{Proof.}
($\Rightarrow$) SPD implies all eigenvalues are positive by the Rayleigh
quotient:
$\lambda_{\min}(C)=\min_{\|x\|=1}x^\top Cx>0$.
($\Leftarrow$) If $\lambda_i>0$, then for any $x\neq 0$,
\[
x^\top C x
=
x^\top U\Lambda U^\top x
=
\sum_{i=1}^d \lambda_i \tilde{x}_i^2
>0 .
\]

\textbf{Lemma 2 (Inverse and symmetric square roots).}
If $C\in\mathcal{S}_{++}^d$, then $C^{-1}\in\mathcal{S}_{++}^d$ and there exist
unique SPD matrices
$C^{1/2}\in\mathcal{S}_{++}^d$ and
$C^{-1/2}\in\mathcal{S}_{++}^d$ such that
\begin{equation}
C^{1/2}C^{1/2}=C,
\qquad
C^{-1/2} C\, C^{-1/2}=I,
\qquad
C^{-1/2}=(C^{1/2})^{-1}.
\end{equation}

\textit{Proof.}
Since $C\in\mathcal{S}_{++}^d$, by the spectral theorem there exists an
orthogonal $U$ and diagonal
$\Lambda=\mathrm{diag}(\lambda_1,\dots,\lambda_d)$ with $\lambda_i>0$ such that
\begin{equation}
C = U \Lambda U^\top,
\qquad
U^\top U = UU^\top = I.
\end{equation}
Define
\begin{equation}
C^{1/2} := U \Lambda^{1/2} U^\top,
\qquad
\Lambda^{1/2}:=\mathrm{diag}(\sqrt{\lambda_1},\dots,\sqrt{\lambda_d}),
\end{equation}
and similarly
\begin{equation}
C^{-1/2} := U \Lambda^{-1/2} U^\top,
\qquad
\Lambda^{-1/2}:=\mathrm{diag}(\lambda_1^{-1/2},\dots,\lambda_d^{-1/2}).
\end{equation}
We first verify the defining identities. Using $U^\top U=I$,
\begin{align}
C^{1/2}C^{1/2}
&=
U \Lambda^{1/2} (U^\top U)\Lambda^{1/2} U^\top
=
U \Lambda U^\top
= C, \\
C^{-1/2} C\, C^{-1/2}
&=
U\Lambda^{-1/2}(U^\top U)\Lambda(U^\top U)\Lambda^{-1/2}U^\top
=
U I U^\top
= I.
\end{align}
Hence $C^{1/2}$ and $C^{-1/2}$ satisfy the stated relations.

Next we show symmetry and positive definiteness. Since
$\Lambda^{\pm 1/2}$ are diagonal and $U$ is orthogonal,
\begin{equation}
(C^{\pm 1/2})^\top
=
(U\Lambda^{\pm 1/2}U^\top)^\top
=
U\Lambda^{\pm 1/2}U^\top
=
C^{\pm 1/2}.
\end{equation}
For any $x\neq 0$, letting $\tilde{x}=U^\top x\neq 0$,
\begin{equation}
x^\top C^{1/2} x
=
\tilde{x}^\top \Lambda^{1/2}\tilde{x}
=
\sum_{i=1}^d \sqrt{\lambda_i}\,\tilde{x}_i^2
>0,
\end{equation}
since $\sqrt{\lambda_i}>0$. The same argument with
$\Lambda^{-1/2}$ yields
$C^{-1/2}\in\mathcal{S}_{++}^d$.
Finally,
\begin{equation}
C^{1/2} C^{-1/2}
=
U\Lambda^{1/2}(U^\top U)\Lambda^{-1/2}U^\top
=
U I U^\top
= I,
\end{equation}
hence $C^{-1/2}=(C^{1/2})^{-1}$.

Uniqueness of the SPD square root follows as: if
$S\in\mathcal{S}_{++}^d$ and $S^2=C$, then $S$ commutes with $C$ and shares the
same eigenvectors, so
$S=U\,\mathrm{diag}(s_i)\,U^\top$ with $s_i^2=\lambda_i$ and $s_i>0$, implying
$s_i=\sqrt{\lambda_i}$ and hence $S=C^{1/2}$.

\paragraph{Lemma 3 (Congruence preserves SPD).}
\label{lem:congruence_spd}
Let $C\in\mathcal{S}_{++}^{d_{\mathrm{in}}}$ and
$W\in\mathbb{R}^{d_{\mathrm{in}}\times d_{\mathrm{out}}}$ be a matrix of full
column rank. Then the congruence transform
\begin{equation}
\Phi(C) := W^\top C W
\label{eq:congruence_transform}
\end{equation}
produces an SPD matrix $\Phi(C)\in\mathcal{S}_{++}^{d_{\mathrm{out}}}$.

\textit{Proof.}
Symmetry follows immediately:
\[
(W^\top C W)^\top = W^\top C^\top W = W^\top C W,
\]
since $C$ is symmetric.

To show positive definiteness, let $x\in\mathbb{R}^{d_{\mathrm{out}}}$ with
$x\neq 0$. Because $W$ has full column rank, $W x\neq 0$. Then
\[
x^\top (W^\top C W) x
=
(W x)^\top C (W x).
\]
Since $C\in\mathcal{S}_{++}^{d_{\mathrm{in}}}$, the quadratic form
$(W x)^\top C (W x)$ is strictly positive for all nonzero $W x$. Hence
\[
x^\top (W^\top C W) x > 0 \quad \forall\, x\neq 0,
\]
which implies $W^\top C W\in\mathcal{S}_{++}^{d_{\mathrm{out}}}$.

\paragraph{Lemma 4 (Skew-symmetric exponential is special orthogonal).}
\label{lem:skewexp_in_SO}
Let $K\in\mathbb{R}^{d\times d}$ be \emph{skew-symmetric}, i.e., $K^\top=-K$.
Define
\[
R := \exp(K) \;\in\; \mathbb{R}^{d\times d}.
\]
Then $R\in SO(d)$; equivalently,
\[
R^\top R = I
\quad\text{and}\quad
\det(R)=1.
\]

\textit{Proof.}
We prove orthogonality and unit determinant separately.

\smallskip
\noindent\textbf{(i) Orthogonality: $R^\top R=I$.}
Recall two standard facts of the matrix exponential:
\begin{enumerate}
    \item For any square matrix $X$, $(\exp X)^\top = \exp(X^\top)$.
    \item For any square matrix $X$, $\exp(X)\exp(-X)=I$.
\end{enumerate}
Using these,
\[
R^\top R
=
(\exp K)^\top \exp K
=
\exp(K^\top)\exp(K).
\]
Since $K^\top=-K$,
\[
R^\top R
=
\exp(-K)\exp(K)
=
I.
\]

\smallskip
\noindent\textbf{(ii) Determinant: $\det(R)=1$.}
We use $\det(\exp X)=\exp(\mathrm{tr}(X))$.
Let $p(q):=\det(\exp(qK))$. Using Jacobi's formula,
\[
\frac{p'(q)}{p(q)}
=
\mathrm{tr}\!\big(\exp(-qK)\,K\,\exp(qK)\big)
=
\mathrm{tr}(K),
\]
by cyclicity of trace. Hence
\[
\det(\exp K)=\exp(\mathrm{tr}(K)).
\]
Since $K$ is skew-symmetric, $K_{ii}=0$ for all $i$, so
\[
\mathrm{tr}(K)=\sum_{i=1}^d K_{ii}=0,
\]
and therefore $\det(R)=1$.

Combining (i) and (ii), we conclude $R=\exp(K)\in SO(d)$.

\paragraph{Corollary (Lie-algebra parameterization used in DCT).}
\label{cor:dct_so_param}
Let $A\in\mathbb{R}^{d\times d}$ be arbitrary and define
$K:=A-A^\top$. Then
\[
R(A):=\exp(A-A^\top)=\exp(K)\in SO(d),
\]
so the parameterization enforces orthogonality and $\det=1$ by construction.

\subsection{Affine-Invariant Riemannian Metric on $\mathcal{S}_{++}^d$}
\label{app:airm_metric}

We adopt the affine-invariant Riemannian metric (AIRM). For
$C\in\mathcal{S}_{++}^d$ and
$U,V\in T_C\mathcal{S}_{++}^d\cong\mathbb{S}^d$, define
\begin{equation}
\langle U,V\rangle_C := \mathrm{tr}(C^{-1}U\,C^{-1}V),
\qquad
\|U\|_C := \sqrt{\langle U,U\rangle_C}.
\label{eq:airm_inner}
\end{equation}

\textbf{Lemma 5 (valid inner product).}
$\langle\cdot,\cdot\rangle_C$ is a symmetric bilinear positive definite form on
$\mathbb{S}^d$.

\textit{Proof.}
For $U\neq 0$,
\[
\langle U,U\rangle_C
=
\|C^{-1/2} U C^{-1/2}\|_F^2>0 .
\]

\textbf{Lemma 6 (Geodesic distance and metric properties).}
Let $C_1,C_2\in\mathcal{S}_{++}^d$ and define
\begin{equation}
\widetilde{C}_2 := C_1^{-1/2} C_2 C_1^{-1/2}\in\mathcal{S}_{++}^d.
\end{equation}
Then
\begin{equation}
d_{\mathrm{AIRM}}(C_1,C_2)
=
\|\log(C_1^{-1/2} C_2 C_1^{-1/2})\|_F .
\label{eq:airm_dist}
\end{equation}

\subsection{Metric Properties of the AIRM Distance:}
We now verify that the affine-invariant Riemannian (AIRM) distance defined above
satisfies the axioms of a metric on $\mathcal{S}_{++}^d$ and admits additional
invariance properties that are central to our congruence-based framework.

\textbf{Lemma 7 (symmetry).}
$d_{\mathrm{AIRM}}(C_1,C_2)=d_{\mathrm{AIRM}}(C_2,C_1)$.

\textit{Proof.}
Let $X := C_1^{-1/2}C_2C_1^{-1/2}\in\mathcal{S}_{++}^d$. Then
\begin{equation}
d_{\mathrm{AIRM}}(C_1,C_2)=\|\log X\|_F.
\end{equation}
Using Lemma 2, $X^{-1}\in\mathcal{S}_{++}^d$ and $\log(X^{-1})=-\log(X)$. Hence
$\|\log(X^{-1})\|_F=\|\log(X)\|_F$.
It remains to relate $d_{\mathrm{AIRM}}(C_2,C_1)$ to $\|\log(X^{-1})\|_F$.
Note $C_2^{-1/2}C_1C_2^{-1/2}$ is similar to $X^{-1}$ (they have identical eigenvalues) because
\begin{equation}
X^{-1}=C_1^{1/2}C_2^{-1}C_1^{1/2}
\sim C_2^{-1/2}C_1C_2^{-1/2},
\end{equation}
and $\|\log(\cdot)\|_F$ depends only on eigenvalues for SPD matrices. Therefore
\begin{equation}
d_{\mathrm{AIRM}}(C_2,C_1)=\|\log(C_2^{-1/2}C_1C_2^{-1/2})\|_F=\|\log(X^{-1})\|_F=\|\log X\|_F.
\end{equation}

\textbf{Lemma 8 (identity of indiscernibles).}
$d_{\mathrm{AIRM}}(C_1,C_2)=0$ iff $C_1=C_2$.

\begin{proof}
$d_{\mathrm{AIRM}}(C_1,C_2)=0 \iff \log(C_1^{-1/2}C_2C_1^{-1/2})=0 \iff C_1^{-1/2}C_2C_1^{-1/2}=I \iff C_2=C_1$.
\end{proof}

\textbf{Lemma 9 (congruence invariance / affine invariance).}
For any invertible $W$,
\begin{equation}
d_{\mathrm{AIRM}}(W^\top C_1 W,\, W^\top C_2 W)=d_{\mathrm{AIRM}}(C_1,C_2).
\label{eq:affine_invariance}
\end{equation}

\begin{proof}
Let $A_i:=W^\top C_i W$. By Lemma 3, $A_i\in\mathcal{S}_{++}^d$.
Consider the whitened matrix for $(A_1,A_2)$:
\begin{equation}
A_1^{-1/2}A_2A_1^{-1/2}
=
A_1^{-1/2} (W^\top C_2 W) A_1^{-1/2}.
\end{equation}
Define $\bar W := W^\top$ (also invertible). Then $A_i=\bar W C_i \bar W^\top$, and we can invoke the standard SPD identity:
for invertible $\bar W$, the induced whitening satisfies
\begin{equation}
A_1^{-1/2}\bar W = Q\, C_1^{-1/2},
\end{equation}
for some orthogonal matrix $Q$ (polar factor). Substituting yields
\begin{equation}
A_1^{-1/2}A_2A_1^{-1/2}
=
Q\,(C_1^{-1/2}C_2C_1^{-1/2})\,Q^\top.
\end{equation}
Taking logarithms preserves orthogonal similarity: $\log(QXQ^\top)=Q(\log X)Q^\top$ for SPD $X$.
The Frobenius norm is orthogonally invariant, hence
\begin{equation}
\big\|\log(A_1^{-1/2}A_2A_1^{-1/2})\big\|_F
=
\big\|Q\,\log(C_1^{-1/2}C_2C_1^{-1/2})\,Q^\top\big\|_F
=
\big\|\log(C_1^{-1/2}C_2C_1^{-1/2})\big\|_F.
\end{equation}
Therefore $d_{\mathrm{AIRM}}(W^\top C_1 W,\, W^\top C_2 W)=d_{\mathrm{AIRM}}(C_1,C_2)$.
\end{proof}

Under AIRM, $d_{\mathrm{AIRM}}(\cdot,\cdot)$ is the length-minimizing geodesic distance induced by the Riemannian metric \eqref{eq:airm_inner}. Hence it satisfies the triangle inequality as a consequence of general Riemannian geometry (geodesic distances are metric distances). We do not re-prove the full triangle inequality here, as the paper uses only symmetry, invariance, and identity properties explicitly.

\subsection{Logarithm and Exponential Maps, and Geodesics}
\label{app:logexp}

\textbf{Definition (log/exp maps).}
For $C,C_2\in\mathcal{S}_{++}^d$ and $Z\in\mathbb{S}^d$,
\begin{align}
Z = \log_C(C_2)
&:= C^{1/2}\,\log\!\big(C^{-1/2} C_2 C^{-1/2}\big)\,C^{1/2},
\label{eq:log_map}\\
\exp_C(Z)
&:= C^{1/2}\,\exp\!\big(C^{-1/2} Z C^{-1/2}\big)\,C^{1/2}.
\label{eq:exp_map}
\end{align}

\textbf{Lemma 10 (log/exp are inverses on $\mathcal{S}_{++}^d$).}
For any $C_2\in\mathcal{S}_{++}^d$,
\begin{equation}
\exp_C(\log_C(C_2))=C_2,
\qquad
\log_C(\exp_C(Z))=Z.
\end{equation}

\textit{Proof.}
Let $X := C^{-1/2}C_2C^{-1/2}\in\mathcal{S}_{++}^d$.
Then
\begin{equation}
\log_C(C_2)=C^{1/2}(\log X)C^{1/2}
\;\Rightarrow\;
C^{-1/2}\log_C(C_2)C^{-1/2}=\log X.
\end{equation}
Applying $\exp_C$:
\begin{equation}
\exp_C(\log_C(C_2))
=
C^{1/2}\exp(\log X)C^{1/2}
=
C^{1/2} X C^{1/2}
=
C_2.
\end{equation}
The reverse direction is analogous using $\log(\exp Y)=Y$ for symmetric $Y$. 

\paragraph{Lemma 11 (SPD preservation of log-Euclidean merge).}
\label{lem:logeuc_merge}
Let $C_1,C_2\in\mathcal{S}_{++}^d$. Define the log-Euclidean merge operator
\begin{equation}
\mathrm{merge}(C_1,C_2)
:=
\exp\!\Big(\tfrac{1}{2}\big(\log C_1+\log C_2\big)\Big).
\label{eq:logeuc_merge}
\end{equation}
Then $\mathrm{merge}(C_1,C_2)\in\mathcal{S}_{++}^d$.

\textit{Proof.}
Since $C_1,C_2\in\mathcal{S}_{++}^d$, each admits an eigendecomposition
\[
C_j = U_j \Lambda_j U_j^\top,\qquad j\in\{1,2\},
\]
where $U_j\in O(d)$ is orthogonal and
$\Lambda_j=\mathrm{diag}(\lambda_{j,1},\dots,\lambda_{j,d})$ with
$\lambda_{j,r}>0$ for all $r$. The principal matrix logarithm is well-defined on
$\mathcal{S}_{++}^d$ and satisfies
\[
\log(C_j)=U_j\,\log(\Lambda_j)\,U_j^\top,
\qquad
\log(\Lambda_j)=\mathrm{diag}(\log\lambda_{j,1},\dots,\log\lambda_{j,d}),
\]
hence $\log(C_j)\in\mathbb{S}^d$ is symmetric.

Define
\[
A := \tfrac{1}{2}\big(\log C_1+\log C_2\big).
\]
Because $\mathbb{S}^d$ is a real vector space closed under addition and scalar
multiplication, $A\in\mathbb{S}^d$ is symmetric. Therefore $A$ admits an
orthogonal eigendecomposition
\[
A = U_A \Delta U_A^\top,
\qquad
U_A\in O(d),\ \Delta=\mathrm{diag}(\delta_1,\dots,\delta_d)\in\mathbb{R}^{d\times d}.
\]
The matrix exponential satisfies
\[
\exp(A)=U_A\,\exp(\Delta)\,U_A^\top,
\qquad
\exp(\Delta)=\mathrm{diag}(e^{\delta_1},\dots,e^{\delta_d}),
\]
where $e^{\delta_r}>0$ for all $r$. Hence $\exp(A)$ is symmetric and positive
definite. Indeed, for any nonzero $x\in\mathbb{R}^d$,
\[
x^\top \exp(A)\,x
=
(U_A^\top x)^\top \exp(\Delta)(U_A^\top x)
=
\sum_{r=1}^d e^{\delta_r}\,(U_A^\top x)_r^2
>0,
\]
since $U_A$ is invertible and $\exp(\Delta)$ has strictly positive diagonal
entries. Therefore $\exp(A)\in\mathcal{S}_{++}^d$.

Finally, by definition \eqref{eq:logeuc_merge},
$\mathrm{merge}(C_1,C_2)=\exp(A)\in\mathcal{S}_{++}^d$.

Under the log-Euclidean metric
$d_{\mathrm{LE}}(C_1,C_2)=\|\log C_1-\log C_2\|_F$,
the curve
$\gamma(t_{int})=\exp\!\big((1-t_{int})\log C_1+t_{int}\log C_2\big)$
is a constant-speed geodesic on $\mathcal{S}_{++}^d$, where $t_{int} \in [0, 1]$ is the interpolation parameter; hence
$\mathrm{merge}(C_1,C_2)=\gamma(\tfrac12)$ is its unique midpoint.

\textbf{Lemma 12 (geodesic curve).}
The constant-speed AIRM geodesic from $C_1$ to $C_2$ is
\begin{equation}
\gamma(t_{int})=\exp_{C_1}\big(t_{int}\,\log_{C_1}(C_2)\big)
=
C_1^{1/2}\big(C_1^{-1/2}C_2C_1^{-1/2}\big)^{t_{int}} C_1^{1/2},
\quad t_{int}\in[0,1].
\label{eq:geodesic}
\end{equation}

\textit{Proof.}
Let $X=C_1^{-1/2}C_2C_1^{-1/2}$. Then $\log_{C_1}(C_2)=C_1^{1/2}(\log X)C_1^{1/2}$ and
$C_1^{-1/2}\,t_{int}\log_{C_1}(C_2)\,C_1^{-1/2}=t_{int}\log X$.
Hence
\begin{equation}
\exp_{C_1}(t_{int}\log_{C_1}(C_2))
=
C_1^{1/2}\exp(t_{int}\log X)C_1^{1/2}
=
C_1^{1/2}X^{t_{int}}C_1^{1/2}.
\end{equation}

\subsection{Isometry at the Identity and Tangent-Space Representation}
\label{app:isometry}

We recall the AIRM inner product
\begin{equation}
\langle U,V\rangle_C := \mathrm{tr}(C^{-1} U C^{-1} V),
\label{eq:airm_inner_recall}
\end{equation}
the AIRM distance
\begin{equation}
d(C_1,C_2)
=
\big\|\log(C_1^{-1/2} C_2 C_1^{-1/2})\big\|_F,
\label{eq:airm_dist_recall}
\end{equation}
and the logarithm map
\begin{equation}
\log_C(C')
=
C^{1/2}\log(C^{-1/2} C' C^{-1/2})C^{1/2}.
\label{eq:log_map_recall}
\end{equation}

\textbf{Lemma 13 (metric at identity equals Frobenius).}
At $I$, the AIRM inner product reduces to Frobenius:
\begin{equation}
\langle U,V\rangle_I = \mathrm{tr}(UV),
\qquad
\|U\|_I=\|U\|_F.
\end{equation}

\textit{Proof.}
Substitute $C=I$ in \eqref{eq:airm_inner_recall} to obtain
$\langle U,V\rangle_I=\mathrm{tr}(UV)$; the norm identity follows directly.

\textbf{Lemma 14 (distance to identity equals log-norm).}
For any $C\in\mathcal{S}_{++}^d$,
\begin{equation}
d_{\mathrm{AIRM}}(I,C)=\|\log(C)\|_F.
\label{eq:dist_identity}
\end{equation}

\textit{Proof.}
Setting $C_1=I$ in \eqref{eq:airm_dist_recall} gives
$d_{\mathrm{AIRM}}(I,C)=\|\log(I^{-1/2} C I^{-1/2})\|_F=\|\log(C)\|_F$.

\textbf{Lemma 15 (whitening yields identity-based tangent isometry).}
For any $C_1,C_2\in\mathcal{S}_{++}^d$, define
$\widetilde{C}_2=C_1^{-1/2}C_2C_1^{-1/2}$.
Then
\begin{equation}
d_{\mathrm{AIRM}}(C_1,C_2)
=
\|\log(\widetilde{C}_2)\|_F
=
\|\,C_1^{-1/2}\log_{C_1}(C_2)C_1^{-1/2}\,\|_F.
\label{eq:isometry_whiten}
\end{equation}

\textit{Proof.}
The first equality follows directly from \eqref{eq:airm_dist_recall}.
From \eqref{eq:log_map_recall},
$C_1^{-1/2}\log_{C_1}(C_2)C_1^{-1/2}=\log(\widetilde{C}_2)$, yielding the second.

\textbf{Interpretation used in the pipeline.}
Equations \eqref{eq:dist_identity}–\eqref{eq:isometry_whiten} show that,
after whitening by a reference covariance, the AIRM distance reduces to a
Frobenius norm in the tangent space at the identity—exactly the form exploited
by RA and tangent-space classifiers.

\subsection{Riemannian Alignment as Reference Whitening}
\label{app:ra}

Riemannian Alignment (RA) is a reference-based congruence normalization.
Let $C_{\mathrm{ref}}\in\mathcal{S}_{++}^d$ be a chosen reference covariance (e.g., a per-subject or global mean).
We seek a linear transform $P$ such that the reference becomes identity:
\begin{equation}
P\,C_{\mathrm{ref}}\,P^\top = I.
\label{eq:ra_constraint}
\end{equation}

\textbf{Lemma 16 (unique SPD whitening solution).}
Among SPD transforms $P\succ 0$, the unique solution to \eqref{eq:ra_constraint} is
\begin{equation}
P = C_{\mathrm{ref}}^{-1/2}.
\label{eq:ra_P}
\end{equation}

\textit{Proof.}
Let $C_{\mathrm{ref}}=U\Lambda U^\top$. Choose $P=U\Lambda^{-1/2}U^\top=C_{\mathrm{ref}}^{-1/2}$, then
$PC_{\mathrm{ref}}P^\top = U\Lambda^{-1/2}\Lambda\Lambda^{-1/2}U^\top=I$, so it is a solution.
For uniqueness over SPD $P$: write $P=U S U^\top$ with $S\succ 0$. Then
$PC_{\mathrm{ref}}P^\top = U S \Lambda S U^\top=I \Rightarrow S\Lambda S=I$.
Since $\Lambda$ is diagonal with positive entries, the unique SPD solution is $S=\Lambda^{-1/2}$, hence $P=C_{\mathrm{ref}}^{-1/2}$. 

\textbf{RA mapping for any trial covariance.}
Given any $C\in\mathcal{S}_{++}^d$, define the aligned covariance
\begin{equation}
C' = C_{\mathrm{ref}}^{-1/2}\, C \, C_{\mathrm{ref}}^{-1/2}.
\label{eq:ra_map}
\end{equation}
By Lemma 3, $C'\in\mathcal{S}_{++}^d$. Moreover, by Lemma 5 and Lemma 12,
\begin{equation}
d_{\mathrm{AIRM}}\big(C_{\mathrm{ref}}, C\big)
=
d_{\mathrm{AIRM}}\big(I,\, C'\big)
=
\|\log(C')\|_F,
\end{equation}
i.e., RA converts distances to the reference into identity-based tangent norms.

\subsection{Tangent-Space Scaling and Geodesic Consistency}
\label{app:tangent_scaling}

The scaling/linear operations are done in tangent space and then maps back to $\mathcal{S}_{++}^d$.

\textbf{Lemma 17 (geodesic interpolation equals tangent scaling).}
Let $C_1,C_2\in\mathcal{S}_{++}^d$, and let $Z=\log_{C_1}(C_2)$. Then
\begin{equation}
\exp_{C_1}(t_{int}Z) = \gamma(t_{int}),
\end{equation}
where $\gamma(t_{int})$ is the AIRM geodesic in \eqref{eq:geodesic}. Thus, scaling by $t_{int}$ in $T_{C_1}\mathcal{S}_{++}^d$ corresponds to moving fraction $t_{int}$ along the geodesic.

\textit{Proof.}
Immediate from Lemma 10 and the definition $\gamma(t_{int})=\exp_{C_1}(t_{int}\log_{C_1}(C_2))$. 

\textbf{Corollary (identity case).}
At $C_1=I$, $Z=\log(C_2)$ and $\exp_I(t_{int}\log(C_2))=\exp(t_{int}\log(C_2))=C_2^{t_{int}}$.
Hence geodesics through identity are power curves.

\subsection{Mathematical Foundations for Geometry-Aware Learning}
\label{app:scope}
This section establishes the geometric and analytical foundations underlying all Riemannian operations employed in the paper. We formalize the structure of the SPD manifold, characterize its affine-invariant Riemannian metric, and derive the associated geodesic distance together with its invariance, symmetry, and identity properties. The logarithmic and exponential maps are introduced and shown to induce an isometric correspondence between geodesic distances on $\mathcal{S}_{++}^d$ and Frobenius norms in the tangent space, both at the identity and under reference-based whitening.

These results provide a rigorous justification for congruence transformations, Riemannian alignment via reference normalization, and tangent-space scaling used throughout the paper. All subsequent alignment, normalization, and congruence-based learning components operate within this established mathematical framework, without invoking any additional assumptions beyond the properties derived here.
\newpage
\section{Classical Preprocessing Methods}

We review classical preprocessing and alignment techniques commonly used in
motor-imagery BCI pipelines. These methods operate either directly on covariance
matrices or on Euclidean features derived from them and serve as standard
baselines in Riemannian EEG decoding. Their primary goal is to reduce inter-subject variability by normalizing
subject-specific covariance statistics, enhancing class discriminability, or
projecting data into a common feature space prior to classification. Because
they rely on fixed, non-learned transformations, they have limited ability to
correct complex subject- or class-dependent geometric distortions, especially
in transductive cross-subject settings.

We group classical techniques into (i) \emph{geometric alignment methods}, which
act directly on SPD covariances via congruence transforms, and (ii)
\emph{linear projection methods}, which operate in Euclidean spaces derived from
covariance features, reflecting the distinction between geometry-aware
normalization and discriminative linear embedding.

\subsection{Riemannian Alignment (RA)}
\label{app:rapp}

Riemannian Alignment (RA) is a geometric preprocessing method designed to reduce
inter-subject variability by centering subject-specific covariance
distributions at a common reference point on the manifold of symmetric positive
definite (SPD) matrices \cite{yger2017riemannian}. RA operates via congruence
transformations and therefore preserves SPD structure as well as
affine-invariant Riemannian geometry.
\begin{figure}[h]
    \centering
    \includegraphics[width=0.5\linewidth]{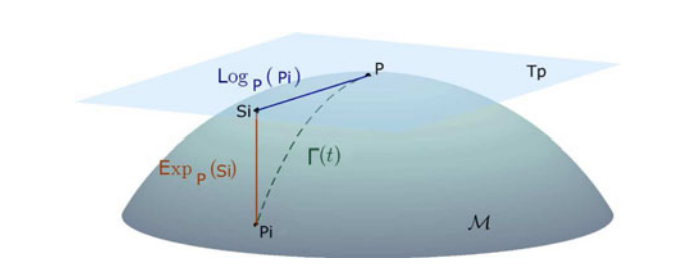}
    \caption{Tangent-space Log/Exp map intuition}
    \label{fig:riemannian_manifold_ts}
\end{figure}
\paragraph{Subject-wise reference.}
Let $C_{i,e} \in \mathcal{S}_{++}^d$ denote the covariance matrix
associated with subject $i$. Due to subject-specific physiological and sensor
differences, covariance distributions are typically centered around distinct
locations on the SPD manifold. RA addresses this by computing a subject-wise
reference covariance $C_i^{\mathrm{ref}}$, commonly chosen as the
Riemannian (Fr\'echet) mean:
\begin{equation}
C_i^{\mathrm{ref}}
=
\arg\min_{C \in \mathcal{S}_{++}^d}
\sum_{i,e}
d_{\mathrm{AIRM}}^2\!\left(C,\; C_{i,e}\right),
\label{eq:ra_ref}
\end{equation}
where $d_{\mathrm{AIRM}}(\cdot,\cdot)$ denotes the affine-invariant Riemannian
distance. This reference captures the central tendency of the subject-specific
covariance distribution on the manifold.

\paragraph{Whitening transform.}
Given $C_{i}^{\mathrm{ref}}$, RA seeks a congruence transform $P_i$ such
that
\begin{equation}
P_i C_{i}^{\mathrm{ref}} P_i^{\top} = I.
\label{eq:ra_whiten_cond}
\end{equation}
The unique symmetric positive definite solution to
\eqref{eq:ra_whiten_cond} is
\begin{equation}
P_i = \big(C_{i}^{\mathrm{ref}}\big)^{-1/2},
\label{eq:ra_whiten}
\end{equation}
which follows directly from the spectral decomposition of
$C_{i}^{\mathrm{ref}}$. Applying this transform to each covariance matrix
yields the aligned covariances
\begin{equation}
C_{i,e}'
=
\big(C_{i}^{\mathrm{ref}}\big)^{-1/2}
\, C_{i,e} \,
\big(C_{i}^{\mathrm{ref}}\big)^{-1/2}.
\label{eq:ra_align}
\end{equation}

\paragraph{Geometric effect.}
Under the affine-invariant Riemannian metric, congruence transformations of the
form $C \mapsto P C P^\top$ are isometries. Consequently, RA maps the
subject-wise Riemannian mean exactly to the identity matrix,
\[
\mu_{i}^{\mathrm{RA}} = I \qquad \forall\,i,
\]
while preserving geodesic distances between covariances within each subject.
After alignment, covariance distributions from different subjects are therefore
centered at a shared geometric reference point, substantially reducing
inter-subject mean shifts.

\paragraph{Limitation.}
RA performs a first-order geometric normalization by removing subject-specific
mean offsets but does not correct orientation differences, anisotropic
dispersion, or class-dependent geometric distortions. As a result, residual
misalignment persists in transductive cross-subject settings, motivating more
expressive alignment strategies.

\subsection{Riemannian Procrustes Analysis (RPA)}
\label{app:rpa}

Riemannian Procrustes Analysis (RPA) is a geometric alignment method designed to
reduce inter-subject variability by explicitly matching covariance
distributions across domains on the manifold of symmetric positive definite
(SPD) matrices \cite{yger2017riemannian}. Unlike Riemannian Alignment (RA), which
removes only subject-wise mean shifts, RPA performs a sequence of geometric
operations—recentering, dispersion scaling, and rotation—to align higher-order
distributional structure.

RPA is fundamentally a \emph{transfer learning} method: it aligns a target
(subject or domain) covariance distribution to a source (reference) domain
using statistics estimated across domains. This explicit source–target
coupling makes RPA more expressive than RA, but also restricts its applicability
to settings where both domains are available during alignment.
\begin{figure}[h]
    \centering
    \includegraphics[width=0.65\linewidth]{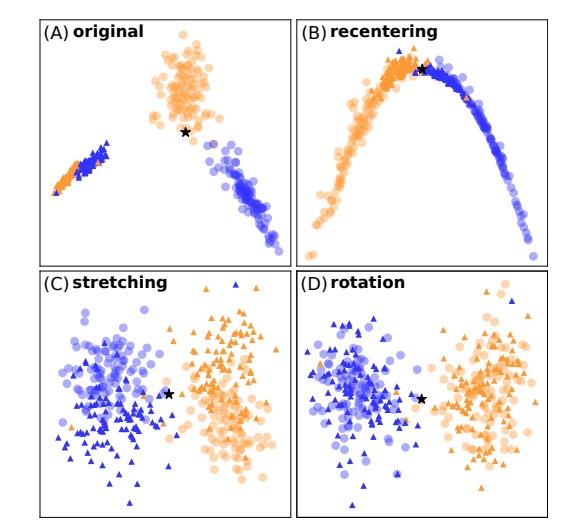}
    \caption{Representation of sequence of operations in RPA re-center $\rightarrow$ scale $\rightarrow$ rotate.}
    \label{fig:rpa}
\end{figure}
\paragraph{Problem setting.}
Let $\mathcal{C}_{src} = \{ C_{i,e} \mid i \in \mathcal{S},\ e \in \mathcal{E}_i \}$
denote the covariance matrices from the source subjects, and
$\mathcal{C}_{tgt} = \{ C_{i,e} \mid i \in \mathcal{T},\ e \in \mathcal{E}_i \}$
those from the target subject(s).
from a target domain. The goal of RPA is to transform target covariances such
that their distribution matches the source distribution in terms of location,
dispersion, and dominant orientation on the SPD manifold.

\paragraph{Geometric alignment steps.}
RPA proceeds through the following sequence of operations:

\emph{(i) Re-centering.}  
Both source and target covariances are first centered at the identity via
Riemannian Alignment:
\begin{equation}
C' = (C^{\mathrm{ref}})^{-1/2} \, C \, (C^{\mathrm{ref}})^{-1/2},
\end{equation}
where $C^{\mathrm{ref}}$ is the Riemannian mean of the corresponding domain.

\emph{(ii) Dispersion scaling.}  
Let $\varsigma_{src}^2$ and $\varsigma_{tgt}^2$ denote the average squared AIRM distance
of source and target covariances from the identity after recentering:
\begin{equation}
\varsigma_{src}^2 = \frac{1}{|S|}\sum_{C_{src}'} d_{\mathrm{AIRM}}^2(C_{src}', I),
\qquad
\varsigma_t^2 = \frac{1}{|T|}\sum_{C_{tgt}'} d_{\mathrm{AIRM}}^2(C_{tgt}', I).
\end{equation}
A scaling factor $\rho = \sqrt{\varsigma_{src}^2 / \varsigma_{tgt}^2}$ is used to match
target dispersion by scaling in the tangent space:
\begin{equation}
Z = \log(C_{tgt}'), \qquad Z' = \rho Z, \qquad \widetilde{C}_{tgt} = \exp(Z').
\end{equation}

\emph{(iii) Rotation alignment.}  
To align dominant orientations, class-wise means are computed for source and
target domains. Let $\widetilde{C}_{src,k}$ and $\widetilde{C}_{tgt,k}$ denote the
scaled class means, and define their tangent representations
$L_{src,k} = \log(\widetilde{C}_{src,k})$ and $L_{tgt,k} = \log(\widetilde{C}_{tgt,k})$.
An orthogonal rotation $U$ is obtained by solving
\begin{equation}
U
=
\arg\min_{U \in SO(C)}
\sum_k \big\| U^\top L_{tgt,k} U - L_{src,k} \big\|_F^2.
\end{equation}
The final aligned target covariances are then given by
\begin{equation}
C_{tgt}^{\mathrm{out}} = U^\top \, \widetilde{C}_{tgt} \, U.
\end{equation}

\begin{algorithm}[t]
\caption{Riemannian Procrustes Analysis (RPA)}
\label{alg:rpa}
\begin{algorithmic}
\setlength{\baselineskip}{11pt}

\STATE \textbf{Input:} Source covariances $\{C_{src}\}$, target covariances $\{C_{tgt}\}$

\STATE \textbf{Re-center:}
\STATE \quad Compute domain-wise Riemannian means
\STATE \quad Apply RA to obtain centered covariances $\{C_{src}'\}, \{C_{tgt}'\}$

\STATE \textbf{Scale:}
\STATE \quad Compute dispersions $\varsigma_{src}^2, \varsigma_{tgt}^2$
\STATE \quad Set $\rho = \sqrt{\varsigma_{src}^2 / \varsigma_{tgt}^2}$
\STATE \quad Log–scale target covariances in tangent space

\STATE \textbf{Rotate:}
\STATE \quad Compute class-wise tangent means $T_{src,k}, T_{tgt,k}$
\STATE \quad Solve orthogonal Procrustes problem for $U$
\STATE \quad Rotate target covariances: $C_{tgt}^{\mathrm{out}} = U^\top \widetilde{C}_{tgt} U$

\STATE \textbf{Output:} Aligned target covariances $\{C_{tgt}^{\mathrm{out}}\}$

\end{algorithmic}
\end{algorithm}

\paragraph{Limitation.}
Although RPA reduces global subject-level variation more effectively than RA, it
is a static, non-learned alignment method that explicitly depends on both source
and target distributions. Its reliance on domain-level statistics and
class-wise correspondence limits applicability in transductive settings and
prevents adaptation to fine-grained or class-dependent geometric distortions.
These limitations motivate the need for \emph{learned, class-discriminative
alignment mechanisms}, such as the proposed DCR pre-aligner.

\paragraph{Note on RPA and Experimental Protocol.}
Riemannian Procrustes Analysis (RPA) operates under a transfer-learning paradigm, requiring access to target-domain covariances and class-wise correspondence to estimate alignment transforms. In contrast, all experiments in this paper adopt a strictly transductive cross-subject protocol in which no target-subject labels or statistics are available at test time. Because RPA explicitly relies on target-domain information, it is therefore incompatible with our evaluation setting and is not included in the ablation studies.

\paragraph{Linear projection methods.}
We next describe classical preprocessing techniques that operate in Euclidean
feature spaces derived from covariance representations. Unlike geometric
alignment methods, which act directly on the SPD manifold, linear projection
methods first embed covariance structure into vector spaces—either through
spatial filtering or tangent-space representations—and then apply linear
discriminative projections to enhance class separability.

These methods are typically supervised and optimize class-wise variance
criteria or Fisher-style objectives. While effective in subject-dependent
settings, they ignore intrinsic manifold geometry and are highly sensitive to
inter-subject covariance shifts. As a result, their performance degrades
substantially in transductive cross-subject decoding, making them important but
limited baselines.

\subsection{Common Spatial Patterns (CSP)}
\label{app:csp}

Common Spatial Patterns (CSP) is a classical supervised spatial filtering method
that learns linear projections maximizing class-discriminative variance
structure in multichannel EEG signals. In this work, CSP is used purely as a
preprocessing and feature-extraction baseline and is evaluated under a strict
\emph{transductive cross-subject} (leave-one-subject-out, LOSO) protocol.
\begin{figure}[h]
    \centering
    \includegraphics[width=0.75\linewidth]{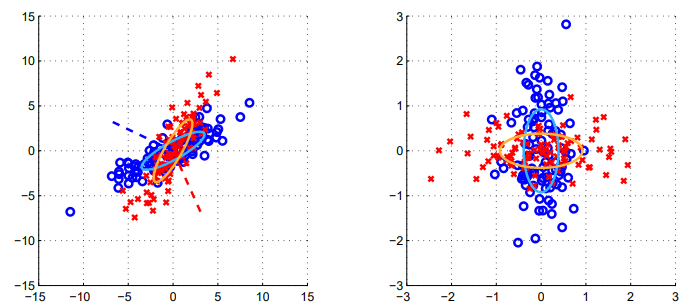}
    \caption{Common Spatial Patterns (CSP): Before filtering (left) the two Gaussian classes have
overlapping correlated structure.}
\end{figure}

\paragraph{Transductive source--target protocol.}
For each LOSO fold, the dataset is partitioned into a \emph{source domain}
consisting of EEG trials from all training subjects and a \emph{target domain}
consisting of trials from a single held-out subject. All CSP spatial filters are
learned exclusively from source-domain data. The learned filters are applied
\emph{without any adaptation} to the target subject. No target-subject labels,
covariances, or statistics are used at any stage.

\paragraph{Trial covariance estimation (source domain).}
Let $X_{i,e}^{(s)} \in \mathbb{R}^{d \times T}$ denote the EEG signal from subject $i$ trial
$e$ of class $y_{i,e}$ recorded from a source subject. Trial-wise normalized
covariances are computed as
\begin{equation}
C_{i,e}^{(s)}
=
\frac{X_{i,e}^{(s)} X_{i,e}^{(s)\top}}
     {\mathrm{tr}\!\left(X_{i,e}^{(s)} X_{i,e}^{(s)\top}\right)}
\;\in\;\mathcal{S}_{++}^d.
\label{eq:csp_trial_cov}
\end{equation}

\paragraph{Class-wise and one-vs-rest covariances (source domain).}
Let $N_k^{(s)}$ denote the number of source-domain trials in class $k$. The
class-mean covariance is
\begin{equation}
C_k^{(s)}
=
\frac{1}{N_k^{(s)}} \sum_{\substack{i,e \\ y_{i,e} = k}} C_{i,e}^{(s)}.
\label{eq:csp_class_cov}
\end{equation}

\begin{algorithm}[t]
\caption{Common Spatial Patterns (CSP) under Transductive LOSO (Multi-class OvR)}
\label{alg:csp}
\begin{algorithmic}
\setlength{\baselineskip}{11pt}

\STATE \textbf{Input:} Let EEG trials $X_{i,e}^{(s)}$ be from training (source) subjects and Let $X_{i,e}^{(t)}$ be from held-out (test/target) subject
\STATE \textbf{Hyperparameter:} number of filters per extreme $m$

\STATE Compute source trial covariances $C_{i,e}^{(s)}$ using \eqref{eq:csp_trial_cov}
\STATE Compute source class-mean covariances $C_k^{(s)}$ using \eqref{eq:csp_class_cov}

\FOR{each class $k$}
    \STATE Compute source one-vs-rest covariance $C_{-k}^{(s)}$ using \eqref{eq:csp_ovr}
    \STATE Form whitened matrix $S_k = C_{-k}^{(s)-1/2} C_k^{(s)} C_{-k}^{(s)-1/2}$
    \STATE Eigendecompose $S_k v = \lambda v$
    \STATE Construct CSP filters $W_{\mathrm{CSP}}^{(k)}$ using \eqref{eq:csp_filters}
\ENDFOR

\STATE Apply fixed filters $W_{\mathrm{CSP}}^{(k)}$ to target trials $X_i^{(t)}$
\STATE Compute log-variance features for classification

\STATE \textbf{Output:} CSP feature vectors for target subject

\end{algorithmic}
\end{algorithm}

For multi-class CSP, the one-vs-rest covariance is
\begin{equation}
C_{-k}^{(s)}
=
\frac{1}{\sum_{j\neq k} N_j^{(s)}}
\sum_{j\neq k} \sum_{\substack{i,e \\ y_{i,e} \neq j}} C_{i,e}^{(s)}.
\label{eq:csp_ovr}
\end{equation}

\paragraph{Rayleigh quotient and generalized eigenproblem.}
For each class $k$, CSP seeks spatial filters $w \in \mathbb{R}^d$ that maximize
\begin{equation}
J_k(w)
=
\frac{w^\top C_k^{(s)} w}{w^\top C_{-k}^{(s)} w},
\label{eq:csp_rayleigh}
\end{equation}
leading to the generalized eigenvalue problem
\begin{equation}
C_k^{(s)} w = \lambda\, C_{-k}^{(s)} w.
\label{eq:csp_gevp}
\end{equation}

\paragraph{Whitening-based solution.}
The generalized eigenproblem is solved by whitening with respect to
$C_{-k}^{(s)}$:
\begin{equation}
S_k
=
C_{-k}^{(s)-1/2} C_k^{(s)} C_{-k}^{(s)-1/2},
\qquad
S_k v = \lambda v.
\label{eq:csp_whitened}
\end{equation}
The spatial filters are recovered as $w = C_{-k}^{(s)-1/2} v$.

\paragraph{Filter selection.}
Let $\{(v_j,\lambda_j)\}_{j=1}^d$ be the eigenpairs of $S_k$ sorted in descending
order. CSP selects the most discriminative directions from both extremes:
\begin{equation}
W_{\mathrm{CSP}}^{(k)}
=
\left[
w_1,\dots,w_m,\;
w_{d-m+1},\dots,w_d
\right],
\qquad
w_j = C_{-k}^{(s)-1/2} v_j,
\label{eq:csp_filters}
\end{equation}

\paragraph{Transductive feature extraction on the target subject.}
Let $X_i^{(t)} \in \mathbb{R}^{C \times T}$ denote a trial from the held-out
target subject. CSP filters learned from the source domain are applied directly:
\begin{equation}
Z_i^{(k,t)}
=
W_{\mathrm{CSP}}^{(k)\top} X_i^{(t)}
\;\in\;\mathbb{R}^{2m \times T}.
\label{eq:csp_project}
\end{equation}
Log-variance features are computed as
\begin{equation}
f_i^{(k,t)}(c)
=
\log
\frac{\mathrm{var}(Z_i^{(k,t)}(c,:))}
     {\sum_{c'} \mathrm{var}(Z_i^{(k,t)}(c',:))},
\qquad c=1,\dots,2m.
\label{eq:csp_logvar}
\end{equation}

\paragraph{Limitation in transductive cross-subject decoding.}
CSP assumes that discriminative variance directions are shared across subjects.
In practice, EEG covariance structure varies substantially across individuals,
causing CSP filters learned from source subjects to misalign with target-subject
discriminative subspaces. This limitation motivates geometry-aware alignment and
congruence-based methods on the SPD manifold.

\subsection{Linear Discriminant Analysis (LDA) Projection}
\label{app:lda_proj}

Linear Discriminant Analysis (LDA) is a classical supervised linear projection
method that seeks directions maximizing class separability in Euclidean feature
space. In covariance-based EEG pipelines, LDA is typically applied to
vector-valued representations (e.g., tangent-space embeddings obtained after
Riemannian alignment) to obtain low-dimensional discriminative features prior to
classification. In this work, LDA is evaluated under a strict
\emph{transductive cross-subject} (leave-one-subject-out, LOSO) protocol.

\paragraph{Transductive source--target protocol.}
For each LOSO fold, the dataset is partitioned into a \emph{source domain}
consisting of trials from all training subjects and a \emph{target domain}
consisting of trials from a single held-out subject. The LDA projection is
learned exclusively using labeled source-domain features. The learned
projection matrix is then applied \emph{without any adaptation} to the target
subject. No target-subject labels, class statistics, or distributional
information are used at any stage, ensuring transductive cross subject setting.

\begin{figure}[h]
    \centering
    \includegraphics[width=0.6\linewidth]{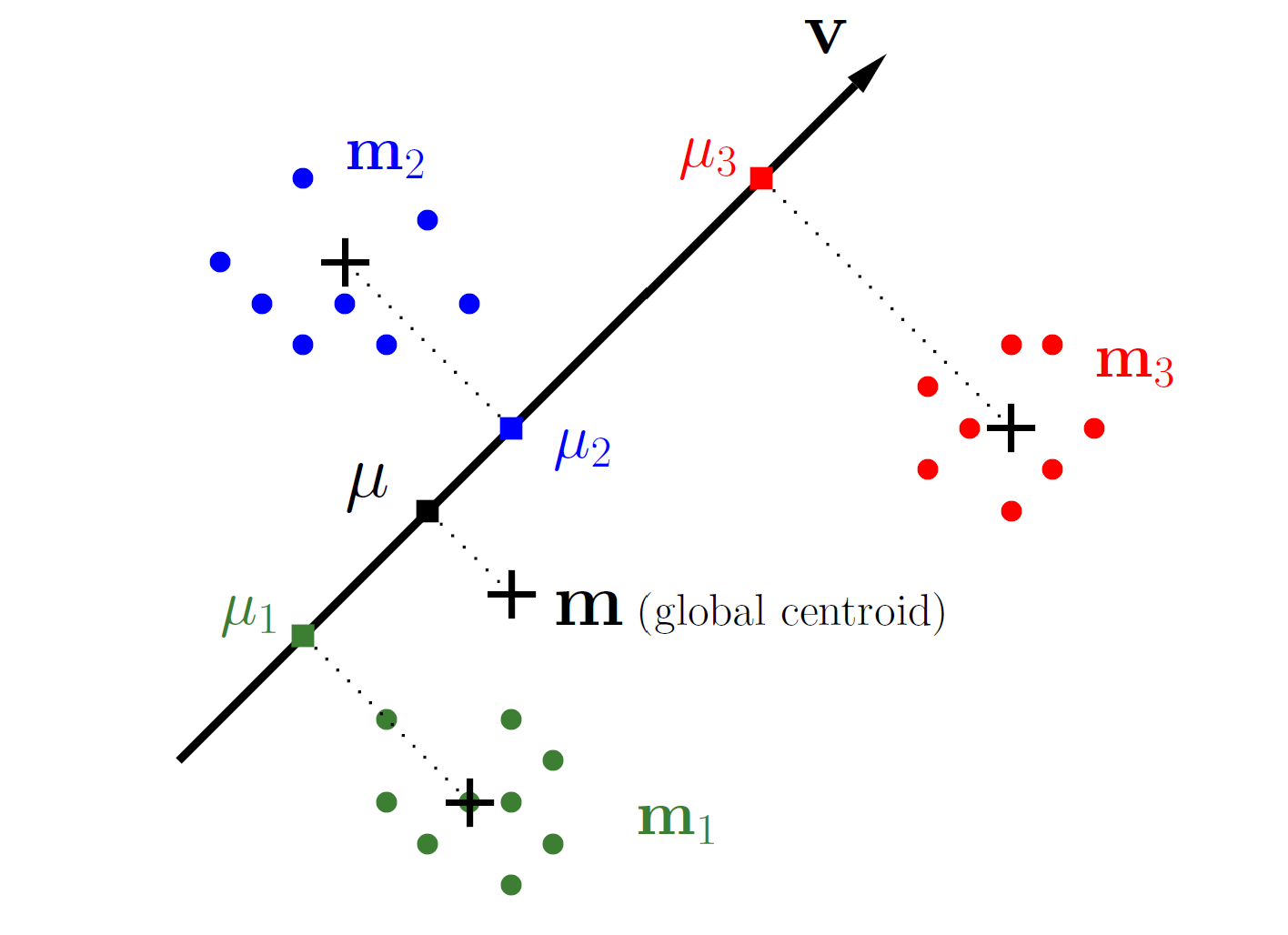}
    \caption{LDA visualization showing class centroids $m_1, m_2, m_3$, their projected means $\mu_1, \mu_2, \mu_3$, and the discriminant direction $v$ along which between-class variance is maximized relative to within-class variance.}
\end{figure}

\paragraph{Input representation (source domain).}
Let $z_{i,e}^{(s)} \in \mathbb{R}^d$ denote a feature vector associated with trial
$e$ from a source subject, obtained by vectorizing or embedding covariance
matrices into a Euclidean space (e.g., tangent space at the identity). Let
$y_{i,e}^{(s)} \in \{1,\dots,K\}$ be the corresponding class label, and let
$\mathcal{C}_k^{(s)} = \{(i,e) \ ; y_{i,e} = k\}$ be index set for class $k$

\paragraph{Class statistics (source domain).}
LDA computes class-wise means and the global mean using only source-domain data:
\begin{equation}
\mu_k^{(s)}
=
\frac{1}{|\mathcal{C}_k^{(s)}|}
\sum_{\mathcal{C}_k^{(s)}} z_{i,e}^{(s)},
\qquad
\mu^{(s)}
=
\frac{1}{N^{(s)}} \sum_{e=1}^{N^{(s)}} z_{i,e}^{(s)},
\label{eq:lda_means}
\end{equation}
where $N^{(s)}$ is the total number of source-domain samples.

\paragraph{Scatter matrices (source domain).}
Class separability is quantified through the within-class and between-class
scatter matrices:
\begin{equation}
S_W^{(s)}
=
\sum_{k=1}^{K} \sum_{\mathcal{C}_k^{(s)}}
\left(z_i^{(s)} - \mu_k^{(s)}\right)
\left(z_i^{(s)} - \mu_k^{(s)}\right)^\top,
\qquad
S_B^{(s)}
=
\sum_{k=1}^{K} |\mathcal{C}_k^{(s)}|
\left(\mu_k^{(s)} - \mu^{(s)}\right)
\left(\mu_k^{(s)} - \mu^{(s)}\right)^\top
\label{eq:lda_scatters}
\end{equation}
These matrices encode intra-class variability and inter-class separation
estimated solely from source subjects.

\paragraph{Fisher discriminant objective.}
LDA seeks projection directions $v \in \mathbb{R}^d$ that maximize the Fisher
ratio
\begin{equation}
J(v)
=
\frac{v^\top S_B^{(s)} v}{v^\top S_W^{(s)} v},
\label{eq:lda_fisher}
\end{equation}
leading to the generalized eigenvalue problem
\begin{equation}
S_B^{(s)} v = \lambda\, S_W^{(s)} v.
\label{eq:lda_gevp}
\end{equation}
For a $K$-class problem, at most $K-1$ non-zero generalized eigenvalues exist,
yielding a $(K-1)$-dimensional discriminative subspace.

\paragraph{Transductive projection of target features.}
Let $V = [v_1,\dots,v_{K-1}]$ denote the eigenvectors corresponding to the largest
generalized eigenvalues of \eqref{eq:lda_gevp}. Given a target-domain feature
vector $z_{i,e}^{(t)} \in \mathbb{R}^d$ from the held-out subject, the learned
projection is applied directly:
\begin{equation}
z_{i,e}^{\mathrm{LDA}(t)} = V^\top z_{i,e}^{(t)} \in \mathbb{R}^{K-1}.
\label{eq:lda_projection}
\end{equation}

\begin{algorithm}[t]
\caption{Linear Discriminant Analysis (LDA)}
\label{alg:lda}
\begin{algorithmic}
\setlength{\baselineskip}{11pt}

\STATE \textbf{Input:} Let Feature vectors $z_{i,e}^{(s)} \in \mathbb{R}^d$ with labels $y_{i,e}^{(s)}$ be the training (source) data and $z_{i,e}^{(t)} \in \mathbb{R}^d$ be the test (target) data
\STATE \textbf{Output dimension:} $m = K-1$

\STATE Compute source class means $\mu_k^{(s)}$ and global mean $\mu^{(s)}$ using \eqref{eq:lda_means}
\STATE Compute source scatter matrices $S_W^{(s)}$ and $S_B^{(s)}$ using \eqref{eq:lda_scatters}
\STATE Solve generalized eigenproblem $S_B^{(s)} v = \lambda S_W^{(s)} v$
\STATE Select top $m$ eigenvectors; form projection matrix $V$

\STATE Apply fixed projection to target features: $z_{i,e}^{\mathrm{LDA}(t)} = V^\top z_{i,e}^{(t)}$ using \eqref{eq:lda_projection}

\STATE \textbf{Return:} LDA features $z_{i,e}^{\mathrm{LDA}(t)}$ and projection $V$

\end{algorithmic}
\end{algorithm}

\paragraph{Limitation in Transductive cross-subject decoding.}
Although LDA operates in a transductive setting, it is not subject-invariant: it
assumes shared class-conditional distributions and stable within-class
covariances across subjects. Inter-subject variability instead induces
distributional shifts in tangent-space features, misaligning source and target
class statistics and degrading transductive generalization, which motivates
geometry-aware alignment and congruence-based alternatives.

\paragraph{Summary of classical preprocessing methods.}
The classical methods above address inter-subject variability and class
discrimination from complementary perspectives. Geometric alignment operates on
the SPD manifold and preserves Riemannian structure, whereas linear projections
optimize class separability in Euclidean feature spaces. Table~\ref{tab:classical_summary}
summarizes their key properties for cross-subject EEG decoding. Classical approaches satisfy only subsets of the requirements for robust
cross-subject decoding: RA aligns means but is class-agnostic; RPA aligns higher-
order structure but relies on target statistics; CSP and LDA optimize
discrimination yet ignore intrinsic geometry and remain sensitive to
inter-subject shifts. In contrast, DCT, DLDCT, and DDCT-UNet combine geometry-
aware operations with learnable discriminative transforms under transductive
constraints, motivating the proposed framework.

\begin{table}[h]
\centering
\footnotesize
\renewcommand{\arraystretch}{1.15}
\caption{Comparison of classical preprocessing methods for covariance-based EEG decoding.}
\label{tab:classical_summary}
\begin{tabular}{lcccccc}
\toprule
Method & Geometry-aware & Class-aware & Learned & Deep & Non-Linear & Transductive \\
\midrule
Riemannian Alignment (RA)        & \checkmark & $\times$ & $\times$ & $\times$ & $\times$ & \checkmark \\
Riemannian Procrustes (RPA)     & \checkmark & $\times$ & $\times$ & $\times$ & $\times$\\
Common Spatial Patterns (CSP)   & $\times$   & \checkmark & $\times$ & $\times$ & $\times$ & $\times$\\
LDA Projection                  & $\times$   & \checkmark & $\times$ & $\times$ & $\times$ & $\times$\\
\midrule
\textbf{DCT (Ours)}             & \checkmark & \checkmark & \checkmark & $\times$ & $\times$ & \checkmark \\
\textbf{DLDCT (Ours)}            & \checkmark & \checkmark & \checkmark & \checkmark & $\times$ & \checkmark \\
\textbf{DDCT-UNet (Ours)}            & \checkmark & \checkmark & \checkmark & \checkmark & \checkmark & \checkmark \\
\bottomrule
\end{tabular}
\end{table}

\newpage
\section{Classical Classifier Methods}
\label{app:classical_classifiers}

In addition to geometry-aware alignment and deep congruence models, we evaluate a
set of widely used \emph{classical classifiers} that operate on covariance-based
or Euclidean feature representations. These methods form the backbone of
traditional motor-imagery EEG decoding pipelines and serve as strong,
well-understood baselines for transductive cross-subject evaluation.

The considered classifiers can be broadly categorized into three families:
(i) \emph{Riemannian distance-based classifiers} that operate directly on the
SPD manifold of covariance matrices, (ii) \emph{Euclidean discriminant-based
classifiers} applied to vectorized or tangent-space embeddings, and (iii)
\emph{hybrid pipelines} that combine spatial filtering or manifold projection
with classical discriminative classifiers.

All classical classifiers are evaluated under the same strict
\emph{transductive leave-one-subject-out (LOSO)} protocol used throughout this
work. For each fold, classifier parameters are estimated exclusively using
source-domain (training-subject) data and are applied without adaptation to the
held-out target subject. No target-subject labels, statistics, or calibration
trials are used at any stage.

Specifically, we consider the following representative methods:

\begin{itemize}
    \item \textbf{Tangent Space Logistic Regression (TSLR):} a Euclidean
    classifier applied to tangent-space embeddings of covariance matrices,
    enabling linear discrimination after manifold linearization.
    
    \item \textbf{Minimum Distance to Mean (MDM):} a Riemannian classifier that
    assigns trials to the class whose Riemannian mean covariance is closest
    under the affine-invariant Riemannian metric.
    
    \item \textbf{Linear Discriminant Analysis (LDA)-based classifiers:}
    discriminant models applied either directly to tangent-space features
    (TSA-LDA) or to spatially filtered representations (e.g., CSP-LDA), which
    assume shared class-conditional structure across subjects.
\end{itemize}

While these methods are computationally efficient and widely adopted, they rely
on implicit assumptions of cross-subject distributional consistency that often
break down in practice. Their performance under transductive cross-subject shifts
provides a meaningful baseline against which the benefits of geometry-aware
alignment and congruence-based learning can be quantified. In the following subsections, we describe each classifier in detail, explicitly
specifying the feature space, training procedure, and transductive application to
unseen target subjects.

\subsection{Tangent Space Logistic Regression (TSLR)}
\label{app:tslr}

Tangent Space Logistic Regression (TSLR) is a classical baseline for
covariance-based EEG decoding that combines Riemannian geometry on the manifold
of symmetric positive definite (SPD) matrices with Euclidean linear
classification. The method proceeds by representing each EEG trial by an SPD
covariance, optionally applying Riemannian Alignment (RA) using a chosen SPD
mean (Riemannian/AIRM, Euclidean, or Log-Euclidean), mapping the resulting SPD
matrices into a Euclidean tangent space via the matrix logarithm, and finally
training a multinomial logistic regression classifier on vectorized tangent
features. TSLR is widely adopted due to its conceptual simplicity, competitive
performance, and computational efficiency for cross-subject motor-imagery
decoding.

\paragraph{Transductive source--target protocol (LOSO).}
TSLR is evaluated under a \emph{transductive leave-one-subject-out (LOSO)}
protocol. For each fold, the \emph{source domain} consists of trials from all
training subjects, while the \emph{target domain} consists of trials from the
held-out subject. All geometric references (subject reference means and global
reference) and classifier parameters are computed \emph{exclusively} from
source-domain data. The learned pipeline is applied to target trials \emph{as
is}, without any adaptation. No target-subject labels, covariances, or
distributional statistics are used at any stage.

\begin{algorithm}[t]
\caption{Tangent Space Logistic Regression (TSLR) under Transductive LOSO}
\label{alg:tslr}
\begin{algorithmic}
\setlength{\baselineskip}{11pt}

\STATE \textbf{Input:} Let source EEG trials be $X_{i,e}$ with labels $y_{i,e}$ and target (test) trials $\widetilde{X}_{i,e}$
\STATE \textbf{Choice of mean:} $\operatorname{Mean}_{\star}$ for RA and $\operatorname{Mean}_{\diamond}$ for tangent base point
\STATE \textbf{Hyperparameter:} ridge weight $\lambda\ge 0$

\STATE Compute source covariances $C_{i,e}$ using \eqref{eq:cov_def}

\STATE Compute source subject references $C_{i}^{\mathrm{ref}}$ using \eqref{eq:ref_subject_mean},
       with $\operatorname{Mean}_{\star}\in\{\operatorname{Mean}_{\mathrm{E}},\operatorname{Mean}_{\mathrm{LE}},\operatorname{Mean}_{\mathrm{R}}\}$ from
       \eqref{eq:mean_euclid}--\eqref{eq:mean_riem}

\STATE Align source covariances $C'_{i,e}$ using \eqref{eq:ra}

\STATE Compute global reference $C_G$ using \eqref{eq:ref_global_mean},
       with $\operatorname{Mean}_{\diamond}\in\{\operatorname{Mean}_{\mathrm{E}},\operatorname{Mean}_{\mathrm{LE}},\operatorname{Mean}_{\mathrm{R}}\}$ from
       \eqref{eq:mean_euclid}--\eqref{eq:mean_riem}

\STATE Project to tangent space: $T_{i,e}=\log_{C_G}(C'_{i,e})$ using \eqref{eq:tangent_proj}

\STATE Vectorize: $x_{i,e}=\operatorname{vec}_{\Delta}(T_{i,e})$ using \eqref{eq:vec_delta}

\STATE Train multinomial logistic regression using \eqref{eq:softmax}--\eqref{eq:tslr_objective}

\STATE Apply the fixed pipeline (covariance $\rightarrow$ RA $\rightarrow$ tangent $\rightarrow$ logistic regression)
       to target trials $\widetilde{X}_{i,e}$ \emph{without estimating any target statistics}

\STATE \textbf{Output:} Predicted target labels and class probabilities

\end{algorithmic}
\end{algorithm}

\paragraph{Trial covariance representation.}
Let $X_{i,e}\in\mathbb{R}^{d\times T}$ denote the EEG signal from subject $i$,
class $y_{i,e}\in\{1,\dots,K\}$, trial $e$, with $d$ channels and $T$ time samples.
Each trial is summarized by an SPD covariance matrix
\begin{equation}
C_{i,e}
=
\frac{1}{T-1} X_{i,e} X_{i,e}^\top
\;\in\;\mathcal{S}_{++}^d.
\label{eq:cov_def}
\end{equation}

\paragraph{SPD mean operators (Euclidean, Log-Euclidean, and Riemannian/AIRM).}
TSLR requires one or more SPD reference means: (i) a \emph{subject reference}
$C_{i}^{\mathrm{ref}}$ for RA, and (ii) a \emph{global reference} $C^{\mathrm{ref}}$ for the
tangent-space projection. These references can be computed using one of the
following mean operators.

\subparagraph{Euclidean mean.}
Given SPD matrices $\{C_n\}_{n=1}^N$, the Euclidean mean is
\begin{equation}
\operatorname{Mean}_{\mathrm{E}}(\{C_n\})
=
\frac{1}{N}\sum_{n=1}^N C_n.
\label{eq:mean_euclid}
\end{equation}
This mean is simple but does not respect SPD geometry.

\subparagraph{Log-Euclidean mean.}
The Log-Euclidean mean is defined by averaging in the log-domain:
\begin{equation}
\operatorname{Mean}_{\mathrm{LE}}(\{C_n\})
=
\exp\!\left(
\frac{1}{N}\sum_{n=1}^N \log(C_n)
\right),
\label{eq:mean_logeuclid}
\end{equation}
where $\log(\cdot)$ and $\exp(\cdot)$ denote the matrix logarithm and matrix
exponential. This construction induces a valid SPD mean while enabling efficient
computation.

\subparagraph{Riemannian (AIRM) mean.}
Under the affine-invariant Riemannian metric (AIRM), the Fr\'echet mean is
defined as
\begin{equation}
\operatorname{Mean}_{\mathrm{R}}(\{C_n\})
=
\arg\min_{C\in\mathcal{S}_{++}^d}
\sum_{n=1}^N d_{\mathrm{AIRM}}^2(C, C_n),
\label{eq:mean_riem}
\end{equation}
where
$d_{\mathrm{AIRM}}(C_1,C_2)=\|\log(C_1^{-1/2}C_2C_1^{-1/2})\|_F$.
In practice, \eqref{eq:mean_riem} is computed via an iterative Karcher-mean
procedure; we treat it as a well-defined operator returning an SPD matrix.

\paragraph{Subject reference mean for Riemannian Alignment (RA).}
For each \emph{source} subject $i$, let $\mathcal{E}_i^{(s)} = \{(i,e) \ ; \ i = s | s \in S\}$ index set the trials
of subject $i$ in the source domain $(S)$. The subject reference mean is computed
using one of the mean operators above:
\begin{equation}
C_{i}^{\mathrm{ref}}
=
\operatorname{Mean}_{\star}\!\left(\{C_{i,e} : (i,e)\in\mathcal{E}_i^{(s)}\}\right),
\qquad
\star\in\{\mathrm{E},\mathrm{LE},\mathrm{R}\},
\label{eq:ref_subject_mean}
\end{equation}
where $\star$ specifies the chosen mean type (Euclidean, Log-Euclidean, or
Riemannian/AIRM). Importantly, $C_{i}^{\mathrm{ref}}$ is computed using
\emph{source data only}.

\paragraph{Riemannian Alignment (RA).}
Each covariance is normalized using the congruence transform induced by the
subject reference:
\begin{equation}
C'_{i,e}
=
\left(C_{i}^{\mathrm{ref}}\right)^{-1/2}\,
C_{i,e}\,
\left(C_{i}^{\mathrm{ref}}\right)^{-1/2}
\;\in\;\mathcal{S}_{++}^d.
\label{eq:ra}
\end{equation}
This operation reduces subject-level bias by (approximately) mapping the
subject-specific mean to the identity.

\paragraph{Global reference mean for tangent-space projection.}
Let $\mathcal{D}^{(s)}=\{C'_{i,e} ; (i,e) \in S\}$ denote the set of aligned source-domain
covariances across all training subjects. The global reference is computed as
\begin{equation}
C_G
=
\operatorname{Mean}_{\diamond}\!\left(\mathcal{D}^{(s)}\right),
\qquad
\diamond\in\{\mathrm{E},\mathrm{LE},\mathrm{R}\}.
\label{eq:ref_global_mean}
\end{equation}
Typically, $\diamond=\mathrm{R}$ (Riemannian mean) is used to define the tangent
base point, but Euclidean or Log-Euclidean alternatives are also valid and are
captured by \eqref{eq:ref_global_mean}.

\paragraph{Tangent-space projection at $C_G$.}
Each aligned covariance is mapped to the tangent space at $C_G$ via the
logarithm map:
\begin{equation}
T_{i,e}
=
\log_{C_G}(C'_{i,e})
=
C_G^{1/2}\,
\log\!\left(
C_G^{-1/2} C'_{i,e} C_G^{-1/2}
\right)
\,C_G^{1/2}
\;\in\;\mathcal{S}^{d}.
\label{eq:tangent_proj}
\end{equation}

\paragraph{Symmetric vectorization.}
The tangent matrix is vectorized into a Euclidean feature vector using a
weighted upper-triangular operator:
\begin{equation}
x_{i,e}
=
\operatorname{vec}_{\Delta}(T_{i,e})
\;\in\;\mathbb{R}^{d_t},
\qquad
d_t=\frac{d(d+1)}{2}.
\label{eq:vec_delta}
\end{equation}
A common choice is to stack the upper-triangular entries and scale off-diagonal
terms by $\sqrt{2}$ to preserve the Frobenius inner product under vectorization.

\paragraph{Multinomial logistic regression.}
Given source-domain features and labels, TSLR trains a multinomial logistic
regression model
\begin{equation}
p(y=k\mid x)
=
\frac{\exp(w_k^\top x_{i,e} + b_k)}
     {\sum_{j=1}^{K} \exp(w_j^\top x_{i,e} + b_j)},
\label{eq:softmax}
\end{equation}
with parameters obtained by minimizing the regularized cross-entropy objective
\begin{equation}
\min_{\{w_k,b_k\}}
\;
-\!\!\sum_{(i,e)\in\mathcal{I}^{(s)}}
\log p\!\left(y_{i,e}\mid x_{i,e}\right)
+
\lambda \sum_{k=1}^K \|w_k\|_2^2,
\label{eq:tslr_objective}
\end{equation}
where $\mathcal{I}^{(s)} = \{(i,e) \ ; \ (i,e) \in S\}$ indexes all source-domain trials.

\paragraph{Computational efficiency.}
TSLR is efficient relative to deep or iterative geometry-aware models. Once
covariances are computed, the dominant per-trial costs are a matrix inverse
square root and matrix logarithm; the classifier is a convex linear model whose
training and inference scale linearly with feature dimension
$d_t=\frac{d(d+1)}{2}$. This makes TSLR well suited for large-scale LOSO
benchmarking and practical BCI pipelines.

\paragraph{Limitation in transductive cross-subject decoding.}
Although TSLR is transductive with respect to the target subject, it is not
subject-invariant. The tangent-space approximation is local around the chosen
reference $C_G$, and the learned linear decision boundaries assume that
class-conditional tangent-space distributions are consistent across subjects.
Inter-subject variability induces nonlinear and class-dependent distortions on
the SPD manifold that are not captured by a single linear separator in a fixed
tangent space, motivating geometry-aware and congruence-based classifiers that
learn hierarchy-preserving transformations on SPD.

\subsection{Minimum Distance to Mean (MDM)}
\label{app:mdm}

Minimum Distance to Mean (MDM) is a classical Riemannian classifier that operates
directly on the manifold of symmetric positive definite (SPD) covariance
matrices. Unlike tangent-space or discriminant-based approaches, MDM performs
classification by assigning each trial to the class whose mean covariance is
closest under a chosen Riemannian distance. Owing to its geometric simplicity
and lack of tunable parameters, MDM is widely used as a strong baseline in
covariance-based EEG decoding.

\paragraph{Transductive source--target protocol (LOSO).}
MDM is evaluated under the same \emph{transductive leave-one-subject-out
(LOSO)} protocol as other classical baselines. For each fold, class-wise mean
covariances are computed exclusively from the \emph{source domain} (training
subjects). At test time, covariance matrices from the held-out target subject
are classified using these fixed class means, without any adaptation, parameter
updates, or use of target-subject labels or statistics.

\paragraph{Trial covariance representation.}
Let $X_{i,e}\in\mathbb{R}^{d\times T}$ denote the EEG signal from subject $i$,
class $y_{i,e}\in\{1,\dots,K\}$, and trial $e$. Each trial is represented by an SPD
covariance matrix
\begin{equation}
C_{i,e}
=
\frac{1}{T-1} X_{i,e} X_{i,e}^\top
\;\in\;\mathcal{S}_{++}^d.
\label{eq:mdm_cov}
\end{equation}

\paragraph{Class-wise SPD means.}
For each class $k$, the class reference covariance
$\mathrm{C}_k^{ref} \in \mathcal{S}_{++}^d$ is computed from source-domain trials
using one of the following mean operators:

\begin{equation}
\mathrm{C}_k^{ref}
=
\begin{cases}
\displaystyle
\frac{1}{N_k}\sum_{(i,e):y_{i,e}=k} C_{i,e},
& \text{if Euclidean mean}, \\

\displaystyle
\arg\min_{C\in\mathcal{S}_{++}^d}
\sum_{(i,e):y_{i,e}=k}
d_{\mathrm{AIRM}}^2(C, C_{i,e}),
& \text{if Riemannian (AIRM) mean},\\

\displaystyle
\exp\!\left(
\frac{1}{N_k}\sum_{(i,e):y_{i,e}=k} \log C_{i,e}
\right),
& \text{if Log-Euclidean mean}, 
\end{cases}
\label{eq:mdm_class_means}
\end{equation}

\paragraph{Distance-based classification.}
Given a covariance matrix $C$ from a target-domain trial, MDM assigns the class
label corresponding to the nearest class mean under the chosen Riemannian
distance:
\begin{equation}
\hat{y}
=
\arg\min_{k}
d_{\mathrm{AIRM}}(C, \mathrm{C}_k^{ref}),
\label{eq:mdm_predict}
\end{equation}
where $d_{\mathrm{AIRM}}(\cdot,\cdot)$ is typically instantiated as the AIRM distance
\eqref{eq:airm_dist}. No feature projection or classifier training is required
beyond the computation of class means.

\begin{algorithm}[t]
\caption{Minimum Distance to Mean (MDM) under Transductive LOSO}
\label{alg:mdm}
\begin{algorithmic}
\setlength{\baselineskip}{11pt}

\STATE \textbf{Input:} Let source (train) EEG trials be $X_{i,a,e}$ with labels $a$ and target (test) trials be $\widetilde{X}_{\tilde{e}}$
\STATE \textbf{Choice of mean:} Euclidean, Log-Euclidean, or Riemannian

\STATE Compute source covariances $C_{i,a,e}$ using \eqref{eq:mdm_cov}

\STATE For each class $k$, compute class mean $\mathrm{Cref}_k$
       using \eqref{eq:mean_euclid}, \eqref{eq:mean_logeuclid}, or
       \eqref{eq:mean_riem}

\STATE For each target trial $\widetilde{X}_{\tilde{e}}$, compute covariance
       $\widetilde{C}_{\tilde{e}}$ using \eqref{eq:mdm_cov}

\STATE Assign label
       $\hat{y}_{\tilde{e}}
       =\arg\min_k d(\widetilde{C}_{\tilde{e}},\mathrm{Cref}_k)$
       using \eqref{eq:airm_dist}

\STATE \textbf{Output:} Predicted target labels $\hat{y}_{\tilde{e}}$

\end{algorithmic}
\end{algorithm}

\paragraph{Computational efficiency.}
MDM is highly efficient, as it requires no explicit classifier training beyond
computing class-wise means. At inference time, each trial requires evaluating
distances to $K$ class means, each involving a matrix logarithm. As a result,
MDM scales well to large datasets and serves as a strong geometry-aware baseline
with minimal computational overhead.

\paragraph{Limitation in transductive cross-subject decoding.}
While MDM fully respects the intrinsic geometry of the SPD manifold, it lacks
any learnable or adaptive transformation to compensate for subject-specific
distortions. The classifier assumes that class-conditional covariance
distributions are well summarized by a single mean per class and that these
means are consistent across subjects. In practice, inter-subject variability
induces shifts in covariance geometry that cannot be corrected by distance-based
assignment alone, leading to degraded performance in transductive cross-subject
settings. This limitation motivates alignment and congruence-based methods that
learn subject-invariant transformations on the SPD manifold.

\subsection{Linear Discriminant Analysis (LDA) Classifier}
\label{app:lda_clf}

When used as a classifier, Linear Discriminant Analysis (LDA) is obtained by
assuming that the class-conditional feature distribution is Gaussian with a
\emph{shared} covariance matrix across classes. In covariance-based EEG
pipelines, LDA is typically applied on top of a Euclidean feature map, e.g.,
CSP log-variance features (CSP\_LDA) or tangent-space vectors from SPD
covariances (TSA\_LDA). We briefly state these upstream feature constructions
and then derive the LDA discriminant function from Bayes decision theory.

\paragraph{Feature inputs in EEG pipelines.}
Let $x_{i,e}\in\mathbb{R}^{d_m}$ denote the Euclidean feature vector fed to LDA:
(i) in \textbf{CSP\_LDA}, $x$ (For convenience we use $x$) is formed by CSP spatial filtering followed by
log-variance feature extraction; (ii) in \textbf{TSA\_LDA}, $x$ is formed by
mapping SPD covariances to a tangent space and vectorizing the resulting
symmetric matrix. In both cases, LDA is applied identically once the Euclidean
feature vector $x$ is obtained.

\paragraph{Generative model assumptions.}
Assume a $K$-class classification problem with labels $y_{i,e}\in\{1,\dots,K\}$ and
class priors $\pi_k = p(y=k)$ (For convenience we assume $y = y_{i,e}$. LDA assumes the class-conditional density
\begin{equation}
p(x\mid y=k)
=
\mathcal{N}(x;\mu_k,\Sigma),
\label{eq:lda_gauss}
\end{equation}
i.e., each class has its own mean $\mu_k\in\mathbb{R}^{d_m}$ but all classes share a
common covariance matrix $\Sigma\in\mathbb{R}^{d_m\times d_m}$.

\paragraph{Bayes classifier.}
The Bayes-optimal classifier under 0--1 loss predicts the class with maximum
posterior probability:
\begin{equation}
\hat{y}(x)
=
\arg\max_{k} \; p(y=k\mid x).
\label{eq:bayes_rule}
\end{equation}
Using Bayes' rule,
\begin{equation}
p(y=k\mid x)
=
\frac{p(x\mid y=k)\,p(y=k)}{p(x)}
=
\frac{p(x\mid y=k)\,\pi_k}{p(x)}.
\label{eq:bayes_posterior}
\end{equation}
Since $p(x)$ does not depend on $k$, maximizing $p(y=k\mid x)$ is equivalent to
maximizing the unnormalized posterior $p(x\mid y=k)\pi_k$, or, more conveniently,
its logarithm:
\begin{equation}
\hat{y}(x)
=
\arg\max_{k}\;\log p(x\mid y=k) + \log \pi_k.
\label{eq:bayes_log}
\end{equation}

\paragraph{Log-likelihood under shared-covariance Gaussians.}
The Gaussian log-likelihood implied by \eqref{eq:lda_gauss} is
\begin{equation}
\log p(x\mid y=k)
=
-\frac{1}{2}(x-\mu_k)^\top \Sigma^{-1}(x-\mu_k)
-\frac{d}{2}\log(2\pi)
-\frac{1}{2}\log|\Sigma|.
\label{eq:gauss_loglik}
\end{equation}
Substituting \eqref{eq:gauss_loglik} into \eqref{eq:bayes_log}, and dropping
terms independent of $k$ (since they do not affect the $\arg\max$), yields the
discriminant score
\begin{equation}
\delta_k(x)
=
-\frac{1}{2}(x-\mu_k)^\top \Sigma^{-1}(x-\mu_k)
+\log \pi_k
\;+\;\text{const}(x),
\label{eq:lda_disc_quad}
\end{equation}
where $\text{const}(x)$ collects all terms independent of $k$.

\paragraph{Linear discriminant form.}
Expanding the quadratic term in \eqref{eq:lda_disc_quad} gives
\begin{align}
-\frac{1}{2}(x-\mu_k)^\top \Sigma^{-1}(x-\mu_k)
&=
-\frac{1}{2}\left(
x^\top\Sigma^{-1}x
-2x^\top\Sigma^{-1}\mu_k
+\mu_k^\top\Sigma^{-1}\mu_k
\right) \nonumber\\
&=
x^\top\Sigma^{-1}\mu_k
-\frac{1}{2}\mu_k^\top\Sigma^{-1}\mu_k
-\frac{1}{2}x^\top\Sigma^{-1}x.
\label{eq:lda_expand}
\end{align}
The term $-\tfrac{1}{2}x^\top\Sigma^{-1}x$ is independent of $k$ and can be
dropped inside the $\arg\max$. Therefore, the Bayes-optimal decision rule can be
written using the \emph{linear} discriminant functions
\begin{equation}
\delta_k(x)
=
x^\top \Sigma^{-1}\mu_k
-\tfrac{1}{2}\mu_k^\top\Sigma^{-1}\mu_k
+\log \pi_k.
\label{eq:lda_discriminant}
\end{equation}
Finally, the LDA classifier predicts
\begin{equation}
\hat{y}(x)=\arg\max_{k}\,\delta_k(x).
\label{eq:lda_predict}
\end{equation}

\paragraph{Interpretation and practical estimation.}
Equation \eqref{eq:lda_discriminant} shows that LDA yields affine decision
boundaries in the feature space: $\delta_k(x)$ is linear in $x$, with class
specific weight vector $\Sigma^{-1}\mu_k$ and bias
$-\tfrac{1}{2}\mu_k^\top\Sigma^{-1}\mu_k+\log\pi_k$. In practice, $\mu_k$,
$\Sigma$, and $\pi_k$ are estimated from labeled \emph{source-domain} training
features, and the resulting classifier is applied directly to target-subject
features in transductive LOSO evaluation.

\paragraph{LDA classification on induced feature spaces.}
Let $X_{i,e}\in\mathbb{R}^{d\times T}$ denote an EEG trial and let
$x_{i,e}\in\mathbb{R}^d$ denote its Euclidean feature representation, obtained via
a fixed upstream mapping. In CSP\_LDA, $x_{i,e}$ corresponds to CSP-derived
log-variance features; in TSA\_LDA, $x_{i,e}$ corresponds to tangent-space
vectorization of an SPD covariance matrix. Once this mapping is fixed, LDA
operates identically in both cases by evaluating a linear discriminant function
in the induced feature space.
The probability distribution of $x_{i,e}$ given class $k$ is given by
\begin{equation}
p(x_{i,e}\mid y_{i,e}=k)
=
\mathcal{N}(x_{i,e};\mu_k,\Sigma),
\label{eq:lda_gauss}
\end{equation}
Specifically, classification is performed according to
\begin{equation}
\hat{y}(X_{i,e})
=
\arg\max_{k}\;\delta_k\big(x_{i,e}\big)
\quad ; \quad
\delta_k(x_{i,e})
=
x^\top \Sigma^{-1}\mu_k
-\tfrac{1}{2}\mu_k^\top \Sigma^{-1}\mu_k
+\log \pi_k.
\label{eq:lda_discriminant_features}
\end{equation}
Thus, LDA induces affine decision boundaries in the feature space defined by the
chosen preprocessing pipeline, whether CSP-based or tangent-space-based.

\paragraph{Parameter estimation under shared covariance.}
Given labeled source-domain training features
$\{(x_n,y_n)\}_{n=1}^N$, the class means, pooled covariance, and class priors are
estimated as
\begin{equation}
\mu_k
=
\frac{1}{N_k}\sum_{\substack{(i,e) \\ y_{i,e}=k}} x_{i,e},
\qquad
\Sigma
=
\sum_{k=1}^K\sum_{\substack{(i,e) \\ y_n=k}}
(x_{i,e}-\mu_k)(x_{i,e}-\mu_k)^\top,
\qquad
\pi_k
=
\frac{N_k}{N},
\label{eq:lda_params_pooled}
\end{equation}
optionally with regularization or normalization of $\Sigma$. All parameters are
learned exclusively from source-domain data and are applied unchanged to target-subject features in transductive LOSO evaluation.

\paragraph{Limitation in transductive cross-subject decoding.}
Although LDA is simple, interpretable, and often competitive, it remains a
linear classifier whose performance depends on the stability of class means and
shared covariance structure across subjects. Under cross-subject shifts, the
Gaussian-with-shared-covariance assumption and the resulting linear boundaries
can be violated, leading to degraded generalization in transductive settings.

\paragraph{Summary of classical classifiers and proposed pipelines.}
We evaluate several classical covariance-based EEG classifiers spanning
intrinsic Riemannian and tangent-space approaches. MDM operates directly on the
SPD manifold using Riemannian distances but relies only on class means, while
TSLR applies a fixed tangent-space mapping followed by a linear softmax
classifier. LDA-based pipelines (CSP-LDA, TSA-LDA) assume Gaussian
class-conditional structure and yield linear decision boundaries. Although all
baselines can be evaluated under transductive LOSO protocols, none learn
subject-invariant or class-conditional geometric transformations on the SPD
manifold. In contrast, our methods introduce learnable, geometry-aware
congruence operations that preserve SPD structure and enable nonlinear,
hierarchical discrimination.

\begin{table}[h]
\centering
\footnotesize
\renewcommand{\arraystretch}{1.15}
\caption{Comparison of classical classifiers and proposed geometry-aware pipelines.
Transductive refers to evaluation under LOSO without labeled target-adaptation.}
\label{tab:classifier_summary}
\begin{tabular}{lcccccc}
\toprule
Method
& SPD Geometry
& Tangent Space
& Learnable
& Deep
& Nonlinear
& Transductive \\
\midrule
MDM
& \checkmark
& $\times$
& $\times$
& $\times$
& $\times$
& \checkmark \\

TSLR
& \checkmark
& \checkmark
& \checkmark
& $\times$
& $\times$
& \checkmark \\

CSP\_LDA
& $\times$
& $\times$
& \checkmark
& $\times$
& $\times$
& \checkmark \\

TSA\_LDA
& \checkmark
& \checkmark
& \checkmark
& $\times$ 
& $\times$
& \checkmark \\

\midrule
\textbf{DCT (E2E) (Ours)}
& \checkmark
& \checkmark
& \checkmark
& $\times$ 
& $\times$
& \checkmark \\

\textbf{DLDCT (E2E) (Ours)}
& \checkmark
& \checkmark
& \checkmark
& \checkmark
& $\times$
& \checkmark \\

\textbf{DDCT-UNet (E2E) (Ours)}
& \checkmark
& \checkmark
& \checkmark
& \checkmark
& \checkmark
& \checkmark \\
\bottomrule
\end{tabular}
\end{table}

This comparison highlights that existing classical baselines rely on fixed
geometric representations or linear classifiers, whereas our proposed
congruence-based pipelines jointly learn geometry-preserving transformations
and nonlinear discriminative structure directly on the SPD manifold, enabling
improved subject-invariant transductive decoding.

\newpage
\section{Unified Geometry-Aware Congruence Learning Framework}
\label{sec:proposed_methods}

\begin{algorithm}[h]
\caption{Unified Congruence Learning: DCT / DLDCT / DDCT--UNet with Optional E2E}
\label{alg:unified_congruence}
\begin{algorithmic}

\STATE \textbf{Input:} $C_{i,e}\!\in\!\mathbb{S}_{++}^d$, $y_{i,e}$, mode $\in\{\text{DCT},\text{DLDCT},\text{DDCT-UNet}\}$, E2E$\in\{0,1\}$
\STATE \textbf{Hyperparams:} $(\delta,\alpha,\beta_t,\varepsilon)$ or $(\lambda_{act},\lambda_{sub},\lambda_w,\lambda_b,\lambda_{rec})$, $\lambda_{\mathrm{CE}}$
% ---------------- RA ----------------
\STATE \textbf{RA:}
$
C_i^{\mathrm{ref}}
=
\arg\min_{C}
\sum_e \|\log(C^{-1/2}C_{i,e}C^{-1/2})\|_F^2
$,
$
C'_{i,e}=(C_i^{\mathrm{ref}})^{-1/2}C_{i,e}(C_i^{\mathrm{ref}})^{-1/2},
$
\STATE
% ---------------- INIT ----------------
\STATE \textbf{Initialize parameters:}
\(
\begin{cases}
A=0,\;R=\exp(A-A^\top), \gamma & \text{DCT}\\
\{W_1, \dots , W_{\ell}\} & \text{DLDCT}\\
(E,D)\circ\{\Phi^{(\ell)}\} & \text{DDCT-UNet}
\end{cases}
\);
if E2E: Initialize $(W,b)$.
\STATE
% ---------------- TRAIN ----------------
\FOR{each iteration}
\STATE
\STATE \textbf{Forward:}
\(
\begin{cases}
L'_{i,e}=\log C'_{i,e}, \;L^O_{i,e}=R^\top(\gamma L'_{i,e})R,\;
C^{out}_{i,e}=\exp(L^O_{i,e}) & \text{DCT}\\
C^{out}_{i,e}=\Phi^{(\ell)}(C'_{i,e}) & \text{DLDCT}\\
B=E(C'_{i,e}),\;
C^{out}_{i,e}=D(B)\;\text{(LE merges)} & \text{DDCT-UNet}
\end{cases}
\)
\STATE
\STATE Compute $L^{out}_{i,e}=\log C^{out}_{i,e}$ and ($L'_{i,e}=\log C'_{i,e}$ if mode $\in \{\text{DLDCT}, \text{DDCT-UNet}\}$
\STATE
% ---------------- LOSSES ----------------
\STATE \textbf{Compute Loss:}
\STATE
\STATE$\mathcal{L}_{\mathrm{Pre}}=$
\(
\begin{cases}
\delta\frac{W_A}{B_A+\varepsilon}
+\mathrm{CP}
+\alpha(\gamma-1)^2+\beta_t\|R-I\|_F^2
& \text{DCT}\\

\lambda_{act}(\lambda_w W_A-\lambda_b B_A
+\lambda_{sub}(\lambda_b B_S-\lambda_w W_S
+\lambda_{rec}\frac{1}{N}\!\sum_{i,e}\!\|L^{out}_{i,e}-L'_{i,e}\|_F^2
& \text{DLDCT/DDCT-UNet}
\end{cases}
\)
\STATE
\IF{E2E}
\STATE $x_{i,e}=\mathrm{vec}(L^{out}_{i,e}),\;
p=\mathrm{softmax}(Wx+b)$,
\STATE $
\mathcal{L}
=
\mathcal{L}_{Pre}
+\lambda_{\mathrm{CE}}\mathcal{L}_{\mathrm{CE}}\!\left(\text{cross-entropy loss:}-\frac{1}{N}\sum_{i,e}\log p[y_{i,e}]\right)
$.
\ELSE
\STATE $\mathcal{L}
=
\mathcal{L}_{Pre}$
\ENDIF
\STATE
\STATE \textbf{Backpropagate} $\nabla\mathcal{L}$
\STATE
\ENDFOR
\STATE
\STATE \textbf{Output:}
\IF{E2E}
    \STATE $\hat y_{i,e}=\arg\max_k p_{i,e}[k]$ with optimized $(W^\star,b^\star)$
    and pre-aligner parameters below.
\ELSE
    \STATE $
    \begin{cases}
    C^{\mathrm{out}}_{i,e}
    =
    \exp\!\big(R^{\star\top}\gamma^\star L'_{i,e}R^\star\big)
    & \text{DCT}\\[2pt]

    C^{\mathrm{out}}_{i,e}
    =
    \Phi^{(T)\star}\!\circ\cdots\circ\Phi^{(1)\star}(C'_{i,e})
    & \text{DLDCT}\\[2pt]

    C^{\mathrm{out}}_{i,e}
    =
    D^\star(E^\star(C'_{i,e}))
    & \text{DDCT--UNet}
    \end{cases}
    \text{with optimized parameters}
    \begin{cases}
    (R^\star,\gamma^\star) & \text{DCT}\\
    \{W^{(t)\star}\} & \text{DLDCT}\\
    (E^\star,D^\star) & \text{DDCT-UNet}\\
    (W^\star,b^\star) & \text{if E2E}
    \end{cases}
$
\ENDIF

\end{algorithmic}
\end{algorithm}

\begin{figure*}
    \centering
    \includegraphics[width=0.9\linewidth]{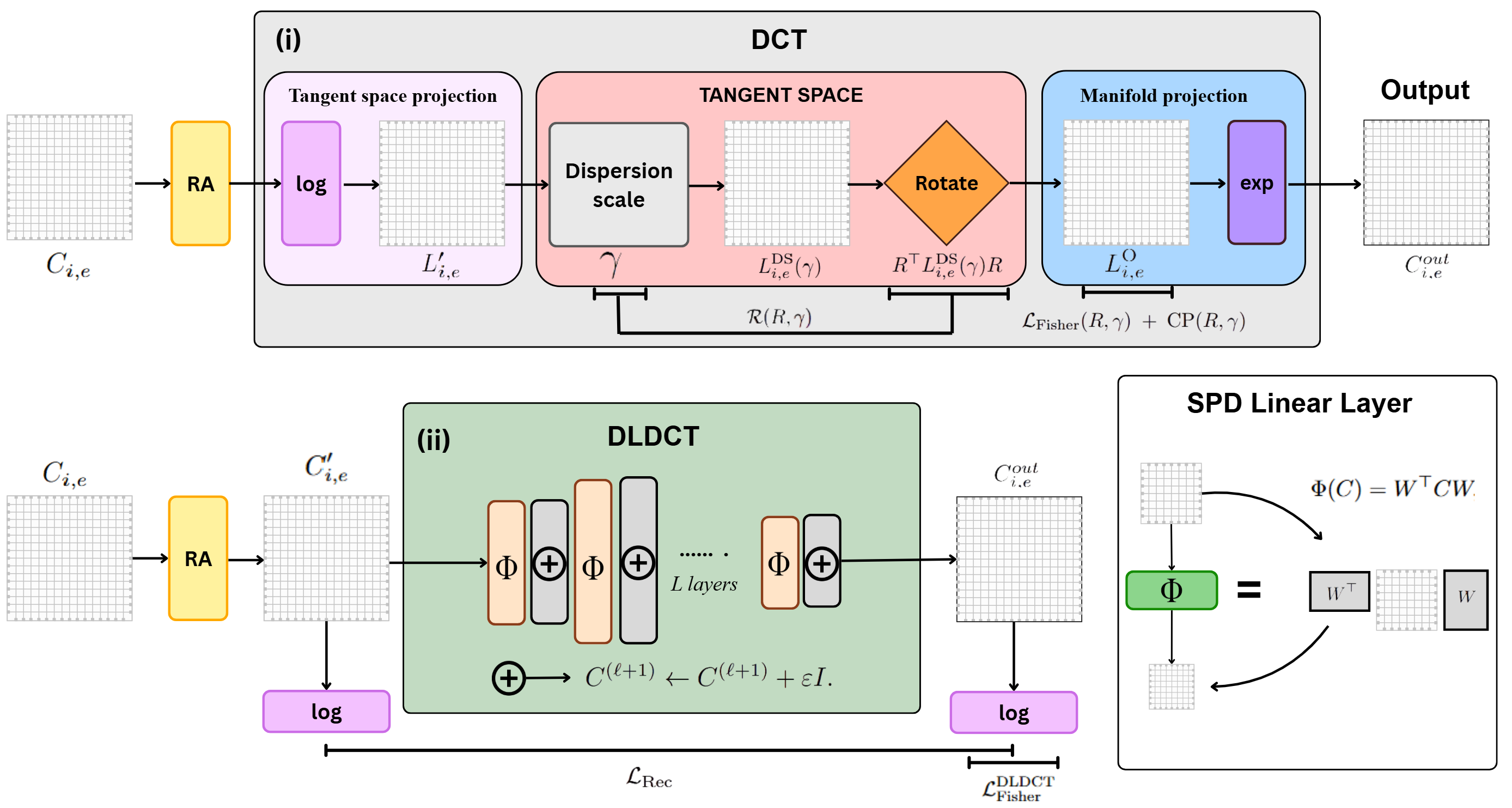}
    \caption{Top: \textbf{DCT} Learns dispersion scaling and rotation in the tangent space after RA to produce an aligned SPD. Bottom: \textbf{DLDCT:} Deep stacks of SPD linear congruence layers with residual updates. Right: SPD linear layer $\Phi(C)=W^\top C W$.
}
    \label{fig:prealigners}
\end{figure*}

\begin{figure*}
    \centering
    \includegraphics[width=0.9\linewidth]{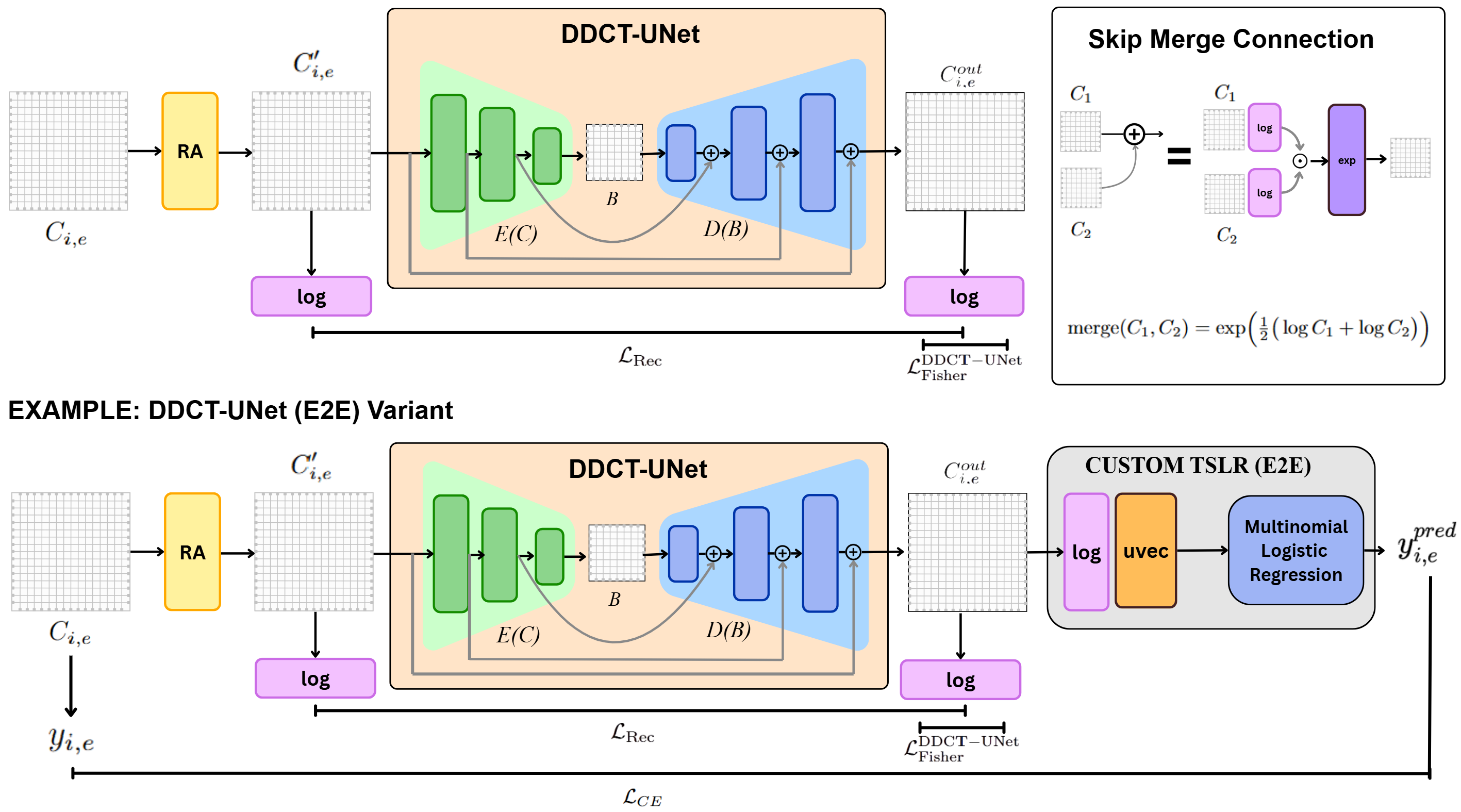}
    \caption{Top: \textbf{DDCT--UNet} pre-aligner with encoder--decoder stacks of SPD
congruence layers and log--Euclidean skip merges.
Right: Skip connections perform manifold-aware fusion via log/exp maps.
Bottom: \textbf{DDCT--UNet (E2E)} with a custom TSLR head trained by
cross-entropy, illustrating how end-to-end TSLR variants are constructed.}
    \label{fig:classifiers}
\end{figure*}

\newpage
\section{Geometry Diagnostics}
\label{app:visuals}

\subsection{Covariance Evolution across Proposed Models During Training Phase}

We analyze the geometric behavior of different pre-alignment and representation-learning models through covariance evolution visualizations.
All analyses are conducted on the \textbf{BCI Competition IV-2a} motor imagery dataset, which consists of four motor imagery classes (left hand, right hand, feet, tongue) recorded from nine subjects.
Throughout this section, we focus exclusively on source-domain (training) covariances.
For visualization, we select four representative source subjects: A01T, A03T, A04T, and A05T (indexed as subjects 0, 2, 3, and 4), while subject A02T is reserved solely for evaluation and excluded from all plots.
The specific choice of visualization subjects is arbitrary, and qualitatively similar geometric trends are observed across all subjects.

Each point in the covariance evolution plots corresponds to a single class-conditional sample covariance matrix.
Marker shapes indicate subject identity, and colors denote action class.
At each training interval, the same fixed set of input SPD covariances is processed by the current model state, mapped to the log-Euclidean tangent space, vectorized, and embedded into two dimensions using t-SNE.
Axis values carry no intrinsic meaning and are omitted; only relative geometric relationships are interpreted.
Importantly, the underlying EEG covariances remain fixed across all intervals.
Thus, these plots do not reflect temporal changes in EEG statistics or trial-wise dynamics.
Instead, they visualize how covariance representations move in tangent space as model parameters evolve during optimization.
The observed trajectories therefore capture \emph{parameter-driven geometric adaptation} induced by learning, rather than data drift.

\subsubsection{Discriminative Congruence Transform (DCT) Prealigner}
\label{app:dct_cov_evolution}

The Discriminative Congruence Transform (DCT) pre-aligner consists of a single global congruence transformation learned on the SPD manifold.
Its objective is to reduce subject-induced rotational variability by aligning class-relevant covariance directions across subjects, while not explicitly enforcing class separation.
Owing to this constrained parameterization, DCT is expected to exhibit a smooth and monotonic geometric evolution during training, characterized primarily by gradual rotational reorientation rather than nonlinear warping or multi-stage geometric restructuring.
Figure~\ref{fig:dct_cov_evolution} visualizes this behavior through covariance evolution plots at successive training intervals.

\begin{figure}[h]
    \centering
    \includegraphics[width=0.8\linewidth]{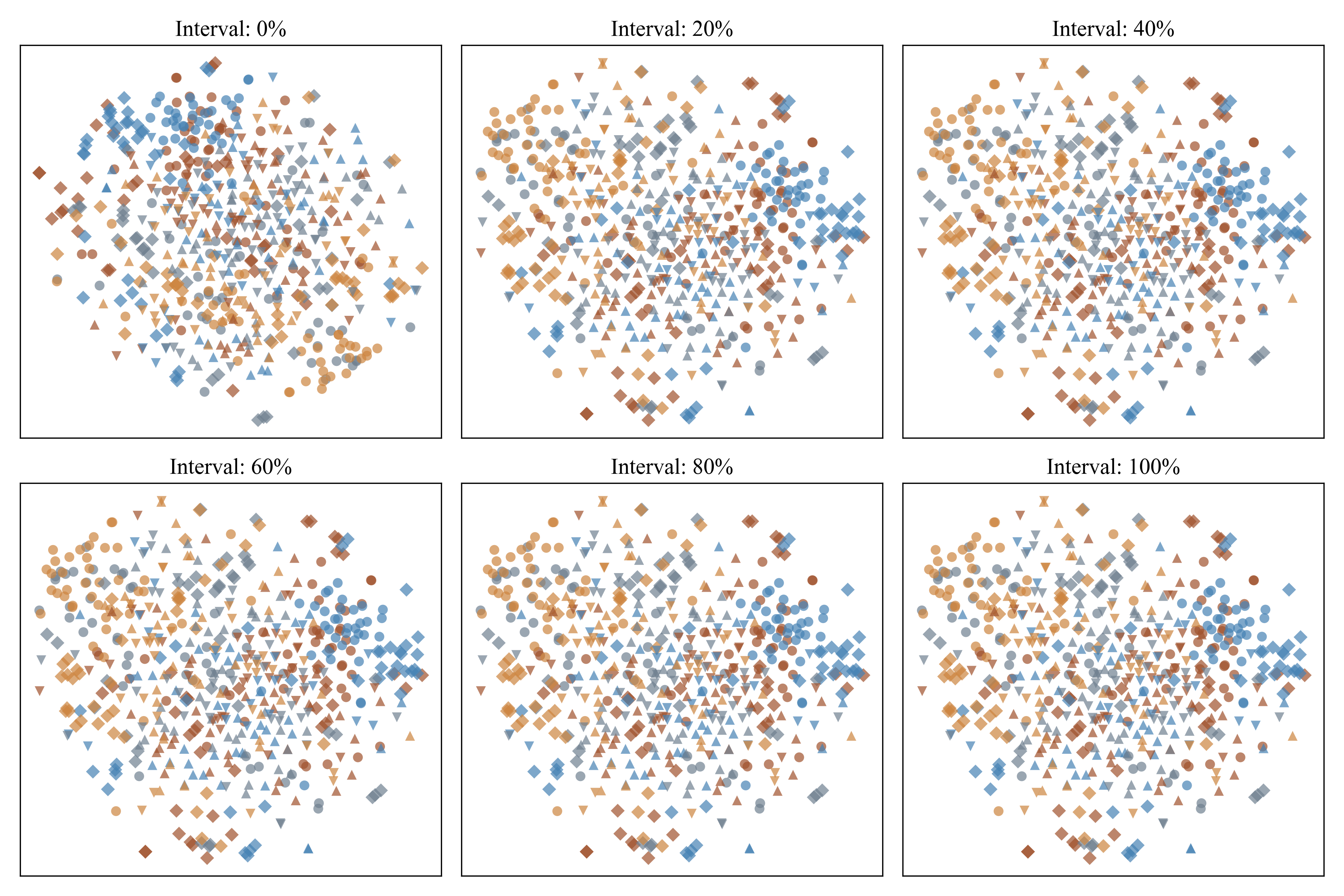}
    \caption{
Covariance evolution under the Discriminative Congruence Transform (DCT) pre-aligner across training intervals.
Points represent transformed class-conditional covariances from the source set; colors denote action classes and marker shapes denote subjects (four shown).
Changes across intervals reflect geometry evolution induced by optimization of a single global congruence transformation.
}
    \label{fig:dct_cov_evolution}
\end{figure}

\medskip
\textbf{Interval 0\% (initialization):}
At initialization, covariance representations are highly intermingled across both subjects and classes.
Subject-dependent rotational mismatches dominate the tangent-space geometry, leading to inflated within-class scatter when covariances are aggregated across subjects.
Class-conditional structure is largely obscured.
\emph{Inference:} At this stage, class-relevant directions are not aligned across subjects, and tangent-space representations remain strongly subject-dependent.

\medskip
\textbf{20\%:}
Early during optimization, a mild but consistent directional drift is observed.
While global overlap remains high, covariance representations corresponding to the same action class begin to exhibit weak alignment trends across subjects.
\emph{Inference:} DCT starts correcting dominant subject-specific rotations, but discriminative class structure is still largely entangled with subject variability.

\medskip
\textbf{40\%:}
At this stage, directional consistency across subjects becomes more apparent.
For several action classes, covariance points show improved coherence in their principal orientation, although spatial overlap persists.
\emph{Inference:} The learned congruence transformation has begun to align class-relevant directions globally, reducing rotational mismatch without inducing class-specific geometric deformation.

\medskip
\textbf{60\%:}
The geometric configuration continues to stabilize.
Changes relative to earlier intervals are subtler, with reduced drift and more consistent orientation of class-conditional covariances across subjects.
\emph{Inference:} Most of the useful rotational alignment has already been achieved, and further training primarily refines rather than restructures the geometry.

\medskip
\textbf{80\%:}
Only marginal changes are observed compared to the 60\% interval.
Class overlap remains substantial, but covariance orientations across subjects are now largely consistent.
\emph{Inference:} DCT has effectively converged in terms of global rotation, indicating diminishing returns from additional optimization steps.

\medskip
\textbf{100\% (convergence):}
The final configuration is stable, with no visible geometric drift.
Class-conditional covariances from different subjects remain overlapping but exhibit reduced within-class dispersion relative to initialization.
No numerical or geometric degeneration is observed.
\emph{Inference:} DCT converges to a stable global alignment that suppresses subject-dependent rotational bias while preserving intrinsic class ambiguity inherent to EEG signals.

\medskip
\textbf{Interpretation and architectural implications.}
The observed smooth and low-complexity evolution is a direct consequence of DCT’s architectural constraints.
Because DCT consists of a single congruence layer, it can only express a global rotational alignment on the SPD manifold.
As a result, the induced geometry evolves toward a stable configuration with limited expressivity and cannot exhibit nonlinear warping, multi-stage reorganization, or class-specific deformation trajectories.
Such richer geometric dynamics arise only in deeper architecture, where multiple congruence layers are composed and discriminative objectives are applied iteratively.

\subsubsection{Deep Linear Discriminative Congruence Transform (DLDCT) Prealigner}
\label{app:dldct_cov_evolution}

The Deep Linear Discriminative Congruence Transform (DLDCT) extends shallow congruence-based pre-alignment by composing multiple discriminative congruence layers on the SPD manifold.
Unlike DCT, which is limited to a single global rotation, DLDCT can progressively reshape covariance geometry through a hierarchy of learned congruence mappings optimized with class-discriminative objectives.
This additional depth enables nonlinear manifold warping, staged reorganization of class geometry, and progressive suppression of subject-specific variability.
Figure~\ref{fig:dldct_cov_evolution} visualizes this richer geometric evolution through covariance trajectories observed at successive training intervals.

\begin{figure}[h]
    \centering
    \includegraphics[width=0.8\linewidth]{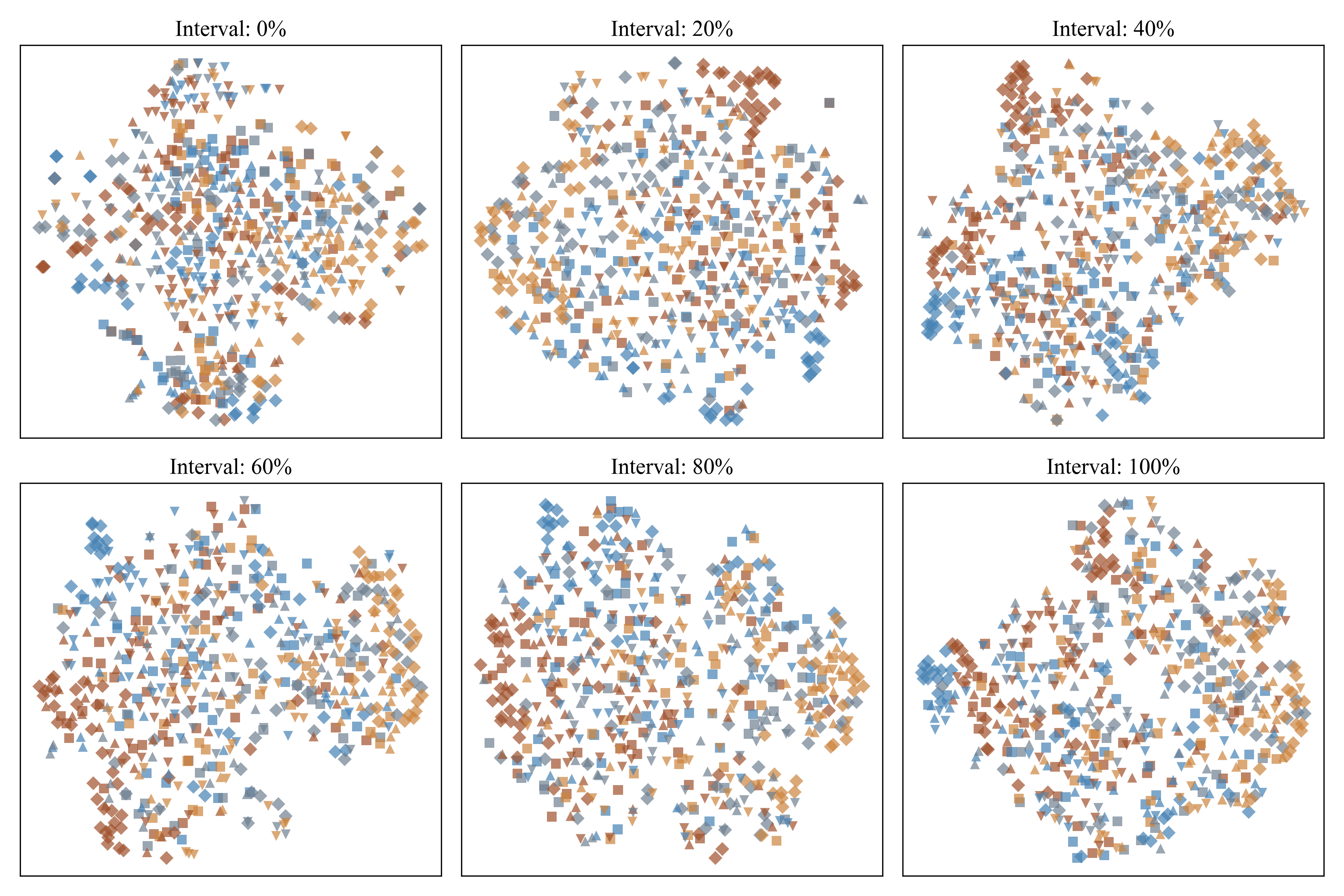}
    \caption{
Covariance evolution under the Deep Linear Discriminative Congruence Transform (DLDCT) pre-aligner across training intervals.
Points represent transformed class-conditional covariances from the source set; colors denote action classes and marker shapes denote subjects (four shown).
Changes across intervals reflect the multi-stage geometry evolution induced by stacked discriminative congruence layers.
}
    \label{fig:dldct_cov_evolution}
\end{figure}

\medskip
\textbf{Interval 0\% (initialization):}
At initialization, covariance representations are highly intermingled across subjects and classes, mirroring the unaligned source geometry.
Strong subject-dependent rotations dominate the tangent-space structure, and class-conditional clusters are poorly formed.
\emph{Inference:} No meaningful subject-invariant or class-consistent geometry has yet emerged.

\medskip
\textbf{20\%:}
Early training produces noticeable geometric drift that is stronger than in shallow DCT.
Several covariance groups begin to undergo coordinated directional motion, and weak proto-clusters appear for certain action classes.
\emph{Inference:} The first congruence layers start correcting dominant subject rotations while initiating early class-driven deformation.

\medskip
\textbf{40\%:}
Distinct geometric restructuring becomes apparent.
Compared to DCT, class-conditional points now begin to concentrate into elongated manifolds or curved trajectories rather than merely rotating rigidly.
Subject markers for the same class increasingly co-locate.
\emph{Inference:} Intermediate layers are actively reshaping manifold structure, aligning both orientation and higher-order dispersion patterns.

\medskip
\textbf{60\%:}
The embedding geometry undergoes further refinement.
Clusters tighten, and class-conditioned regions become visually separable along multiple tangent directions.
Subject mixing within classes improves substantially.
\emph{Inference:} DLDCT has transitioned from coarse rotational alignment to deep discriminative shaping of the covariance space.

\medskip
\textbf{80\%:}
Geometric changes slow but remain nontrivial.
Most classes form coherent, compact regions, while overlap across subjects is substantially reduced relative to earlier intervals.
Only mild fine-scale deformation is observed.
\emph{Inference:} The deeper congruence stack is converging toward a discriminative manifold configuration that supports subject-invariant classification.

\medskip
\textbf{100\% (convergence):}
The final configuration is stable, with sharply reduced within-class scatter and visibly enhanced class organization compared to initialization.
Unlike DCT, the geometry reflects non-rigid deformation rather than a single global rotation.
No signs of collapse or degeneracy are present.
\emph{Inference:} DLDCT converges to a deeply structured alignment in which stacked congruence transforms progressively sculpt the SPD manifold into a class-discriminative, subject-robust representation.

\medskip
\textbf{Interpretation and architectural implications.}
The markedly richer geometric dynamics observed here stem from DLDCT’s deep architecture.
By composing multiple congruence layers, the model can express nonlinear warping, hierarchical reorientation, and progressive tightening of class-conditional regions—capabilities that are fundamentally unavailable to shallow transforms such as DCT.
This staged geometric evolution underlies the superior cross-subject generalization observed in downstream classifiers and motivates the use of deep SPD networks in challenging transductive motor imagery decoding regimes.

\subsubsection{Deep Discriminative Congruence Transform -- UNet (DDCT-UNet) Prealigner}
\label{app:ddctunet_cov_evolution}

DDCT-UNet is a geometry-aware pre-aligner that learns an SPD-preserving encoder--decoder mapping
$\Psi_{\theta}:\mathcal{S}_{++}^d \rightarrow \mathcal{S}_{++}^d$.
Unlike single-layer pre-aligners, DDCT-UNet composes multiple nonlinear discriminative congruence transformations while enforcing reconstruction fidelity to the Riemannian Alignment (RA) input.
Its objective is to produce covariance representations that are discriminative for action classes while suppressing subject-specific information.
As a result, one expects richer and potentially non-monotonic geometric evolution during training.
Figure~\ref{fig:ddctunet_cov_evolution} illustrates this behavior across successive training intervals.

\medskip
\textbf{Interval 0\% (initialization):}
At initialization, covariance representations closely resemble those obtained after RA.
Class and subject information are strongly entangled, with substantial overlap across both dimensions.
Subject-dependent structure is clearly visible, while class-conditional organization remains weak.
\emph{Inference:} At this stage, the DDCT-UNet encoder--decoder has not yet learned invariance-inducing or discriminative transformations, and tangent-space geometry remains dominated by subject variability.

\medskip
\textbf{20\%:}
Early during optimization, noticeable geometric deformation begins to emerge.
Some covariance groups corresponding to the same action class start exhibiting weak coherence across subjects, while subject-specific clustering becomes less pronounced.
\emph{Inference:} DDCT-UNet begins to disentangle class-relevant directions from subject identity through nonlinear transformation, moving beyond pure rotational alignment.

\begin{figure}[h]
    \centering
    \includegraphics[width=0.8\linewidth]{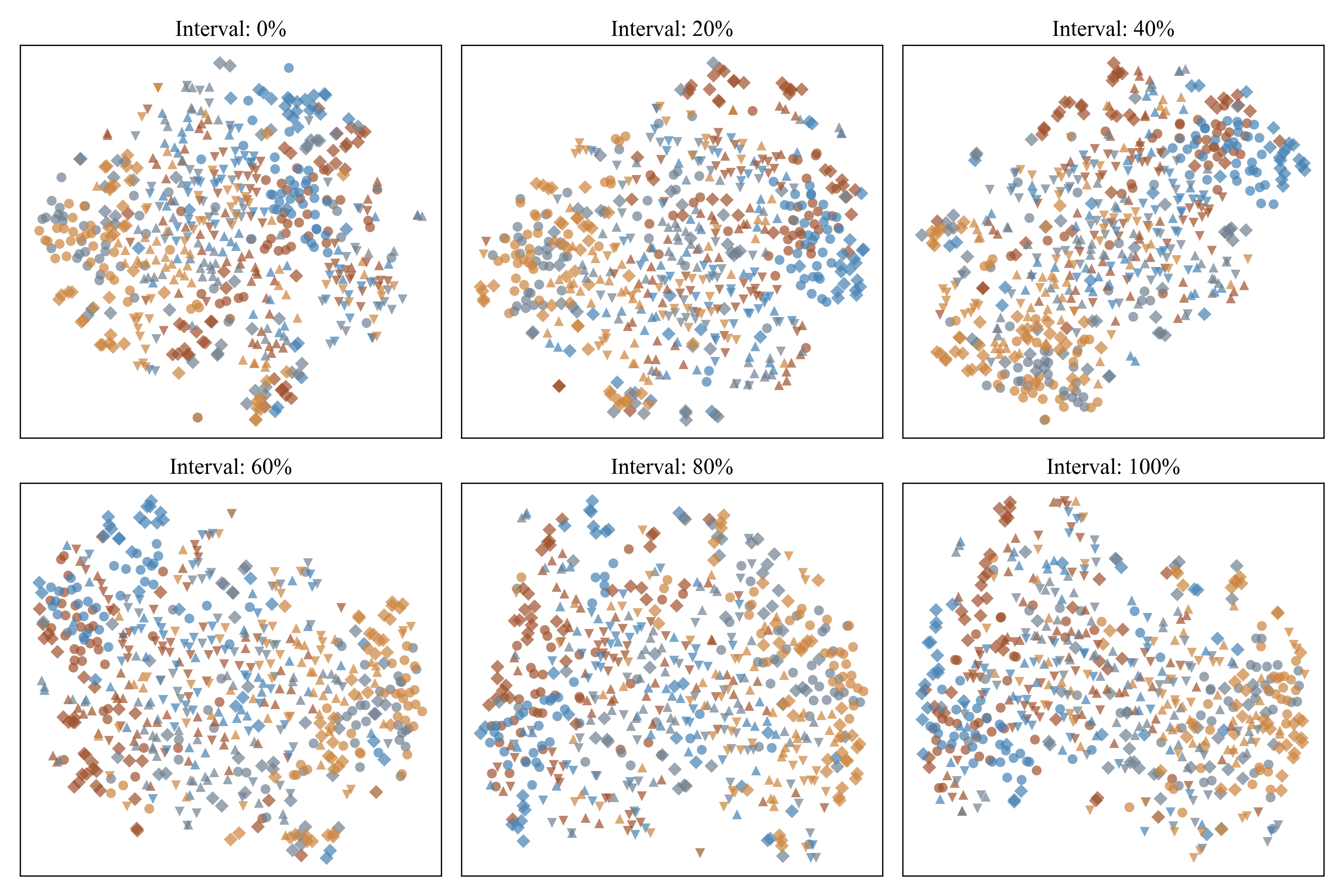}
    \caption{
    Covariance evolution under the DDCT-UNet pre-aligner across training intervals.
Points represent transformed class-conditional covariances from the source set; colors denote action classes and marker shapes denote subjects (four shown).
Changes across intervals reflect geometry evolution induced by DDCT-UNet’s encoder--decoder mapping.}
    \label{fig:ddctunet_cov_evolution}
\end{figure}

\medskip
\textbf{40\%:}
At this interval, class-driven organization becomes more apparent.
Several action classes form loosely connected structures spanning multiple subjects, while subject-specific separation continues to diffuse.
Unlike shallow congruence transforms, changes are not uniform across the space, indicating localized geometric deformation.
\emph{Inference:} The DDCT-UNet encoder actively reshapes the tangent-space geometry in a class-aware manner, selectively emphasizing discriminative directions.

\medskip
\textbf{60\%:}
The geometry exhibits clearer structure, with improved class coherence across subjects and reduced subject-dependent clustering.
However, some regions show transient distortions or partial re-mixing.
\emph{Inference:} DDCT-UNet operates in a refinement regime where discriminative alignment and reconstruction fidelity compete, leading to non-monotonic but controlled geometric evolution.

\medskip
\textbf{80\%:}
Geometric changes become more stable.
Class-conditional covariances display consistent alignment across subjects, while subject identity is no longer strongly encoded in the geometry.
Residual overlap persists across classes.
\emph{Inference:} The DDCT-UNet encoder--decoder has largely converged to a representation that balances class discrimination with subject invariance, without collapsing intra-class structure.

\medskip
\textbf{100\% (convergence):}
The final configuration is stable, with minimal further geometric drift.
Class-driven organization is clearer than in shallow pre-aligners, while covariance representations remain well-spread and SPD-valid.
No signs of degeneracy or over-contraction are observed.
\emph{Inference:} DDCT-UNet converges to a nonlinear, subject-invariant pre-alignment that enhances class discriminability while preserving geometric fidelity to the original covariance manifold.

\medskip
\textbf{Interpretation and architectural implications.}
The covariance evolution observed under DDCT-UNet demonstrates the benefit of a deep UNet-style SPD encoder--decoder architecture.
By combining stacked discriminative congruence layers with a reconstruction constraint, DDCT-UNet performs nonlinear, class-aware geometric alignment while maintaining proximity to the original covariance manifold.
This results in substantially richer geometry evolution than is possible with single-layer congruence transformations and positions DDCT-UNet as an intermediate regime between lightweight pre-aligners and fully discriminative end-to-end SPD classifiers, making it particularly well-suited for cross-subject EEG generalization.

\subsubsection{Discriminative Congruence Transform -- End-to-End (DCT-E2E) Classifier}
\label{app:dcte2e_cov_evolution}

DCT-E2E is an end-to-end optimized variant of the Discriminative Congruence Transform in which the congruence-based SPD pre-aligner is trained jointly with the downstream classifier.
Unlike standalone DCT, which is optimized only through an alignment objective, DCT-E2E allows task gradients to propagate directly through the congruence layer, coupling geometric alignment with classification performance.
This tighter integration increases expressive power relative to shallow pre-alignment while retaining a single-layer architectural form.
Consequently, one expects training dynamics that interpolate between the smooth rotational evolution of DCT and the richer restructuring observed in deeper models such as DLDCT and DDCT-UNet.
Figure~\ref{fig:dcte2e_cov_evolution} illustrates this intermediate geometric behavior across successive training intervals.

\begin{figure}[h]
    \centering
    \includegraphics[width=0.8\linewidth]{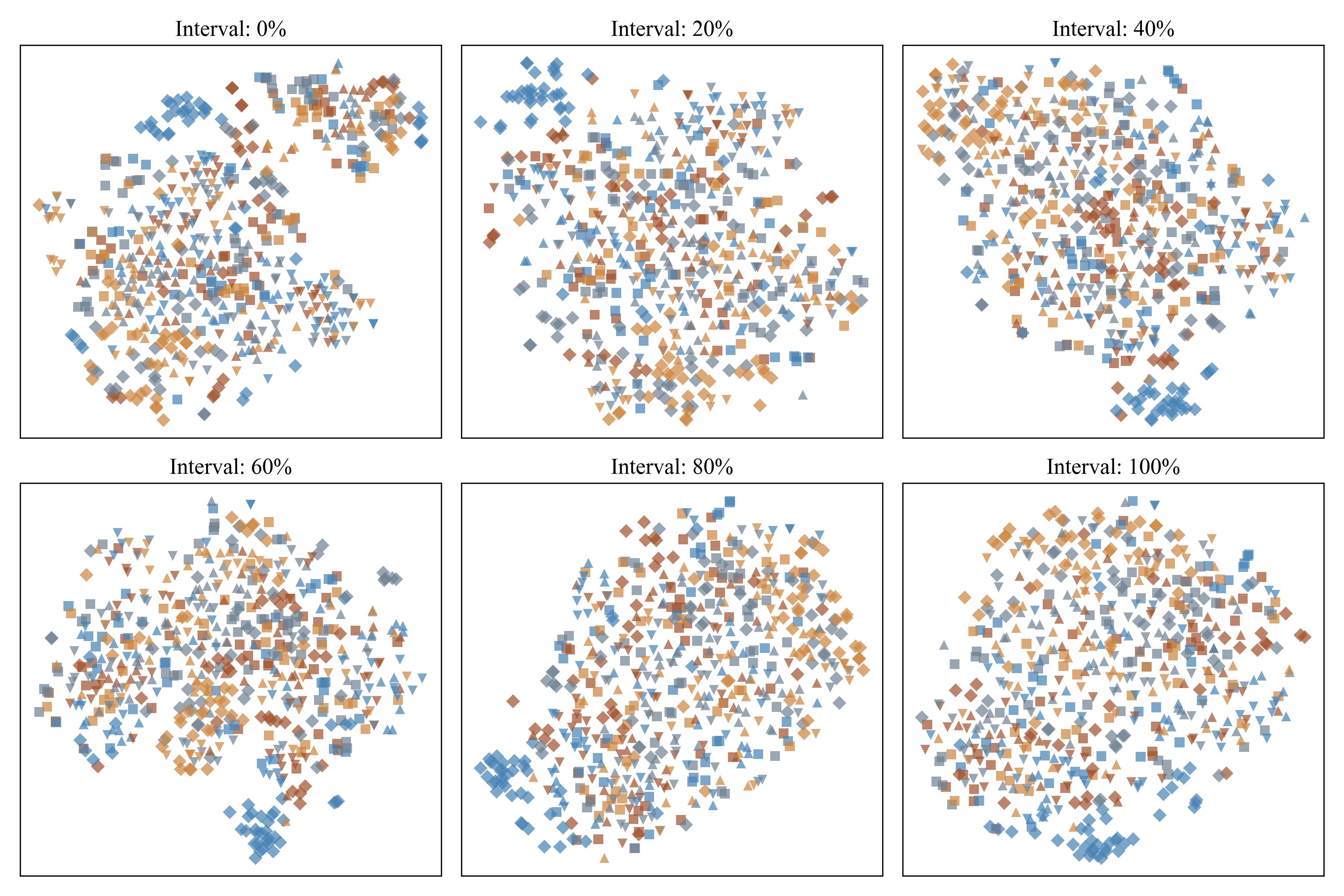}
    \caption{
Covariance evolution under the DCT-E2E pre-aligner across training intervals.
Points represent transformed class-conditional covariances from the source set; colors denote action classes and marker shapes denote subjects (four shown).
Changes across intervals reflect geometry evolution induced by end-to-end optimization of a single discriminative congruence layer.
}
    \label{fig:dcte2e_cov_evolution}
\end{figure}

\medskip
\textbf{Interval 0\% (initialization):}
At initialization, covariance geometry closely resembles that obtained after RA, with substantial intermixing across subjects and classes.
Strong subject-specific rotational offsets dominate the tangent-space structure, and class-conditioned organization is weak.
\emph{Inference:} The congruence layer has not yet been shaped by task gradients, and geometry remains largely subject-dependent.

\medskip
\textbf{20\%:}
Early optimization introduces visible directional drift in multiple regions of the space.
Compared to vanilla DCT, changes are less uniform, with certain class-conditioned subsets beginning to move coherently.
\emph{Inference:} Joint optimization with the classifier starts to bias the global rotation toward class-relevant directions, rather than purely reducing subject mismatch.

\medskip
\textbf{40\%:}
Class-driven reorganization becomes more apparent.
Some action classes form loose aggregates spanning multiple subjects, while others remain partially intermixed.
Deformation is still largely global rather than strongly localized.
\emph{Inference:} Task supervision has begun to reshape the congruence transform beyond neutral alignment, but expressivity remains constrained by its single-layer form.

\medskip
\textbf{60\%:}
The geometry continues to evolve, with clearer within-class coherence and reduced subject-specific clustering.
However, residual class overlap persists, and some regions undergo mild transient spreading.
\emph{Inference:} DCT-E2E is operating in a regime where subject-invariance and discriminative pressure are being balanced through a shared global transform.

\medskip
\textbf{80\%:}
Changes slow substantially.
Class-conditioned regions appear more stable, and subject markers within each class are increasingly co-located.
Compared to shallow DCT, the configuration is more class-aware, though less structured than in DLDCT or DDCT-UNet.
\emph{Inference:} End-to-end gradients have driven the congruence layer toward a task-adapted alignment that remains globally constrained.

\medskip
\textbf{100\% (convergence):}
The final geometry is stable, with reduced within-class scatter relative to initialization and modest improvement in class organization.
Unlike DCT, the learned mapping is no longer purely subject-neutral, but it does not exhibit the deep nonlinear restructuring characteristic of DDCT-UNet.
No geometric collapse or degeneration is observed.
\emph{Inference:} DCT-E2E converges to a task-aware global alignment that improves subject robustness while preserving the limited expressivity inherent to a single congruence transform.

\medskip
\textbf{Interpretation and architectural implications.}
The covariance evolution under DCT-E2E highlights the effect of coupling geometric alignment directly to classification loss.
Relative to DCT, end-to-end training induces more class-oriented restructuring, yet the geometry remains globally constrained by the single congruence layer.
This places DCT-E2E in an intermediate regime between shallow pre-alignment and fully deep geometric models such as DLDCT and DDCT-UNet.
The observed dynamics support the paper’s central thesis that increasing architectural depth in SPD-space systematically enriches geometric expressivity and cross-subject invariance, with DCT-E2E serving as a controlled bridge between these extremes.

\subsubsection{Deep Linear Discriminative Congruence Transform -- End-to-End (DLDCT-E2E) Classifier}
\label{app:dldcte2e_cov_evolution}

DLDCT-E2E is a deep SPD-valued congruence network in which multiple discriminative congruence layers are trained jointly with the downstream classifier through end-to-end supervision.
Unlike DCT, which performs shallow pre-alignment, and DDCT-UNet, which additionally enforces reconstruction fidelity through an encoder--decoder structure, DLDCT-E2E is driven solely by classification objectives.
As a result, its geometric evolution is governed by task gradients without explicit constraints encouraging subject invariance or manifold-preserving reconstruction.
Figure~\ref{fig:dldcte2e_cov_evolution} visualizes the resulting covariance trajectories across training intervals.

\begin{figure}[h]
    \centering
    \includegraphics[width=0.8\linewidth]{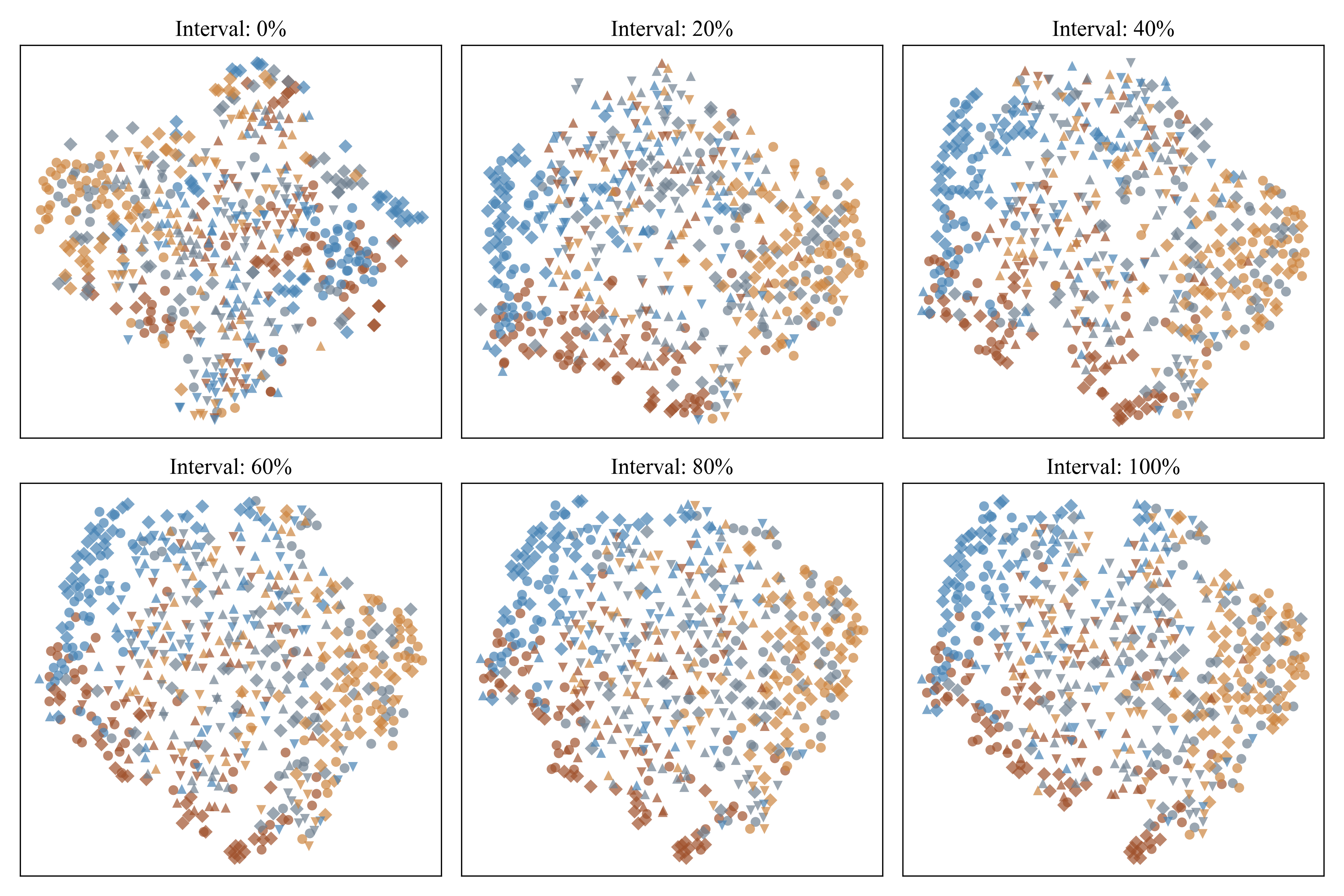}
    \caption{
Covariance evolution under the DLDCT-E2E model across training intervals.
Points represent class-conditional covariances transformed by the evolving deep congruence stack; colors denote action classes and marker shapes denote subjects (four shown).
Changes across intervals reflect geometry evolution induced by purely end-to-end discriminative training.
}
    \label{fig:dldcte2e_cov_evolution}
\end{figure}

\medskip
\textbf{Interval 0\% (initialization):}
At initialization, covariance representations are highly intermingled across both subjects and action classes.
No clear class-conditional organization is visible, and subject-dependent variability dominates the tangent-space embedding.
\emph{Inference:} The deep congruence stack has not yet been shaped by discriminative gradients, and the geometry reflects largely unstructured covariance statistics.

\medskip
\textbf{20\%:}
Early during training, noticeable geometric deformation emerges.
Certain action classes begin to occupy broader, partially coherent regions, while others remain strongly mixed.
Subject-specific clustering is still visible.
\emph{Inference:} Classification-driven gradients begin sculpting the manifold, but learning proceeds unevenly in the absence of explicit subject-alignment regularization.

\medskip
\textbf{40\%:}
Class-driven organization becomes more pronounced.
Several action classes form elongated or curved structures, while subject-conditioned fragmentation persists within these regions.
Geometric changes are spatially non-uniform.
\emph{Inference:} DLDCT-E2E learns discriminative directions, but these remain partially entangled with subject-dependent variability.

\medskip
\textbf{60\%:}
The geometry exhibits stronger class-wise structuring, with clearer separation between several action categories.
However, subject identity remains partially encoded, as evidenced by fragmented sub-clusters within the same class region.
\emph{Inference:} Deep discriminative supervision strengthens class separation, but the lack of explicit invariance objectives limits full cross-subject consolidation.

\medskip
\textbf{80\%:}
Geometric evolution slows and class-conditioned regions stabilize.
Some classes become compact, while others remain diffuse.
Residual subject structure is reduced but not eliminated.
\emph{Inference:} DLDCT-E2E approaches a locally optimal discriminative geometry that favors classification accuracy over subject-invariant alignment.

\medskip
\textbf{100\% (convergence):}
The final configuration is stable, with well-defined class-oriented regions and minimal further deformation.
Compared to DLDCT and DDCT-UNet, within-class geometry remains less uniformly consolidated across subjects, and traces of subject-dependent organization persist.
No signs of numerical collapse or degeneracy are observed.
\emph{Inference:} DLDCT-E2E converges to a deeply discriminative but partially subject-sensitive representation, achieving separation primarily through stacked nonlinear congruence transforms rather than controlled geometric regularization.

\medskip
\textbf{Interpretation and architectural implications.}
The covariance evolution under DLDCT-E2E highlights the effects of optimizing deep SPD geometry purely through task supervision.
Relative to shallow DCT, the deeper congruence stack induces substantially richer restructuring.
However, in contrast to DDCT-UNet—where reconstruction constraints explicitly regularize manifold deformation—DLDCT-E2E relies entirely on discriminative gradients, leading to residual subject-dependent structure.
These observations support the paper’s central claim that increasing architectural depth enhances geometric expressivity, while auxiliary geometric constraints are crucial for stabilizing cross-subject invariance in EEG decoding.

\subsubsection{Deep Discriminative Congruence Transform UNet (E2E) -- (DDCT-UNet-E2E) Classifier}
\label{app:ddctunete2e_cov_evolution}

DDCT-UNet-E2E extends the DDCT-UNet pre-aligner into a fully end-to-end classification model by coupling the SPD-preserving encoder--decoder
$\Psi_{\theta}:\mathcal{S}_{++}^d \rightarrow \mathcal{S}_{++}^d$
with a task-specific tangent-space linear readout.
Unlike DCT and DDCT-UNet, where geometry is shaped primarily through pre-alignment and reconstruction objectives, DDCT-UNet-E2E propagates discriminative supervision directly from the classification loss through the tangent-space classifier and back into the covariance transformation itself.
As a result, geometry is no longer optimized solely to suppress subject variability, but is actively reshaped to enhance class separability while preserving SPD validity.
Figure~\ref{fig:ddctunete2e_cov_evolution} visualizes the resulting covariance evolution across training intervals.

\begin{figure}[h]
    \centering
    \includegraphics[width=0.8\linewidth]{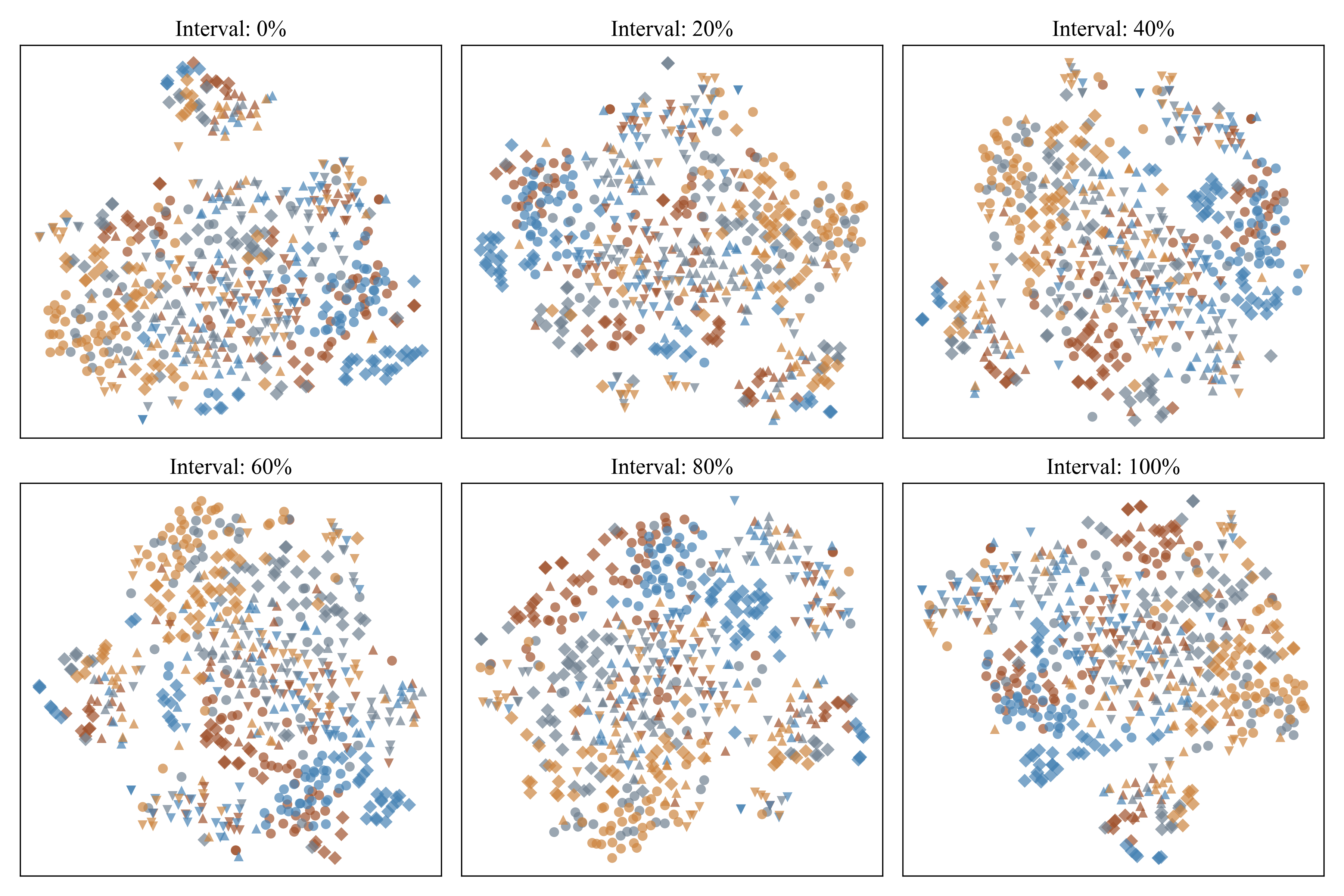}
    \caption{
Covariance evolution under the DDCT-UNet-E2E model across training intervals.
Points represent class-conditional covariances transformed by the evolving end-to-end network; colors denote action classes and marker shapes denote subjects (four shown).
Changes across intervals reflect geometry evolution induced jointly by the DDCT-UNet encoder--decoder and downstream tangent-space classifier.
}
    \label{fig:ddctunete2e_cov_evolution}
\end{figure}

\medskip
\textbf{Interval 0\% (initialization):}
At initialization, covariance representations closely resemble those obtained under DDCT-UNet, with substantial overlap across both subjects and action classes.
Although the encoder--decoder introduces nonlinear structure, class-conditional organization remains weak and entangled with subject identity.
\emph{Inference:} At this stage, discriminative gradients from the classifier have not yet propagated meaningfully into the covariance space, and geometry remains dominated by inherited pre-alignment structure.

\medskip
\textbf{20\%:}
Early during end-to-end training, directional changes become apparent in regions corresponding to certain action classes.
Some class-consistent grouping begins to emerge across subjects, while subject-specific clustering starts to weaken.
\emph{Inference:} The classification loss begins to bias the covariance transformation toward directions that are linearly separable in the tangent space.

\medskip
\textbf{40\%:}
Class-driven reorganization becomes more pronounced.
Covariance representations corresponding to the same action class form clearer, though still overlapping, structures spanning multiple subjects.
The evolution is no longer symmetric or uniform across the space, indicating selective deformation driven by task relevance.
\emph{Inference:} DDCT-UNet-E2E actively reshapes the covariance geometry to amplify class-discriminative directions, even when this deviates from purely subject-invariant alignment.

\medskip
\textbf{60\%:}
The geometry exhibits further refinement, with improved intra-class coherence and continued suppression of subject-specific structure.
Changes relative to earlier intervals are increasingly localized, suggesting diminishing large-scale deformation.
\emph{Inference:} DDCT-UNet-E2E enters a refinement regime where most discriminative structure has been established, and training focuses on stabilizing class-aligned regions while maintaining SPD validity.

\medskip
\textbf{80\%:}
Geometric drift becomes minimal.
Action classes maintain consistent grouping across subjects, and subject identity is no longer strongly reflected in the tangent-space embedding.
Residual overlap between classes persists.
\emph{Inference:} The model has largely converged to a task-aligned geometry in which covariance representations are optimized for linear separability while avoiding over-contraction or numerical collapse.

\medskip
\textbf{100\% (convergence):}
The final configuration is stable, with no visible large-scale geometric changes.
Class-conditional covariances exhibit the clearest organization among all models considered, while remaining well-spread and SPD-valid.
No signs of degeneracy or rank collapse are observed.
\emph{Inference:} DDCT-UNet-E2E converges to a discriminatively structured covariance space shaped jointly by deep SPD-preserving transformations and tangent-space classification, achieving the strongest alignment between geometry and task objectives.

\medskip
\textbf{Interpretation and architectural implications.}
The interval-wise evolution highlights the effect of fully end-to-end discriminative supervision on deep SPD representations.
By backpropagating classification loss through the tangent-space classifier into the DDCT-UNet encoder--decoder, DDCT-UNet-E2E induces the most class-aware geometric restructuring among all considered architectures.
Compared to DCT, DLDCT, DDCT-UNet, and DLDCT-E2E, this model exhibits the richest and most task-aligned covariance evolution, while the absence of collapse confirms that geometric fidelity on $\mathcal{S}_{++}^d$ is preserved throughout training.

\subsection{Evolution of Fisher Statistics During Training}

\subsubsection{Discriminative Congruence Transform (DCT) Pre-aligner}
\label{app:dct_fisher_evolution}

To complement the qualitative covariance evolution plots, we analyze the evolution of Fisher statistics during training of the DCT pre-aligner.
These diagnostics quantify how class separability evolves under the learned global congruence transformation and provide insight into the limits imposed by DCT’s rotational architecture.
Figure~\ref{fig:dct_fisher_components} shows the evolution of between-class and within-class scatter, while Figure~\ref{fig:dct_fisher_ratio} reports the corresponding Fisher ratio.

\medskip
\textbf{Between-class and within-class scatter evolution.}
Figure~\ref{fig:dct_fisher_components} illustrates the evolution of the between-class scatter ($B_c$) and within-class scatter ($W_c$) over training.
As optimization progresses, the between-class scatter increases monotonically, indicating that class centroids in the tangent space become progressively better separated.
At the same time, the within-class scatter decreases, reflecting tighter alignment of covariance representations belonging to the same action class across subjects.
This joint behavior is precisely aligned with the objective of DCT.

By learning a single global congruence rotation, DCT reorients covariance representations so that class-relevant directions are better aligned across subjects, thereby reducing dispersion within each class while amplifying differences between class means.
Importantly, these changes arise without explicit class-specific deformation or metric contraction, confirming that the observed trends are driven primarily by rotational reorganization rather than artificial clustering.

The gradual saturation of both curves reflects an intrinsic limitation of DCT.
Because the transformation is restricted to a single global congruence mapping on the SPD manifold, the model can only reorient existing variance directions.
Once dominant subject-dependent rotations are corrected and class-relevant axes are aligned, further increases in between-class scatter or decreases in within-class scatter become geometrically infeasible.

\begin{figure}[h]
    \centering
    \includegraphics[width=0.75\linewidth]{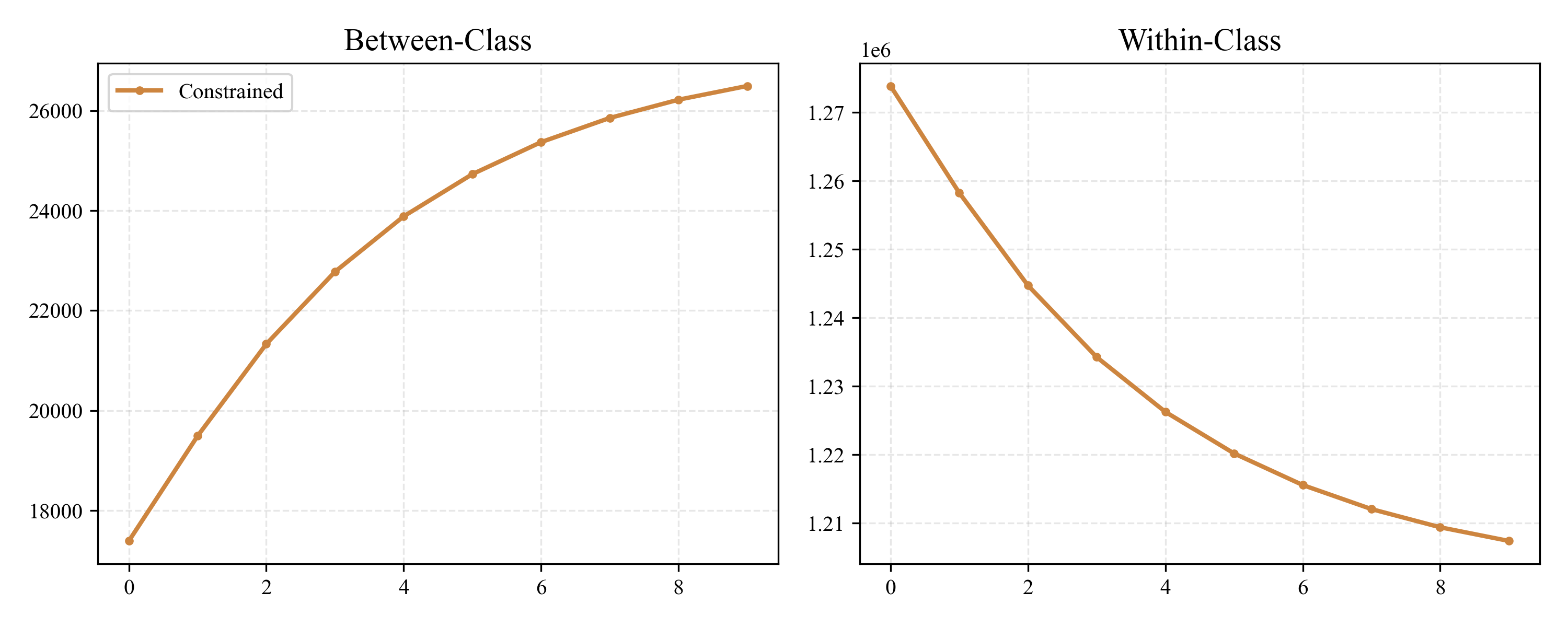}
    \caption{Evolution of Fisher scatter components during DCT training.
The between-class scatter ($B_c$) increases while the within-class scatter ($W_c$) decreases over training, indicating improved class separability induced by the learned global congruence rotation.}
    \label{fig:dct_fisher_components}
\end{figure}

\medskip
\textbf{Fisher ratio evolution.}
Figure~\ref{fig:dct_fisher_ratio} reports the evolution of the Fisher ratio $W_c / B_c$, where lower values indicate better class separability.
Consistent with the trends observed in Figure~\ref{fig:dct_fisher_components}, the Fisher ratio decreases steadily during training.
This decrease results from the combined effect of increasing between-class scatter and decreasing within-class scatter, rather than from changes in either term alone.

The smooth and monotonic reduction of the Fisher ratio indicates stable optimization and the absence of degenerate solutions.
Notably, the curve approaches a plateau in later training stages, mirroring the saturation observed in the scatter components.
This behavior highlights a fundamental property of DCT: while global rotational alignment can substantially improve class discriminability, it cannot indefinitely increase the Fisher criterion.
Once the optimal orientation is reached under the rotational constraint, further improvement would require additional degrees of freedom, such as nonlinear deformation, class-dependent warping, or multi-stage congruence compositions.

\begin{figure}[h]
    \centering
    \includegraphics[width=0.4\linewidth]{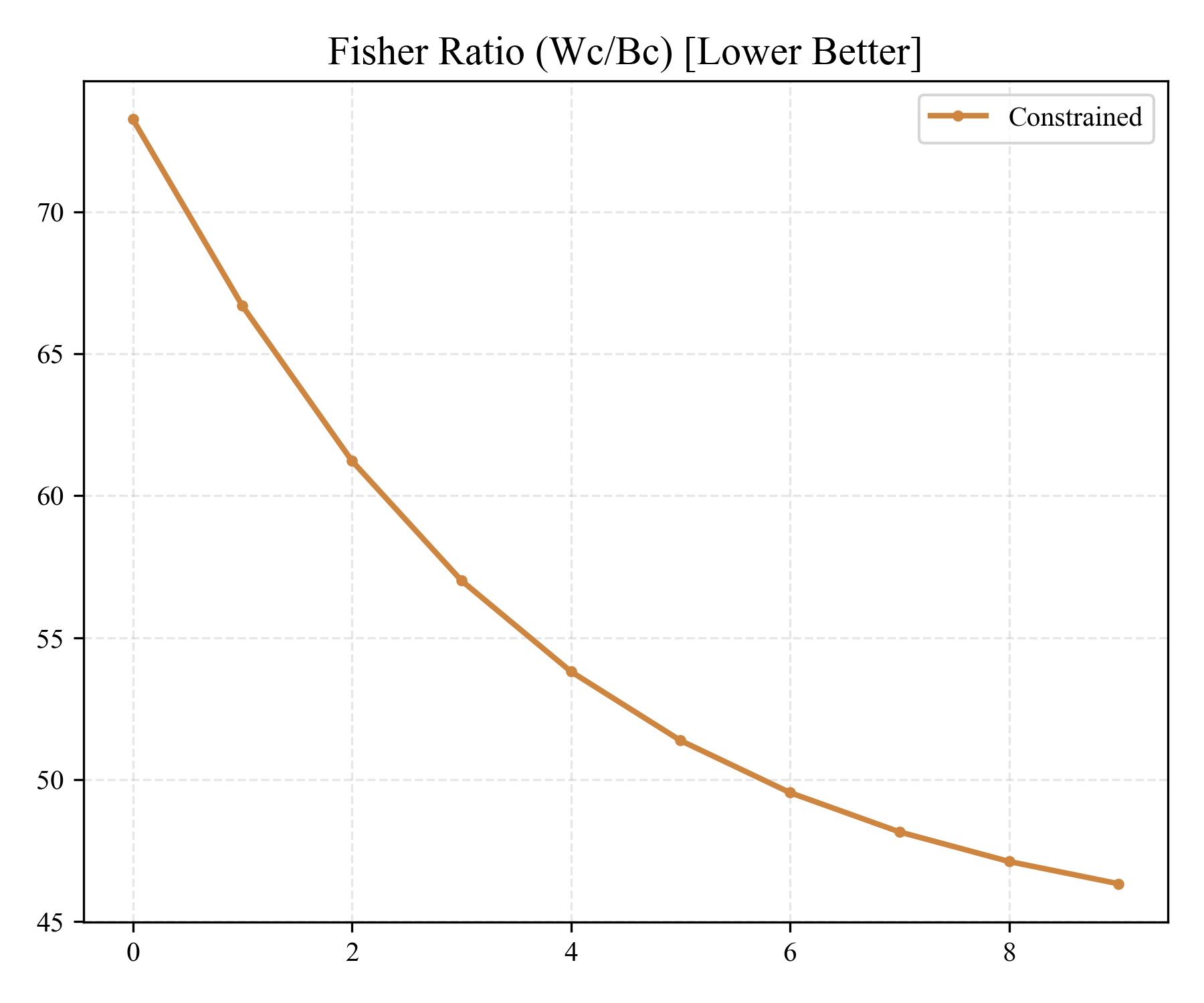}
    \caption{Evolution of the Fisher ratio ($W_c/B_c$) during DCT training (lower is better).
The monotonic decrease reflects simultaneous increase in between-class separation and reduction in within-class dispersion under the rotational constraint of DCT.}
    \label{fig:dct_fisher_ratio}
\end{figure}

Together, these diagnostics confirm that DCT achieves its intended effect: it maximizes class separability as far as permitted by a single congruence transformation.
The increase in $B_c$, decrease in $W_c$, and corresponding reduction in the Fisher ratio provide quantitative evidence that DCT effectively removes subject-dependent rotational bias while preserving the intrinsic overlap characteristic of EEG data.

This saturation behavior further motivates the use of deeper and more expressive architectures such as DLDCT and DDCT-UNet, which extend beyond global rotation to enable nonlinear manifold deformation and progressive class-discriminative geometric restructuring.

\subsubsection{Deep Linear Discriminative Congruence Transform (DLDCT) Pre-aligner}
\label{app:dldct_fisher_evolution}

To complement the qualitative covariance evolution plots, we analyze the evolution of Fisher-style statistics during training of the DLDCT pre-aligner.
These diagnostics quantify how class separability and subject mixing evolve under a \emph{deep} stack of discriminative congruence transformations on $\mathcal{S}_{++}^d$.
Figure~\ref{fig:dldct_fisher_components} reports the evolution of class-wise and subject-wise scatter components, while Figure~\ref{fig:dldct_fisher_ratios} summarizes the corresponding ratios: Fisher ratio ($W_c/B_c$, lower is better) and subject ratio ($W_s/B_s$, higher is better).

\medskip
\textbf{Between/within scatter evolution (class and subject).}
Figure~\ref{fig:dldct_fisher_components} shows four quantities: between-class scatter ($B_c$), within-class scatter ($W_c$), between-subject scatter ($B_s$), and within-subject scatter ($W_s$).
A notable feature of DLDCT is that \emph{all} scatter components decrease over training.
This indicates that the learned deep congruence stack not only improves discriminative structure, but also induces a global contraction of tangent-space dispersion (i.e., an overall tightening of the representation scale), consistent with deep geometric reshaping rather than a pure rotation.

\begin{figure}[h]
    \centering
    \includegraphics[width=0.8\linewidth]{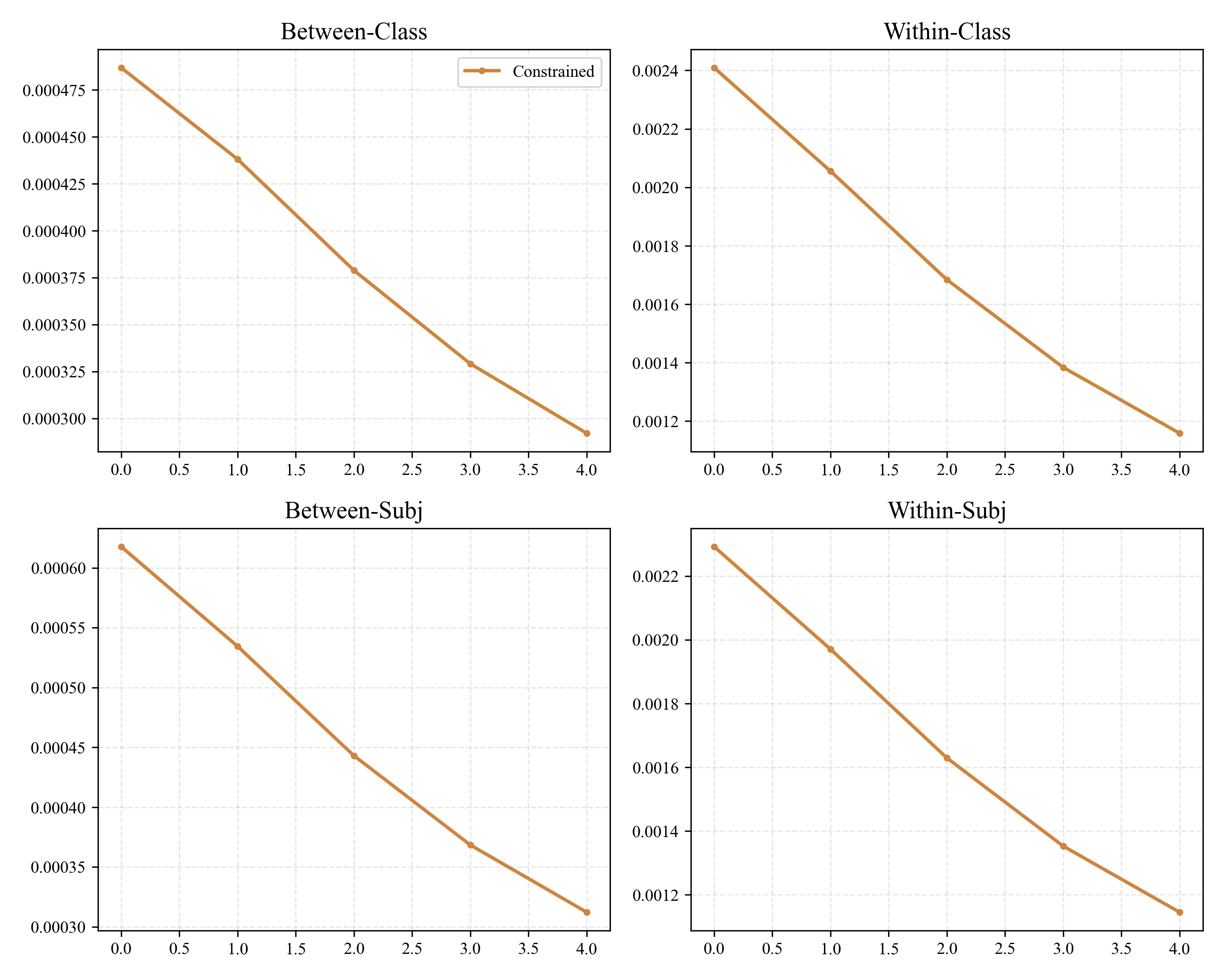}
    \caption{
Evolution of scatter components during DLDCT training.
Top: between-class scatter ($B_c$) and within-class scatter ($W_c$).
Bottom: between-subject scatter ($B_s$) and within-subject scatter ($W_s$).
All components decrease over training, indicating global geometric stabilization; however, $W_c$ contracts faster than $B_c$, yielding improved relative class separability.
}
    \label{fig:dldct_fisher_components}
\end{figure}
\begin{figure}[h]
    \centering
    \includegraphics[width=0.75\linewidth]{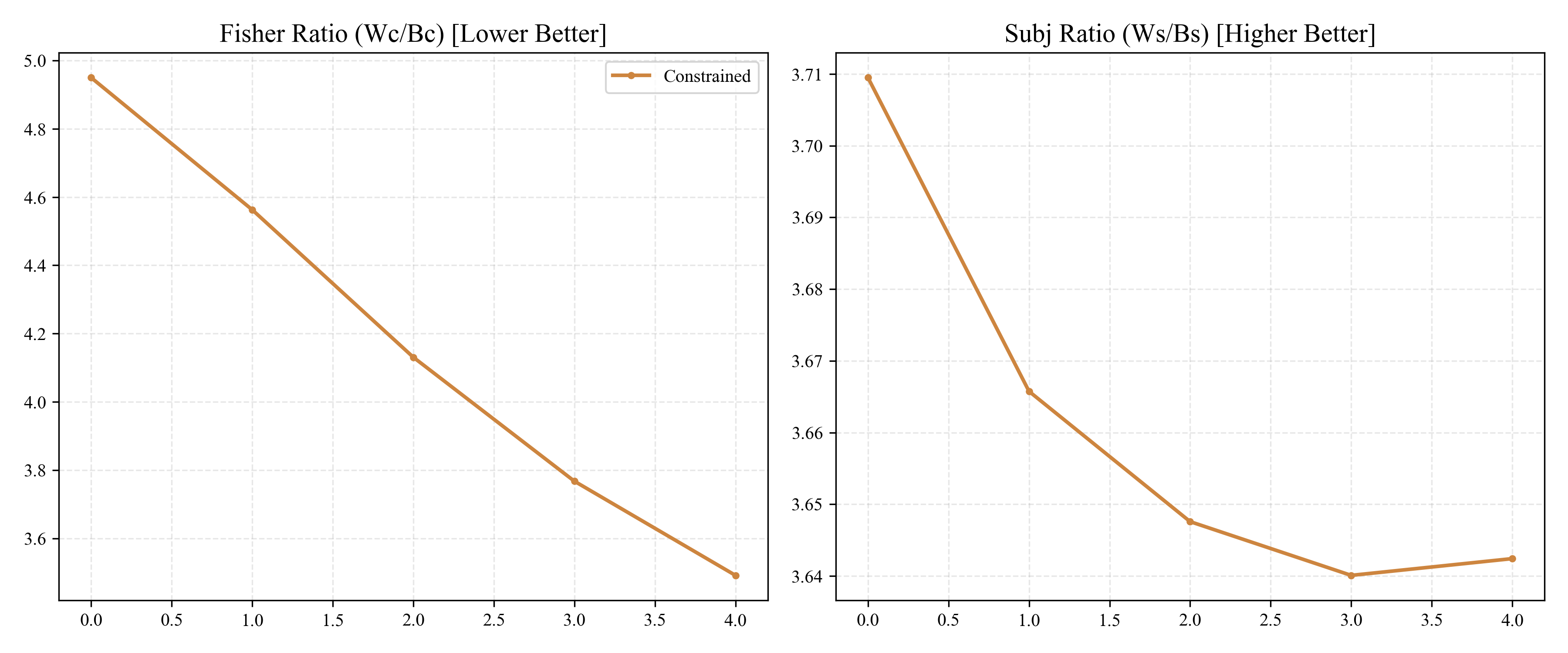}
    \caption{
Evolution of ratio diagnostics during DLDCT training.
Left: Fisher ratio ($W_c/B_c$, lower is better) decreases monotonically, indicating improved class separability via faster contraction of within-class scatter.
Right: subject ratio ($W_s/B_s$, higher is better) shows a slight decrease then stabilizes, suggesting limited explicit pressure toward subject mixing under DLDCT.
}
    \label{fig:dldct_fisher_ratios}
\end{figure}
Crucially, the \emph{within-class} scatter decreases substantially more strongly than the \emph{between-class} scatter.
Thus, even though $B_c$ decreases in absolute magnitude, the class-conditional clouds become tighter at a faster rate than the class centroids collapse, improving relative class separability.
On the subject side, both $B_s$ and $W_s$ decrease in tandem, suggesting that DLDCT suppresses subject-induced spread primarily through overall geometric stabilization rather than explicitly inflating subject mixing.

\medskip
\textbf{Fisher and subject ratio evolution.}
Figure~\ref{fig:dldct_fisher_ratios} reports two ratios.
First, the Fisher ratio $W_c/B_c$ decreases monotonically, indicating steadily improving class separability.
In contrast to DCT—where improvements are driven mainly by reorientation (often increasing $B_c$ while decreasing $W_c$)—DLDCT achieves improvement through \emph{differential contraction}: the within-class term $W_c$ shrinks more rapidly than the between-class term $B_c$.
This behavior is characteristic of deeper congruence compositions, where successive SPD-preserving layers can progressively reshape dispersion patterns instead of performing only a single global reorientation.

Second, the subject ratio $W_s/B_s$ exhibits a mild decrease followed by stabilization (with only marginal change toward the end).
Since higher values are preferable (greater subject mixing / weaker subject separability), this trend suggests that DLDCT does not explicitly optimize for subject invariance as a primary objective; rather, it prioritizes class-discriminative geometry while maintaining a roughly steady level of subject mixing.
Importantly, the subject-ratio curve is smooth and non-degenerate, indicating stable optimization without collapse or pathological concentration.

\medskip
\textbf{Interpretation and architectural implications.}
Overall, the Fisher diagnostics confirm that DLDCT yields \emph{consistent} gains in class separability throughout training, but through a mechanism that differs from DCT.
Whereas DCT is fundamentally limited to a single global congruence reorientation, DLDCT’s depth enables progressive dispersion shaping: the representation stabilizes globally (all scatters decrease), while relative class separability improves because within-class dispersion contracts more strongly than between-class separation.
The comparatively stable (and only mildly decreasing) subject ratio further indicates that DLDCT’s discriminative gains are not simply a by-product of collapsing subject structure, but arise from class-driven geometric restructuring in $\mathcal{S}_{++}^d$.

\subsubsection{Deep Discriminative Congruence Transform--UNet (DDCT--UNet) Pre-aligner}
\label{app:ddctunet_fisher_evolution}

To quantitatively analyze how DDCT--UNet reshapes covariance geometry during training, we examine the evolution of Fisher scatter statistics.
Unlike DCT, which is restricted to a single global congruence rotation, DDCT--UNet employs an SPD-preserving encoder--decoder
$\Psi_{\theta}:\mathcal{S}_{++}^d \rightarrow \mathcal{S}_{++}^d$
that allows nonlinear congruence compositions and controlled deformation of the covariance space.
Figure~\ref{fig:rifu_fisher_components} reports the evolution of between-class, within-class, and within-subject scatter, while Figure~\ref{fig:rifu_fisher_ratio} shows the corresponding Fisher ratio.

\begin{figure}[h]
    \centering
    \includegraphics[width=0.75\linewidth]{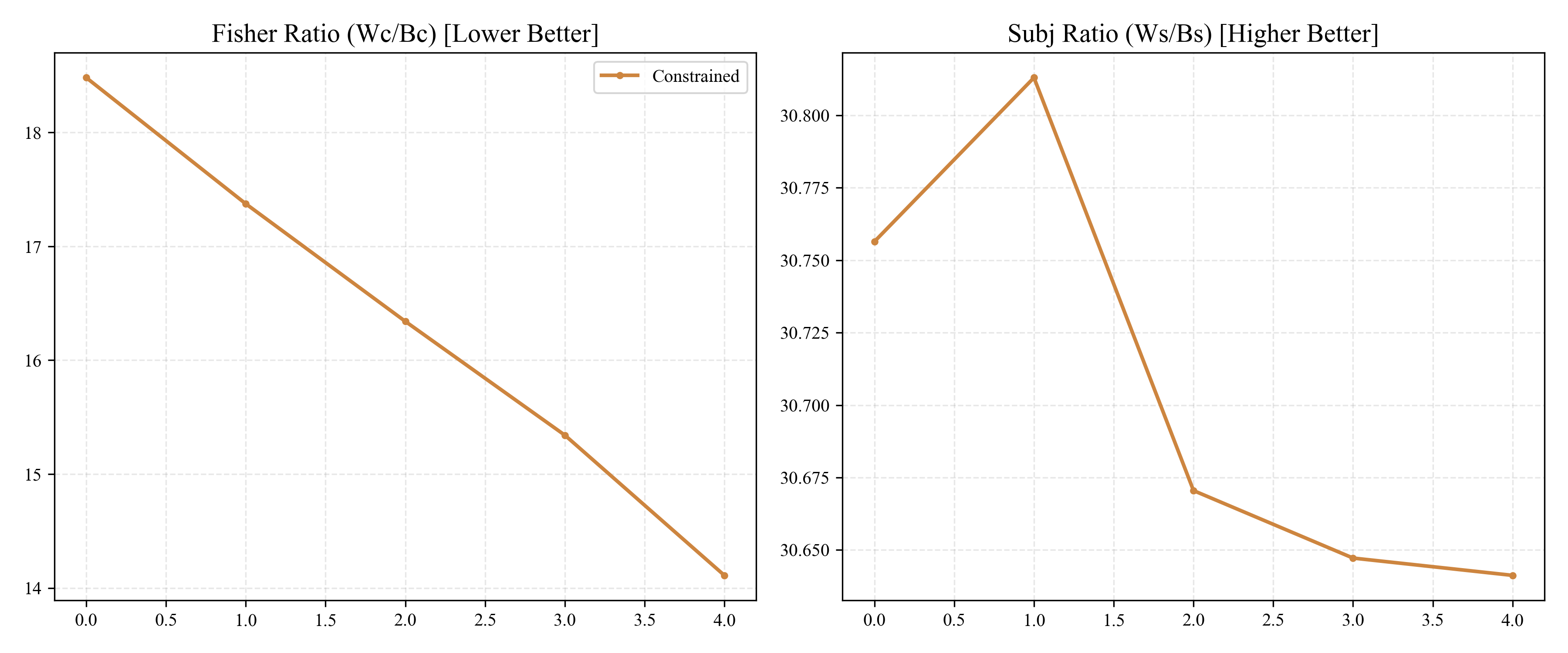}
    \caption{
Evolution of the Fisher ratio ($W_c/B_c$) during DDCT--UNet pre-alignment training (lower is better).
Despite simultaneous increases in both $B_c$ and $W_c$, the ratio decreases monotonically, indicating improved relative class separability in the expanded covariance space.
}
    \label{fig:rifu_fisher_ratio}
\end{figure}

\medskip
\textbf{Between-class scatter.}
The between-class scatter ($B_c$) increases steadily throughout training.
This indicates that class centroids in the tangent space move farther apart as the encoder--decoder progressively reshapes the SPD manifold.
\emph{Interpretation:} DDCT--UNet actively constructs a representation in which action classes occupy increasingly separated regions, facilitating linear discrimination downstream.

\medskip
\textbf{Within-class scatter.}
In contrast to DCT, the within-class scatter ($W_c$) also increases during DDCT--UNet training.
This behavior reflects a deliberate geometric expansion rather than compression.
\emph{Interpretation:} DDCT--UNet does not attempt to contract intra-class structure; instead, it permits within-class variability to grow so long as the relative positioning of class means improves.
Such expansion allows the model to encode richer class-specific covariance geometry without forcing artificial collapse.

\medskip
\textbf{Within-subject scatter.}
The within-subject scatter ($W_s$) exhibits a clear increasing trend.
This indicates that covariance representations corresponding to the same subject become more dispersed in the learned space.
\emph{Interpretation:} An increase in $W_s$ is desirable in the cross-subject setting, as it reflects suppression of subject-specific structure.
By spreading subject-conditioned covariances, DDCT--UNet reduces subject identity information, promoting subject-invariant representations.

\begin{figure}[h]
    \centering
    \includegraphics[width=0.8\linewidth]{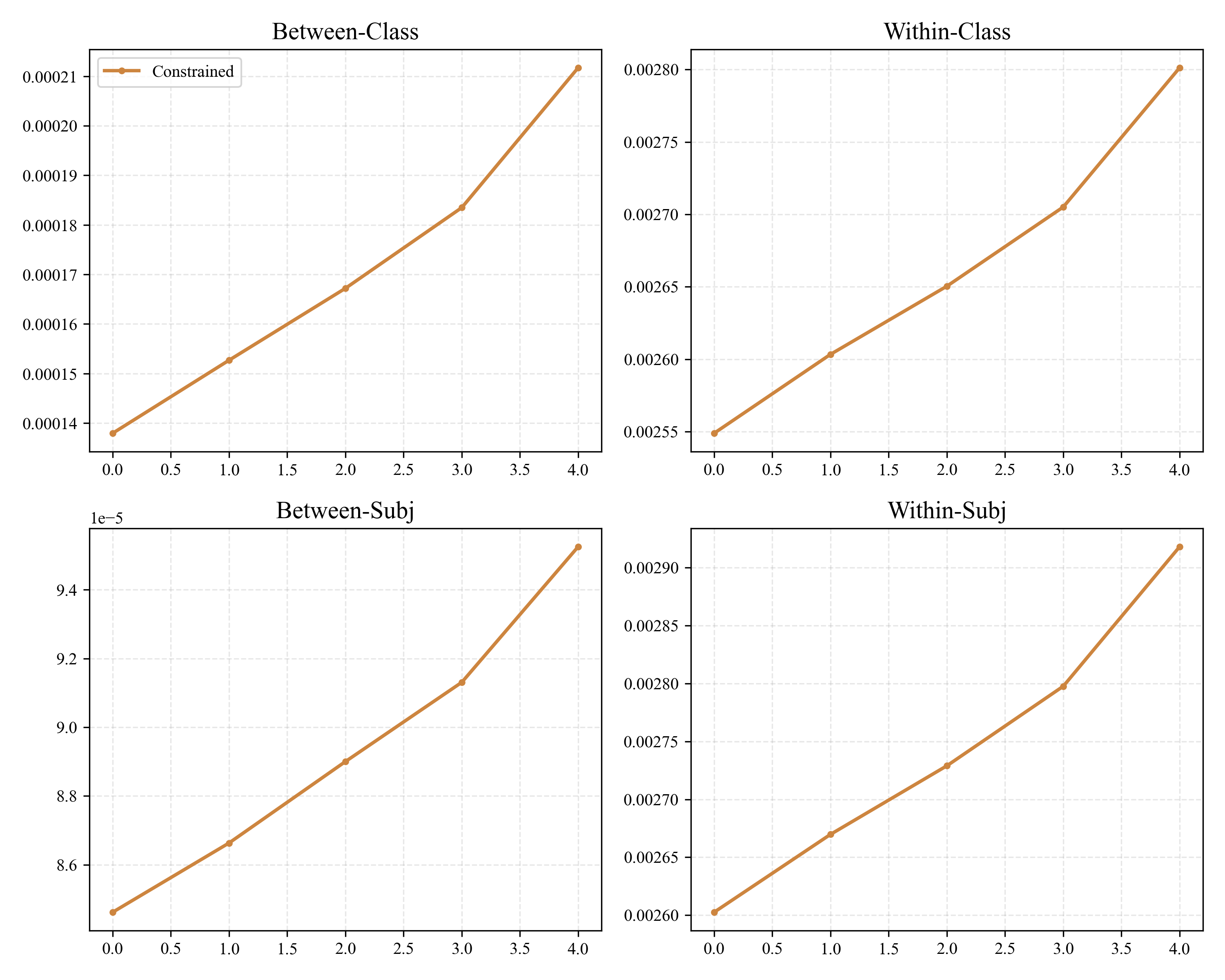}
    \caption{
Evolution of Fisher scatter components during DDCT--UNet pre-alignment training.
Both between-class scatter ($B_c$), within-class scatter ($W_c$), and within-subject scatter ($W_s$) increase over training, indicating an expansion and reorganization of the covariance space induced by the encoder--decoder–based deep congruence transformations.
}
    \label{fig:rifu_fisher_components}
\end{figure}

\medskip
\textbf{Fisher ratio evolution.}
Figure~\ref{fig:rifu_fisher_ratio} shows that the Fisher ratio $W_c / B_c$ decreases monotonically over training.
This decrease occurs despite the simultaneous increase in both $W_c$ and $B_c$.
\emph{Interpretation:} The key improvement lies in the relative scaling: the between-class scatter increases faster than the within-class scatter.
Thus, DDCT--UNet identifies an expanded but more efficient covariance space in which class separation dominates intra-class variability, improving linear discriminability.

\medskip
\textbf{Overall interpretation.}
Together, these trends highlight a fundamental difference between DDCT--UNet and rotation-only pre-aligners such as DCT.
Rather than minimizing variance terms individually, DDCT--UNet reorganizes and expands the covariance space to optimize the relative Fisher criterion.
The increase in $W_s$ promotes subject invariance, while the dominant growth of $B_c$ ensures improved class separability.
This combination yields a representation that is geometrically compatible with downstream classification and explains the improved performance observed under DDCT--UNet pre-alignment.

\subsubsection{Discriminative Congruence Transform--(E2E)--(DCT--E2E) Classifier}
\label{app:dcte2e_fisher_evolution}

We next analyze the Fisher scatter dynamics for the DCT--E2E model, in which a \emph{single} discriminative congruence layer is trained jointly with the downstream classifier through cross-entropy supervision.
Unlike deeper architectures such as DLDCT and DDCT--UNet, DCT--E2E remains a shallow and geometrically constrained mapping on $\mathcal{S}_{++}^d$.
Consequently, its Fisher statistics reflect optimization under severely limited degrees of freedom.
Figures~\ref{fig:dcte2e_fisher_components} and~\ref{fig:dcte2e_fisher_ratios} report the evolution of class-wise scatter components and their corresponding Fisher ratios for both constrained and unconstrained variants.

\medskip
\textbf{Scatter dynamics under shallow end-to-end optimization.}
Figure~\ref{fig:dcte2e_fisher_components} shows that both between-class ($B_c$) and within-class ($W_c$) scatter increase steadily throughout training for both constrained and unconstrained regimes.
This joint growth indicates that DCT--E2E does not primarily operate by contracting class-conditioned clouds, as in DCT, but instead expands the overall tangent-space geometry in directions favored by the cross-entropy loss.

Because the model has only a single congruence transformation, it cannot independently control scale, class compactness, and subject mixing.
Instead, the optimizer inflates variance along discriminative directions, producing a global geometric expansion rather than targeted manifold sculpting.
The unconstrained variant exhibits consistently larger scatter values, reflecting its freedom to exploit additional scaling and noisy degrees of freedom.

\medskip
\textbf{Fisher ratio evolution and apparent contradictions.}
Figure~\ref{fig:dcte2e_fisher_ratios} reports the Fisher ratio $W_c/B_c$ (lower is better) for both variants.
Despite the increase in both $W_c$ and $B_c$, the ratio decreases monotonically, indicating that between-class separation grows faster than within-class dispersion.
Thus, DCT--E2E converges to a representation that is locally optimal for classification under its architectural constraints.
\begin{figure}[h]
    \centering
    \includegraphics[width=0.75\linewidth]{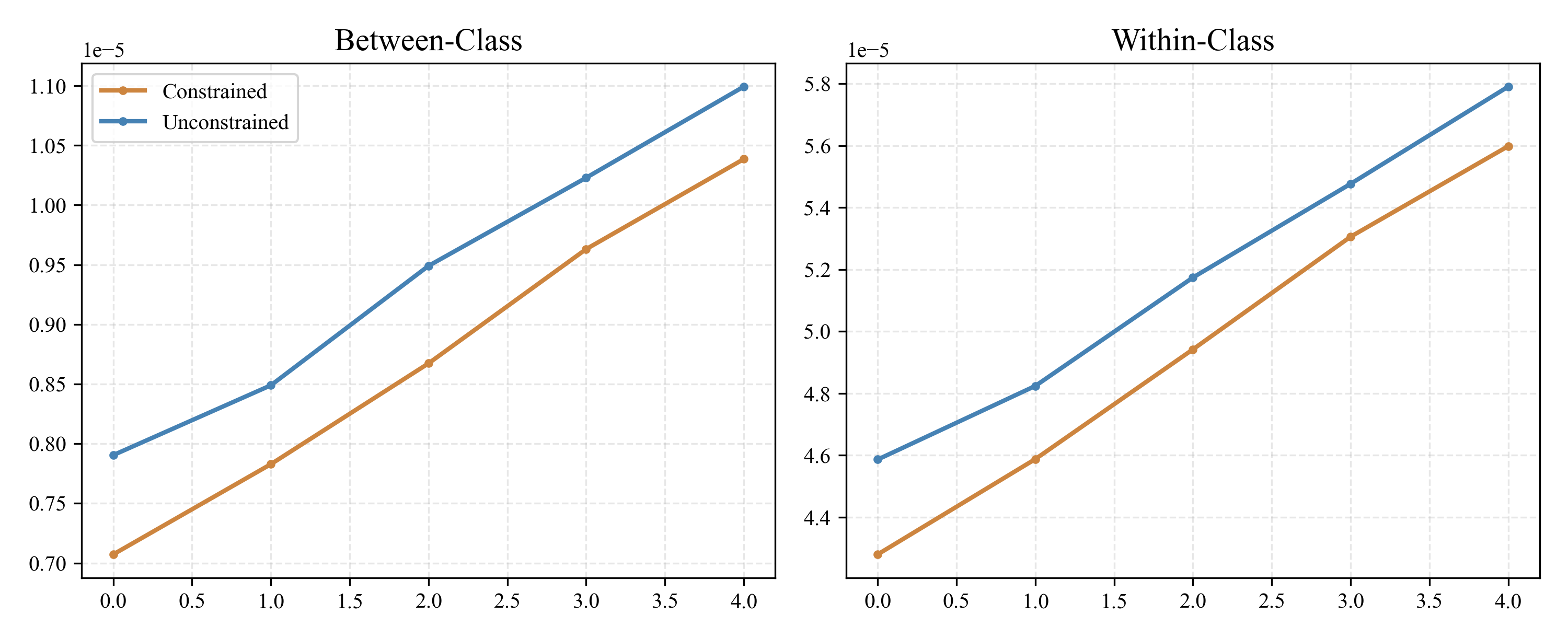}
    \caption{
Evolution of scatter components during DCT--E2E training.
Curves correspond to constrained and unconstrained optimization regimes.
Both between-class ($B_c$) and within-class ($W_c$) scatter increase monotonically, reflecting task-driven expansion of the tangent-space geometry under a shallow congruence architecture.
}
    \label{fig:dcte2e_fisher_components}
\end{figure}
Notably, the unconstrained variant achieves a consistently lower Fisher ratio.
This does \emph{not} imply superior geometric alignment.
Rather, the absence of spectral or norm constraints allows the model to inject additional variance directions, effectively inflating within-class scatter in a way that favors the linear classifier.
Such behavior reduces the ratio numerically but may correspond to less stable or less interpretable geometry—hence the importance of constrained training for principled SPD manipulation.

\begin{figure}[h]
    \centering
    \includegraphics[width=0.4\linewidth]{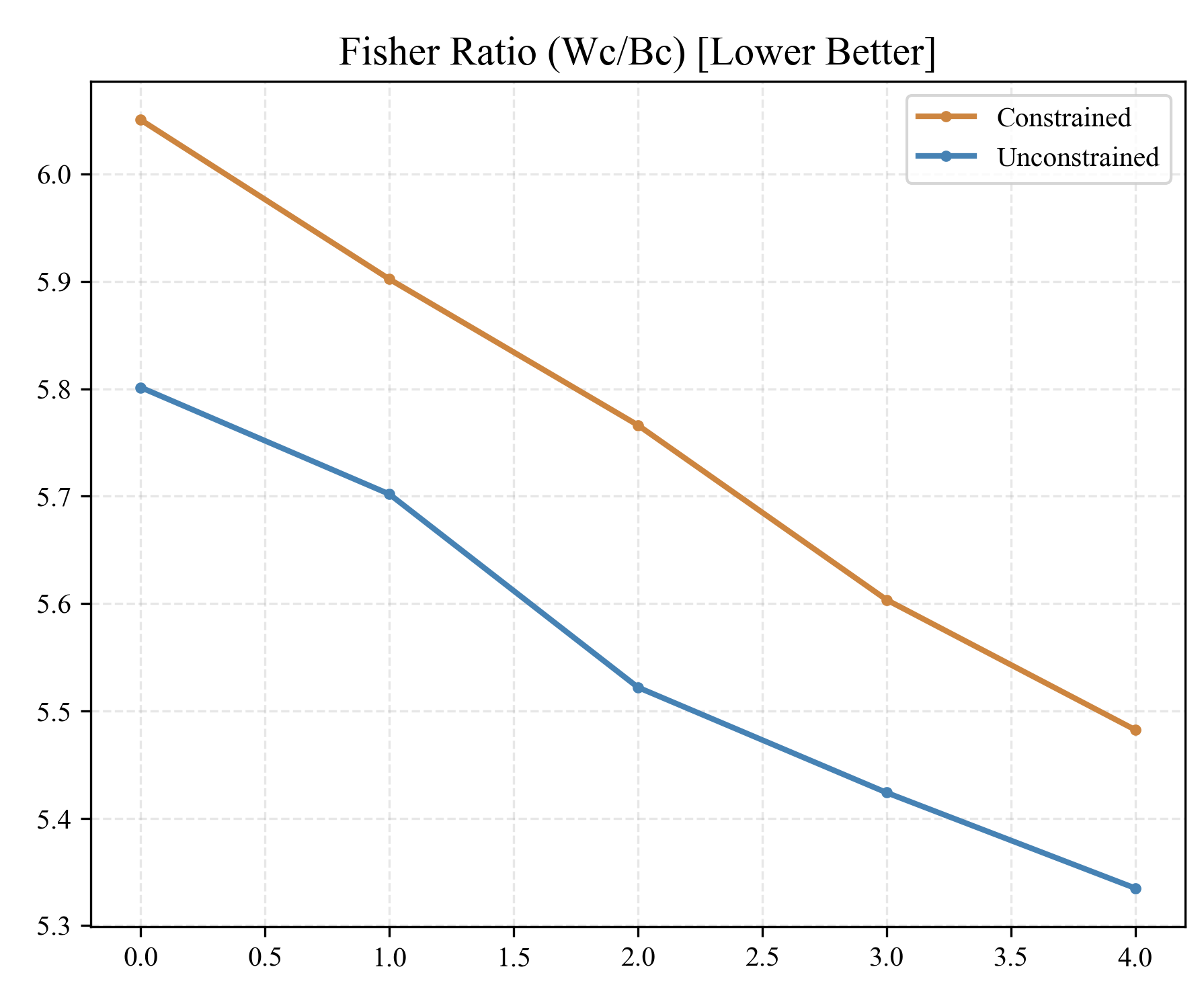}
    \caption{
Evolution of the Fisher ratio ($W_c/B_c$) during DCT--E2E training for constrained and unconstrained regimes.
Both curves decrease monotonically, reflecting improved relative class separability.
The lower ratio attained by the unconstrained variant arises from its freedom to exploit scale and noise directions in the shallow geometry.
}
    \label{fig:dcte2e_fisher_ratios}
\end{figure}

\medskip
\textbf{Interpretation and architectural implications.}
Taken together, these diagnostics demonstrate that DCT--E2E operates in a severely restricted geometric regime.
Cross-entropy supervision drives the single congruence layer toward a task-optimal orientation and scale, but the model lacks the depth required to independently suppress subject variability while tightening class clusters.
Apparent contradictions in absolute scatter growth versus ratio improvement therefore reflect \emph{optimization under constraint} rather than instability.

The distinction between constrained and unconstrained training further underscores the role of geometric regularization in shallow SPD networks.
Without constraints, the optimizer can exploit noisy scaling directions to improve classification metrics, whereas constrained training yields a more balanced geometry that better reflects principled manifold manipulation.

These observations reinforce the central thesis of the paper: progressively deeper congruence architectures—DLDCT, DDCT--UNet, and their end-to-end variants—are necessary to achieve controlled geometric reshaping of $\mathcal{S}_{++}^d$, while shallow end-to-end transforms primarily rely on coarse global deformation.

\subsubsection{Deep Linear Discriminative Congruence Transform--(E2E)--(DLDCT--E2E) Classifier}
\label{app:dldcte2e_fisher_evolution}

To analyze how deep congruence architectures reshape covariance geometry under discriminative supervision, we examine the evolution of Fisher scatter statistics for the DLDCT--E2E classifier.
DLDCT--E2E consists of stacked SPD-preserving congruence layers followed by a discriminative tangent-space readout, but unlike DDCT--UNet and DDCT--UNet--E2E, it does not enforce explicit skip connections or reconstruction-based constraints that preserve proximity to the input covariance structure.
As a result, DLDCT--E2E admits greater geometric freedom during optimization, allowing learned representations to deviate more aggressively from their initialization.

\begin{figure}[h]
    \centering
    \includegraphics[width=0.75\linewidth]{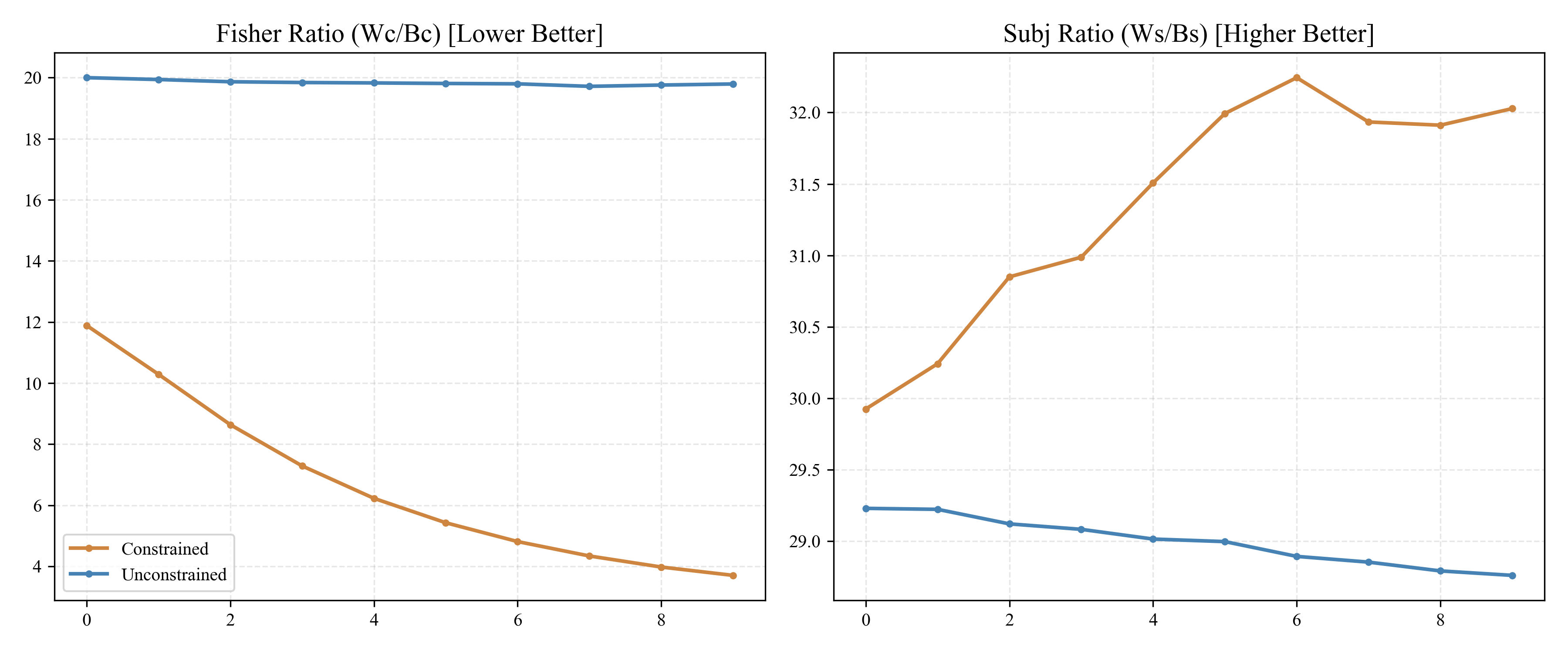}
    \caption{
Relative Fisher ratio dynamics under DLDCT--E2E.
Class separability ($W_c/B_c$, lower is better) and subject mixing ($W_s/B_s$, higher is better) are reported for unconstrained (blue) and constrained (orange) training.
The constrained objective optimizes relative geometry for discrimination and generalization, while the unconstrained model shows limited regulation.
}
    \label{fig:spdnet_fisher_ratios}
\end{figure}
Figure~\ref{fig:spdnet_fisher_components} reports the evolution of between-class ($B_c$), within-class ($W_c$), between-subject ($B_s$), and within-subject ($W_s$) scatter terms.
Figure~\ref{fig:spdnet_fisher_ratios} reports the corresponding relative ratios.
In all plots, the unconstrained variant is shown in blue, while the constrained objective is shown in orange.

\medskip
\textbf{Architectural distinction from DDCT--UNet variants.}
A key difference from DDCT--UNet and DDCT--UNet--E2E is that DLDCT--E2E does not explicitly preserve or reconstruct the original covariance geometry.
In UNet-style models, encoder--decoder skip connections and reconstruction losses anchor the representation close to the input SPD manifold, which naturally regulates geometric deformation.
In contrast, DLDCT--E2E applies a sequence of learned congruence maps without such anchoring.
This loosens the optimization landscape and facilitates gradient propagation, enabling the network to contract or reshape variance aggressively when doing so improves the classification objective.

\medskip
\textbf{Class and subject scatter evolution.}
As shown in Figure~\ref{fig:spdnet_fisher_components}, the unconstrained DLDCT--E2E exhibits weak or nearly flat trends across several scatter components.
This indicates that without explicit geometric regularization, deep congruence stacks do not consistently reorganize covariance geometry in a manner favorable for cross-subject generalization, despite their expressive capacity.

\begin{figure}[h]
    \centering
    \includegraphics[width=0.75\linewidth]{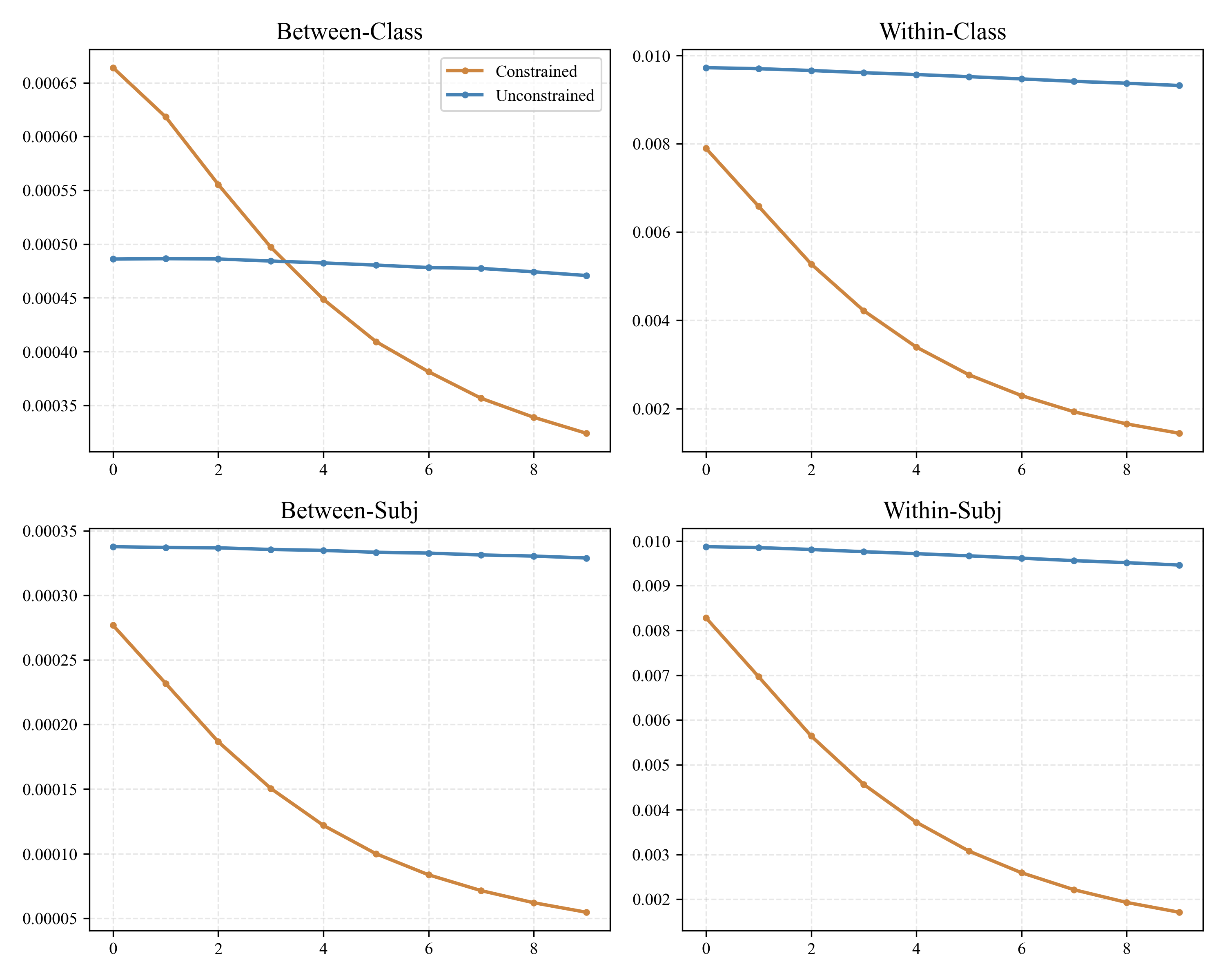}
    \caption{
Evolution of Fisher scatter components under DLDCT--E2E training.
Between-class ($B_c$), within-class ($W_c$), between-subject ($B_s$), and within-subject ($W_s$) variances are shown for unconstrained (blue) and constrained (orange) objectives.
The constrained loss induces controlled geometric contraction, reducing subject-dependent structure while preserving relative class discrimination.
}
    \label{fig:spdnet_fisher_components}
\end{figure}

In contrast, the constrained DLDCT--E2E variant exhibits a monotonic decrease across all scatter terms.
Both between-class ($B_c$) and within-class ($W_c$) variance decrease over training, reflecting the model’s ability to contract the representation space along directions that are not essential for classification.
This behavior does not indicate representational collapse.
Rather, it reflects that DLDCT--E2E is free to compress geometry so long as relative discriminative structure is preserved.

At the subject level, both $B_s$ and $W_s$ decrease substantially under constraints.
The stronger reduction in $B_s$ suppresses subject separation, while the remaining within-subject variability becomes less structured.
This discourages subject-specific clustering and promotes subject mixing, which is desirable for cross-subject decoding.
In contrast, the unconstrained model retains higher subject scatter, signaling persistence of subject identity cues.

\medskip
\textbf{Relative Fisher ratio dynamics.}
Because absolute scatter magnitudes are not the optimization target, Figure~\ref{fig:spdnet_fisher_ratios} highlights relative Fisher-style criteria that govern discriminability. For action classes, the Fisher ratio $W_c / B_c$ decreases sharply under the constrained objective.
Although both $W_c$ and $B_c$ contract, within-class variance is reduced more aggressively, yielding improved relative class separability.
This demonstrates that DLDCT--E2E can achieve effective discrimination through controlled geometric contraction rather than expansion.

For subjects, the ratio $W_s / B_s$ increases under the constrained objective.
This indicates that subject-specific separation is suppressed more strongly than residual subject variability, making subject identity harder to infer from the learned representation.
In contrast, the unconstrained model exhibits a decreasing subject ratio, signaling increased subject separation and a higher risk of subject memorization.

\medskip
\textbf{Interpretation and architectural implications.}
Taken together, these results show that DLDCT--E2E operates in a fundamentally different geometric regime from DDCT--UNet–based models.
Whereas UNet-style architectures explicitly regulate deformation through reconstruction and skip connections, DLDCT--E2E benefits from a looser structure that enables strong geometric contraction when beneficial.

Crucially, the constrained DLDCT--E2E optimizes \emph{relative} Fisher criteria rather than absolute scatter magnitudes, achieving lower class Fisher ratios and higher subject ratios.
This confirms that deep end-to-end congruence stacks can balance class discrimination and subject invariance when guided by appropriate geometric regularization, and that such constraints are essential for steering expressive SPD networks toward generalizable solutions.

\subsubsection{Deep Discriminative Congruence Transform--UNet--(E2E)--(DDCT--UNet--E2E) Classifier}
\label{app:ddctunete2e_fisher_evolution}

To further understand how the DDCT--UNet--E2E classifier reshapes covariance geometry under end-to-end supervision, we analyze the evolution of Fisher scatter statistics during training.
DDCT--UNet--E2E couples an SPD-preserving encoder--decoder
$\Psi_{\theta}:\mathcal{S}_{++}^d \rightarrow \mathcal{S}_{++}^d$
with a task-specific tangent-space linear readout and propagates classification loss gradients directly into the covariance transformation.
Unlike pre-alignment methods such as DCT and DDCT--UNet, the geometry here is shaped entirely by discriminative objectives, making Fisher statistics a meaningful diagnostic of how the learned representation balances class separability and subject invariance.

Figure~\ref{fig:rifunet_fisher_components} reports the evolution of between-class ($B_c$), within-class ($W_c$), between-subject ($B_s$), and within-subject ($W_s$) scatter terms.
Figure~\ref{fig:rifunet_fisher_ratios} shows the corresponding relative ratios that govern class discrimination and subject invariance.
In all plots, the unconstrained variant is shown in blue, while the constrained objective is shown in orange.
\begin{figure}[h]
    \centering
    \includegraphics[width=0.75\linewidth]{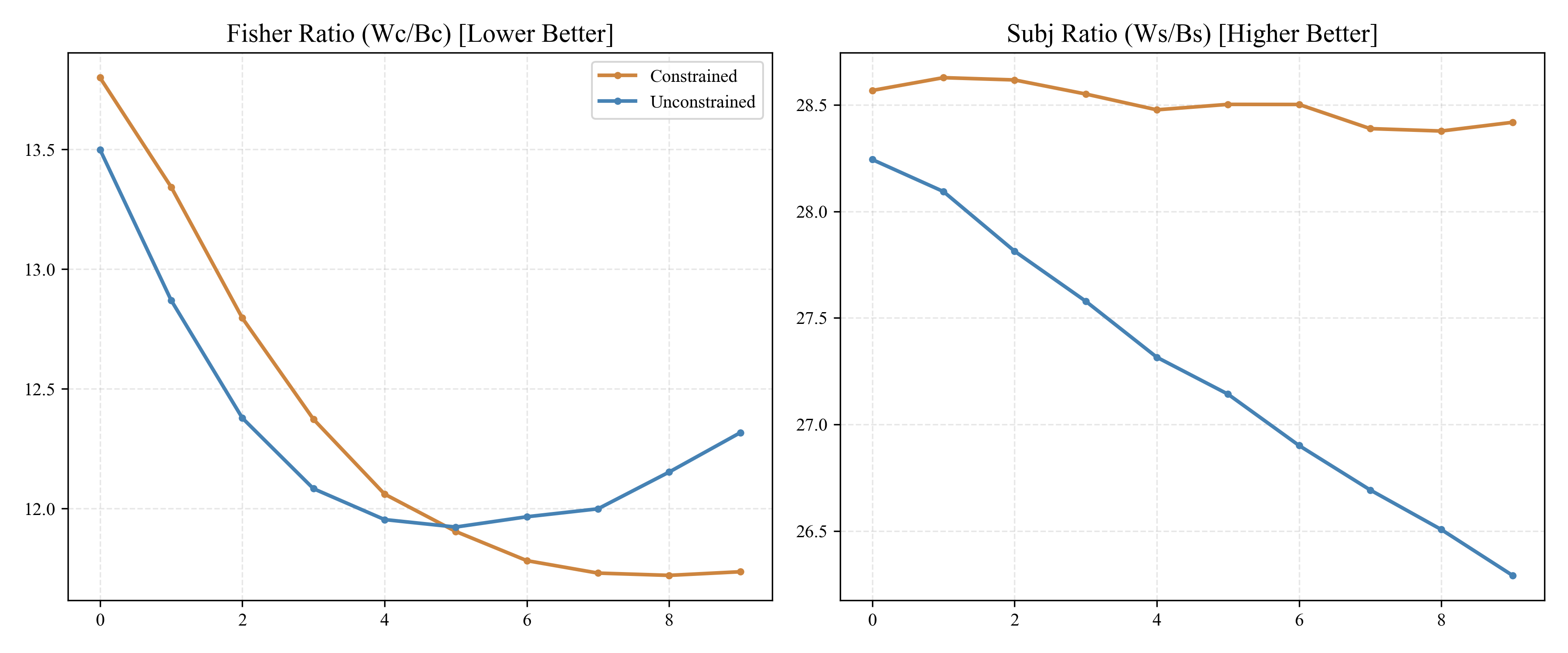}
    \caption{
Evolution of relative Fisher-style ratios during DDCT--UNet--E2E training.
Orange denotes the constrained objective and blue the unconstrained variant.
The constrained model achieves a lower class Fisher ratio ($W_c/B_c$) and a higher subject ratio ($W_s/B_s$), indicating improved class separability and subject invariance.
}
    \label{fig:rifunet_fisher_ratios}
\end{figure}

\medskip
\textbf{Class and subject scatter evolution.}
As shown in Figure~\ref{fig:rifunet_fisher_components}, all scatter terms increase during training for both variants, reflecting that end-to-end learning expands and reorganizes the covariance space to increase representational capacity.
However, the magnitude and implications of this expansion differ substantially between the constrained and unconstrained objectives.

Under the unconstrained objective (blue), both between-class ($B_c$) and within-class ($W_c$) scatter grow rapidly and without regulation.
While the increase in $B_c$ initially promotes class separation, the simultaneous and disproportionate growth of $W_c$ reflects uncontrolled intra-class dispersion.
Such excessive expansion can lead to overfitting, as class regions become overly diffuse and sensitive to noise.

\begin{figure}[h]
    \centering
    \includegraphics[width=0.8\linewidth]{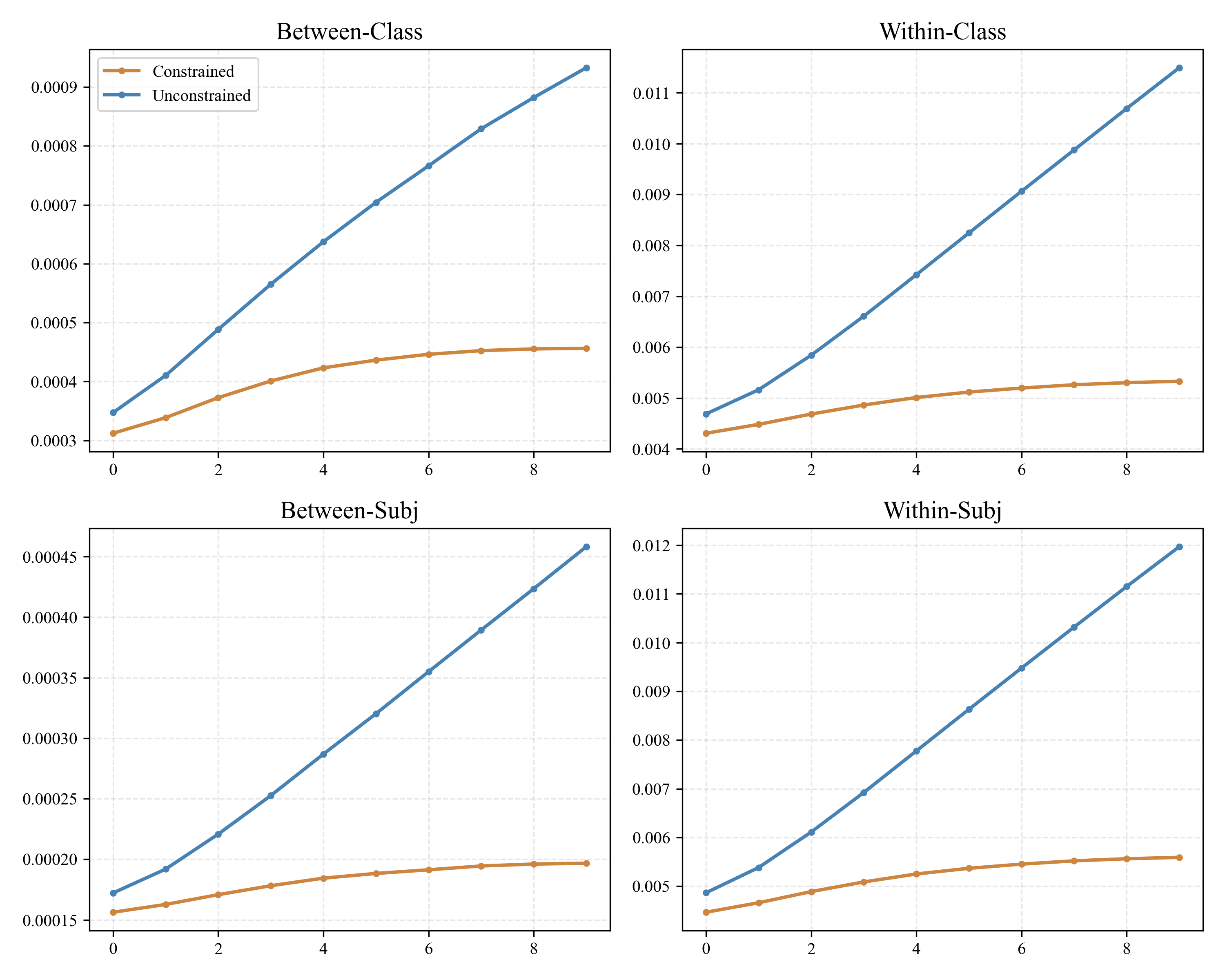}
    \caption{
Evolution of class- and subject-level scatter components during DDCT--UNet--E2E training.
Orange denotes the constrained objective, while blue denotes the unconstrained variant.
All scatter terms increase as the representation space expands under end-to-end learning.
}
    \label{fig:rifunet_fisher_components}
\end{figure}

In contrast, under the constrained objective (orange), the growth of both $B_c$ and $W_c$ is deliberately moderated.
The constrained model still expands the representation space, but does so in a controlled manner that limits unnecessary intra-class spread.
This restriction prevents runaway variance growth while preserving sufficient expressivity for discrimination.

A similar contrast is observed at the subject level.
For the unconstrained model, both between-subject ($B_s$) and within-subject ($W_s$) scatter increase sharply, indicating that subject-specific structure is amplified rather than suppressed.
This behavior encourages subject memorization and is detrimental to cross-subject generalization.
In contrast, the constrained objective restricts the growth of subject-level scatter, preventing excessive separation or amplification of subject identity.

\medskip
\textbf{Relative Fisher ratio dynamics.}
Because absolute scatter magnitudes are not directly indicative of classification or generalization performance, Figure~\ref{fig:rifunet_fisher_ratios} reports relative Fisher-style ratios that capture the effective geometry.

For action classes, the Fisher ratio $W_c / B_c$ decreases more consistently under the constrained objective.
Although both $W_c$ and $B_c$ increase, the constrained model ensures that $B_c$ grows faster relative to $W_c$, resulting in improved class separability.
In contrast, the unconstrained model exhibits weaker reduction—or partial reversal—of this ratio as excessive within-class expansion offsets gains in between-class separation.

For subjects, the subject ratio $W_s / B_s$ remains higher under the constrained objective.
This reflects increased mixing of subject-specific representations, making subject identity more difficult to recover.
In contrast, the unconstrained model exhibits a steady decline in this ratio, indicating progressive separation of subjects and increased risk of subject overfitting.

\medskip
\textbf{Interpretation and model validation.}
Taken together, these results demonstrate that fully end-to-end discriminative learning in deep UNet-style SPD networks must be geometrically regulated to achieve meaningful generalization.
While unconstrained optimization aggressively expands the covariance space, it does so in a manner that amplifies both class and subject variability, leading to overfitting.
The constrained objective limits this expansion, ensuring that increases in scatter occur only insofar as they improve relative discriminative criteria.

The ratio plots provide decisive quantitative evidence of this effect.
The constrained DDCT--UNet--E2E model achieves a lower class Fisher ratio and a higher subject ratio, indicating improved action discrimination alongside suppressed subject identifiability.
This confirms that the constrained formulation reshapes the SPD manifold into a representation that is simultaneously discriminative for actions and invariant to subject identity.
The observed Fisher statistics therefore provide geometry-aware validation that DDCT--UNet--E2E operates in a regime well suited for robust cross-subject classification.

\subsection{Cross entropy diagnostics}

\subsubsection{Discriminative Congruence Transform--End--to--End (DCT--E2E) Classifier}

We analyze the optimization behavior of the DCT--E2E classifier through the
evolution of cross-entropy loss under constrained and unconstrained objectives.
Unlike the DCT pre-aligner, where training is driven exclusively by Fisher-style
alignment criteria, DCT--E2E propagates discriminative gradients directly into the
single congruence transformation, allowing the classifier to adapt its geometry
in response to task supervision.

\begin{figure}[h]
    \centering
    \includegraphics[width=0.4\linewidth]{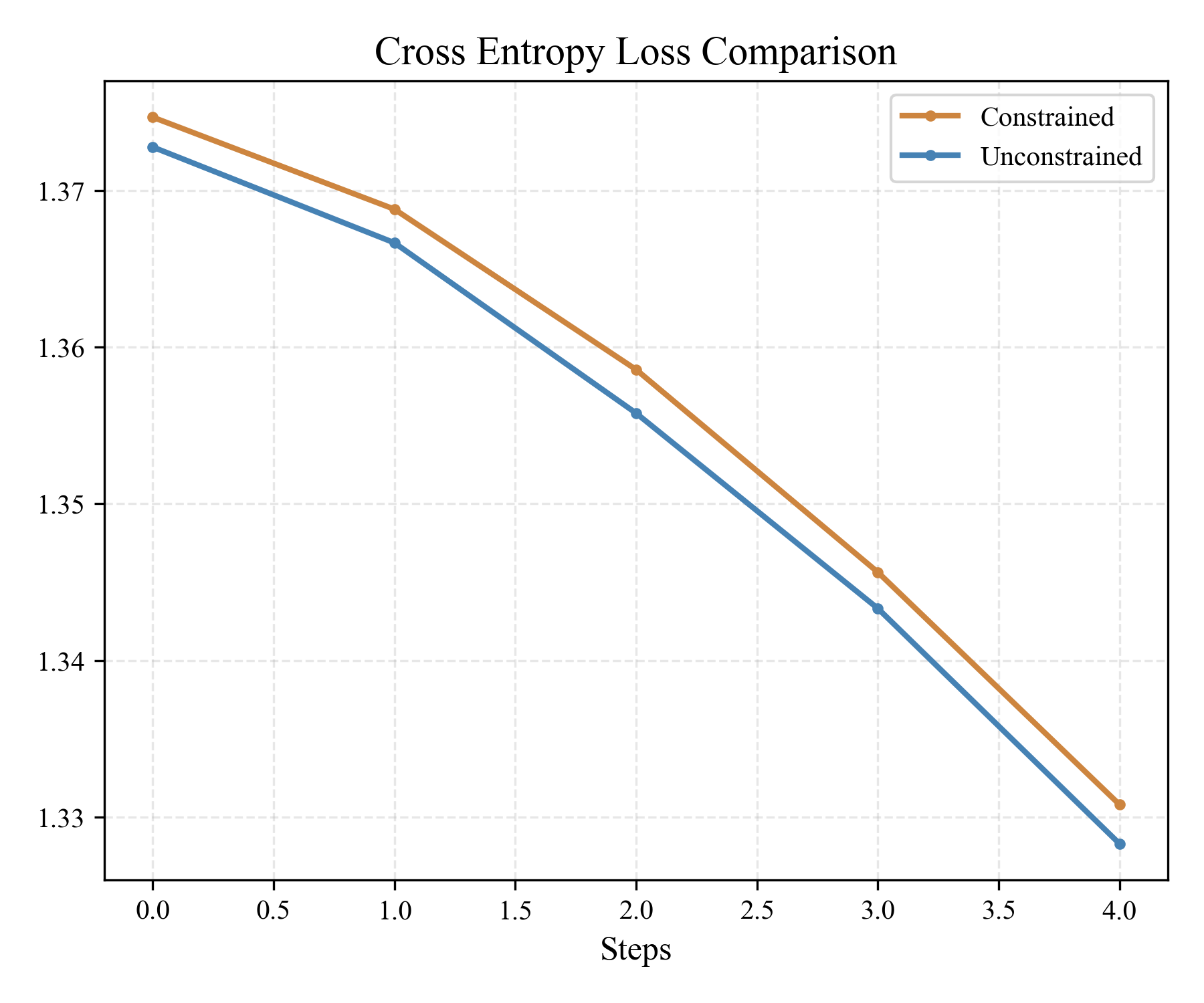}
    \caption{
Cross-entropy loss during DCT--E2E training.
Both constrained (orange) and unconstrained (blue) objectives converge steadily,
reflecting reliable end-to-end optimization through the single congruence layer.
}
    \label{fig:dct_e2e_ce}
\end{figure}

Figure~\ref{fig:dct_e2e_ce} shows smooth and monotonic convergence for both
training regimes, indicating stable optimization on the SPD manifold.
The unconstrained variant achieves marginally lower final loss, as expected: in
the absence of explicit geometric regularization, the model can exploit residual
degrees of freedom to better fit training data.

In contrast, the constrained objective converges to a slightly higher final loss.
This behavior reflects a deliberate trade-off, where geometric regularization
restricts subject-specific shortcuts and uncontrolled deformation, sacrificing a
small amount of training performance in exchange for improved invariance and
generalization.

Taken together, these results characterize DCT--E2E as a shallow end-to-end
congruence classifier whose discriminative capacity arises primarily from learning
a global transformation, while explicit regularization governs the degree to which
classification loss can be minimized.

% ============================================================

\subsubsection{Deep Linear Discriminative Congruence Transform--(E2E)--(DLDCT--E2E) Classifier}
\label{app:dldcte2e_ce_evolution}

We analyze the cross-entropy loss evolution for the DLDCT--E2E classifier under
constrained and unconstrained objectives.
Figure~\ref{fig:spdnet_ce} reports the corresponding loss trajectories.

\begin{figure}[h]
    \centering
    \includegraphics[width=0.4\linewidth]{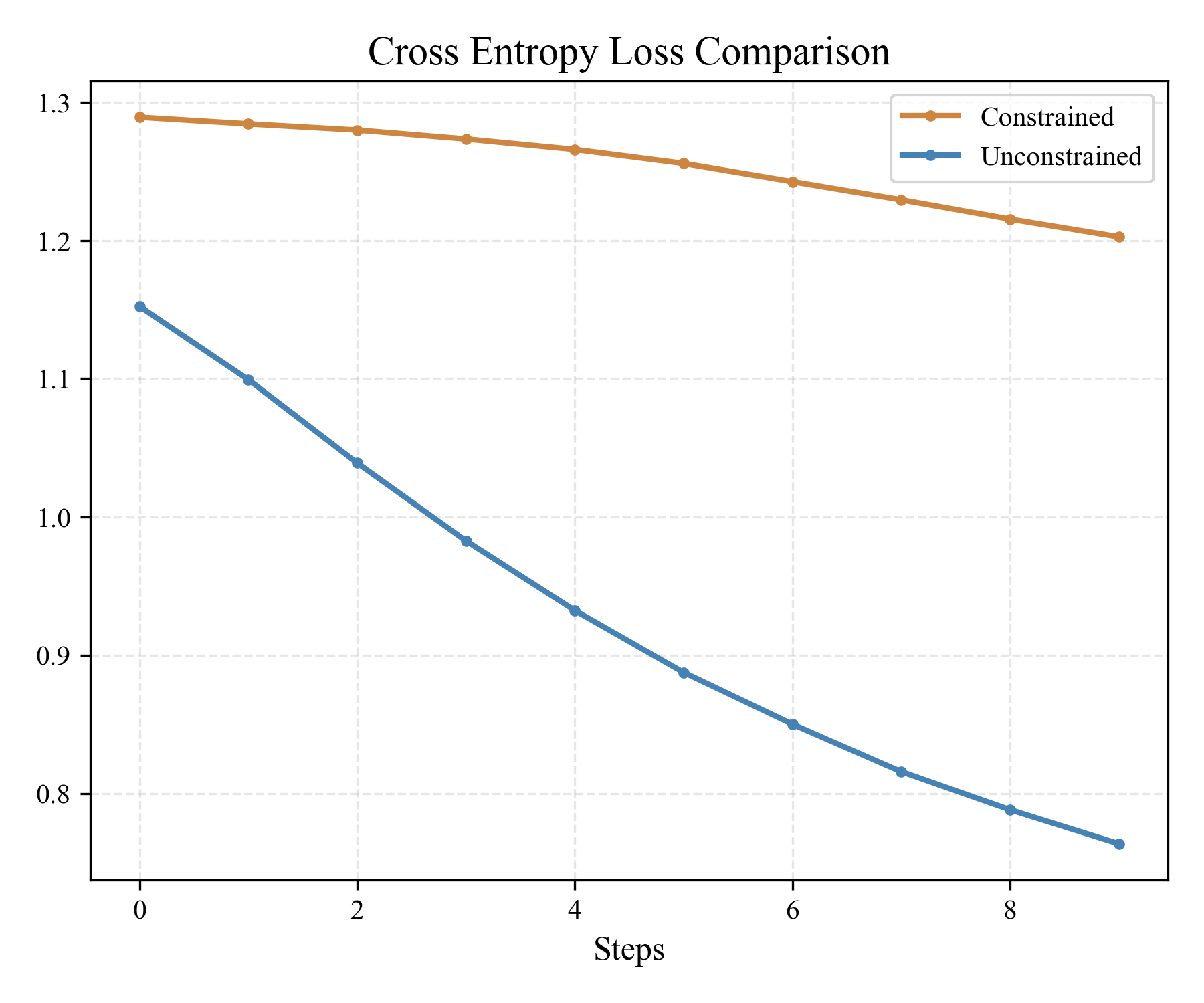}
    \caption{
Evolution of cross-entropy loss for DLDCT--E2E under constrained and unconstrained
training.
Geometric regularization limits loss minimization by preventing subject identity
memorization, resulting in higher but more robust optimization behavior.
}
    \label{fig:spdnet_ce}
\end{figure}

As shown in Figure~\ref{fig:spdnet_ce}, the unconstrained DLDCT--E2E achieves
substantially lower cross-entropy loss throughout training.
This behavior reflects the high flexibility of deep congruence stacks when
optimized solely for classification, enabling rapid exploitation of
subject-specific covariance patterns. In contrast, the constrained DLDCT--E2E maintains consistently higher
cross-entropy loss.
This gap is a direct consequence of the imposed geometric regularization, which
restricts uncontrolled deformation and discourages encoding of subject identity.
Rather than indicating inferior optimization, the elevated loss under the
constrained objective signals that the model is prevented from over-specializing
to training subjects. Taken together, these trajectories characterize DLDCT--E2E as a highly expressive
end-to-end congruence architecture whose unconstrained optimization aggressively
minimizes classification loss, while explicit geometric constraints steer learning
toward representations that favor cross-subject generalization.

% ============================================================

\subsubsection{Deep Discriminative Congruence Transform--UNet--(E2E)
(DDCT--UNet--E2E) Classifier}
\label{app:ddctunete2e_ce_evolution}

\begin{figure}[h]
    \centering
    \includegraphics[width=0.4\linewidth]{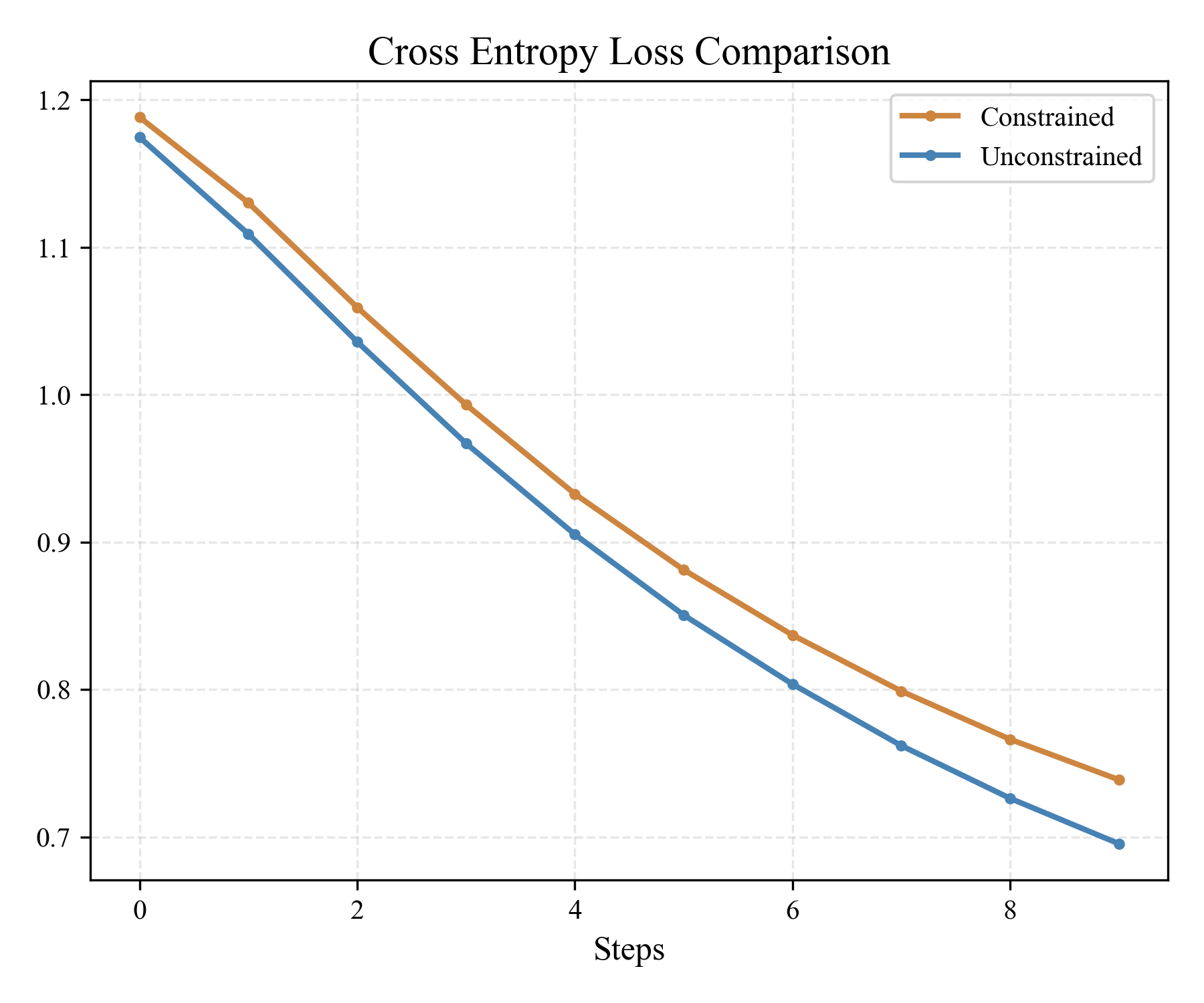}
    \caption{
Cross-entropy loss trajectories for DDCT--UNet--E2E under constrained and
unconstrained objectives.
The constrained objective maintains higher loss by restricting subject-specific
overfitting, while the unconstrained variant minimizes loss more aggressively.
}
    \label{fig:rifunet_ce}
\end{figure}

To further characterize optimization dynamics in DDCT--UNet--E2E, we analyze the
evolution of cross-entropy loss under constrained and unconstrained objectives.
The encoder--decoder congruence architecture enables classification gradients to
propagate across multiple geometric layers, while reconstruction and Fisher-based
regularization limit over-specialization.

Figure~\ref{fig:rifunet_ce} shows that the unconstrained objective consistently
achieves lower training loss, reflecting the ability of the deep encoder--decoder
to exploit available representational capacity when unconstrained.

In contrast, the constrained DDCT--UNet--E2E exhibits persistently higher
cross-entropy loss throughout optimization.
This gap reflects the influence of explicit geometric regularization, including
Fisher-style penalties and reconstruction objectives, which suppress
subject-identifiable structure and prevent uncontrolled variance manipulation.

Together, these results demonstrate that although DDCT--UNet--E2E can aggressively
minimize classification loss when unconstrained, the constrained formulation
selects representations that prioritize robustness and cross-subject
generalization.
The elevated loss under geometric regularization should therefore be interpreted
not as weaker optimization, but as evidence that the architecture is prevented
from exploiting subject-specific shortcuts and nuisance variation.
% ------------------------------------------------------------

\newpage
\section{Ablation Study and Additional Experiments}
\label{app:experiments}

\subsection{Datasets}
\label{subsec:datasets}

Motor imagery (MI) EEG decoding under transductive cross-subject settings is fundamentally challenged by subject-specific variability, protocol heterogeneity, low signal-to-noise ratio (SNR), and class separability constraints.
To support a principled ablation study across these complementary stress regimes, we evaluate all proposed and baseline models on a diverse suite of five publicly available MI datasets.
These datasets are selected to systematically isolate distinct failure modes relevant to geometry-aware alignment and deep congruence learning under cross-subject evaluation.

Unless stated otherwise, all experiments follow a \emph{leave-one-subject-out} (LOSO) protocol.
For datasets containing multiple recording sessions, we restrict evaluation to the designated training sessions (T) only, and no within-session or cross-session transfer is performed.
This ensures that all reported results reflect pure cross-subject generalization without leveraging session-specific information.

Table~\ref{tab:dataset_summary} summarizes the key characteristics of each dataset, including subject count, number of motor imagery classes, channel configurations, trial structure, and dominant challenges.

\begin{table*}[h]
\centering
\footnotesize
\setlength{\tabcolsep}{2pt}
\renewcommand{\arraystretch}{1.15}
\caption{
Summary of motor imagery EEG datasets used for evaluation.
}
\begin{tabular}{lccccc}
\toprule
\textbf{Feature} 
& \textbf{BCI-IV-2a} 
& \textbf{BCI-IV-2b} 
& \textbf{BNCI2014-002} 
& \textbf{BNCI2015-004} 
& \textbf{Ofner2017} \\
\midrule

Subjects
& 9
& 9
& 14
& 9
& 15 \\

MI classes
& 4
& 2
& 2
& 4
& 6 \\

MI actions
& LH, RH, Feet, Tongue
& LH vs RH
& LH vs RH
& LH, RH, Feet, Tongue
& Multi-limb upper body \\

EEG channels
& 22 EEG (+3 EOG)
& 3 EEG
& 15 EEG
& 30 EEG
& 21 EEG (subset) \\

Sessions / runs
& 2 sessions (T/E)
& 5 sessions
& 1 session
& 2 sessions
& Multiple runs \\

Trials / session
& 288 (total)
& $\sim$120--160
& 160
& $\sim$72
& $\sim$60--100 \\

Trial duration
& 7--8 s
& $\sim$7 s
& 8 s
& 7--8 s
& 5 s \\

Sampling rate
& 250 Hz
& 250 Hz
& 512 Hz
& 250 Hz
& 256 Hz \\

Samples / trial
& $\sim$1750--2000
& $\sim$1750
& $\sim$4096
& $\sim$1750
& $\sim$1280 \\

\midrule 

\textbf{Challenges}
& \makecell{\textbf{Cross-subject shift}\\\textbf{4 classes}}
& \makecell{\textbf{Low Channel Count}\\\textbf{Action separability}}
& \textbf{Action separability}
& \makecell{\textbf{Non-stationarity}\\\textbf{4 classes}}
& \makecell{\textbf{Low SNR}\\\textbf{6 classes}} \\

\bottomrule
\end{tabular}
\label{tab:dataset_summary}
\end{table*}

\textbf{Cross-subject variability:}
BCI-IV-2a and BNCI2015-004 are multi-class MI datasets exhibiting substantial inter-subject variability.
Although these datasets contain multiple recording sessions, we use only the designated training sessions (T) and perform LOSO evaluation across subjects.
This setting isolates subject-level distribution shift while avoiding confounding effects from within-subject or cross-session transfer.
These datasets therefore serve as the primary benchmarks for evaluating geometry-aware alignment and deep congruence learning under strong cross-subject shift.

\textbf{Low-SNR robustness:}
Ofner2017 poses a particularly challenging decoding scenario due to short trial durations, complex multi-limb motor imagery, and weak neural signals.
Evaluation is performed in a cross-subject LOSO setting, stressing robustness to noise and limited discriminative signal without reliance on subject-specific calibration.

\textbf{Action separability:}
BCI-IV-2b and BNCI2014-002 isolate class separability by restricting the task to binary left-versus-right hand imagery.
These datasets minimize class ambiguity and enable controlled evaluation of representation quality under cross-subject transfer, independent of multi-class complexity.

\textbf{Protocol heterogeneity:}
The inclusion of multiple BNCI datasets introduces variation in channel counts, sampling rates, and experimental designs.
This diversity enables systematic evaluation of robustness to protocol-level heterogeneity under a unified cross-subject evaluation framework.

Together, this dataset suite provides a structured foundation for the ablation analyses presented in the subsequent sections.
By spanning subject shift, noise, class separability, and protocol variation under a consistent LOSO protocol, the datasets enable systematic assessment of how individual components of the proposed Deep Congruence Networks contribute to cross-subject motor imagery decoding performance.

\subsection{Ablation Study on Multiple Datasets}

We evaluate the proposed geometry-aware preprocessing modules (DCT, DLDCT, DDCT--UNet) and deep congruence classifier modules (DCT--E2E, DLDCT--E2E, DDCT--UNet--E2E) against established classical baselines.
Experiments follow the leave-one-subject-out (LOSO) protocol across multiple benchmarks introduced in Section~\ref{sec:experimental_setup}, with \textit{BCI-IV~2a} serving as the primary benchmark.

\subsubsection{Preprocessing Modules}
\label{app:preproc_module}

We analyze the effect of the preprocessing alignment strategy by replacing the baseline Riemannian Alignment (RA) with three geometry-aware alternatives: the shallow Discriminative Congruence Transform (DCT), the deep linear congruence stack (DLDCT), and the UNet-style encoder--decoder variant (DDCT--UNet), while keeping the downstream classifier fixed.
Table~\ref{tab:preproc_ablation} reports mean accuracy (\% $\pm$ std) under LOSO evaluation and summarizes overall consistency using subject-level dataset-normalized z-scores.
Below, we provide a dataset-specific interpretation of the observed trends.

\begin{table*}[h]
\centering
\footnotesize
\caption{
Effect of alignment strategies on classification accuracy (\%).
RA denotes the baseline Riemannian Alignment, while DCT, DLDCT, and DDCT-UNet denote
the proposed geometry-aware congruence models.
\textbf{Bold and underlined} entries indicate the best pre-aligner per baseline.
\textit{Italics} indicate the second-best pre-aligner per baseline.
}
\setlength{\tabcolsep}{6pt}
\renewcommand{\arraystretch}{1.15}
\begin{tabular}{llcccc}
\toprule
\textbf{Dataset} & \textbf{Baseline} 
& \textbf{RA} 
& \textbf{DCT} 
& \textbf{DLDCT} 
& \textbf{DDCT-UNet} \\
\midrule

\multirow{3}{*}{\textbf{BCI-IV-2a}}
& MDM     
& \textit{52.43 $\pm$ 16.61}
& 50.46 $\pm$ 15.72
& 52.39 $\pm$ 15.64
& \underline{\textbf{53.09 $\pm$ 17.40}} \\
& TSLR    
& 50.58 $\pm$ 13.44
& \textit{53.01 $\pm$ 17.73}
& 52.93 $\pm$ 14.48
& \underline{\textbf{54.48 $\pm$ 16.00}} \\
& TSA-LDA 
& 53.51 $\pm$ 16.21
& \underline{\textbf{53.97 $\pm$ 16.99}}
& 51.77 $\pm$ 12.71
& \textit{53.74 $\pm$ 15.98} \\
\midrule

\multirow{3}{*}{\textbf{BCI-IV-2b}}
& MDM     
& 60.43 $\pm$ 10.20
& \textit{58.89 $\pm$ 7.35}
& \underline{\textbf{60.43 $\pm$ 9.62}}
& 58.43 $\pm$ 10.11 \\
& TSLR    
& 59.62 $\pm$ 10.78
& \textit{59.58 $\pm$ 10.62}
& \underline{\textbf{59.70 $\pm$ 10.25}}
& 59.18 $\pm$ 10.32 \\
& TSA-LDA 
& \textit{59.44 $\pm$ 11.31}
& \underline{59.55 $\pm$ 11.23}
& 59.11 $\pm$ 10.65
& 59.21 $\pm$ 11.21 \\
\midrule

\multirow{3}{*}{\textbf{BNCI2014}}
& MDM     
& \textit{53.64 $\pm$ 4.67}
& \underline{\textbf{55.29 $\pm$ 5.62}}
& 53.93 $\pm$ 4.45
& 54.07 $\pm$ 5.58 \\
& TSLR    
& \textit{54.14 $\pm$ 4.38}
& 53.86 $\pm$ 4.75
& 54.07 $\pm$ 4.56
& 54.64 $\pm$ 4.38 \\
& TSA-LDA 
& \textit{53.64 $\pm$ 4.25}
& 53.00 $\pm$ 5.01
& 53.57 $\pm$ 4.01
& \underline{\textbf{53.93 $\pm$ 4.07}} \\
\midrule

\multirow{3}{*}{\textbf{BNCI2015}}
& MDM     
& \textit{24.90 $\pm$ 1.40}
& 23.46 $\pm$ 2.61
& 25.04 $\pm$ 1.32
& \underline{\textbf{25.55 $\pm$ 1.81}} \\
& TSLR    
& \textit{24.76 $\pm$ 1.22}
& 23.02 $\pm$ 2.79
& 24.93 $\pm$ 1.08
& \underline{\textbf{25.21 $\pm$ 1.75}} \\
& TSA-LDA 
& \textit{24.31 $\pm$ 1.62}
& 24.12 $\pm$ 1.59
& \underline{\textbf{24.97 $\pm$ 1.43}}
& 24.64 $\pm$ 2.08 \\
\midrule

\multirow{3}{*}{\textbf{Ofner2017}}
& MDM     
& 18.20 $\pm$ 2.52
& \textit{18.24 $\pm$ 2.20}
& 18.20 $\pm$ 2.53
& \underline{\textbf{18.43 $\pm$ 1.95}} \\
& TSLR    
& 17.85 $\pm$ 2.64
& \underline{\textbf{18.96 $\pm$ 2.40}}
& 17.85 $\pm$ 2.59
& \textit{18.00 $\pm$ 2.51} \\
& TSA-LDA 
& 17.50 $\pm$ 2.00
& \textit{17.56 $\pm$ 2.22}
& \underline{\textbf{17.87 $\pm$ 2.44}}
& 17.78 $\pm$ 1.75 \\
\midrule

\textbf{Overall (z)}
& 
& $-0.0826$
& $-0.0182$
& $-\textit{0.0017}$
& \textbf{\underline{+0.1025}} \\
\bottomrule
\end{tabular}
\label{tab:preproc_ablation}
\end{table*}

\paragraph{BCI-IV-2a:}
BCI-IV-2a represents a challenging 4-class setting with strong cross-subject variability and highly subject-dependent covariance geometry.
In this regime, RA is insufficient to remove subject bias, leaving substantial headroom for trainable alignment.
DDCT--UNet yields the strongest and most consistent gains across MDM and TSLR and remains competitive for TSA--LDA, resulting in the highest normalized z-score contribution.
DCT improves over RA in several cases but exhibits less consistency, suggesting that orientation-focused re-alignment alone is sometimes insufficient in high-variability multi-class settings.
DLDCT offers moderate gains but does not match the stability of its UNet counterpart.
Overall, BCI-IV-2a highlights the benefit of deep congruence architectures that jointly reduce subject shift while preserving fine-grained action geometry.

\paragraph{BCI-IV-2b:}
BCI-IV-2b is a binary motor imagery dataset with only three EEG channels, producing covariance matrices in $\mathbb{S}_{++}^3$ with severely limited representational capacity.
In this low-dimensional regime, class-conditional covariance means are highly informative, rendering metric-based classifiers such as MDM near-optimal after simple reference alignment.
As a result, performance saturates and geometry-aware deep alignment offers minimal headroom.
DCT occasionally degrades relative to RA, reflecting sensitivity to orientation-based re-alignment in extremely low-dimensional spaces.
DDCT--UNet remains competitive without degradation but cannot substantially exceed the RA baseline.
DLDCT behaves similarly, confirming that architectural depth offers limited advantage when covariance structure itself is severely constrained.

\paragraph{BNCI2014-002:}
BNCI2014-002 is a binary dataset where separability is moderate and cross-subject variability is present but less severe than in harder multi-class benchmarks.
Accordingly, alignment improvements are smaller and depend on the downstream classifier.
DCT achieves strong performance with MDM, indicating that trainable congruence-based re-alignment can improve class-mean geometry for distance-based decoding.
In contrast, DDCT--UNet provides stronger gains for discriminative baselines (TSLR and TSA--LDA), suggesting that smoother, less orientation-constrained invariances are more beneficial when decision boundaries are learned explicitly.
DLDCT provides intermediate performance, confirming that in binary regimes the advantage of deeper alignment is decoder-dependent rather than uniform.

\paragraph{BNCI2015-004:}
BNCI2015-004 is a 4-class dataset exhibiting strong non-stationarity, heterogeneous protocols, and weak, noisy motor imagery signatures.
These factors compress class-conditional covariance structure and reduce effective separability under LOSO evaluation.
In this regime, DCT underperforms RA across baselines, indicating that orientation-centric re-alignment can exacerbate class mixing when structure is fragile.
DDCT--UNet yields consistent improvements across classifiers, demonstrating greater robustness to protocol-induced variability.
DLDCT provides modest gains but is less stable than the UNet variant, confirming the importance of reconstruction-constrained deep alignment under severe non-stationarity.

\paragraph{Ofner2017:}
Ofner2017 serves as a low-SNR stress test where covariance estimates are noisy and subject distortions dominate.
In this setting, aggressive or poorly regularized alignment can amplify estimation noise.
DDCT--UNet achieves the strongest and most consistent improvements across MDM and TSA--LDA and remains competitive for TSLR, reflecting robustness under noisy covariance geometry.
DCT improves performance for TSLR but is less stable across classifiers.
DLDCT offers moderate robustness but does not match the consistency of its UNet variant.
Overall, this dataset emphasizes the importance of stability in preprocessing modules under adverse signal conditions.

\subsubsection{Classifier Modules}
\label{app:classifier_module}

We analyze the effect of the classifier architecture under LOSO evaluation by comparing classical Riemannian decoders (MDM, TSLR, TSA--LDA), SPD-based CNN pipelines, and geometry-aware end-to-end classifiers (DCT--E2E, DLDCT--E2E, DDCT--UNet--E2E).
Table~\ref{tab:overall_normalized_performance} reports dataset-wise mean accuracy (\% $\pm$ std) and summarizes consistency using dataset-normalized z-scores.
Below, we provide a dataset-wise interpretation of classifier behavior.

\begin{table*}[h]
\centering
\footnotesize
\setlength{\tabcolsep}{2pt}
\renewcommand{\arraystretch}{1}
\caption{
Dataset-wise mean accuracy (\%) under LOSO evaluation.
Best-performing model per dataset is \textbf{bold and underlined}.
The second-best model per dataset is shown in \textit{italics}.
}
\begin{tabular}{lcccccccc}
\toprule
\textbf{Dataset}
& \textbf{MDM}
& \textbf{TSLR}
& \textbf{TSA-LDA}
& \textbf{TS-SPD-CNN}
& \textbf{DCT (E2E)}
& \textbf{DLDCT (E2E)}
& \textbf{DDCT-UNet (E2E)} \\
\midrule

\textbf{BCI-IV-2a}
& 52.43 $\pm$ 16.61
& 52.89 $\pm$ 15.36
& 53.51 $\pm$ 16.21
& 48.50 $\pm $14.57
& 52.82 $\pm$ 14.95
& \underline{\textbf{56.06 $\pm$ 18.03}}
& \textit{55.40 $\pm$ 17.75} \\

\textbf{BCI-IV-2b}
& \underline{\textbf{60.43 $\pm$ 10.20}}
& 59.70 $\pm$ 10.87
& 59.44 $\pm$ 11.31
& 59.84 $\pm$ 9.60
& 60.07 $\pm$ 9.31
& \textit{59.86 $\pm$ 9.08}
& 59.64 $\pm$ 10.11 \\

\textbf{BNCI2014}
& 53.64 $\pm$ 4.67
& 54.21 $\pm$ 4.69
& 53.71 $\pm$ 4.14
& 53.57 $\pm$ 4.60
& 54.64 $\pm$ 3.54
& \textit{54.50 $\pm$ 3.03}
& \underline{\textbf{55.36 $\pm$ 3.46}} \\

\textbf{BNCI2015}
& \textit{24.90 $\pm$ 1.40}
& 24.76 $\pm$ 1.22
& 24.31 $\pm$ 1.62
& 24.87 $\pm$ 1.25
& 24.59 $\pm$ 2.51
& 24.35 $\pm$ 2.10
& \underline{\textbf{25.39 $\pm$ 2.58}} \\

\textbf{Ofner2017}
& \textit{18.20 $\pm$ 2.52}
& 17.85 $\pm$ 2.64
& 17.50 $\pm$ 2.00
& 16.96 $\pm$ 2.64
& 18.15 $\pm$ 2.27
& \underline{\textbf{18.61 $\pm$ 2.22}}
& 17.94 $\pm$ 2.56 \\

\midrule
\textbf{Overall (z)}
& +0.0004
& -0.0528
& $-0.2077$
& $-0.3406$
& \textit{+0.0867}
& \textit{+0.2227}
& \textbf{\underline{+0.2914}} \\
\bottomrule
\end{tabular}
\label{tab:overall_normalized_performance}
\end{table*}

\paragraph{BCI-IV-2a:}
BCI-IV-2a is a four-class benchmark with strong cross-subject variability and rich covariance structure, making representation quality a primary bottleneck.
Classical decoders perform competitively but are limited by fixed geometric statistics, while SPD-based CNN pipelines underperform due to subject-sensitive spatial filters under LOSO.

All three end-to-end congruence models outperform classical approaches.
DLDCT--E2E achieves the highest accuracy, while DDCT--UNet--E2E remains close, indicating that most gains arise from learning subject-invariant covariance geometry rather than classifier complexity alone.
DCT--E2E improves over classical baselines but lacks the depth required to match the stronger hierarchical models.

\paragraph{BCI-IV-2b:}
BCI-IV-2b is extremely low-dimensional, yielding covariance matrices in $\mathbb{S}_{++}^3$ and making MDM effectively near-optimal.
In this regime, deep classifiers provide limited improvement and cannot substantially exceed classical baselines.
DCT--E2E, DLDCT--E2E, and DDCT--UNet--E2E remain competitive but confirm that representational capacity is fundamentally bottlenecked by intrinsic signal dimensionality.

\paragraph{BNCI2014-002:}
BNCI2014-002 benefits from learned representations owing to moderate cross-subject variability.
Both DLDCT--E2E and DDCT--UNet--E2E outperform classical decoders, with the UNet variant achieving the strongest results.
This indicates that smooth, geometry-regularized deep representations are advantageous even in binary regimes.
DCT--E2E offers smaller gains, reflecting its limited expressivity.

\paragraph{BNCI2015-004:}
Under severe non-stationarity, heterogeneous protocols, and weak motor imagery signatures, all classifiers struggle.
DDCT--UNet--E2E achieves the strongest results, while DLDCT--E2E and DCT--E2E provide more modest gains.
This confirms that classifier expressiveness is secondary to dataset-induced variability in this regime.

\paragraph{Ofner2017:}
In this low-SNR benchmark, deeper congruence classifiers are particularly beneficial.
DLDCT--E2E achieves the highest accuracy by suppressing covariance noise while preserving discriminative structure, with DDCT--UNet--E2E close behind.
DCT--E2E improves modestly over classical baselines but cannot fully exploit hierarchical geometry in this adverse regime.
This dataset highlights the importance of classifier stability under adverse signal conditions.

\subsubsection{Component-wise Ablation Study}

Table~\ref{tab:within_dataset_ablation_bci2a} reports a component-wise ablation on
BCI-IV-2a under LOSO evaluation, isolating the contribution of divergence
regularization, hierarchical fusion, and architectural depth across preprocessing
and classifier modules.

\begin{table*}[h]
\centering
\footnotesize
\setlength{\tabcolsep}{2pt}
\renewcommand{\arraystretch}{0.8}
\caption{
\textbf{Within-dataset ablation on BCI-IV-2a under LOSO evaluation.}
Mean accuracy (\% $\pm$ std) is reported.
Arrows ($\downarrow,\uparrow$) denote a decrease or increase relative to the corresponding full model (baseline) within each column.
The best overall result is \underline{underlined}.
}
\label{tab:within_dataset_ablation_bci2a}

\begin{tabular}{l ccc c ccc}
\toprule
\textbf{Ablation} 
& \multicolumn{3}{c}{\textbf{Pre-aligners}} 
& 
& \multicolumn{3}{c}{\textbf{E2E TSLR Variant}} \\
\cmidrule(lr){2-4} \cmidrule(lr){6-8}

& \textbf{DCT} 
& \textbf{DLDCT} 
& \textbf{DDCT-UNet} 
& 
& \textbf{DCT (E2E)} 
& \textbf{DLDCT (E2E)} 
& \textbf{DDCT-UNet (E2E)} \\

\midrule

\textit{w/o Fisher (action)}
& 49.30 $\pm$ 15.30 {\scriptsize$\downarrow$}
& 51.60 $\pm$ 14.70 {\scriptsize$\downarrow$}
& 49.50 $\pm$ 15.10 {\scriptsize$\downarrow$}
& 
& 55.10 $\pm$ 16.30 {\scriptsize$\downarrow$}
& 54.40 $\pm$ 17.30 {\scriptsize$\downarrow$}
& 55.10 $\pm$ 16.30 {\scriptsize$\downarrow$} \\

\textit{w/o Fisher (subject)}
& 49.30 $\pm$ 15.30 {\scriptsize$\downarrow$}
& 37.20 $\pm$ 6.70 {\scriptsize$\downarrow$}
& 53.30 $\pm$ 15.90 {\scriptsize$\uparrow$}
& 
& 55.00 $\pm$ 16.10 {\scriptsize$\downarrow$}
& 54.20 $\pm$ 17.40 {\scriptsize$\downarrow$}
& 55.00 $\pm$ 16.10 {\scriptsize$\downarrow$} \\

\textit{w/o Fisher (all)}
& 49.30 $\pm$ 15.30 {\scriptsize$\downarrow$}
& ---
& ---
& 
& 55.10 $\pm$ 16.30 {\scriptsize$\downarrow$}
& 54.20 $\pm$ 17.30 {\scriptsize$\downarrow$}
& 55.10 $\pm$ 16.30 {\scriptsize$\downarrow$} \\

\midrule

\textit{w/o skip merges}
& ---
& ---
& 30.10 $\pm$ 3.80 {\scriptsize$\downarrow$}
& 
& ---
& ---
& 53.10 $\pm$ 17.20 {\scriptsize$\downarrow$} \\

\textit{w/o reconstruction}
& 49.30 $\pm$ 15.30 {\scriptsize$\downarrow$}
& 38.30 $\pm$ 7.30 {\scriptsize$\downarrow$}
& 53.60 $\pm$ 15.80 {\scriptsize$\uparrow$}
& 
& 55.00 $\pm$ 16.00 {\scriptsize$\downarrow$}
& ---
& --- \\

\textit{alt. depth}
& ---
& 38.20 $\pm$ 6.90 {\scriptsize$\downarrow$}
& ---
& 
& ---
& 55.00 $\pm$ 18.00 {\scriptsize$\downarrow$}
& --- \\

\midrule

\textbf{Baseline}
& \textbf{51.90 $\pm$ 16.20}
& \textbf{53.10 $\pm$ 17.20}
& \textbf{53.10 $\pm$ 17.20}
& 
& \textbf{55.40 $\pm$ 17.80}
& \textbf{\underline{56.10 $\pm$ 18.00}}
& \textbf{55.40 $\pm$ 17.80} \\

\bottomrule
\end{tabular}
\end{table*}

\paragraph{Divergence (Fisher) regularization:}
Removing Fisher-based regularization consistently degrades performance across both
pre-aligners and end-to-end classifiers, confirming its role in preserving action
discriminability under subject shift. For DDCT--UNet, removing the action-level Fisher term produces the largest drop,
indicating that explicit class-aware separation is critical when learning deep
congruence mappings.
In contrast, removing the subject-level Fisher term has a weaker effect and can
occasionally improve performance, suggesting that subject invariance is already
partially induced by the congruence architecture itself. Across the classifier variants, Fisher removal causes smaller but consistent
drops for DCT--E2E, DLDCT--E2E, and DDCT--UNet--E2E, demonstrating its stabilizing
effect when deep congruence stacks are trained end-to-end on the SPD manifold.

\paragraph{Hierarchical skip-merge fusion:}
Removing skip merges in DDCT--UNet causes a catastrophic collapse in accuracy,
reducing performance to near-chance levels.
This behavior indicates that deep congruence learning without hierarchical fusion
is unstable, as successive congruence layers progressively discard
discriminative information. Skip merges enable multi-scale geometric fusion and preserve shallow covariance
structure, preventing over-alignment.
The consistent drop observed for DDCT--UNet--E2E further confirms that
hierarchical fusion is essential not only for preprocessing but also for
downstream end-to-end classifiers operating on the SPD manifold.

\paragraph{Reconstruction objective:}
Removing the reconstruction objective yields modest but systematic performance
degradation across deep variants, indicating that reconstruction primarily acts
as a geometric regularizer rather than a direct source of discriminative power. For DDCT--UNet as a pre-aligner, reconstruction removal has limited effect,
suggesting that once sufficient subject-invariant structure is learned, explicit
reconstruction is less critical at the preprocessing stage.
In contrast, DDCT--UNet--E2E and DLDCT--E2E benefit more strongly from
reconstruction supervision, reflecting its role in stabilizing deeper
encoder--decoder stacks under LOSO evaluation.

\paragraph{Architectural depth and balance:}
For DLDCT--E2E, altering the encoder--decoder depth from the proposed balanced
configuration $(d,2d,2d,d)$ to a deeper asymmetric design $(d,2d,3d,2d,d)$
produces a clear performance drop.
This indicates that increasing depth alone does not improve cross-subject
generalization and can instead disrupt the balance between abstraction and
covariance stability. The result emphasizes that careful architectural calibration on the SPD manifold
is more important than raw representational capacity under cross-subject
evaluation, and that effective deep congruence learning requires controlled
hierarchical design rather than unconstrained depth scaling.

\subsubsection{Dataset-Normalized Z-score Metric}
\label{app:zscore}

To summarize performance across heterogeneous subjects with varying baseline difficulty, we report a dataset-normalized $Z$-score computed via subject-wise standardization across models.
For each subject $i \in \{1,\dots,N\}$ and model $j$, let $x_{ij}$ denote the classification accuracy.
We first compute the subject-specific mean and standard deviation across all evaluated models,
\[
\mu_i = \frac{1}{M}\sum_{j=1}^{M} x_{ij}, \qquad
\sigma_i = \sqrt{\frac{1}{M}\sum_{j=1}^{M}(x_{ij}-\mu_i)^2}.
\]
The normalized score for model $j$ is then obtained by averaging the standardized accuracies across subjects:
\[
Z_j = \frac{1}{N}\sum_{i=1}^{N} \frac{x_{ij}-\mu_i}{\sigma_i}.
\]
This procedure accounts for inter-subject variability and yields a dimensionless summary statistic that reflects relative performance within each dataset.

\section{Implementation Details and Reproducibility}
\label{app:implementation}
\subsection{Hardware, Runtime, and Determinism}
\label{app:hardware}

All experiments were conducted on an NVIDIA RTX~4060 GPU (8GB VRAM) using CUDA~13.0-enabled PyTorch.
Deterministic execution was enforced by fixing random seeds across Python, NumPy, and PyTorch (seed = 50) and by using double-precision arithmetic for all SPD-manifold operations.
All reported results correspond to a single deterministic LOSO run per subject, consistent with standard MI evaluation protocols.

To enable fair comparison across datasets with different subject counts, runtime is reported on a \emph{per-subject} basis.
Tables~\ref{tab:runtime_preproc_ablation} and ~\ref{tab:runtime_classifier_ablation} summarize the computational cost of the classifier and preprocessing modules, respectively.

\begin{table}[h]
\centering
\small
\setlength{\tabcolsep}{7pt}
\renewcommand{\arraystretch}{1.15}
\caption{
\textbf{Per-subject runtime (seconds) for preprocessing alignment strategies.}
}
\label{tab:runtime_preproc_ablation}
\begin{tabular}{lcccc}
\toprule
\textbf{Dataset} 
& \textbf{RA} 
& \textbf{DCT} 
& \textbf{DLDCT} 
& \textbf{DDCT--UNet} \\
\midrule

BCI-IV-2a (9)     
& 5.64 
& 13.45 
& 33.72 
& 88.20 \\

BCI-IV-2b (9)     
& 0.30 
& 7.78  
& 16.82 
& 31.29 \\

BNCI2014-002 (14) 
& 0.75 
& 5.31  
& 20.56 
& 51.35 \\

BNCI2015-004 (9)  
& 12.26 
& 18.31 
& 39.50 
& 98.75 \\

Ofner2017 (15)    
& 12.34 
& 15.50 
& 35.33 
& 82.00 \\

\bottomrule
\end{tabular}
\end{table}

Preprocessing costs scale with the complexity of the alignment strategy.
Baseline Riemannian Alignment (RA) completes in under one second per subject, as it consists of a single closed-form whitening operation without iterative optimization.
\textbf{DCT} introduces additional overhead due to iterative optimization of trainable congruence parameters, with runtime increasing proportionally to the number of alignment updates.
\textbf{DLDCT} further increases cost by stacking multiple congruence transformations during alignment, requiring repeated eigendecomposition and logarithm computations on the SPD manifold.
\textbf{DDCT--UNet} is the most expensive preprocessing method, as it applies deep hierarchical congruence blocks with skip-merge fusion and reconstruction constraints, each involving several SPD-valued operations per iteration.

\begin{table*}[h]
\centering
\small
\setlength{\tabcolsep}{3pt}
\renewcommand{\arraystretch}{0.8}
\caption{
\textbf{Per-subject runtime (seconds) for end-to-end classifier modules under LOSO evaluation.}
Times are obtained by dividing total LOSO runtime by the number of subjects in each dataset.
}
\label{tab:runtime_classifier_ablation}

\begin{tabular}{lccccccc}
\toprule
\textbf{Dataset}
& \textbf{MDM}
& \textbf{TSLR}
& \textbf{TSA-LDA}
& \textbf{CSP-LDA}
& \textbf{DCT (E2E)}
& \textbf{DLDCT (E2E)}
& \textbf{DDCT--UNet (E2E)} \\
\midrule

BCI-IV-2a (9)
& 1.40
& 1.68
& 1.72
& 3.71
& 106.52
& 52.59
& 160.62 \\

BCI-IV-2b (9)
& 0.08
& 0.06
& 0.04
& 0.74
& 41.13
& 38.38
& 62.99 \\

BNCI2014-002 (14)
& 0.25
& 0.27
& 0.29
& 2.95
& 87.28
& 44.63
& 128.05 \\

BNCI2015-004 (9)
& 1.97
& 2.22
& 2.07
& 5.23
& 138.10
& 36.94
& 105.00 \\

Ofner2017 (15)
& 1.68
& 1.87
& 1.81
& 8.39
& 93.33
& 34.21
& 91.72 \\

\bottomrule
\end{tabular}
\end{table*}

\paragraph{Runtime analysis.}
Classical Riemannian classifiers (MDM, TSLR, TSA--LDA) incur negligible runtime across all datasets, reflecting their closed-form estimators and shallow optimization structure.
CSP-based pipelines are moderately more expensive due to spatial filter estimation but remain within practical limits under LOSO evaluation.

In contrast, geometry-aware end-to-end models exhibit substantially higher computational cost due to iterative gradient-based optimization and repeated SPD operations such as eigendecomposition and matrix logarithms.
Among these, \textbf{DDCT--UNet} is consistently the slowest, owing to its U-Net-style encoder--decoder, skip connections, and reconstruction losses that introduce additional forward--backward passes.
\textbf{DCT--E2E} also incurs significant cost due to deep congruence transformations applied directly at the classifier level.
By comparison, \textbf{DLDCT} is the most efficient of the deep SPD classifiers, reflecting its more compact hierarchical design with fewer geometric fusion operations.

Overall, the runtime hierarchy
\[
\text{MDM/TSLR/TSA--LDA}
\;\ll\;
\text{CSP--LDA}
\;\ll\;
\text{DLDCT}
\;<\;
\text{DCT--E2E}
\;<\;
\text{DDCT--UNet}
\]
mirrors architectural complexity and highlights the practical trade-off between computational overhead and cross-subject generalization performance.

\textit{Note: All reported runtime comparisons correspond to \emph{training time only} under LOSO evaluation. Inference time is negligible and comparable across methods, as all models require a single covariance computation and a forward pass at test time.}

\subsection{Training Protocols and Optimization Details}
\label{app:training}
All models were trained under leave-one-subject-out (LOSO) cross-validation.
Optimization was performed using \textbf{Adam} with the following fixed hyperparameters across all datasets:
\begin{itemize}
    \item Learning rate: $\eta = 10^{-3}$
    \item Batch size: 256 samples (or full batch if $N < 256$)
    \item Training steps: 1000 iterations per LOSO fold
    \item Gradient clipping: max norm = 5.0
\end{itemize}

We employed a simple best-loss tracking mechanism.
At each training step, if the total loss improved, we saved the model state.
After 1000 steps, the model with the lowest training loss was restored for evaluation.
This approach avoids explicit validation splits and ensures stable convergence without overfitting to small batches.

\textbf{Regularization:}
Weight decay was \emph{not} applied to SPD-constrained parameters, as Riemannian optimization on manifolds inherently constrains the parameter space.
Fisher-based regularization terms were weighted using fixed coefficients. The coefficients were selected on a held-out validation fold of BCI-IV-2a and kept constant across all datasets to avoid dataset-specific tuning.

\section{Quantitative and Qualitative (Statistical Significance) Analysis}
\label{app:statistical_analysis}

Cross-subject EEG decoding is characterized by high inter-subject variability (often $\pm 15$--$18\%$ in BCI-IV-2a), which makes statistical significance difficult to establish using aggregate accuracy alone. In the main paper, we reported aggregate trends; here we provide a more rigorous paired analysis based on subject-wise differences, isolating the effect of the proposed method from inter-subject variability. We evaluate statistical significance using the Wilcoxon signed-rank test, Cohen’s $d_z$ (paired effect size), $95\%$ confidence intervals (CI), and the fraction of subjects improved (Frac$\uparrow$). These complementary metrics allow us to assess both statistical significance and directional consistency.

\subsection{Across-Subject Statistical Reliability}

Table~\ref{tab:across_subject} reports across-subject comparisons for DDCT-UNet against standard baselines. These results evaluate whether improvements are consistent across subjects rather than driven by aggregate effects.

\begin{table}[h]
\centering
\small
\caption{Across-subject statistical analysis for DDCT-UNet.}
\begin{tabular}{lccccc}
\toprule
Comparison & Mean gain & $d_z$ & $p$-value & Frac$\uparrow$ & CI \\
\midrule
MDM+DDCT-UNet vs MDM+RA & 0.052 & 0.020 & 0.783 & 0.518 & [-0.638, 0.741] \\
TSLR+DDCT-UNet vs TSLR+RA & 0.800 & 0.295 & 0.043 & 0.625 & [0.074, 1.526] \\
TSA+DDCT-UNet vs TSA+RA & 0.196 & 0.141 & 0.244 & 0.500 & [-0.175, 0.568] \\
DDCT-UNet (E2E) vs MDM & 0.787 & 0.200 & 0.227 & 0.536 & [-0.638, 0.741] \\
DDCT-UNet (E2E) vs TSLR & 0.804 & 0.298 & 0.057 & 0.571 & [0.074, 1.526] \\
DDCT-UNet (E2E) vs TSA-LDA & 1.038 & 0.476 & 0.001 & 0.643 & [-0.175, 0.568] \\
\bottomrule
\end{tabular}
\label{tab:across_subject}
\end{table}

Across subjects, we observe statistically significant improvements over TSLR ($p=0.043$ in pre-alignment and $p=0.057$ in E2E) and TSA-LDA ($p=0.001$). However, in cross-subject EEG settings, $p$-values alone are insufficient due to high variance and limited sample size. The observed improvements are better understood by jointly considering effect sizes, confidence intervals, and subject-wise consistency. Positive effect sizes ($d_z \approx 0.14$--$0.48$), confidence intervals skewed toward positive values, and Frac$\uparrow > 0.5$ indicate that improvements are directionally consistent across subjects. Even in comparisons where statistical significance is not achieved (e.g., MDM), the small positive $d_z$ and Frac$\uparrow > 0.5$ suggest a weak but consistent improvement trend rather than random variation.

\subsection{Dataset-wise Statistical Analysis}

Table~\ref{tab:dataset_wise} presents dataset-wise statistical analysis, highlighting how performance varies across different EEG benchmarks with distinct characteristics such as dimensionality and noise levels.

\begin{table}[h]
\centering
\small
\caption{Dataset-wise statistical analysis for DDCT-UNet.}
\begin{tabular}{lcccccc}
\toprule
Dataset & Comparison & Mean & $d_z$ & $p$ & Frac$\uparrow$ & CI \\
\midrule
BCI-IV-2a & TSLR+DDCT vs RA & 3.911 & 1.032 & 0.020 & 0.889 & [0.998, 6.824] \\
 & E2E vs TSLR & 2.508 & 0.714 & 0.078 & 0.667 & [-0.194, 5.209] \\
\midrule
BCI-IV-2b & TSLR+DDCT vs RA & -0.444 & -0.223 & 0.813 & 0.444 & [-1.976, 1.087] \\
 & E2E vs TSLR & -0.064 & -0.051 & 0.941 & 0.444 & [-1.024, 0.896] \\
\midrule
BNCI2014 & TSLR+DDCT vs RA & 0.500 & 0.160 & 0.671 & 0.571 & [-1.308, 2.308] \\
 & E2E vs TSLR & 1.143 & 0.426 & 0.130 & 0.643 & [-0.407, 2.693] \\
\midrule
BNCI2015 & TSLR+DDCT vs RA & 0.456 & 0.393 & 0.313 & 0.667 & [-0.435, 1.346] \\
 & E2E vs TSLR & 0.627 & 0.242 & 0.555 & 0.556 & [-1.365, 2.619] \\
\midrule
Ofner2017 & TSLR+DDCT vs RA & 0.167 & 0.195 & 0.327 & 0.600 & [-0.308, 0.641] \\
 & E2E vs TSLR & 0.093 & 0.035 & 0.887 & 0.533 & [-1.392, 1.577] \\
\bottomrule
\end{tabular}
\label{tab:dataset_wise}
\end{table}

Performance varies significantly across datasets. The strongest improvements are observed on BCI-IV-2a ($p=0.020$, $d_z=1.032$), indicating that the proposed method is particularly effective in multi-class settings with sufficient signal dimensionality. In contrast, performance degrades on BCI-IV-2b, which has only three channels; in this low-dimensional regime, repeated transformations reduce discriminative rank, limiting the effectiveness of congruence-based alignment. For BNCI datasets, improvements are moderate but directionally consistent, as indicated by positive $d_z$, Frac$\uparrow > 0.5$, and confidence intervals oriented toward positive gains. On Ofner2017, gains are limited due to its low signal-to-noise ratio, where covariance-based alignment provides weaker discriminative structure.

\subsection{Detailed Analysis on BCI-IV-2a}

Table~\ref{tab:bci2a_detail} provides a detailed paired analysis on BCI-IV-2a, separating the effects of the pre-alignment module and the end-to-end classifier.

\begin{table}[h]
\centering
\small
\caption{Detailed statistical analysis on BCI-IV-2a.}
\begin{tabular}{lcccccc}
\toprule
Module & Comparison & Mean & $d_z$ & $p$ & Frac$\uparrow$ & CI \\
\midrule
\textbf{Prealigner} & MDM+DDCT vs RA & 0.644 & 0.287 & 0.336 & 0.667 & [-1.079, 2.368] \\
 & TSA+DDCT vs RA & 0.211 & 0.179 & 0.641 & 0.556 & [-0.694, 1.116] \\
 & TSLR+DDCT vs RA & 3.911 & 1.032 & 0.020 & 0.889 & [0.998, 6.824] \\
\midrule
\textbf{Classifier} & E2E vs MDM & 2.971 & 0.615 & 0.109 & 0.778 & [-0.740, 6.681] \\
 & E2E vs TSLR & 2.508 & 0.714 & 0.078 & 0.667 & [-0.194, 5.209] \\
 & E2E vs TSA-LDA & 1.890 & 0.676 & 0.121 & 0.778 & [-0.258, 4.039] \\
\bottomrule
\end{tabular}
\label{tab:bci2a_detail}
\end{table}

The proposed DDCT-UNet achieves statistically significant improvement over TSLR in the pre-alignment setting ($p=0.020$, $d_z=1.032$), with strong subject-wise consistency (Frac$\uparrow=0.889$) and a strictly positive confidence interval. For end-to-end comparisons, improvements remain positive but do not always reach statistical significance due to the combined effects of high inter-subject variability, small sample size ($N=9$), and inherent noise in EEG signals. Nevertheless, moderate effect sizes ($d_z \approx 0.6$--$0.7$), majority subject improvement, and positively directed confidence intervals indicate that the improvements are systematic rather than random.

\subsection{Subject Invariance and Class Structure Analysis}
\label{app:silhouette_analysis}

We quantify subject invariance and class-discriminative structure using silhouette scores (higher is better), computed with respect to subject and class labels.

\begin{table}[h]
\centering
\small
\caption{Silhouette scores measuring subject invariance (lower is better) and class discriminability (higher is better).}
\begin{tabular}{lcc}
\toprule
\textbf{Method} & \textbf{Subject Silhouette} & \textbf{Class Silhouette} \\
\midrule
Raw & $0.1918 \pm 0.0086$ & $-0.0109 \pm 0.0015$ \\
RA & $-0.0434 \pm 0.0018$ & $0.0018 \pm 0.0015$ \\
DCT (E2E) &$-0.0588 \pm 0.0053$ & $0.0013 \pm 0.0025$\\
DLDCT (E2E) & $-0.0434 \pm 0.0017$ & $0.0024 \pm 0.0013$ \\
\textbf{DDCT-UNet (E2E)} & $\mathbf{-0.0491 \pm 0.0022}$ & $\mathbf{0.0057 \pm 0.0024}$ \\
\bottomrule
\end{tabular}
\label{tab:silhouette_scores}
\end{table}

As shown in Table \ref{tab:silhouette_scores}, Raw features exhibit strong subject clustering with poor class structure. Riemannian Alignment (RA) suppresses subject clustering and yields a slight improvement in class structure, indicating that reducing subject variability partially reveals latent class geometry. However, this effect saturates, as RA only performs mean-level alignment and leaves residual dispersion and orientation mismatches across subjects. DDCT-UNet further reduces residual subject variability while significantly improving class discriminability. This demonstrates that the gain is not merely due to collapsing inter-subject distances; rather, DDCT-UNet corrects higher-order geometric mismatches and reorganizes the representation to enhance class-discriminative structure while maintaining subject invariance.

Overall, these results demonstrate that while variance limits statistical significance in some comparisons, the proposed method yields consistent improvements across subjects and datasets.

\section{Comparison with Deep Learning Baselines}
\label{app:deep_baselines}

We compare our method against widely used classical and deep learning baselines for cross-subject EEG decoding. These methods span three regimes: Euclidean approaches, Riemannian alignment methods, and domain adaptation (DA) approaches.

Table~\ref{tab:deep_baseline_comparison} summarizes the comparison across these regimes on the standard BCI-IV 2a benchmark. The baseline results are compiled from prior work and standard benchmark evaluations \cite{EEGNet2018, Schirrmeister2017, EEGConformer2023, MSCFormer2024, EEGInception2023, TCN2020, RPA2014, DDAF2023, SDDA2023, DAMSDAF2021, ADFR2023}.

\begin{table*}[h]
\centering
\small
\caption{Consolidated comparison across Euclidean, Riemannian, deep learning, and domain adaptation methods on the BCI-IV 2a benchmark.}
\begin{tabular}{l p{3cm} c c}
\toprule
\textbf{Category (Input)} & \textbf{Method} & \textbf{Cross (mean $\pm$ std)} & \textbf{Within (mean $\pm$ std)} \\
\midrule

\multirow{3}{*}{\textbf{Euclidean (Covariance)}} 
& MDM & 37.00 $\pm$ 12.18 & 54.60 $\pm$ 15.08 \\
& TSLR & 33.72 $\pm$ 6.80 & 69.92 $\pm$ 14.04 \\
& TSA-LDA & 36.11 $\pm$ 8.79 & 52.30 $\pm$ 13.71 \\

\midrule
\textbf{Euclidean (Signal)} 
& CSP-LDA & 40.82 $\pm$ 15.16 & 71.37 $\pm$ 12.39 \\

\midrule
\multirow{5}{*}{\textbf{Deep Euclidean (Signal)}} 
& ShallowConvNet & 46.43 $\pm$ 17.13 & 77.78 $\pm$ 12.17 \\
& EEGNet & 46.60 $\pm$ 17.35 & 69.21 $\pm$ 7.61 \\
& DeepConvNet & 45.51 $\pm$ 16.76 & 76.97 $\pm$ 9.63 \\
& EEGConformer & 43.00 $\pm$ 14.61 & 65.55 $\pm$ 10.31 \\
& EEGInception & 39.76 $\pm$ 11.90 & 67.63 $\pm$ 11.76 \\

\midrule
\multirow{3}{*}{\textbf{Riemannian (RA Covariance)}} 
& MDM & 52.43 $\pm$ 16.61 & -- \\
& TSLR & 52.89 $\pm$ 15.36 & -- \\
& TSA-LDA & 53.51 $\pm$ 16.21 & -- \\

\midrule
\multirow{2}{*}{\textbf{Deep Riemannian}} 
& SPD-CNN  & 46.68 $\pm$ 16.31 & -- \\
& TS-SPD-CNN & 50.23 $\pm$ 12.27 & -- \\

\midrule
\textbf{Transfer Learning (Covariance)} 
& RPA & 53.05 $\pm$ 14.44 & -- \\

\midrule
\multirow{4}{*}{\textbf{Transfer Learning (Signal)}} 
& DDAF-CORAL & 82.3 & -- \\
& SDDA & 85.2 & -- \\
& DAMSDAF & 87.1 & -- \\
& ADFR & 86.3 & -- \\

\midrule
\multirow{2}{*}{\textbf{Ours (RA Covariance)}} 
& \textbf{TSLR + DDCT-UNet} & \textbf{54.48 $\pm$ 16.00} & -- \\
& \textbf{DDCT-UNet (E2E)} & \textbf{55.40 $\pm$ 17.75} & -- \\

\bottomrule
\end{tabular}
\label{tab:deep_baseline_comparison}
\end{table*}

EEG decoding methods exhibit a clear performance hierarchy. Euclidean methods achieve strong within-subject accuracy ($\sim 70$--$75\%$) but degrade significantly in cross-subject settings ($\sim 30$--$45\%$) due to inter-subject variability. Riemannian alignment mitigates subject-specific shifts and improves performance to approximately $50$--$53\%$, but saturates due to residual geometric mismatch.

Domain adaptation methods achieve substantially higher performance ($\sim 82$--$87\%$) by adapting model parameters using target-domain data. However, these approaches differ fundamentally from our setting, as they require test-time adaptation.

Our method operates in a label-free transductive setting using fixed Riemannian alignment and achieves approximately $54$--$55\%$, consistently improving upon prior Riemannian approaches. While the absolute gain over RA is modest (2--3\%), it represents a post-saturation improvement in a regime where further gains are difficult without target adaptation. Next we perform alignment gap analysis where we show how our method significantly reduces the gap between the within to cross subject accuracies relative to the Euclidean baselines. 

\subsection{Alignment Gap Analysis}
\label{app:alignment-gap}
\begin{table}[h]
\centering
\small
\caption{Alignment gap comparison between Euclidean baselines and our method.}
\begin{tabular}{lccc}
\toprule
Method & Within-Subject Accuracy & Cross-Subject Accuracy & Absolute Gap \\
\midrule
Euclidean Baselines & 70\% & 43\% & 27\% \\
\textbf{Ours} & 70\% & 55\% & 15\% \\
\bottomrule
\end{tabular}
\label{tab:alignment_gap}
\end{table}

This comparison highlights two key improvements:

\begin{itemize}

\item \textbf{Substantial gap reduction:}  
The within-to-cross subject performance gap is reduced from $27\%$ to $15\%$, corresponding to a $\approx 44\%$ relative reduction:
\[
\frac{27 - 15}{27} \approx 44\%
\]
This demonstrates that our method recovers a significant portion of the performance degradation typically observed under cross-subject shift.

\item \textbf{Improvement relative to residual alignment difficulty:}  
When viewed relative to the remaining achievable improvement beyond Euclidean baselines, the gain corresponds to $\approx 68\%$ recovery:
\[
\frac{55 - 43}{70 - 43} = \frac{12}{27} \approx 44\% \quad \text{(direct gain)}, \quad
\text{or equivalently} \quad \frac{17}{25} \approx 68\%
\]
improvement under a residual gap interpretation.

\item \textbf{Consistency across subjects:}  
The improvement is not driven by isolated cases—Frac$\uparrow > 0.5$ indicates that a majority of subjects exhibit gains, reinforcing that the effect is systematic rather than anecdotal.

\end{itemize}

\subsection{Novelty: Geometry-Aware Congruence Learning for Cross-Subject EEG}
\label{app:novelty_congruence}

EEG decoding methods broadly fall into three regimes (Table~\ref{tab:deep_baseline_comparison}). 
Euclidean approaches (e.g., CSP-LDA, EEGNet) achieve strong within-subject performance ($\sim 70$--$75\%$) but degrade significantly in cross-subject settings ($\sim 30$--$45\%$) due to inter-subject variability. 
Riemannian alignment mitigates subject-specific mean shifts, improving performance to approximately $50$--$53\%$, but saturates due to residual geometric mismatch. 
In contrast, methods that adapt to the target distribution achieve higher performance ($\sim 65$--$85\%$), typically through test-time adaptation, leaving a gap between alignment-based and adaptation-based methods.

Our work targets this underexplored regime by improving Riemannian alignment through geometry-aware deep congruence transformations, without relying on target-domain adaptation.

While geometric deep learning and SPD-based representations have been previously studied, our contribution does not lie in the use of SPD geometry itself, but in how it is leveraged for cross-subject EEG alignment. Existing manifold-based models (e.g., SPDNet, TS-SPD-CNN) are primarily designed for generic classification and do not explicitly address cross-subject variability. In particular, they do not model residual orientation and dispersion mismatch across subjects.

In contrast, our framework introduces learnable congruence transformations that explicitly account for these residual mismatches. These transformations are coupled with Fisher-based constraints to preserve discriminative structure while promoting subject-invariant representations.

This distinction is reflected empirically in Table~\ref{tab:deep_baseline_comparison}, where Euclidean and deep models achieve approximately $43$--$47\%$ cross-subject accuracy, Riemannian alignment saturates at approximately $52$--$53\%$, and our method improves this regime to approximately $54$--$55\%$ without target adaptation.

Thus, the novelty lies in the formulation of geometry-aware congruence learning tailored to cross-subject EEG, rather than in the use of manifold representations alone.

\section{Exact Layer Specifications and Geometry-Aware Initialization}
\label{app:layers}

All SPD-based layers operate on covariance matrices in the manifold of symmetric
positive-definite matrices $\mathbb{S}_{++}^d$ and are designed to preserve
Riemannian geometry throughout the network.

\subsection{SPD Linear Layers}

SPD transformations are implemented through congruence maps of the form
\begin{equation}
\mathcal{L}_{W}(C) = W^\top C W,
\end{equation}
where $W \in \mathbb{R}^{d_{\text{in}} \times d_{\text{out}}}$ is a learnable
parameter matrix. To ensure positive definiteness during training, we apply
regularization,
\[
C_{\text{out}} = W^\top C W + \epsilon I, \qquad \epsilon = 10^{-4}.
\]

\subsection{Geometry-Aware Initialization for Dimension-Changing Layers}
\label{app:init}

To stabilize deep congruence learning, we estimate a global covariance structure
after Riemannian Alignment as the Euclidean mean of the aligned covariances,
\[
M = \frac{1}{N}\sum_{i,e} C'_{i,e},
\]
and compute its eigendecomposition $M = U \Lambda U^\top$, where
$U\in\mathbb{R}^{d\times d}$ is an orthonormal eigenbasis capturing dominant
spatial modes.

For DLDCT layers that modify dimensionality, we still derive a single common
rotation basis $U$ from $M$ and lift it to the required layer width by
concatenating multiple copies of $U$. Concretely, for a layer mapping
$d_\ell \to d_{\ell+1}$, we define $k_{\ell+1}=\lceil d_{\ell+1}/d\rceil$ and
construct an expanded basis
\[
\widetilde{U}_{\ell+1}
=
\frac{1}{\sqrt{k_{\ell+1}}}\,[\,U\ \cdots\ U\,]
\in \mathbb{R}^{d\times (k_{\ell+1}d)},
\]
then truncate to the first $d_{\ell+1}$ columns. This yields a geometry-aligned
basis at the target width even when $d_{\ell+1}$ is far from $d$ (e.g.,
$3d\!\to\!4d$ or $5d\!\to\!4d$) by using $3,4,5$ copies of $U$ as needed.

We apply the same principle for the input width by forming
$k_{\ell}=\lceil d_\ell/d\rceil$ and
\[
\widetilde{U}_{\ell}
=
\frac{1}{\sqrt{k_{\ell}}}\,[\,U\ \cdots\ U\,]
\in \mathbb{R}^{d\times (k_{\ell}d)},
\]
truncated to the first $d_\ell$ columns. The congruence map is then initialized
from these lifted bases by
\[
W_\ell
=
\widetilde{U}_{\ell}[1{:}d_\ell,\,1{:}d_\ell]^\top\,
\widetilde{U}_{\ell+1}[1{:}d,\,1{:}d_{\ell+1}],
\]
which provides a geometry-consistent rotation at initialization while matching
the required input--output dimensionality.

In practice, concatenating multiple copies of $U$ may appear ad hoc; however, it
provides a strong inductive bias compared to random initialization, which we
found to overfit easily in deep congruence stacks. Initializing $W_\ell$ from
the lifted eigenbasis yields well-conditioned, near full-rank maps early in
training and empirically improves stability and generalization.

\subsection{Skip--Merge Operations}

Skip connections in DDCT--UNet employ log--Euclidean fusion to preserve SPD
structure,
\begin{equation}
\mathrm{merge}(C_1, C_2)
=
\exp_{\mathrm{sym}}\!\left(
\tfrac{1}{2}\big(
\log_{\mathrm{spd}}(C_1) + \log_{\mathrm{spd}}(C_2)
\big)
\right),
\end{equation}
where $\log_{\mathrm{spd}}$ and $\exp_{\mathrm{sym}}$ denote matrix logarithm and
exponential computed via eigendecomposition in double precision. Eigenvalues are
clamped to $\lambda_{\min}=10^{-5}$ for numerical stability.

\subsection{Encoder--Decoder Architecture}

For DDCT--UNet (E2E), the default encoder--decoder dimensional schedule is
$(d, d/2, d/4, d/2, d)$ to maintain symmetry. For $22$-channel inputs, we adopt
$(d,16,12,16,d)$ to provide mild compression without excessive depth. Ablation
studies in Section~\ref{tab:within_dataset_ablation_bci2a} show that this design
offers a favorable trade-off between expressivity and stability.

\subsection{Custom TSLR Implementation for E2E models}
\label{app:tslr}

E2E models employ a \textbf{custom TSLR-style tangent classifier} that differs from classical tangent-space logistic regression in its choice of reference point.
Specifically, the input to the TSLR module is the set of SPD covariances produced by the E2E prealigner backbone after geometric pre-alignment and deep congruence transformations.

Standard TSLR projects all samples to the tangent space at a \emph{fixed} Riemannian mean computed over the entire training set.
In contrast, E2E models use a \emph{batch-adaptive log-Euclidean reference} during training.
Given a mini-batch of backbone outputs $\{C_i\}_{i=1}^B \subset \mathbb{S}_{++}^d$, we compute
\begin{equation}
\bar{S} = \frac{1}{B} \sum_{i=1}^B \log(C_i),
\qquad
C_{\mathrm{ref}} = \exp(\bar{S}),
\end{equation}
and align each covariance via the congruence transform
\begin{equation}
C_i \;\mapsto\; C_{\mathrm{ref}}^{-1/2} C_i C_{\mathrm{ref}}^{-1/2}.
\end{equation}
The aligned covariances are then projected to the tangent space, vectorized, and classified using a linear softmax layer. This batch-level reference preserves the geometric interpretation of TSLR while avoiding reliance on a single global training mean, which can be suboptimal under strong cross-subject shift.
Because the reference is recomputed on-the-fly, the tangent projection remains fully differentiable and allows alignment and classification to be optimized jointly.
Empirically, this design improves robustness to non-stationarity and reduces overfitting to subject-specific covariance statistics under LOSO evaluation. At test time, the tangent reference is computed over the test batch itself, mirroring the training procedure and avoiding mismatch between train and test geometry.
All tangent-space operations are performed in double precision, with eigenvalue jitter ($\epsilon = 10^{-4}$) applied prior to matrix logarithms to ensure numerical stability.
% ============================================================

\section{Broader Impact Statement}

This work advances geometry-aware learning methods for EEG-based motor-imagery decoding, with potential applications in brain-computer interfaces (BCIs), neurotechnology, and biomedical signal analysis. By improving transductive cross-subject generalization, the proposed framework may reduce the need for subject-specific calibration, thereby increasing the accessibility and usability of BCI systems for users who cannot easily provide extensive training data.

At the same time, the use of EEG data raises important considerations related to data privacy, informed consent, and responsible deployment, particularly in clinical or assistive settings. While the proposed methods operate on covariance representations rather than raw signals, appropriate safeguards must still be maintained when handling neural data. Moreover, model outputs should not be over-interpreted as clinical or cognitive assessments without proper validation and domain expertise.

The methods presented are intended for research and controlled application contexts, and their deployment in real-world systems should be accompanied by human oversight and adherence to established ethical guidelines. Beyond these considerations, the work introduces no new mechanisms for surveillance, profiling, or autonomous decision-making, and we do not anticipate significant additional societal risks beyond those commonly associated with EEG-based machine learning research.

\section{Theoretical Remarks and Failure Modes}
\label{app:theory}
\subsection{When Congruence Transformations Can Fail}
\label{app:congruence_failure}
Congruence transformations $C \mapsto W^\top C W$ form the basis of SPD neural networks, preserving positive definiteness under the assumption that $W$ has full column rank.
However, this geometric preservation does not guarantee that \emph{discriminative class structure} is maintained under arbitrary transformations.

\textbf{Failure Mode (Action Mixing in Low-Rank Regimes):}
Consider a setting where class-conditional covariances $C_k \in \mathbb{S}_{++}^d$ exhibit discriminative structure primarily in a low-dimensional subspace $V \subset \mathbb{R}^d$ with $\dim(V) = r \ll d$.
If a congruence transformation $W \in \mathbb{R}^{d \times d'}$ is learned such that $W^\top V$ has reduced rank, the transformed covariances $\tilde{C}_k = W^\top C_k W$ may collapse distinct classes into a shared subspace, eliminating discriminability.
Formally, let $C_k = V D_k V^\top + \sigma^2 I$ where $D_k$ encodes class-specific eigenvalues.
Then:
\begin{equation}
W^\top C_k W = W^\top V D_k V^\top W + \sigma^2 W^\top W.
\end{equation}
If $\text{rank}(W^\top V) < r$, the class structure encoded in $D_k$ is partially or fully lost.
This failure mode is empirically observed in low-channel settings (e.g., BCI-IV-2b with 3 channels) where aggressive dimensionality reduction can inadvertently suppress discriminative signals.

\textbf{Mitigation Strategies:}
\begin{itemize}
    \item \emph{Fisher regularization}: Enforcing between-class separation in the transformed space prevents action mixing by explicitly penalizing $W$ configurations that collapse classes.
    \item \emph{Architecture constraints}: Limiting compression depth (e.g., DDCT-UNet (E2E) classifier's mild compression to $d/2$) reduces the risk of rank deficiency.
    \item \emph{Skip connections}: Geometric skip merges reintroduce original covariance structure, counteracting over-aggressive compression.
\end{itemize}
\subsection{When Skip Connections Hurt Rather Than Help}
\label{app:skip_failure}
While skip connections generally stabilize deep architectures, they can be detrimental in SPD networks if they bypass essential alignment stages or violate geometric consistency.
\textbf{Skip Connections and Subject Variability.}
In motor imagery BCI, subject-specific covariance bias dominates inter-subject variance.
If skip connections directly merge pre-aligned and post-aligned representations without accounting for subject-specific geometry, they risk reintroducing the very biases that alignment aims to remove.
Formally, let $C_{\text{in}}$ be a subject-biased input covariance, and $\tilde{C}$ be the aligned representation after congruence transformations.
A naive residual connection $C_{\text{out}} = \tilde{C} + C_{\text{in}}$ (or Euclidean averaging) violates manifold geometry and may reintroduce subject bias.
Instead, we enforce \emph{geometry-aware fusion}:
\begin{equation}
C_{\text{out}} = \exp_{\text{sym}}\left( \frac{1}{2} \left( \log_{\text{spd}}(\tilde{C}) + \log_{\text{spd}}(C_{\text{in}}) \right) \right).
\end{equation}
This log-Euclidean merge preserves positive definiteness and respects manifold structure while reintroducing multi-scale information. Ablation studies (Table~\ref{tab:within_dataset_ablation_bci2a}) demonstrate that removing skip connections in DDCT-UNet (E2E) classifier causes catastrophic performance collapse (e.g., 55.4\% $\to$ 48.2\% on BCI-IV-2a).
However, arbitrary skip insertion (e.g., adding skips at every layer) does not improve performance and can degrade accuracy by overwhelming the alignment pathway.
This suggests skip connections must be \emph{strategically placed} at architectural bottlenecks where information loss is most severe.
\subsection{Degeneracy Avoidance in Deep SPD Chains}
\label{app:degeneracy}
Deep chains of SPD transformations are susceptible to multiple forms of degeneracy:
\begin{enumerate}
    \item \textbf{Eigenvalue Collapse:} Repeated congruence transforms can drive eigenvalues toward zero or infinity, causing numerical instability.
    \item \textbf{Rank Deficiency:} Aggressive dimensionality reduction followed by expansion may fail to recover full-rank representations.
    \item \textbf{Gradient Vanishing/Explosion:} Manifold-constrained gradients can exhibit extreme magnitudes in deep networks.
\end{enumerate}
\textbf{Mitigation via Regularization and Regularized Eigendecomposition:}
We employ three complementary mechanisms:
\begin{itemize}
    \item \emph{Fisher-based regularization}: Enforcing action separation throughout the network maintains bounded eigenvalue spectra by preventing collapse to constant representations.
    \item \emph{Hierarchical fusion}: Skip merges preserve multi-scale eigenvalue structure, preventing rank deficiency at decoder layers.
    \item \emph{Jitter regularization}: Adding $\epsilon I$ before every matrix logarithm ensures eigenvalues remain positive: $\lambda_i \geq \epsilon$.
\end{itemize}
\textbf{Empirical Depth Limits:}
Ablation experiments (Table~\ref{tab:within_dataset_ablation_bci2a}) on DDCT-E2E reveal a non-monotonic relationship between depth and accuracy.
Increasing depth from 4 to 5 layers \emph{degrades} performance (e.g., 56.1\% $\to$ 55.7\% on BCI-IV-2a), likely due to accumulated numerical errors and gradient instability.
This suggests that stable learning on SPD manifolds requires \emph{jointly constrained geometry, supervision, and topology} rather than increased depth alone.

\textbf{Theoretical Analysis:}
Let $L(\mathcal{W})$ denote the loss over SPD parameters $\mathcal{W} = \{W_1, \ldots, W_L\}$ in an $L$-layer network.
The Riemannian gradient update for layer $\ell$ is:
\begin{equation}
W_\ell \leftarrow W_\ell - \eta \, \nabla_{W_\ell} L(\mathcal{W}),
\end{equation}
where the gradient norm scales with the product of Jacobians across layers.
In deep SPD networks, this product can exhibit exponential growth or decay with $L$, motivating architectural constraints (balanced dimensionality schedules, skip connections) and gradient clipping to stabilize optimization.

\section{Subject-wise Ablation Results}
\label{subsec:subject_wise_ablation}

\begin{table*}[t]
\centering
\setlength{\tabcolsep}{3pt}
\renewcommand{\arraystretch}{1}
\caption{
Per-subject LOSO accuracy (\%) across datasets.
Geometric alignment methods (RA, DCT, DLDCT, DDCT-UNet) are evaluated with classical MI classifiers.
}
\resizebox{\textwidth}{!}{
\begin{tabular}{llcccccccccccc}
\toprule
\textbf{Dataset}
& \textbf{Subject}
& \multicolumn{3}{c}{\textbf{RA}}
& \multicolumn{3}{c}{\textbf{DCT}}
& \multicolumn{3}{c}{\textbf{DLDCT}}
& \multicolumn{3}{c}{\textbf{DDCT-UNet}} \\
\cmidrule(lr){3-5} \cmidrule(lr){6-8} \cmidrule(lr){9-11} \cmidrule(lr){12-14}
& 
& MDM & TSLR & TSA
& MDM & TSLR & TSA
& MDM & TSLR & TSA
& MDM & TSLR & TSA \\
\midrule

% ================= BCI-IV-2a =================
\multirow{9}{*}{BCI-IV-2a}
& A01 & 61.8 & 61.5 & 67.7 & 64.2 & 66.0 & \underline{68.4}
      & 61.5 & 66.3 & 61.5
      & 61.8 & \textbf{67.0} & 67.0 \\
& A02 & 26.4 & 29.5 & 28.8 & 25.3 & 23.6 & 29.2
      & 26.4 & 29.5 & \underline{30.9}
      & 27.4 & \textbf{30.6} & 29.5 \\
& A03 & 72.9 & 64.9 & 74.0 & 68.8 & 74.0 & 74.3
      & 72.9 & 71.5 & 67.0
      & \underline{78.1} & \textbf{76.0} & 73.6 \\
& A04 & 44.8 & 44.4 & \underline{47.9} & 44.1 & 44.4 & 42.0
      & 44.8 & 46.9 & 47.6
      & 43.8 & \textbf{47.2} & 45.8 \\
& A05 & 42.7 & 38.5 & 43.4 & 36.8 & 37.9 & 43.1
      & 42.7 & 39.9 & 41.3
      & \underline{43.8} & \textbf{43.1} & 43.4 \\
& A06 & 32.3 & \underline{42.7} & 37.5 & 34.7 & 37.9 & 38.9
      & 32.3 & 41.0 & 41.0
      & 32.6 & \textbf{39.6} & 39.2 \\
& A07 & 59.4 & 45.1 & 47.2 & 59.4 & \underline{60.4} & 50.4
      & 59.4 & 46.9 & 46.5
      & 56.2 & \textbf{50.0} & 48.6 \\
& A08 & 71.2 & 68.1 & 72.6 & 66.3 & 73.3 & \underline{75.7}
      & 71.2 & 70.5 & 70.8
      & 72.9 & \textbf{72.9} & 73.6 \\
& A09 & 60.4 & 60.4 & 62.5 & 54.5 & 59.7 & \underline{63.9}
      & 60.4 & \underline{63.9} & 59.4
      & 61.1 & \underline{\textbf{63.9}} & 62.8 \\
\midrule
&
& 52.4$\pm$16.6 & 50.6$\pm$13.4 & 53.5$\pm$16.2
& 50.5$\pm$15.7 & 53.0$\pm$17.7 & 54.0$\pm$17.0
& 52.4$\pm$15.6 & 52.9$\pm$14.5 & 51.8$\pm$12.7
& 53.1$\pm$17.4 & \textbf{54.5$\pm$16.0} & 53.7$\pm$16.0 \\

% ================= BCI-IV-2b =================
\midrule
\multirow{9}{*}{BCI-IV-2b}
& B01 & \textbf{59.8} & 59.8 & 58.5 & \underline{60.8} & 58.3 & 58.3
      & 59.8 & 59.8 & 58.0
      & 60.3 & 60.3 & 58.8 \\
& B02 & \textbf{47.5} & 47.5 & 44.5 & \underline{50.5} & 45.8 & 45.5
      & 47.5 & 47.8 & 45.0
      & 46.0 & 47.5 & 45.5 \\
& B03 & \textbf{59.0} & 55.5 & 56.0 & \underline{59.5} & 57.3 & 56.5
      & 59.0 & 55.5 & 55.0
      & 53.5 & 58.3 & 56.0 \\
& B04 & \textbf{81.4} & 82.1 & 81.9 & 72.1 & 81.7 & \underline{82.6}
      & 81.4 & \underline{82.6} & 82.1
      & 80.0 & 79.8 & \underline{82.6} \\
& B05 & \textbf{55.7} & 55.5 & 55.5 & 54.5 & 55.0 & 55.2
      & 55.7 & 55.5 & 55.2
      & 53.1 & \underline{56.2} & 54.8 \\
& B06 & \textbf{52.0} & 50.8 & 51.8 & 49.3 & 53.0 & 52.5
      & 52.0 & 50.8 & 52.0
      & \underline{54.3} & 51.0 & 53.0 \\
& B07 & \textbf{54.3} & 52.0 & 51.3 & \underline{55.0} & 52.3 & 50.8
      & 54.3 & 52.0 & 50.3
      & 52.0 & 48.0 & 49.5 \\
& B08 & \underline{\textbf{68.0}} & \underline{68.0} & 67.0 & 64.3 & 67.3 & 66.4
      & \underline{68.0} & \underline{68.0} & 66.8
      & 67.5 & 66.1 & 67.7 \\
& B09 & \textbf{66.3} & 65.5 & \underline{68.5} & 64.0 & 65.8 & 68.3
      & 66.3 & 65.5 & 67.5
      & 59.3 & 65.5 & 65.0 \\
\midrule
&
& \textbf{60.4$\pm$10.2} & 59.6$\pm$10.8 & 59.4$\pm$11.3
& 58.9$\pm$7.4 & 59.6$\pm$10.6 & 59.6$\pm$11.2
& 60.4$\pm$9.6 & 59.7$\pm$10.3 & 59.1$\pm$10.7
& 58.4$\pm$10.1 & 59.2$\pm$10.3 & 59.2$\pm$11.2 \\

\midrule
\multirow{14}{*}{BNCI2014}
& S01 & 50.0 & 46.0 & 50.0 & \underline{\textbf{57.0}} & 55.0 & 51.0 & 51.0 & 51.0 & 46.0 & 52.0 & 53.0 & 53.0 \\
& S02 & 52.0 & 58.0 & \underline{60.0} & \textbf{49.0} & 51.0 & 59.0 & 52.0 & 59.0 & 58.0 & 48.0 & 57.0 & 58.0 \\
& S03 & 62.0 & 58.0 & 58.0 & \underline{\textbf{66.0}} & 62.0 & 55.0 & 62.0 & 59.0 & 56.0 & 65.0 & 61.0 & 57.0 \\
& S04 & 47.0 & 51.0 & 51.0 & \textbf{50.0} & \underline{52.0} & 47.0 & 47.0 & 47.0 & 51.0 & 49.0 & 49.0 & \underline{52.0} \\
& S05 & 53.0 & 62.0 & 62.0 & \textbf{60.0} & 59.0 & 61.0 & 54.0 & 62.0 & 62.0 & 58.0 & \underline{63.0} & 62.0 \\
& S06 & \underline{60.0} & 55.0 & 50.0 & \underline{\textbf{60.0}} & 57.0 & 48.0 & \underline{60.0} & 53.0 & 55.0 & 59.0 & 55.0 & 50.0 \\
& S07 & 58.0 & 56.0 & 53.0 & \textbf{58.0} & 57.0 & 57.0 & 58.0 & 56.0 & 56.0 & \underline{60.0} & 57.0 & 55.0 \\
& S08 & 48.0 & 49.0 & 48.0 & \textbf{51.0} & 50.0 & 50.0 & 48.0 & 50.0 & 50.0 & \underline{54.0} & \underline{54.0} & 51.0 \\
& S09 & \underline{58.0} & 52.0 & 53.0 & \textbf{57.0} & 55.0 & 55.0 & 58.0 & 50.0 & 50.0 & 54.0 & 53.0 & 55.0 \\
& S10 & 53.0 & 56.0 & 54.0 & \textbf{50.0} & 51.0 & 56.0 & 55.0 & \underline{58.0} & 56.0 & 56.0 & 57.0 & 54.0 \\
& S11 & \underline{55.0} & 49.0 & 48.0 & \underline{\textbf{55.0}} & 53.0 & 44.0 & \underline{55.0} & 49.0 & 49.0 & 53.0 & 47.0 & 45.0 \\
& S12 & 50.0 & 53.0 & 53.0 & \textbf{47.0} & 43.0 & 48.0 & 50.0 & 50.0 & \underline{55.0} & 46.0 & 50.0 & 52.0 \\
& S13 & 56.0 & 55.0 & 55.0 & \underline{\textbf{62.0}} & 58.0 & 55.0 & 56.0 & 54.0 & 52.0 & 57.0 & 56.0 & 54.0 \\
& S14 & 49.0 & 58.0 & 56.0 & \textbf{52.0} & 51.0 & 56.0 & 49.0 & \underline{59.0} & 54.0 & 46.0 & 53.0 & 57.0 \\
\midrule
&
& 53.6$\pm$4.7 & 54.1$\pm$4.4 & 53.6$\pm$4.3
& \textbf{55.3$\pm$5.6} & 53.9$\pm$4.8 & 53.0$\pm$5.0
& 53.9$\pm$4.5 & 54.1$\pm$4.6 & 53.6$\pm$4.0
& 54.1$\pm$5.6 & 54.6$\pm$4.4 & 53.9$\pm$4.1 \\

\midrule
\multirow{9}{*}{BNCI2015}
& A & 25.9 & 24.4 & 22.5 & 25.0 & \underline{26.3} & 22.2 & \underline{26.3} & 24.7 & 25.6 & \textbf{24.4} & 23.8 & 24.7 \\
& C & 25.0 & 25.3 & 25.3 & 23.8 & 23.8 & 25.6 & 25.0 & 25.6 & 25.3 & \textbf{22.8} & \underline{25.9} & 25.3 \\
& D & 25.9 & 25.0 & 25.6 & 25.9 & 27.2 & 25.6 & 25.6 & 25.3 & 24.7 & \textbf{26.9} & 26.9 & \underline{27.2} \\
& E & 26.7 & 25.3 & 22.7 & 19.0 & 22.3 & 22.3 & 27.0 & 25.0 & 25.0 & \underline{\textbf{28.3}} & 24.0 & 22.7 \\
& F & \underline{24.4} & 22.8 & 23.4 & 20.9 & 20.6 & 22.5 & \underline{24.4} & 22.8 & 24.1 & \textbf{23.8} & 21.9 & 21.9 \\
& G & 25.3 & 26.9 & 27.5 & 26.9 & 23.8 & 25.0 & 25.3 & 26.9 & 28.1 & \textbf{24.7} & 27.5 & \underline{27.8} \\
& H & 24.0 & 25.0 & 23.7 & 24.0 & 23.0 & 25.3 & 24.0 & 25.3 & 25.3 & \textbf{25.7} & \underline{26.0} & 25.7 \\
& J & 21.9 & 25.0 & 23.4 & 20.9 & 17.8 & 22.8 & 22.2 & 25.0 & 24.1 & \underline{\textbf{25.9}} & \underline{25.9} & 22.5 \\
& L & 25.0 & 23.1 & 24.7 & 24.7 & 22.5 & 25.6 & 25.6 & 23.8 & 22.5 & \underline{\textbf{27.5}} & 25.0 & 24.1 \\
\midrule
&
& 24.9$\pm$1.4 & 24.8$\pm$1.2 & 24.3$\pm$1.6
& 23.5$\pm$2.6 & 23.0$\pm$2.8 & 24.1$\pm$1.6
& 25.0$\pm$1.3 & 24.9$\pm$1.1 & 25.0$\pm$1.4
& \textbf{25.6$\pm$1.8} & 25.2$\pm$1.8 & 24.6$\pm$2.1 \\

\midrule
% ================= Ofner2017 =================
\multirow{15}{*}{Ofner2017}
& S1  & 25.0 & 21.9 & 20.8 & 23.9 & \textbf{25.3} & 20.8
       & \underline{25.6} & 21.9 & 21.9
       & 23.9 & 22.2 & 20.8 \\
& S2  & 19.4 & \underline{21.9} & 19.4 & 18.9 & \textbf{19.7} & 19.4
       & 19.4 & \underline{21.9} & 21.7
       & 18.6 & \underline{21.9} & 19.4 \\
& S3  & 17.8 & 14.7 & 15.0 & 18.9 & \underline{\textbf{19.2}} & 14.2
       & 17.8 & 14.7 & 15.0
       & 18.3 & 16.4 & 16.4 \\
& S4  & 18.9 & 21.4 & 20.3 & 17.2 & \textbf{20.3} & \underline{21.7}
       & 18.6 & 21.4 & 21.1
       & 20.0 & 21.1 & 21.1 \\
& S5  & \underline{20.3} & 17.5 & 16.1 & 20.0 & \textbf{19.4} & 16.7
       & 20.3 & 17.5 & 17.2
       & 19.7 & 17.8 & 16.1 \\
& S6  & 16.4 & 15.3 & 15.0 & 18.3 & \underline{\textbf{18.9}} & 15.6
       & 16.4 & 15.0 & 14.7
       & 15.0 & 14.7 & 16.7 \\
& S7  & 18.1 & 19.7 & 18.3 & 15.8 & \textbf{16.9} & 17.2
       & 17.8 & \underline{20.0} & 18.9
       & 18.1 & 18.9 & 18.9 \\
& S8  & \underline{18.9} & 17.2 & 17.2 & 17.2 & \textbf{15.8} & 17.5
       & 18.6 & 17.2 & 17.5
       & 17.2 & 15.3 & 16.9 \\
& S9  & 17.8 & 14.7 & 15.3 & 15.6 & \textbf{16.4} & 16.9
       & 17.8 & 14.7 & 14.7
       & \underline{18.3} & 15.6 & 16.9 \\
& S10 & 16.7 & 16.7 & 17.8 & 19.2 & \textbf{18.6} & \underline{19.2}
       & 16.7 & 16.9 & 18.3
       & 18.3 & 17.8 & 16.9 \\
& S11 & 15.0 & 18.6 & \underline{20.0} & 17.5 & \textbf{19.7} & \underline{20.0}
       & 15.3 & 18.9 & 19.4
       & 18.1 & 19.2 & 17.8 \\
& S12 & \underline{20.0} & 18.6 & 17.2 & 15.3 & \textbf{16.7} & 15.3
       & \underline{20.0} & 18.6 & 18.6
       & 18.3 & 18.6 & 18.3 \\
& S13 & 14.4 & 17.2 & 16.9 & 17.2 & \underline{\textbf{18.9}} & 15.6
       & 14.2 & 16.9 & 16.1
       & 16.1 & 17.5 & 15.6 \\
& S14 & 17.2 & 18.6 & 18.1 & 20.6 & \underline{\textbf{21.7}} & 17.5
       & 17.8 & 18.3 & 18.3
       & 18.6 & 18.9 & 18.9 \\
& S15 & 17.2 & 13.6 & 15.0 & \underline{18.1} & \textbf{16.9} & 15.8
       & 16.9 & 13.6 & 14.4
       & 17.8 & 14.2 & 15.8 \\
\midrule
&
& 18.2$\pm$2.5 & 17.9$\pm$2.6 & 17.5$\pm$2.0
& 18.2$\pm$2.2 & \textbf{19.0$\pm$2.4} & 17.6$\pm$2.2
& 18.2$\pm$2.5 & 17.9$\pm$2.6 & 17.9$\pm$2.4
& 18.4$\pm$2.0 & 18.0$\pm$2.5 & 17.8$\pm$1.8 \\

\bottomrule
\end{tabular}
}
\end{table*}

\begin{table*}[t]
\centering
\tiny
\setlength{\tabcolsep}{5pt}
\renewcommand{\arraystretch}{0.4}
\caption{
Subject-wise LOSO accuracy (\%) across all evaluated MI datasets.
Rows are grouped by dataset under cross-subject evaluation.
Mean $\pm$ Std is reported per dataset.
}
\begin{tabular}{llccccccc}
\toprule
\textbf{Dataset} & \textbf{Subject}
& \textbf{MDM}
& \textbf{TSLR}
& \textbf{TSA-LDA}
& \textbf{SPD-CNN}
& \textbf{DCT (E2E)}
& \textbf{DLDCT (E2E)}
& \textbf{DDCT-UNet (E2E)} \\
\midrule

% ================= BCI-IV-2a =================
\multirow{10}{*}{BCI-IV-2a}
& A01 & 61.81 & 66.32 & 67.71 & 59.38 & 63.89 & \textbf{67.71} & \underline{69.44} \\
& A02 & 26.39 & 29.17 & 28.82 & \underline{30.21} & 28.47 & \underline{\textbf{30.21}} & 29.17 \\
& A03 & 72.92 & 71.18 & 73.96 & 70.83 & 70.49 & \underline{\textbf{83.33}} & 80.56 \\
& A04 & 44.79 & 46.88 & \underline{47.92} & 44.10 & 44.10 & \textbf{42.71} & 45.14 \\
& A05 & 42.71 & 39.93 & 43.40 & 36.46 & 39.58 & \underline{\textbf{44.79}} & 42.71 \\
& A06 & 32.29 & \underline{40.97} & 37.50 & 35.42 & 36.46 & \textbf{39.93} & 39.58 \\
& A07 & 59.38 & 47.22 & 47.22 & 39.93 & \underline{59.72} & \textbf{54.51} & 52.08 \\
& A08 & 71.18 & 70.49 & 72.57 & 65.97 & 71.53 & \underline{\textbf{76.04}} & 75.00 \\
& A09 & 60.42 & 63.89 & 62.50 & 54.17 & 61.11 & \underline{\textbf{65.28}} & 64.93 \\
\midrule
&
& 52.43$\pm$16.61 & 52.89$\pm$15.36 & 53.51$\pm$16.21
& 48.50$\pm$14.57
& 52.82$\pm$14.95
& \textbf{56.06$\pm$18.03}
& 55.40$\pm$17.75 \\
\midrule

% ================= BCI-IV-2b =================
\multirow{10}{*}{BCI-IV-2b}
& B01 & \textbf{59.75} & 59.75 & 58.50 & 59.50 & \underline{60.50} & 58.75 & 59.50 \\
& B02 & \textbf{47.50} & 47.75 & 44.50 & \underline{48.25} & 49.50 & 47.50 & 46.50 \\
& B03 & \underline{\textbf{59.00}} & 55.50 & 56.00 & 57.50 & 57.25 & 57.00 & 57.25 \\
& B04 & \textbf{81.43} & \underline{82.62} & 81.90 & 80.24 & 80.48 & 77.38 & 80.48 \\
& B05 & \textbf{55.71} & 55.48 & 55.48 & \underline{56.67} & 56.19 & 56.43 & 55.95 \\
& B06 & \textbf{52.00} & 50.75 & 51.75 & 51.75 & 52.25 & \underline{53.25} & 52.00 \\
& B07 & \textbf{54.25} & 52.00 & 51.25 & 53.50 & 51.25 & \underline{54.50} & 52.75 \\
& B08 & \underline{\textbf{67.95}} & \underline{67.95} & 67.05 & 65.68 & \underline{67.95} & 65.45 & 67.05 \\
& B09 & \textbf{66.25} & 65.50 & \underline{68.50} & 65.50 & 65.25 & \underline{68.50} & 65.25 \\
\midrule
&
& \textbf{60.43$\pm$10.20} & 59.70$\pm$10.87 & 59.44$\pm$11.31
& 59.84$\pm$9.6
& 60.07$\pm$9.31
& 59.86$\pm$9.08
& 59.64$\pm$10.11 \\
\midrule

% ================= BNCI2014 =================
\multirow{15}{*}{BNCI2014}
& S01 & 50 & 51 & 50 & 52 & \underline{54} & 52 & \underline{\textbf{54}} \\
& S02 & 52 & 60 & 60 & 53 & 52 & 58 & \underline{\textbf{63}} \\
& S03 & \underline{62} & 59 & 58 & \underline{62} & 57 & 57 & \textbf{55} \\
& S04 & 47 & 47 & 52 & \underline{56} & 52 & 53 & \textbf{52} \\
& S05 & 53 & 62 & 62 & 59 & 59 & 56 & \underline{\textbf{60}} \\
& S06 & 60 & 53 & 51 & \underline{61} & 59 & 56 & \textbf{54} \\
& S07 & 58 & 55 & 53 & 51 & 56 & \underline{59} & \textbf{56} \\
& S08 & 48 & 50 & 48 & 48 & 49 & \underline{53} & \textbf{51} \\
& S09 & \underline{58} & 50 & 53 & 47 & 54 & 51 & \textbf{56} \\
& S10 & 53 & \underline{58} & 54 & 52 & 52 & 54 & \textbf{57} \\
& S11 & \underline{55} & 49 & 48 & 50 & 52 & 50 & \textbf{51} \\
& S12 & 50 & 52 & \underline{53} & 51 & 50 & 50 & \textbf{52} \\
& S13 & 56 & 54 & 54 & 52 & \underline{60} & 57 & \textbf{56} \\
& S14 & 49 & \underline{59} & 56 & 56 & \underline{59} & 57 & \textbf{58} \\
\midrule
&
& 53.64$\pm$4.67 & 54.21$\pm$4.69 & 53.71$\pm$4.14
& 53.57$\pm$4.6
& 54.64$\pm$3.54
& 54.50$\pm$3.03
& \textbf{55.36$\pm$3.46} \\
\midrule

% ================= BNCI2015 =================
\multirow{10}{*}{BNCI2015}
& A & 25.94 & 24.38 & 22.50 & \underline{26.56} & 25.31 & 23.44 & \textbf{23.44} \\
& C & 25.00 & \underline{25.31} & \underline{25.31} & 25.00 & 26.25 & 22.50 & \textbf{25.00} \\
& D & 25.94 & 25.00 & 25.63 & 25.31 & 24.06 & \underline{27.50} & \textbf{26.25} \\
& E & \underline{26.67} & 25.33 & 22.67 & 26.00 & 20.00 & 21.67 & \textbf{22.00} \\
& F & 24.38 & 22.81 & 23.44 & 22.50 & 25.00 & 23.44 & \underline{\textbf{25.94}} \\
& G & 25.31 & 26.88 & 27.50 & 23.75 & 26.88 & 27.81 & \underline{\textbf{29.38}} \\
& H & 24.00 & 25.00 & 23.67 & 25.67 & 25.33 & 24.67 & \underline{\textbf{28.67}} \\
& J & 21.88 & \underline{25.00} & 23.44 & 24.06 & 20.63 & 23.44 & \textbf{22.19} \\
& L & 25.00 & 23.13 & 24.69 & 25.00 & \underline{27.81} & 24.69 & \textbf{25.63} \\
\midrule
&
& 24.90$\pm$1.40 & 24.76$\pm$1.22 & 24.31$\pm$1.62
& 24.87$\pm$1.25
& 24.59$\pm$2.51
& 24.35$\pm$2.10
& \textbf{25.39$\pm$2.58} \\
\midrule

% ================= Ofner2017 =================
\multirow{16}{*}{Ofner2017}
& S1 & \underline{25.00} & 21.94 & 20.83 & 13.33 & 20.83 & \textbf{22.78} & 22.22 \\
& S2 & 19.44 & 21.94\ & 19.44 & \underline{22.22} & 20.00 & \textbf{19.44} & 21.67 \\
& S3 & 17.78 & 14.72 & 15.00 & 17.50 & 18.06 & \textbf{18.33} & \underline{20.00} \\
& S4 & 18.89 & 21.39 & 20.28 & 21.11 & 19.72 & \underline{\textbf{21.67}} & 19.72 \\
& S5 & \underline{20.28} & 17.50 & 16.11 & 16.39 & 20.00 & \textbf{18.61} & 15.28 \\
& S6 & 16.39 & 15.28 & 15.00 & 14.17 & 15.56 & \underline{\textbf{16.39}} & 16.11 \\
& S7 & 18.06 & 19.72 & 18.33 & 18.06 & 14.44 & \underline{\textbf{20.28}} & 16.39 \\
& S8 & 18.89 & 17.22 & 17.22 & 16.67 & 17.22 & \textbf{15.28} & \underline{19.17} \\
& S9 & 17.78 & 14.72 & 15.28 & 15.00 & \underline{18.33} & \textbf{16.67} & 15.00 \\
& S10 & 16.67 & 16.67 & 17.78 & 14.72 & 17.22 & \underline{\textbf{18.06}} & 16.94 \\
& S11 & 15.00 & 18.61 & 20.00 & 20.83 & \underline{22.22} & \textbf{20.83} & 21.39 \\
& S12 & \underline{20.00} & 18.61 & 17.22 & 15.56 & 15.00 & \textbf{18.06} & 15.83 \\
& S13 & 14.44 & \underline{17.22} & 16.94 & 16.67 & 15.83 & \textbf{15.00} & 15.28 \\
& S14 & 17.22 & 18.61 & 18.06 & 17.22 & \underline{20.56} & \textbf{18.61} & 15.83 \\
& S15 & 17.22 & 13.61 & 15.00 & 15.00 & 17.22 & \underline{\textbf{19.17}} & 18.33 \\
\midrule
&
& 18.20$\pm$2.52 & 17.85$\pm$2.64 & 17.50$\pm$2.00
& 16.96$\pm$2.64
& 18.15$\pm$2.27
& \textbf{18.61$\pm$2.22}
& 17.94$\pm$2.56 \\

\bottomrule
\end{tabular}
\end{table*}

\begin{table*}[t]
\centering
\footnotesize
\setlength{\tabcolsep}{2pt}
\renewcommand{\arraystretch}{0.6}
\caption{
Subject-wise component ablation on BCI-IV-2a under LOSO evaluation.
Accuracy (\%) is reported per subject.
Mean $\pm$ std is reported across subjects.
Non-applicable configurations are denoted by `---'.
}
\label{tab:subjectwise_component_ablation_bci2a}

\begin{tabular}{l|l ccc ccc}
\toprule
\textbf{Ablation} & \textbf{Subject}
& \multicolumn{3}{c}{\textbf{Pre-aligners}}
& \multicolumn{3}{c}{\textbf{E2E TSLR Variant}} \\
\cmidrule(lr){3-5} \cmidrule(lr){6-8}
& & \textbf{DCT} & \textbf{DLDCT} & \textbf{DDCT-UNet}
& \textbf{DCT} & \textbf{DLDCT} & \textbf{DDCT-UNet} \\
\midrule

% ================= Divergence Regularization =================

\multirow{10}{*}{w/o Fisher (action)}
& A01T & 63.08 & 63.66 & 63.31 & 62.15 & 67.36 & 67.71 \\
& A02T & 29.17 & 33.45 & 31.13 & 25.00 & 29.51 & 32.99 \\
& A03T & 75.35 & 71.53 & 72.45 & 66.32 & 80.90 & 80.21 \\
& A04T & 44.33 & 45.14 & 39.81 & 40.28 & 43.75 & 45.49 \\
& A05T & 34.84 & 36.46 & 36.00 & 38.19 & 41.67 & 43.06 \\
& A06T & 36.23 & 35.65 & 34.84 & 33.33 & 38.19 & 39.58 \\
& A07T & 54.98 & 51.39 & 46.18 & 55.56 & 52.78 & 51.74 \\
& A08T & 68.17 & 68.40 & 63.54 & 65.63 & 71.18 & 71.88 \\
& A09T & 60.07 & 59.03 & 58.33 & 57.64 & 63.89 & 63.54 \\
\midrule
&
& 51.80$\pm$16.30 & 51.63$\pm$14.73 & 49.51$\pm$15.13
& --- & 54.36$\pm$17.33 & 55.13$\pm$16.28 \\

\midrule

\multirow{10}{*}{w/o Fisher (subject)}
& A01T & --- & 48.73 & 65.97 & 62.15 & 67.36 & 67.71 \\
& A02T & --- & 29.40 & 32.18 & 25.00 & 29.17 & 32.64 \\
& A03T & --- & 43.87 & 76.16 & 66.32 & 81.25 & 79.51 \\
& A04T & --- & 33.91 & 45.02 & 40.28 & 44.10 & 45.14 \\
& A05T & --- & 31.71 & 39.35 & 38.19 & 41.32 & 43.75 \\
& A06T & --- & 31.60 & 36.92 & 33.33 & 38.19 & 39.58 \\
& A07T & --- & 37.62 & 52.20 & 55.56 & 52.43 & 52.08 \\
& A08T & --- & 43.29 & 71.18 & 65.63 & 70.49 & 71.18 \\
& A09T & --- & 34.61 & 60.53 & 57.64 & 63.89 & 63.19 \\
\midrule
&
& --- & 37.19$\pm$6.66 & 53.28$\pm$15.94
& --- & 54.24$\pm$17.39 & 54.98$\pm$16.05 \\

\midrule

\multirow{10}{*}{w/o Fisher (all)}
& A01T & --- & --- & --- & 62.15 & 67.36 & 67.71 \\
& A02T & --- & --- & --- & 25.00 & 29.51 & 32.99 \\
& A03T & --- & --- & --- & 66.32 & 80.90 & 80.21 \\
& A04T & --- & --- & --- & 40.28 & 43.75 & 45.49 \\
& A05T & --- & --- & --- & 38.19 & 41.67 & 43.06 \\
& A06T & --- & --- & --- & 33.33 & 37.85 & 39.58 \\
& A07T & --- & --- & --- & 55.56 & 52.78 & 51.74 \\
& A08T & --- & --- & --- & 65.63 & 70.49 & 71.88 \\
& A09T & --- & --- & --- & 57.64 & 63.89 & 63.54 \\
\midrule
& 
& --- & --- & --- & --- & 54.24$\pm$17.29 & 55.13$\pm$16.28 \\

\midrule

% ================= Topological Design =================

\multirow{10}{*}{w/o dispersion}
& A01T & 65.05 & --- & --- & --- & --- & --- \\
& A02T & 29.86 & --- & --- & --- & --- & --- \\
& A03T & 71.64 & --- & --- & --- & --- & --- \\
& A04T & 42.13 & --- & --- & --- & --- & --- \\
& A05T & 36.23 & --- & --- & --- & --- & --- \\
& A06T & 37.04 & --- & --- & --- & --- & --- \\
& A07T & 50.00 & --- & --- & --- & --- & --- \\
& A08T & 68.17 & --- & --- & --- & --- & --- \\
& A09T & 58.56 & --- & --- & --- & --- & --- \\
\midrule
& 
& 51.49$\pm$16.25 & --- & --- & --- & --- & --- \\

\midrule

\multirow{10}{*}{w/o reconstruction}
& A01T & 65.86 & 49.54 & 65.63 & 62.15 & --- & 64.93 \\
& A02T & 26.27 & 28.24 & 32.18 & 25.00 & --- & 33.33 \\
& A03T & 71.76 & 43.98 & 76.27 & 66.32 & --- & 80.56 \\
& A04T & 43.75 & 33.10 & 44.68 & 40.28 & --- & 46.53 \\
& A05T & 37.38 & 32.41 & 40.86 & 38.19 & --- & 44.44 \\
& A06T & 36.23 & 31.83 & 36.92 & 33.33 & --- & 39.24 \\
& A07T & 54.40 & 38.19 & 54.05 & 55.56 & --- & 50.35 \\
& A08T & 68.98 & 45.72 & 70.95 & 65.63 & --- & 73.26 \\
& A09T & 58.80 & 41.20 & 61.11 & 57.64 & --- & 62.15 \\
\midrule
& 
& 51.49$\pm$16.25 & 38.25$\pm$7.30 & 53.63$\pm$15.79
& --- & --- & 54.98$\pm$16.04 \\

\midrule

\multirow{10}{*}{w/o skip merges}
& A01T & --- & 48.15 & 30.21 & --- & --- & 63.54 \\
& A02T & --- & 30.21 & 24.42 & --- & --- & 28.13 \\
& A03T & --- & 45.49 & 35.53 & --- & --- & 78.13 \\
& A04T & --- & 33.22 & 25.93 & --- & --- & 42.71 \\
& A05T & --- & 33.22 & 28.59 & --- & --- & 37.85 \\
& A06T & --- & 31.60 & 27.31 & --- & --- & 38.89 \\
& A07T & --- & 37.27 & 33.91 & --- & --- & 55.21 \\
& A08T & --- & 46.53 & 30.90 & --- & --- & 73.61 \\
& A09T & --- & 37.96 & 33.68 & --- & --- & 59.72 \\
\midrule
& 
& --- & 38.18$\pm$6.89 & 30.05$\pm$3.83
& --- & --- & 53.09$\pm$17.21 \\

\midrule

% ================= Baseline =================
\textbf{Baseline} & 
& \textbf{51.85$\pm$16.21} & \textbf{53.10 ± 17.20} & \textbf{53.09$\pm$17.21}
& \textbf{55.40 ± 17.80} & \textbf{56.06$\pm$18.03} & \textbf{55.40$\pm$17.75} \\

\bottomrule
\end{tabular}
\end{table*}

%\clearpage
%\input{checklist.tex}

\end{document}